\documentclass[11pt,letterpaper]{article}
\usepackage[margin=1in]{geometry}
\usepackage{setspace}
\usepackage{fancyhdr}
\usepackage{iftex}
\ifPDFTeX
\usepackage[utf8]{inputenc}
\fi
\usepackage[T1]{fontenc}

\usepackage[authoryear]{natbib}
\bibpunct[, ]{(}{)}{,}{a}{}{,}
\usepackage{url}

\usepackage{amsmath}
\usepackage{mathtools}
\usepackage{amssymb}
\usepackage{amsfonts}
\usepackage{amsthm}

\usepackage{graphicx}
\usepackage{booktabs}
\usepackage{array}
\usepackage{multirow}

\newcounter{algorithm}
\usepackage{tikz}
\usetikzlibrary{decorations.pathreplacing}

\usepackage{nicefrac}
\usepackage{microtype}
\usepackage{xcolor}
\colorlet{revcolor}{black}

\newtheorem{theorem}{Theorem}[section]
\newtheorem{lemma}{Lemma}[section]
\newtheorem{proposition}{Proposition}[section]
\newtheorem{corollary}{Corollary}[section]
\newtheorem{assumption}{Assumption}[section]

\newtheorem{remark}{Remark}[section]
\theoremstyle{definition}
\newtheorem{example}{Example}[section]
\theoremstyle{plain}

\DeclareMathOperator*{\argmax}{arg\,max}
\DeclareMathOperator*{\argmin}{arg\,min}
\DeclareMathOperator{\dist}{dist}
\DeclareMathOperator{\KL}{KL}

\DeclareMathOperator{\tr}{tr}

\newcommand{\X}{\mathcal X}
\newcommand{\Hk}{\mathcal H_k}
\newcommand{\R}{\mathbb R}

\newcommand{\eps}{\varepsilon}
\newcommand{\Econf}[1][\delta]{\mathcal E_{#1}}

\newcommand{\bbB}{\overline B}

\newcommand{\lo}{\underline{f}}
\newcommand{\hi}{\overline{f}}
\newcommand{\doi}[1]{\href{https://doi.org/#1}{doi:\nolinkurl{#1}}}

\usepackage{hyperref}
\hypersetup{hidelinks,pdfauthor={Yilin Zheng, Haowei Wang, Wenjia Wang, Szu Hui Ng},pdftitle={Identifying Near-Optimal Regions in Continuous Simulation Optimization: Minimax Sample Complexity and an Adaptive Allocation}}

\begin{document}

\thispagestyle{empty}
% Title-only adaptation of the arxiv.sty visual style: thick rules and centered author blocks.
% The remainder of the full-length manuscript retains its original layout.
\begingroup
\setstretch{1}
\noindent\rule{\linewidth}{2pt}\par
\vspace{0.75em}
\begin{center}
{\LARGE\bfseries\scshape Simultaneous Near-Optimal Region Identification in Continuous Simulation Optimization: Minimax Complexity and Adaptive Sampling\par}
\end{center}
\vspace{0.4em}
\noindent\rule{\linewidth}{2pt}\par
\vspace{2.4em}
\begin{center}
\begin{tabular}{@{}c@{\hspace{3.8em}}c@{}}
\textbf{Yilin Zheng} & \textbf{Haowei Wang}\\
National University of Singapore & National University of Singapore\\
Singapore & Singapore\\[1.6em]
\textbf{Wenjia Wang} & \textbf{Szu Hui Ng}\\
National University of Singapore & National University of Singapore\\
Singapore & Singapore
\end{tabular}
\end{center}
\vspace{1.8em}
\endgroup
\noindent\textbf{Abstract.} Simulation optimization typically returns a single estimated optimal point. When decisions also depend on criteria evaluated independently of simulation, such as resource cost, the decision maker instead needs the set of designs whose expected performance lies within a tolerance of the optimum, often choosing the tolerance only after seeing the tradeoffs. We study how to identify these near-optimal regions on a continuous design space, simultaneously for all tolerances up to a prescribed maximum $T$, with region estimation error at most $\eta$. We show that region identification is not a by-product of optimization. Specifically, when $\alpha>\nu$, any procedure, even one given the optimal design in advance, needs at least $\Omega\bigl(\eta^{-(2+d/\nu)}\min\{1,KT^{d/\alpha}\}\bigr)$ simulation evaluations, where $d$ is the dimension, $\nu$ the smoothness, and $(K,\alpha)$ the covering parameters that describe the geometry of the near-optimal designs. We propose SASORI, which returns an estimated optimal point and the regions for all tolerances from the same evaluations, matching this lower bound up to logarithmic factors. SASORI also achieves the optimal tradeoff between region estimation error and cumulative regret, and reduces to a minimax-optimal simulation optimization procedure at zero tolerance. Numerical experiments illustrate the effectiveness of our proposed method.
\par

% \medskip
% \noindent\textbf{Area of review:} Simulation.

% \noindent\textbf{Subject classifications:}
% Simulation: design of experiments---adaptive allocation of simulation effort;
% Simulation: statistical analysis---subset selection over a continuum;
% Statistics: nonparametric---minimax sample complexity.


\section{Introduction}\label{sec:intro}

Stochastic simulation is a standard tool for evaluating designs of complex stochastic systems whose performance cannot be computed in closed form, such as service systems, queueing networks, manufacturing systems, and supply chains. In simulation optimization, each simulation evaluation at a design point consists of one replication and returns a noisy observation of expected performance. Because evaluations are costly, the analyst seeks a design with the best expected performance within a fixed evaluation budget. Ranking-and-selection procedures address this problem when the designs form a finite list \citep{kim2001fully,chen2000ocba,fan2016indifference}, search and metamodeling methods address large or continuous design spaces \citep{hong2006compass,ankenman2010stochastic,frazier2008knowledge,salemi2019gmrf,fan2018surrogate}, and Bayesian optimization fits a Gaussian process metamodel and chooses each new design by optimizing an acquisition function \citep{frazier2018tutorial}. In all these methods, the output is an estimated optimal point, often accompanied by a statistical statement of how close its performance is to the best achievable.

Such an output suffices when simulated performance is the only criterion. In practice, however, the decision maker often implements not the simulated optimum but the most attractive design among those whose simulated performance is acceptable, according to downstream requirements, such as resource availability and operational restrictions, that may not be fully specified during the optimization stage \citep{maus2022discovering,miller2024expected}. Related needs arise in molecular design \citep{verhellen2020illuminating}, chemical process optimization with multiple objectives \citep{schweidtmann2018machine}, and engine control, where \citet{miller2024expected} keep several control configurations of comparable performance because restrictions on fuel blends and energy consumption vary across operating conditions.

As an example, consider a contact center whose design variables $x$ are the staffing levels by period and the routing probabilities. Simulation estimates $f(x)$, the expected fraction of calls answered within a service target, whereas the resource cost is $c_0(x)$, including wages, licenses, and trunk capacity. Management will deploy the least-cost design whose service level lies within an acceptable loss $\tau$ of the best achievable, that is, a minimizer of $c_0$ over the acceptable region
\begin{equation*}
 \mathcal S_\tau=\{x\in\mathcal X:f^\star-f(x)<\tau\},\qquad f^\star=\max_{x\in\mathcal X}f(x).
\end{equation*}
The simulation study must therefore identify a region of the design space rather than a single point. Moreover, $\tau$ is a managerial choice that is best made after the tradeoff between service and cost has been seen, so management needs the acceptable regions for all losses it may consider, which an estimated optimal point alone cannot provide.

The simulation literature has long recognized the value of set-valued output when the alternatives are finite. Subset selection returns a bounded-size subset that contains a near-best system with prescribed probability \citep{sullivan1989restricted}, screening procedures eliminate clearly inferior systems before selection \citep{nelson2001simple}, multiple comparisons with the best construct simultaneous confidence intervals for all systems to identify the systems that could be the best \citep{matejcik1995twostage}, and feasibility determination classifies systems against a stated threshold, permitting either classification within an indifference zone around it \citep{andradottir2010feasibility,batur2010feasible}. These procedures do not extend directly to continuous design variables such as service rates, routing probabilities, or reorder points, for which the acceptable region has unknown shape, extent, and number of components, and infinitely many designs must be classified. Thresholding bandits and kernel level-set estimation classify designs against a prescribed level \citep{locatelli2016thresholding,mason2022levelset,gotovos2013levelset,bogunovic2016truncated}, and continuous-domain methods exploit boundary geometry or local search \citep{shekhar2019levelset,bachoc2021levelset,ngo2024continuous,ngo2026trust}. In our problem, the threshold $f^\star-\tau$ depends on the unknown optimal value. Methods that target designs within an additive loss of an unknown maximum, on a continuous domain \citep{nguyen2021unifying} or among finitely many alternatives \citep{mason2020allgood,almarjani2023allbest}, address a single loss specified in advance (``all'' in the all-$\epsilon$-good-arm formulation refers to the arms at one fixed $\epsilon$), and only the finite-alternative methods come with sample-complexity guarantees. Related work estimates the set of maximizers \citep{minsker2013extreme} or combines region learning with optimization \citep{ewald2026cashomon,zhang2023regions}. Thus neither a continuous domain nor an unknown optimum alone distinguishes our problem. Because $\tau$ is chosen after the study, our output must instead be valid simultaneously for all tolerances in $[0,T]$, where the maximum tolerance $T$ is fixed before sampling. These features raise two questions that have not been answered:
\begin{itemize}
\item[Q1:] \textit{Can one simulation study deliver both an estimated optimal point and the acceptable regions, and if so, at what cost to the efficiency with which the simulation budget is allocated?}
\item[Q2:] \textit{How many simulation evaluations are needed to identify the acceptable regions for all tolerances in a range, to a prescribed accuracy?}
\end{itemize}

To address Q1, we propose Shared-confidence Adaptive Sampling for Optimization and Region Identification (SASORI), a sequential procedure that returns an estimated optimal point together with inner and outer sets $I_\tau\subseteq\mathcal S_\tau\subseteq O_\tau$ for all tolerances $\tau\in[0,T]$ from the same evaluations. With high probability, simultaneously for all these tolerances, all points in $I_\tau$ is acceptable, all points outside $O_\tau$ is unacceptable, and all points of the undetermined set $O_\tau\setminus I_\tau$, which shrinks as evaluations accumulate, have performance within $\eta$ of the threshold $f^\star-\tau$. The region estimation error $\eta$ plays a role similar to the indifference-zone parameter in ranking and selection \citep{kim2001fully,nelson2001simple}, and the manager may choose $\tau$ after seeing the output, without additional evaluations and with the same probability guarantee. SASORI fits a Gaussian process metamodel \citep{rasmussen2006gaussian} to batches of evaluations and maintains a single pair of confidence bounds on the response surface, shared by two tracks: an optimization track refines the estimated optimal point, and a region track identifies the acceptable regions. Each track proceeds in phases with geometrically decreasing target accuracy and samples the maximizer of an acquisition function that is large only at points that are still relevant to the track's tolerance and whose confidence interval is still too wide. Each completed phase tightens the shared bounds, so evaluations spent on the regions also sharpen the estimated optimal point, and vice versa. SASORI does not require the covering parameters of the near-optimal set as inputs and reduces to a simulation optimization procedure when $T=0$.

Our analysis combines a smoothness assumption that is standard in kernel-based metamodeling \citep{ankenman2010stochastic,vakili2021optimal,li2022batches}, with smoothness $\nu$ in dimension $d$, and a condition on the geometry of the near-optimal designs in the spirit of \citet{bubeck2011xarmed} and \citet{shekhar2022instance}, where $K$ is a uniform bound on the number of balls, of radius proportional to $u^{1/\alpha}$, needed to cover the designs with performance gap at most $u$, over a specified range of $u$. Thus $K$ can reflect the number of separated near-optimal peaks, while $\alpha$ controls how the covering radius decreases with the performance gap. For example, $\alpha=2$ if $f$ is locally strongly concave near each peak. Our contributions are as follows.

\begin{enumerate}
\item \textbf{Formulation.} We formulate near-optimal region identification on a continuous design space as a set-valued output whose guarantees hold simultaneously for all tolerances up to $T$, relative to an unknown optimal value. This brings the set-valued outputs of ranking and selection to continuous design spaces and allows the tolerance to be chosen after the simulation study (Section~\ref{sec:relaxed-model}).

\item \textbf{Evaluation complexity of region identification (Q2).} When $\alpha>\nu$ and $\eta$ is small compared with $T$, any procedure, even one given the optimal point and its value in advance, needs a number of evaluations of order $\eta^{-(2+d/\nu)}\min\{1,KT^{d/\alpha}\}$ (Theorem~\ref{thm:region-minimax-lower}), and SASORI achieves this order up to logarithmic factors (Corollary~\ref{cor:region-minimax-matching}). The exponent of $\eta^{-1}$ is the one required to approximate the entire response surface uniformly to error $\eta$, and the geometry enters only through the factor $\min\{1,KT^{d/\alpha}\}$, which is small when there are few near-optimal peaks and the near-optimal designs occupy a small part of the design space, and which measures the saving from concentrating evaluations on the near-optimal part of the design space. Solving the optimization problem exactly therefore leaves region identification essentially untouched, and thus region identification is not a by-product of optimization and must be addressed in its own right. The reason is that the two tasks need information in different places. An efficient optimization procedure stops simulating a design as soon as it can tell that the design is not optimal, following the logic of screening \citep{nelson2001simple,kim2001fully}, whereas region identification must determine membership for designs whose performance gap is close to $\tau$, even when the optimum is already known. For a single tolerance fixed before sampling, we give a separate lower bound and a boundary-sensitive specialization of SASORI that can require substantially fewer evaluations under regular boundary geometry (Theorems~\ref{thm:fixed-tolerance-lower} and~\ref{thm:fixed-tolerance-boundary}).

\item \textbf{One procedure for both outputs (Q1).} We measure the quality of the estimated optimal point by its simple regret and the efficiency of the allocation by the cumulative regret, which is the total performance gap of all simulated points. With $T>0$, as long as at least half of the evaluations serve optimization, the estimated optimal point keeps the minimax order of simple regret. Cumulative regret, in contrast, must increase as the target region estimation error decreases. Any procedure with a valid region guarantee incurs cumulative regret at least of order $T$ times the number of evaluations shown in our contributions Point~2, because designs near each tolerance boundary must be simulated even though they are known to be suboptimal (Theorem~\ref{thm:joint-regret-cost}). When $\alpha>\nu$, SASORI achieves the resulting optimal tradeoff between region estimation error and cumulative regret, up to logarithmic factors (Theorem~\ref{thm:budgeted-ccucb} and Corollary~\ref{cor:joint-frontier}). At one end of this tradeoff, it keeps the cumulative regret at its optimal rate while the region estimation error still decreases to zero, although not at the fastest rate; at the other end, it achieves the fastest region identification rate, at the cost of cumulative regret that grows linearly in the number of evaluations.

\item \textbf{Localized minimax rates for simulation optimization.} Of independent interest, with $T=0$, the simple and cumulative regret of SASORI achieve the minimax rates, up to logarithmic factors, in two senses (Theorems~\ref{thm:new-regret}, \ref{thm:covering-lower-alpha-large}, and~\ref{thm:restored-critical-lower}). Without any condition on the near-optimal points, these are the global Mat\'ern rates $N^{-\nu/(2\nu+d)}$ and $N^{(\nu+d)/(2\nu+d)}$, where $N$ is the number of evaluations \citep{scarlett2017lower}. Under the covering condition with $\alpha>\nu$, the simple regret is of order $\min\{(K/N)^{1/(2+d/\nu-d/\alpha)},N^{-\nu/(2\nu+d)}\}$, uniformly in $K$, and the cumulative regret is $N$ times this quantity, so the rates improve when $K$ is small and the global rates take over once $K$ is large. Matching rates also hold at $\alpha=\nu\ge2$, whereas for $2\le\alpha<\nu$ we establish lower bounds (Section~\ref{subsec:other-covering-regimes}).
\end{enumerate}

Our regret analysis relates to the kernel bandit literature. The global Mat\'ern lower bounds are due to \citet{scarlett2017lower}, and matching simple and cumulative regret upper bounds, up to logarithmic factors, are obtained by variance maximization and batched elimination, respectively \citep{vakili2021optimal,li2022batches}. 
% Our sampling analysis builds on these methods and on the radius-dependent information-gain bound of \citet{iwazaki2025improved}. 
Near-optimal geometry also governs regret in continuum bandits \citep{bubeck2011xarmed,wang2018localminimax}.
% , where the geometric restriction applies to a reference function and nearby alternatives need not preserve it. 
The closest RKHS analysis is that of \citet{shekhar2022instance}, who derive instance-dependent cumulative regret bounds for policies that are required to be consistent on the entire RKHS ball. Our covering-class regret analysis results instead concern a minimax problem in which the infimum ranges over all policies, without a consistency requirement on a larger ball, and all alternatives satisfy the same covering condition at all near-optimality levels. The results cover both simple and cumulative regret, keep $K$ explicit, and, for $\alpha>\nu$, match the rates through the transition to the global rates as $K$ grows.

The rest of this work is arranged as follows. Section~\ref{sec:relaxed-model} introduces the model, the region outputs, and the smoothness model. Section~\ref{sec:new-lower-region} establishes the lower bounds for region identification, Section~\ref{sec:new-algorithm} presents SASORI, and Section~\ref{sec:new-regret} collects its guarantees. Section~\ref{sec:numerical} presents the numerical study on Branin-2 and Hartmann-3, Section~\ref{sec:conclusions} concludes, and all proofs and additional numerical results are in the appendices.

\section{Preliminaries}\label{sec:relaxed-model}

\subsection{Simulation Model and Region Guarantees}

\begingroup\color{black}
We consider an expensive stochastic simulation with expected response $f:\X\to\R$, where $\X\subset\R^d$ is a compact design space. Our objective is to maximize $f$, and we write $f^\star=\max_{x\in\X}f(x)$. A design point $x$ may specify, for example, staffing levels, service rates, routing probabilities, or reorder points. In order to find a maximizer of $f$, we select design points sequentially under a total budget of $N$ evaluations. At time $t$, we choose $x_t$ using the preceding observations and observe $y_t=f(x_t)+\varepsilon_t$, where $\varepsilon_t$ is simulation noise with conditional mean zero. After the final evaluation, the procedure \textcolor{black}{returns the estimated optimal point} $\widehat x_N$. We assess the procedure through two criteria: the simple regret $r_N=f^\star-f(\widehat x_N)$ and the cumulative regret $R_N=\sum_{t=1}^N\{f^\star-f(x_t)\}$. Simple regret measures the performance lost by implementing the \textcolor{black}{estimated optimal point}, whereas cumulative regret measures the performance lost at the design points evaluated during the procedure.
% In an offline study it measures
% allocation efficiency; in a learn-while-operating setting it can also
% represent an operating cost.
\par\endgroup

\paragraph{Near-Optimal Regions and Their Confidence Bounds}\label{subsec:certificate-def}

\begingroup\color{black}
In addition to an \textcolor{black}{\textcolor{black}{estimated optimal point}} $\widehat x_N$, we seek to identify the near-optimal regions of the design space. For a tolerance $\tau\ge0$, define the acceptable region
\[
 \mathcal S_\tau=\mathcal S_\tau(f)=\{x\in\X:f^\star-f(x)<\tau\},
\]
which consists of the design points whose expected response is less than $\tau$ below the maximum $f^\star$. We seek an inner set $I_\tau$ consisting only of acceptable design points and an outer set $O_\tau$ containing all of them. \textcolor{black}{For a target region estimation error $\eta>0$}, the desired guarantee is
\begin{align}\label{eq_region_require}
 I_\tau\subseteq\mathcal S_\tau\subseteq O_\tau,\qquad
 O_\tau\setminus I_\tau
 \subseteq\{x\in\X:|f^\star-f(x)-\tau|\le\eta\}.
\end{align}
\textcolor{black}{Here $\tau$ is the acceptable performance loss, whereas $\eta$ bounds the region estimation error measured in response value. Under the desired guarantee, every unclassified design $x\in O_\tau\setminus I_\tau$ satisfies $|f^\star-f(x)-\tau|\le\eta$.} We require \eqref{eq_region_require} to hold simultaneously for all $0<\tau\le T$ on a single event of probability at least $1-\delta$, where $T$ is the largest performance loss under consideration and $\delta$ is a prespecified \textcolor{black}{upper bound on the probability that the simultaneous guarantee does not hold}. On that event, all design points in $I_\tau$ meet the performance tolerance, and all design points outside $O_\tau$ fail it. The observations leave membership undetermined only in $O_\tau\setminus I_\tau$. The second inclusion confines these points to a band of width $\eta$ on either side of the value threshold $f^\star-\tau$; it bounds response differences rather than spatial distances. For example, with $\tau=0.10$ and $\eta=0.01$, a point with performance gap below $0.09$ must belong to $I_\tau$, while one with performance gap above $0.11$ must lie outside $O_\tau$. Because a single event covers all tolerances, a user may choose $\tau\in(0,T]$ after seeing the output \textcolor{black}{while keeping probability at least $1-\delta$ that the set inclusions and \textcolor{black}{region estimation error} bound in~\eqref{eq_region_require} hold at the selected tolerance}. 
% Choosing $\tau$ after sampling does not reduce the achieved $\eta$ or allow a choice beyond $T$; the range $T$ is specified before sampling.
\par\endgroup

\begingroup\color{black}
At $\tau=0$, the strict inequality in the definition requires a negative performance gap, or equivalently $f(x)>f^\star$, which is impossible. Thus the region is known exactly, $\mathcal S_0=I_0=O_0=\varnothing$, and no observations are needed for this set-valued output. A maximizer belongs to $\mathcal S_\tau$ for any positive $\tau$. When SASORI uses $T=0$, it devotes all of its sampling to improving the \textcolor{black}{\textcolor{black}{estimated optimal point}}.
\par\endgroup

% For $\tau>0$, valid function envelopes $L\le f\le U$ and an optimal-value bracket
% $\lo\le f^\star\le\hi$ yield the sets
% $I_\tau=\{x:L(x)>\hi-\tau\}$ and
% $O_\tau=\{x:U(x)>\lo-\tau\}$. Their \textcolor{black}{region estimation error} is controlled by the
% optimal-value uncertainty and the local envelope width. The complete
% deterministic statement and proof are given in
% Proposition~\ref{thm:general-region-certificate} of the supplementary
% \textcolor{black}{Appendix}~\ref{ec:region-envelopes}.

\subsection{From Region Bounds to Decisions}\label{subsec:region-decisions}

The inner and outer sets support decisions involving a known secondary criterion and quantify how much of the design space remains \textcolor{black}{unclassified}. In this subsection, we provide an example illustrating both uses through a service-level and resource-cost decision.
% These uses follow from the region guarantee itself.

Return to the service-system example of Section~\ref{sec:intro}, where $x$ specifies service rates and routing probabilities, $f(x)$ is the expected fraction of customers meeting the service target, and $c_0(x)$ is a known resource cost. Before sampling, the manager specifies a largest acceptable service loss $T$. After observing the simultaneous region guarantees, the manager may choose any $\tau\in(0,T]$ based on the resulting service--cost tradeoff, and can change it after seeing the performance of decision, without simulating the system again. This data-dependent choice \textcolor{black}{keeps} the same confidence guarantee, because the region guarantee holds for any such tolerance on a single event.

For the chosen tolerance, the operating-cost decision is bracketed by
\[
 \inf_{x\in O_\tau}c_0(x)
 \;\le\;
 \inf_{x\in\mathcal S_\tau}c_0(x)
 \;\le\;
 \inf_{x\in I_\tau}c_0(x),
 \qquad \inf\varnothing=+\infty.
\]
A nearly least-cost inner-set configuration is guaranteed to meet the service tolerance, while the outer-set value bounds the further cost reduction that could still be available among \textcolor{black}{unclassified} design points. Infima are used because the strict regions need not contain a cost-minimizing configuration. % If the inner set is empty, the region guarantee supplies no feasible configuration at that tolerance.

% \paragraph{Region size and unresolved volume.}
% For measurable output envelopes, let $\mu$ be a finite Borel measure
% and suppose the following margin condition holds:
% \begin{equation}\label{eq:region-margin-condition}
%  \mu\{x:|f^\star-f(x)-\tau|\le v\}\le C_\tau v^{\kappa_\tau},
%  \qquad 0<v\le v_\tau,
% \end{equation}
% where $C_\tau,\kappa_\tau,v_\tau>0$. Any region guarantee with \textcolor{black}{region estimation error}
% $\eta\le v_\tau$ then gives
% \[
%  \mu(O_\tau\setminus I_\tau)\le C_\tau\eta^{\kappa_\tau},
%  \qquad
%  \mu(I_\tau)\le\mu(\mathcal S_\tau)\le\mu(O_\tau).
% \]
% The margin condition is additional: RKHS membership and the
% near-optimal covering assumption do not themselves provide it at
% an arbitrary positive performance threshold.

% No Hausdorff, boundary topology, component-count, or exact finite-time
% membership conclusion follows without further boundary assumptions.

\subsection{Kernel Ridge Regression and Smoothness}\label{subsec:kernel-smoothness}

To recover the response surface from noisy observations, we use the Gaussian process (GP) regression predictor, whose posterior mean coincides with the kernel ridge regression estimator for the same kernel and regularization parameter.  Throughout, we treat $f$ as a fixed deterministic function in a reproducing kernel Hilbert space (RKHS), rather than as a sample path of a GP prior.

For a positive definite kernel $k$, its RKHS $\Hk(\X)$ is a Hilbert space of functions on $\X$ such that $k(x,\cdot)\in\Hk(\X)$ and, for any $g\in\Hk(\X)$, $g(x)=\langle g,k(x,\cdot)\rangle_{\Hk},\qquad x\in\X$, where $\langle \cdot,\cdot\rangle_{\Hk}$ denotes the inner product of RKHS. In addition, $\|\cdot\|_{\Hk}$ denotes the RKHS norm. For detailed introduction, we refer to \cite{wendland2004scattered}. 

We use a normalized Mat\'ern kernel with fixed smoothness $\nu>0$ and length scale $\ell>0$. For $r=\|x-z\|>0$, its form is
\begin{equation}\label{eq:matern-kernel}
 k_{\nu,\ell}(x,z)=\frac{2^{1-\nu}}{\Gamma(\nu)}
 \left(\frac{\sqrt{2\nu}\,r}{\ell}\right)^\nu
 K_\nu\!\left(\frac{\sqrt{2\nu}\,r}{\ell}\right),
 \qquad k_{\nu,\ell}(x,x)=1,
\end{equation}
where $\Gamma$ is the gamma function and $K_\nu$ is the modified Bessel function of the second kind. The parameter $\nu$ controls smoothness, and $\ell$ determines the spatial correlation scale \citep{rasmussen2006gaussian}. We assume $\|f\|_{\Hk(\X)}\le B$ for a fixed and known $B>0$. The Mat\'ern RKHS is norm-equivalent to the Sobolev space $H^s(\X)$ with $s=\nu+d/2$ on the domain $\X$ \citep{wendland2004scattered}.
Define the kernel metric
\begin{equation}\label{eq:new-metric}
 d_k(x,z)=\|k(x,\cdot)-k(z,\cdot)\|_{\Hk}=\sqrt{k(x,x)+k(z,z)-2k(x,z)}.
\end{equation}
The reproducing property and the Cauchy--Schwarz inequality give
\begin{equation}\label{eq:new-modulus}
 |f(x)|\le B, |f(x)-f(z)|\le B d_k(x,z).
\end{equation}
Since the Mat\'ern spectral density $\widehat k$ satisfies
\citep{rasmussen2006gaussian}
\[
 \widehat k(\omega)\asymp (2\nu/\ell^2+\|\omega\|^2)^{-(\nu+d/2)},
\]
we have 
\[
\begin{aligned}
 d_k(x,z)^2
 &=\int_{\R^d}|e^{i\omega^\top(x-z)}-1|^2\widehat k(\omega)\,d\omega\le C_\theta\|x-z\|^{2\theta}
       \int_{\R^d}\|\omega\|^{2\theta}\widehat k(\omega)\,d\omega
 =L_\theta^2\|x-z\|^{2\theta},
\end{aligned}
\]
\textcolor{black}{also because} $1-\cos t\le C_\theta|t|^{2\theta}$ and the integral is finite for $\theta<\nu$. Hence, for any fixed $0<\theta<\min\{\nu,1\}$, there is a constant $L_\theta$ such that $d_k(x,z)\le L_\theta\|x-z\|^\theta$. \textcolor{black}{This Hölder continuity bound is used to choose the confidence-bound parameters of the algorithm} in Section~\ref{sec:new-algorithm}.

\section{The Irreducible Cost of Region Learning}\label{sec:new-lower-region}

Before introducing the algorithm, we first ask how much simulation effort is unavoidable when the desired output is a family of acceptable regions. We show that, even when an optimal design point and the optimal value are known, learning near-optimal region can still be expensive, which shows the essential difference between region learning and regret minimization. 

Suppose an optimal design point $x^\star$ maximizes $f$, with value $f^\star=f(x^\star)$. Knowing $(x^\star,f^\star)$ fixes the threshold $f^\star-\tau$, but leaves the responses at other design points unknown. For example, two surfaces can share this optimizer and value while assigning a particular design point performance gaps $\tau-4\eta$ and $\tau+4\eta$, respectively. A procedure must distinguish these two cases to construct valid inner and outer sets. We show that this task can remain expensive even when $(x^\star,f^\star)$ is known before sampling. Specifically, in the worst case over the fixed smoothness and covering class with $\alpha>\nu$, any procedure needs at least a constant multiple of $\eta^{-(2+d/\nu)}\min\{1,KT^{d/\alpha}\}$ evaluations at fixed
confidence. Thus, knowing the best design point and its value does not remove the simulation effort needed to identify acceptable alternatives.

The simultaneous guarantee concerns all tolerances in $[0,T]$ together. If instead, one only consider a single positive $\tau$ fixed before sampling, the procedure only needs to \textcolor{black}{determine} membership relative to the one threshold $f^\star-\tau$. Once a design point is classified to be acceptable or unacceptable, its response need not be estimated more precisely for further decisions. If $f^\star$ is unknown, however, additional evaluations at a potentially optimal design point may still be needed to locate the threshold accurately. Thus learning one prespecified region and learning all regions over a range of tolerances impose different requirements. We define their evaluation budgets separately and give a lower bound for each.

% \paragraph{The covering class.}
The lower bounds are stated on a function class defined by a geometric covering condition, imposed on the same family of regions that forms the output. Let
\[
 \mathcal S_{u+}(f):=\bigcap_{v>u}\mathcal S_v(f)
       =\{x\in\X:f^\star-f(x)\le u\},\qquad u\ge0.
\]
Thus, the subscript $u+$ adds the points with performance gap exactly $u$. In particular, $\mathcal S_0(f)$ is empty, whereas $\mathcal S_{0+}(f)$ is the set of maximizers. Let $\mathcal N(A,r)$ be the smallest number of closed Euclidean radius-$r$ balls covering $A$. For fixed $B,c_\rho,u_0,\alpha>0$ and integer $K\ge1$, define $\mathcal F_K(B,c_\rho,u_0,\alpha)$ to consist of functions satisfying
\[
 \|f\|_{\Hk(\X)}\le B,\qquad \mathcal N(\mathcal S_{u+}(f),c_\rho u^{1/\alpha})\le K,  \qquad 0<u\le u_0.
\]
% The condition is imposed at all these value scales, with the same constants.
The same $B,c_\rho,u_0,\alpha$ apply to all functions in the class and remain fixed as $K$ varies. Here, $K$ bounds the number of balls of radius $c_\rho u^{1/\alpha}$ needed to cover the $u$-near-optimal set at any level $u\le u_0$.
% The class is independent of the tolerance
% range and is assumed nonempty.

For $T>0$, $\eta>0$, and $0<\delta<1/2$, we call a procedure $(\eta,T,\delta)$-valid on $\mathcal F_K$ if, for any $f\in\mathcal F_K$, with probability at least $1-\delta$, it returns sets $(I_\tau,O_\tau)_{0\le\tau\le T}$ satisfying, simultaneously,
\begin{equation}\label{eq:region-minimax-loss}
 I_\tau\subseteq\mathcal S_\tau\subseteq O_\tau,
 \qquad
 O_\tau\setminus I_\tau
 \subseteq\{x:|f^\star-f(x)-\tau|\le\eta\},
 \qquad 0\le\tau\le T.
\end{equation}
Let $n_{\rm all}(\eta,T,\delta;\mathcal F_K)$ be the smallest
deterministic evaluation budget for which such a procedure exists. The requirement in~\eqref{eq:region-minimax-loss} combines correct set inclusions with an accuracy bound on \textcolor{black}{undetermined} membership. The following lemma shows how much information this requirement provides when it holds simultaneously over the tolerance range.

\begin{lemma}\label{lem:region-clipped-gap}
Let $T>0$ and $\eta>0$, and let $(I_\tau,O_\tau)_{0\le\tau\le T}$
be random output sets with measurable membership for all fixed design points and tolerances. Suppose that, on a measurable event $\mathcal E_f$, \eqref{eq:region-minimax-loss} holds for all $\tau\in[0,T]$ and all $x\in\X$. Define
\[
 \widehat d(x)=\inf\bigl(
    \{q\in\mathbb Q\cap[0,T]:x\in I_q\}\cup\{T\}\bigr).
\]
Then $\widehat d(x)$ is a measurable random variable for each fixed $x$. On $\mathcal E_f$,
\begin{equation}\label{eq:region-clipped-gap}
 \min\{f^\star-f(x),T\}\le\widehat d(x)
       \le\min\{f^\star-f(x)+\eta,T\}, x\in\X.
\end{equation}
In particular, for an $(\eta,T,\delta)$-valid procedure, these inequalities hold simultaneously over $\X$ with probability at least $1-\delta$.
\end{lemma}

\textcolor{black}{Lemma~\ref{lem:region-clipped-gap} shows that a} simultaneous region \textcolor{black}{guarantee determines the performance gap truncated at $T$, namely the function $x\mapsto\min\{f^\star-f(x),T\}$, uniformly} to accuracy $\eta$. This is more information than either \textcolor{black}{an estimated optimal point} or membership in a single acceptable region provides, since the output distinguishes performance levels throughout the part of the design space \textcolor{black}{within $T$ of the optimum}. This is particularly useful in practice when the acceptable performance loss is chosen after the simulation results are available, for example when a manager compares staffing costs under several service-level tolerances before selecting a final design.

The following theorem quantifies the evaluations required for simultaneous region accuracy on the fixed covering class. Its last \textcolor{black}{claim} shows that the bound persists even when the optimizing design point and the optimal value are the same for all functions in the construction and are known in advance.

\begin{theorem}[Simultaneous-region minimax lower bound]
\label{thm:region-minimax-lower}
Fix a compact $\X\subset\mathbb R^d$ containing an interior Euclidean ball, a normalized Mat\'ern kernel with $\nu>0$, $\alpha>\nu$, and positive $B,c_\rho,u_0,\sigma$. Consider the fixed covering class $\mathcal F_K(B,c_\rho,u_0,\alpha)$ and independent Gaussian observation noise with variance $\sigma^2$. There are constants $\textcolor{black}{K_{\mathrm{reg}}}\in\mathbb N$ and $\textcolor{black}{T_{\mathrm{reg}}},c>0$, depending only on these fixed parameters, such that, for any integer $K\ge \textcolor{black}{K_{\mathrm{reg}}}$, $0<T\le \textcolor{black}{T_{\mathrm{reg}}}$, $0<\eta\le T/128$, and $0<\delta\le1/10$,
\begin{equation}\label{eq:region-minimax-lower}
 n_{\rm all}(\eta,T,\delta;\mathcal F_K)
 \ge c\sigma^2\operatorname{kl}(\delta,1-\delta)
      \eta^{-2-d/\nu}\min\{1,K T^{d/\alpha}\},
\end{equation}
where
$\operatorname{kl}(p,q)=p\log(p/q)+(1-p)\log((1-p)/(1-q))$.
% All constants are uniform in $K,T,\eta,\delta$. 
The bound also holds on a subclass in which a fixed design point $x^\star$ is the unique global maximizer and $f(x^\star)=f^\star=0$ for all functions, even when the procedure is given $(x^\star,0)$ before sampling.
\end{theorem}
% \paragraph{Proof idea.}
% Fix a baseline response surface with a single peak, and consider the shell
% of design points whose performance gap lies between $T/8$ and $3T/4$. Place disjoint
% bumps of height $8\eta$ and radius proportional to $\eta^{1/\nu}$ within
% that shell, subject to the fixed norm budget. Each bump stays below the
% optimal value, so all alternatives share the same optimizer and optimum.
% Nevertheless, it moves its center from $4\eta$ outside to $4\eta$ inside
% the acceptable set at a suitable tolerance $\tau_i$. An accurate inner/outer
% pair must distinguish these two cases, which requires order $\eta^{-2}$
% evaluations in that bump. The supports are disjoint, so these costs add.
% The number of bumps is proportional to the shell volume divided by
% $\eta^{d/\nu}$, and that volume is of order
% $\min\{1,KT^{d/\alpha}\}$. Multiplying gives the lower rate.
% \textcolor{black}{Appendix}~\ref{ec:region-lower} verifies the norm bound,
% \textcolor{black}{covering condition at every near-optimality level}, and adaptive testing argument.\hfill$\square$

Theorem~\ref{thm:region-minimax-lower} shows that the lower bound separates the dependence on the target error $\eta$ from the effect of near-optimal geometry, and remains valid even when the optimizer and its value are known. First, the factor $\eta^{-(2+d/\nu)}$ depends only on the accuracy and has the same exponent of $\eta^{-1}$ as estimating $f$ to within $\eta$ at under sup-norm, which requires of order $\eta^{-(2+d/\nu)}$ evaluations up to logarithmic factors. The geometric factor $\min\{1,KT^{d/\alpha}\}$ describes the reduction associated with the size of the requested region. Second, the covering condition does not reduce this exponent, in contrast to the \textcolor{black}{simple regret} rates of Section~\ref{sec:covering-minimax}, because the \textcolor{black}{band where classification remains undetermined} is a shell inside the near-optimal set rather than a shrinking neighborhood of the optimizer. Third, the bound is unaffected by knowing the optimizer. In the lower-bound construction, the baseline and all perturbed functions attain their \textcolor{black}{unique maximum} at the same point ${\color{black}x^\star}$, with value zero, so the procedure may be \textcolor{black}{given} the coordinates of ${\color{black}x^\star}$ and the value zero before collecting any observations. However, these quantities contain no information about which perturbation is present at the other design points. The lower bound therefore measures the cost of learning the regions after the optimization problem has been solved.

\paragraph{A lower bound for one prespecified tolerance.}
Let $n_{\rm one}(\eta,\tau,\delta;\mathcal F_K)$ be the smallest deterministic evaluation budget for a procedure that, for any $f\in\mathcal F_K$, satisfies~\eqref{eq_region_require} at a single prespecified $\tau>0$, simultaneously over all design points with probability at least $1-\delta$. Since one pair of sets can be extracted from a simultaneous output, $n_{\rm one}(\eta,\tau,\delta;\mathcal F_K) \le n_{\rm all}(\eta,\tau,\delta;\mathcal F_K)$.

\begin{theorem}[A fixed-tolerance lower bound]
\label{thm:fixed-tolerance-lower}
Fix the domain, normalized Mat\'ern kernel, and class parameters of
Theorem~\ref{thm:region-minimax-lower}, including $\alpha>\nu$, and use
independent Gaussian observation noise with variance $\sigma^2$.
There are constants $K_1\in\mathbb N$ and
$\tau_0,c>0$, depending only on those fixed parameters, such that,
for any integer $K\ge K_1$, $0<\tau\le\tau_0$,
$0<\eta\le\tau/32$, and $0<\delta\le1/10$,
\begin{equation}\label{eq:fixed-tolerance-lower}
 n_{\rm one}(\eta,\tau,\delta;\mathcal F_K)
 \ge c\sigma^2\operatorname{kl}(\delta,1-\delta)
       \eta^{-2-(d-1)/\nu}
       \bigl(\min\{1,K\tau^{d/\alpha}\}\bigr)^{(d-1)/d}.
\end{equation}
The bound holds on a subclass with a common unique global maximizer $x^\star$ and value zero, even when $(x^\star,0)$ is \textcolor{black}{provided} before sampling.
\end{theorem}

% \paragraph{Proof idea.}
% Use the preceding construction with $T=4\tau$, but place the perturbation
% centers only on the sphere where the baseline performance gap equals
% $\tau+4\eta$. All perturbed centers then have performance gap $\tau-4\eta$,
% so one fixed tolerance distinguishes all alternatives. There are order
% $(R/\eta^{1/\nu})^{d-1}$ disjoint supports along this sphere, where
% $R^d$ is comparable to $\min\{1,K\tau^{d/\alpha}\}$; in one dimension
% use its two endpoints. Each support requires order $\eta^{-2}$ noisy
% evaluations. Appendix~\ref{ec:fixed-tolerance-lower} gives the
% packing, class-membership, and adaptive-testing details.\hfill$\square$

For fixed admissible $K,\tau,\delta$, \textcolor{black}{the exponent of $\eta^{-1}$ in this lower bound} is $2+(d-1)/\nu$, which is smaller by $1/\nu$ than the simultaneous exponent $2+d/\nu$. This reduction \textcolor{black}{occurs because the perturbations are placed along a single level set of the baseline response surface, which is $(d-1)$-dimensional,} instead of throughout a \textcolor{black}{$d$-dimensional set of designs}.

\begingroup\color{black}
\paragraph{Region learning versus regret minimization.}
Simple regret measures the performance gap of the \textcolor{black}{estimated optimal point}, whereas cumulative regret sums these gaps over all evaluated design points \citep{scarlett2017lower}. In batched kernel optimization, for example, a design point can be eliminated once its upper confidence bound falls below the largest lower confidence bound \citep{li2022batches}. However, such a point may still require refinement to attain the requested region accuracy. Theorems~\ref{thm:region-minimax-lower} and \ref{thm:fixed-tolerance-lower} together separate region learning from both \textcolor{black}{simple regret and cumulative regret} minimization. \textcolor{black}{On the subclass used to establish each lower bound}, a procedure given the common optimizer can \textcolor{black}{select} it and allocate all evaluations to it, attaining $r_N=R_N=0$. Yet such observations do not distinguish the response surfaces away from that optimizer, and valid region learning still requires the evaluations prescribed by the theorems. This remains true for a single tolerance fixed in advance, where the information that remains to be determined lies near the corresponding performance boundary. For simultaneous outputs, that information extends throughout a shell as the tolerance varies. Hence neither \textcolor{black}{a design with a small optimality gap} nor small cumulative regret guarantees an accurate acceptable region. The cost of obtaining both outputs jointly is quantified in Section~\ref{subsec:joint-tradeoff}.
\par\endgroup

\section{Shared-confidence Adaptive Sampling for Optimization and Region Identification (SASORI)}\label{sec:new-algorithm}

\begingroup\color{black}
Section~\ref{sec:new-lower-region} shows that optimizing a response surface does not by itself identify its near-optimal regions. To pursue both objectives under a \textcolor{black}{single evaluation budget}, we introduce Shared-confidence Adaptive Sampling for Optimization and Region Identification (SASORI). The policy allocates effort explicitly to each objective: a region quota $m\in\{0,\ldots,N\}$ reserves $m$ evaluations for region refinement, and the remaining $N-m$ evaluations serve optimization. Both tracks share their completed confidence bands, so each can improve the other's output. The quota therefore controls the tradeoff between region accuracy and
cumulative regret. For a specified maximum tolerance $T\ge0$, the policy returns an \textcolor{black}{\textcolor{black}{estimated optimal point}} $\widehat x_N$ and inner and outer sets for all tolerances $0\le\tau\le T$.
\par\endgroup

\begingroup\color{black}
SASORI collects evaluations in batches, each assigned to optimization or region learning. Optimization batches seek better designs, while region-learning batches determine which designs lie within the requested performance tolerances of the optimum. Each completed nonempty batch tightens confidence bounds used by both tracks. These tighter bounds can improve the \textcolor{black}{estimated optimal point} and the estimated regions, regardless of which track collected it. Algorithm~\ref{alg:continuous-ccucb} schedules the phases and calls the sample-allocation and band-sharing subroutines in Algorithms~\ref{alg:sample-allocation} and~\ref{alg:band-sharing}, respectively. Both subroutines use its fixed inputs and calibrated $(\beta_N,\zeta_N)$. State variables updated by either subroutine are maintained by the main algorithm.
\par\endgroup

% BEGIN ccucb_algorithm
% All three algorithms use the same fixed inputs and persistent state.

% Compact algorithm layout
\newcommand{\sasoriline}[3]{%
 \noindent\makebox[1.8em][r]{#1:}\hspace{.5em}%
 \parbox[t]{\dimexpr\linewidth-2.3em\relax}{%
  \normalsize\setstretch{1}%
  \hangindent=#2em\hangafter=0 #3\strut}\par}

\newenvironment{sasorialgorithm}[2]{%
 \par\medskip\noindent\begin{minipage}{\linewidth}%
 \begingroup\color{black}\normalsize\setstretch{1}%
 \refstepcounter{algorithm}\label{#2}%
 \noindent\hrulefill\par
 \noindent\textbf{Algorithm~\thealgorithm\quad #1}\par\medskip
}{%
 \noindent\hrulefill\par
 \endgroup\end{minipage}\par\medskip}

\begin{sasorialgorithm}{SASORI}{alg:continuous-ccucb}
\noindent\textbf{Require:} $N\ge2$, $\vartheta>1$, $T\ge0$,
$m\in\{0,\ldots,N\}$ ($m=0$ if $T=0$),
$\X,k,B,\sigma,\lambda,\delta,L_\theta,D_{\X}$, $x_0\in\X$,
and a fixed rule for breaking ties.\par
\noindent\textbf{Ensure:} The latest saved estimated optimal point and region bounds for all $0\le\tau\le T$.\par\medskip
\sasoriline{1}{0}{Compute $(\beta_N,\zeta_N)$ by~\eqref{eq:new-confidence-parameters}, with $\theta=\min\{\nu,1\}/2$.}
\sasoriline{2}{0}{Set $(N_0,N_1)=(N-m,m)$, $(T_0^{\rm alg},T_1^{\rm alg})=(0,T)$, and $h_j=t_j=0$, $\bar\eps_j=+\infty$ for $j=0,1$.}
\sasoriline{3}{0}{Set $\ell=-B,u=B$, $\lo=-B,\hi=B$, $\widehat x=x_0$; save the initial output and finish tracks assigned no evaluations.}
\sasoriline{4}{0}{\textbf{while} some track is unfinished \textbf{do}}
\sasoriline{5}{1}{Select unfinished track $j$, starting with 0 and alternating whole phases when both remain; set $\eps=2B\vartheta^{-h_j}$.}
\sasoriline{6}{1}{\textbf{If} $\eps<2\zeta_N$, allocate the remaining $N_j-t_j$ evaluations to $\widehat x$; set $t_j=N_j$, finish track $j$, and \textbf{continue}.}
\sasoriline{7}{1}{Fix $(\lo,W,G)$ by~\eqref{eq:new-frozen-cap}, using $T_j^{\rm alg}$ and the current shared bands.}
\sasoriline{8}{1}{Call Algorithm~\ref{alg:sample-allocation} for this phase.}
\sasoriline{9}{1}{\textbf{If} the phase is incomplete, discard its batch, finish track $j$, and \textbf{continue}.}
\sasoriline{10}{1}{Call Algorithm~\ref{alg:band-sharing} with the completed batch.}
\sasoriline{11}{1}{\textbf{If} Algorithm~\ref{alg:band-sharing} returns ``inconsistent,'' allocate the remaining $N-t_0-t_1$ evaluations to $x_0$ and \textbf{return} $x_0$ and ``no region guarantee.''}
\sasoriline{12}{0}{\textbf{end while}}
\sasoriline{13}{0}{\textbf{return} the latest saved output.}
\end{sasorialgorithm}
% END ccucb_algorithm

\paragraph{\textcolor{black}{Sample allocation.}}
The inputs $B$, $\sigma$, $k$, $\lambda$, and $L_\theta$ (norm bound, noise level, kernel, regularization parameter, and \textcolor{black}{H\"older constant of the kernel from Section~\ref{subsec:kernel-smoothness}}) are specified in Section~\ref{subsec:theory-assumptions}, and are fixed. 
Each phase uses only its own observations for the GP fit:
\[
 \mu_n(x)=k_n(x)^\top(K_n+\lambda I)^{-1}y_{1:n},\qquad
 s_n^2(x)=k(x,x)-k_n(x)^\top(K_n+\lambda I)^{-1}k_n(x),
\]
where $k_n(x)=(k(x_1,x),\ldots,k(x_n,x))^\top$, $K_n=(k(x_i,x_j))_{i,j=1}^n$, and $y_{1:n}$ contains the responses from the current phase. The function $\mu_n$ is the kernel ridge regression predictor, and $s_n^2$ is the associated GP posterior variance. Within a phase, \textcolor{black}{the choice of sampling points} and the stopping test use the bands from completed phases and $s_n$, which depends only on the sampled locations. The current responses $y_{1:n}$ are withheld from these decisions. If the phase completes, those responses are then used to compute $\mu_n$ and update the shared bands.

For $N\ge2$ and $\delta\in(0,1)$, fix $\theta\in(0,\min\{\nu,1\})$ and use the deterministic \textcolor{black}{net-size} bound \(M_N=(1+2D_{\X}N^{4/\theta})^d\), \textcolor{black}{which bounds the number of points in a finite $N^{-4/\theta}$-net of $\X$,} where \(D_{\X}=\sup_{x,z\in\X}\|x-z\|\) is the diameter of $\X$. The confidence parameters are
\begin{equation}\label{eq:new-confidence-parameters}
\begin{split}
 \beta_N&=B+\frac{\sigma}{\sqrt\lambda}
       \sqrt{2\log(4M_NN^2/\delta)},\\
 Y_N&=B+\sigma\sqrt{2\log(4N/\delta)},\\
 \zeta_N&=L_\theta N^{-4}
       \left(B+\frac{\sqrt N Y_N}{2\sqrt\lambda}+\beta_N\right).
\end{split}
\end{equation}
For example, if $\delta=N^{-2}$, then $\beta_N=O(\sqrt{\log N})$ and $\zeta_N=\widetilde O(N^{-7/2})$.

\begingroup\color{black}
To describe one phase of a given track, use $T$ for that track's tolerance, $h$ for its local phase counter, and $\eps_h=2B\vartheta^{-h}$ for its accuracy level. The fixed parameter $\vartheta>1$ determines the reduction in accuracy level between successive phases. For the analysis, write $L_\vartheta=4(\vartheta+1)$; this equals $12$ when $\vartheta=2$. Let $\ell_{h-1},u_{h-1}$ be the shared confidence bounds available at the start of the phase, including the contributions of any intervening completed phases of the other track, with initial values $-B$ and $B$, respectively. \textcolor{black}{The following quantities are held fixed throughout the phase:}
\begin{equation}\label{eq:new-frozen-cap}
 \lo_{h-1}=\max_{x\in\X}\ell_{h-1}(x), W_{h-1}=u_{h-1}-\ell_{h-1}, G_h(x)=[u_{h-1}(x)-\lo_{h-1}+T+4\eps_h]_+.
\end{equation}
Then the next sampling point is obtained via maximizing the acquisition function
\begin{equation}\label{eq:new-full-domain-score}
 a_{h,n}(x)=\min\{G_h(x),W_{h-1}(x), 2\beta_Ns_n(x)+2\zeta_N\},
\end{equation}
i.e., $x_{h,n+1}\in\argmax_{x\in\X}{a_{h,n}}(x).$ The \textcolor{black}{screening function} $G_h$ screens out design points that already perform too poorly to matter at tolerance $T$, the maintained width $W_{h-1}$ preserves the information in completed batches, and the \textcolor{black}{confidence-band width from the current batch} $2\beta_Ns_n+2\zeta_N$ directs new evaluations toward locations where the current batch still leaves large uncertainty. 
% The offset $4\eps_h$ in \textcolor{black}{the screening function} provides a controlled transition around the value threshold needed for localization. 
The phase ends, possibly without any evaluation, when
\begin{equation}\label{eq:new-phase-stop}
 \max_{x\in\X}a_{h,n}(x)\le2\eps_h.
\end{equation}
% Neither the acquisition function nor this test uses the current responses.
% Lemma~\ref{lem:new-active} in \textcolor{black}{Appendix}~\ref{ec:completed-bands}
% shows that every query in phase $h$ has performance gap at most
% $\min\{2B,T+L_\vartheta\eps_h\}$.
\par\endgroup

\begingroup\color{black}
Algorithm~\ref{alg:sample-allocation} implements the sampling loop for one phase. It checks the stopping condition before the remaining track budget, so a phase may complete at the last permitted evaluation or without any new evaluation. Its decisions use the fixed phase quantities and the variance determined by the current batch locations.
\par\endgroup
\begin{sasorialgorithm}{Sample allocation}{alg:sample-allocation}
\noindent\textbf{Require:} Track $j$, accuracy level $\eps=\eps_{h_j}$, fixed $(G,W)$, evaluation count $t_j$, quota $N_j$, and the common inputs of Algorithm~\ref{alg:continuous-ccucb}.\par
\noindent\textbf{Ensure:} Updated $t_j$, batch $(n,\mathcal D,s_n)$, and status ``completed'' or ``incomplete.''\par\medskip
\sasoriline{1}{0}{Set $n=0$, $\mathcal D=\varnothing$, and $s_0=1$.}
\sasoriline{2}{0}{\textbf{loop}}
\sasoriline{3}{1}{Choose $x\in\argmax_{z\in\X}a_{h_j,n}(z)$ using~\eqref{eq:new-full-domain-score}; set $M=a_{h_j,n}(x)$.}
\sasoriline{4}{1}{\textbf{If} $M\le2\eps$, \textbf{return} the batch with status ``completed.''}
\sasoriline{5}{1}{\textbf{If} $t_j=N_j$, \textbf{return} the batch with status ``incomplete.''}
\sasoriline{6}{1}{Observe $y=f(x)+\varepsilon_t$, append $(x,y)$ to $\mathcal D$, and increment $n$ and $t_j$.}
\sasoriline{7}{1}{Update $s_n$ from the batch locations alone, without using the batch responses.}
\sasoriline{8}{0}{\textbf{end loop}}
\end{sasorialgorithm}

\begin{remark}[Selecting the next design point]
\label{rem:threshold-query}
Exact maximization of $a_{h_j,n}$ in Algorithm~\ref{alg:sample-allocation} is sufficient but not necessary for selecting a point to evaluate. It is enough to find any $x\in\X$ such that $a_{h,n}(x)>2\eps_h$. For $\eps_h\ge2\zeta_N$, this implies $G_h(x)>2\eps_h$, $W_{h-1}(x)>2\eps_h$, and
$s_n(x)>(\eps_h-\zeta_N)/\beta_N\ge\eps_h/(2\beta_N)$. These are the properties used in the proof of
Lemma~\ref{lem:cap-driven-phase-cost} and the localization argument of Proposition~\ref{prop:phase-accuracy}. The same bounds therefore apply if point selection does not use responses from the current phase and the stopping
condition~\eqref{eq:new-phase-stop} is retained.
\end{remark}

\paragraph{Band sharing and output updates.}
For a completed nonempty phase, let $w_n=\beta_Ns_n+\zeta_N$ and form the band $[\mu_n-w_n,\mu_n+w_n]$. Define the shared confidence event
\begin{equation}\label{eq:shared-confidence-event}
\begin{split}
 \Econf=\bigl\{&|f(x)-\mu_n(x)|\le\beta_Ns_n(x)+\zeta_N,
 \text{ for all phases,}\\
 &\text{\textcolor{black}{all sample sizes obtained within a phase by time} }N,\text{ and all }x\in\X\bigr\}.
\end{split}
\end{equation}
On $\Econf$, bands from both tracks can be intersected without excluding the true response surface.
% When the bands are calibrated at \textcolor{black}{error probability} $\bar\delta$, we write $\Econf[\bar\delta]$ for the same event with the corresponding confidence parameters.

When a nonempty phase completes, its band is intersected with the \textcolor{black}{maintained confidence bounds} on all of $\X$, while an empty phase leaves them unchanged. With the resulting \textcolor{black}{confidence bounds} $\ell_h,u_h$, set
\begin{equation}\label{eq:new-elimination}
 \lo_h=\max_{x\in\X}\ell_h(x),\qquad \hi_h=\max_{x\in\X}u_h(x),\qquad
 \widehat x_h\in\argmax_{x\in\X}\ell_h(x),
\end{equation}
and the output regions in this phase are
% For the joint region output, $T$ again denotes the maximum requested
% tolerance, rather than the selected track's tolerance, even
% when an optimization phase supplies the latest band:
\begin{equation}\label{eq:ccucb-joint-output}
 I_{h,\tau}=\{x:\ell_h(x)>\hi_h-\tau\},\qquad
 O_{h,\tau}=\{x:u_h(x)>\lo_h-\tau\},\qquad 0<\tau\le T.
\end{equation}
Set $I_{h,0}=O_{h,0}=\varnothing$. Note that these \textcolor{black}{confidence bounds apply to} all tolerances simultaneously, so $\tau\le T$ can be chosen after observing the output.

\begingroup\color{black}
Algorithm~\ref{alg:band-sharing} performs these updates only after a completed phase. A completed empty phase leaves the confidence bounds unchanged but still records the achieved accuracy level and advances the phase counter. If the updated lower and upper bounds are inconsistent, it returns that status before updating extrema or saved outputs; Algorithm~\ref{alg:continuous-ccucb} then applies the stated rule for the remaining total evaluation budget.
\par\endgroup
\begin{sasorialgorithm}{Band sharing}{alg:band-sharing}
\noindent\textbf{Require:} Track $j$, accuracy level $\eps$, completed batch $(n,\mathcal D,s_n)$, and the shared state and fixed inputs of Algorithm~\ref{alg:continuous-ccucb}.\par
\noindent\textbf{Ensure:} Updated shared state and saved output, or status ``inconsistent.''\par\medskip
\sasoriline{1}{0}{\textbf{if} $n>0$ \textbf{then}}
\sasoriline{2}{1}{Fit $\mu_n$ using $\mathcal D$ and set $w_n=\beta_Ns_n+\zeta_N$.}
\sasoriline{3}{1}{On $\X$, set $\ell=\max\{\ell,\mu_n-w_n\}$ and $u=\min\{u,\mu_n+w_n\}$.}
\sasoriline{4}{0}{\textbf{end if}}
\sasoriline{5}{0}{\textbf{If} $\ell(x)>u(x)$ for some $x\in\X$, \textbf{return} status ``inconsistent.''}
\sasoriline{6}{0}{Set $\bar\eps_j=\eps$ and $h_j\gets h_j+1$.}
\sasoriline{7}{0}{Update $(\lo,\hi,\widehat x)$ by~\eqref{eq:new-elimination}, using the fixed rule for breaking ties.}
\sasoriline{8}{0}{Save $(\widehat x,\lo,\hi,\ell,u,\bar\eps_0,\bar\eps_1)$ and the regions in~\eqref{eq:ccucb-joint-output} for the full tolerance range $0<\tau\le T$; set $I_0=O_0=\varnothing$.}
\sasoriline{9}{0}{\textbf{return} status ``consistent.''}
\end{sasorialgorithm} 
% The \textcolor{black}{\textcolor{black}{estimated optimal point}} need not be a queried point. Compactness and continuity ensure that the extrema exist.

\begingroup\color{black}
On $\Econf$, a completed phase of track $j$ with accuracy level $\bar\eps_j$ yields $\hi-\lo\le2\bar\eps_j$ and an \textcolor{black}{\textcolor{black}{estimated optimal point}} with performance gap at most $2\bar\eps_j$. A completed region phase with local index $h$ also gives the simultaneous region guarantee \eqref{eq_region_require} with accuracy $\eta=4\bar\eps_1=4\eps_h=8B\,\vartheta^{-h}$. These guarantees remain valid after later intersections, since the upper confidence bound can only decrease, the lower confidence bound can only increase, and the set where a given track needs a small width can only shrink. The following proposition makes this precise.
\par\endgroup

\begin{proposition}
\label{lem:shared-invariant}
Suppose track $j$ completed a phase \textcolor{black}{with accuracy level $\bar\eps$, leaving confidence bounds} $(\ell^{\rm old},u^{\rm old})$ with $\lo^{\rm old}=\max_\X\ell^{\rm old}$, and that on {$\Econf$}
\begin{equation}\label{eq:shared-invariant-old}
 u^{\rm old}(x)-\ell^{\rm old}(x)\le2\bar\eps \text{ whenever }u^{\rm old}(x)>\lo^{\rm old}-T_j-2\bar\eps, f^\star-\lo^{\rm old}\le2\bar\eps,
\end{equation}
where $T_j=T_j^{\rm alg}$. Let $(\ell,u)$ be the \textcolor{black}{confidence bounds} at any later phase boundary, obtained by intersecting \textcolor{black}{confidence bands from completed batches}, and $\lo=\max_\X\ell$. Then the same two \textcolor{black}{claims} hold for $(\ell,u,\lo)$ with the same $\bar\eps$.
\end{proposition}

\begingroup\color{black}
Consequently, both tracks can use the same observations and confidence event $\Econf$, without a separate data set or additional union bound. This is the sharing property used in our theoretical developments in Section~\ref{sec:new-regret}.
% Theorems~\ref{thm:budgeted-ccucb} and~\ref{thm:region-learning-budget}.
An incomplete final phase returns the latest saved output, and any empty phases after quota exhaustion leave the confidence bounds and \textcolor{black}{\textcolor{black}{estimated optimal point determined by the fixed rule for breaking ties}} unchanged. \textcolor{black}{The band-sharing subroutine in Algorithm~\ref{alg:band-sharing} checks whether the maintained lower and upper confidence bounds are inconsistent after each intersection, that is, whether $\ell(x)>u(x)$ for some $x\in\X$. If so, Algorithm~\ref{alg:continuous-ccucb}} uses $x_0$ for the remaining evaluations and returns it without a region guarantee. This safeguard makes the policy well defined for any observation sequence. On $\Econf$, all maintained bands contain $f(x)$, so $\ell(x)\le f(x)\le u(x)$ and the safeguard is never triggered. Its triggering probability is therefore at most $\delta$. \textcolor{black}{The absence of an inconsistency does not establish that the bounds contain the true response surface.}
\par\endgroup

\begin{remark}
    Algorithm~\ref{alg:continuous-ccucb} concerns all tolerances $\tau\in [0,T]$ simultaneously. For a single tolerance fixed before sampling, Section~\ref{subsec:fixed-tolerance-screening} specifies an \textcolor{black}{alternative screening function} and single-tolerance output within the same fixed-horizon phase mechanism.
\end{remark}

In summary, SASORI combines a fixed budget split with information sharing through confidence \textcolor{black}{bounds}. The \textcolor{black}{acquisition function} concentrates evaluations on design points that remain uncertain and relevant to the selected track's tolerance, while completed nonempty batches can improve both outputs. Thus, SASORI supports \textcolor{black}{both design selection and region decisions} within a single sequential simulation study, with the quota setting the allocation of evaluations between the two \textcolor{black}{tracks}.

% The next section quantifies the costs of these phases and the resulting tradeoff between simultaneous region accuracy and cumulative regret.

% \textcolor{black}{Optional tightening of the confidence bounds using earlier observations} and approximate acquisition with global bounds are
% described in Section~\ref{subsec:cap-driven-variant}.
% All displayed extrema use \textcolor{black}{routines returning exact global extrema} on
% $\X$: the analysis counts simulator evaluations, not \textcolor{black}{the computation performed by the optimization routines}.

\section{Theoretical Guarantees for Joint Region Learning and Optimization}
\label{sec:new-regret}

This section presents the theoretical guarantees. We first state the assumptions and establish evaluation budgets for optimization and simultaneous region learning in Section~\ref{subsec:theory-assumptions} and Section~\ref{subsec:cap-driven-variant}, respectively. Section~\ref{subsec:optimization-benchmarks} then introduces the pure-optimization regret benchmark and its matching lower bounds. Using those results, Section~\ref{subsec:joint-tradeoff} quantifies the regret cost of region learning and the accuracy compatible with efficient optimization. Section~\ref{subsec:fixed-tolerance-screening} specializes the phase mechanism to a single tolerance fixed before sampling, and Section~\ref{subsec:other-covering-regimes} completes the analysis with the critical endpoint $\alpha=\nu$, \textcolor{black}{the case $2\le\alpha<\nu$}, and a sufficient range for \textcolor{black}{the Mat\'ern lower rates without the covering condition}. 

\begingroup\color{black}
\subsection{Assumptions for the Theoretical Results}\label{subsec:theory-assumptions}
\textcolor{black}{Our theoretical analysis imposes the following assumptions. Throughout the upper bounds for SASORI and their proofs, the phase ratio $\vartheta>1$ is fixed. Their multiplicative constants may depend on $\vartheta$, in addition to the model and covering parameters specified in each result; the rate orders are unchanged for each fixed $\vartheta$.} 
% The first condition specifies domain regularity; the next two control simulation error and response-surface smoothness. The final condition provides the optional geometric refinement.
\par\endgroup

\begin{assumption}\label{as:new-domain}
The design space $\X\subset\R^d$ is the closure of a bounded convex Lipschitz domain with nonempty interior and diameter $D_{\X}=\sup_{x,z\in\X}\|x-z\|$.
\end{assumption}

\begin{assumption}\label{as:new-noise}
For some known $\sigma>0$, the noise $\varepsilon_t$ is sub-Gaussian, i.e., $\mathbb E[\exp(u\varepsilon_t)\mid\mathcal F_{t-1},x_t] \le\exp(u^2\sigma^2/2)$ for all $u\in\R$, where $\mathcal F_{t-1}$ denotes the history of the first $t-1$ design points and responses.
\end{assumption}

% \begin{remark}[Unknown noise scale]
%  A plug-in estimate alone need not preserve confidence coverage; any data-driven calibration must account for uncertainty in that bound. Such calibration is outside the present analysis.
% \end{remark}

\begin{assumption}\label{as:new-rkhs}
The kernel $k=k_{\nu,\ell}$ is the normalized Mat\'ern kernel~\eqref{eq:matern-kernel} with fixed smoothness $\nu>0$ and fixed length scale $\ell>0$, and thus $k(x,x)=1$. The objective function satisfies $\|f\|_{\Hk(\X)}\le B$ for some known $B>0$. The regularization $\lambda>0$ is fixed. 
\end{assumption}

\begin{remark}[Fixed regularization]
For a nonempty batch ($n\ge1$), the predictor $\mu_n$ is \textcolor{black}{indeed} the kernel ridge-regression solution
\[
 \mu_n=\argmin_{g\in\Hk(\X)}\left\{\sum_{i=1}^n(y_i-g(x_i))^2+\lambda\|g\|_{\Hk(\X)}^2\right\}.
\]
In the matched Mat\'ern RKHS setting, \textcolor{black}{$\mu_n$ with} a fixed $\lambda$ \textcolor{black}{can lead to} the optimal rate of uniform ($L_\infty$) estimation under suitable space-filling designs, up to logarithmic factors.

% A fixed $\lambda$ also agrees with the GP posterior mean under fixed hyperparameters \citep{rasmussen2006gaussian}, while in this work, we . 
% This interpretation does not impose a GP prior on the deterministic $f$ in our analysis. Positivity of $\lambda$ ensures that $K_n+\lambda I$ is invertible even when sampling sites repeat, while keeping $\lambda$ independent of $N$ avoids additional horizon dependence in the confidence and information-gain constants.
\end{remark}

% The right-limit regions $\mathcal S_{u+}$ and Euclidean covering numbers
% $\mathcal N(A,r)$ were defined in Section~\ref{sec:new-lower-region}.

\begin{assumption}\label{as:new-cover}
For fixed $\alpha,c_\rho,u_0>0$ and integer $K\ge1$,
\begin{equation}\label{eq:new-cover}
 \mathcal N\bigl(\mathcal S_{u+},c_\rho u^{1/\alpha}\bigr)\le K,
 \qquad 0<u\le u_0.
\end{equation}
% The constants are uniform over the class and independent of $N$.
% Cover centers may depend on $u$ and need not be maxima.
\end{assumption}

Assumption~\ref{as:new-domain} fixes the design space and makes continuous band extrema well defined. Convexity is assumed for mathematical simplicity, but we note that the upper bounds only needs compactness. Assumption~\ref{as:new-noise} is a standard assumption that controls the tails of the simulation error conditional on previous observations. The same upper bound $\sigma$ can be used for unequal replication variances. When $\sigma$ is unknown, a conservative upper bound may be derived from the simulation model or calibrated using pilot replications under additional distributional assumptions. Assumption~\ref{as:new-rkhs} expresses response-surface smoothness through a fixed kernel and norm bound. 
% A bound on $B$ that is too small can invalidate coverage, while a conservative larger bound widens the bands. An inconsistency check detects conflicting bands
% but cannot validate the entire declared function class from finite data.
% Pilot-based calibration with its coverage uncertainty accounted for remains
% a practical question for Section~\ref{sec:numerical}.
Assumption~\ref{as:new-cover} bounds the number of spatial neighborhoods needed to cover all $u$-near-optimal design points at radius $c_\rho u^{1/\alpha}$, uniformly over $0<u\le u_0$. The exponent $\alpha$ determines how the radius shrinks with the value level $u$, while $K$ bounds the number of neighborhoods. A simple sufficient condition is the following.
% It permits overlapping balls, nonquadratic growth, and clustering. The parameters are uniform over the class; they enter localized bounds and class-informed quota choices. 
% \textcolor{black}{Without Assumption~\ref{as:new-cover}, the same algorithm still has the global Mat\'ern guarantees obtained by bounding information gain over all of $\X$, including the regional budget~\eqref{eq:region-global-cost} and the regret rates~\eqref{eq:new-global-rates}; only the additional improvement based on near-optimal-set geometry requires the covering condition.} 
For a nonempty set $A\subseteq\X$, define $\dist(x,A)=\inf_{a\in A}\|x-a\|$. For a fixed set $\mathcal M$ of at most $K$ maximum points, the error bound
\begin{equation}\label{eq:new-mode-error}
 f^\star-f(x)\ge c_g\dist(x,\mathcal M)^\alpha \quad\text{whenever }f^\star-f(x)\le u_0
\end{equation}
implies~\eqref{eq:new-cover} with $c_\rho=c_g^{-1/\alpha}$, where $c_g$ may depend on $K$. 
% For an individual function, $c_g$ may depend on $K$; comparisons over the fixed covering classes require a common positive lower bound on $c_g$ as $K$ varies.
% . \textcolor{black}{The growth constant $c_g$ may depend on $K$ in a particular family of response surfaces. Our covering-class comparisons keep $c_\rho,u_0,\alpha$ fixed as $K$ varies; certifying such a class through~\eqref{eq:new-mode-error} therefore requires $c_g\ge c_\rho^{-\alpha}$ uniformly. The covering condition itself is weaker and may hold even when this stronger sufficient condition does not.} 
Conversely, \eqref{eq:new-cover} can hold without any finite mode decomposition. Another sufficient condition is growth around finitely many separated local optima, as formalized in Proposition \ref{prop:separated-optima-cover}. 
% All assertions about a covering class presume that it is nonempty.

\begin{proposition}\label{prop:separated-optima-cover}
Let $\X\subset\R^d$ be compact and $f:\X\to\R$ continuous.
Fix $\alpha>0$ and $\tau\ge0$. Suppose the local optima form a finite nonempty set \(\mathcal M_\tau=\{m_\ell\}_{\ell=1}^{K}, \Delta_\ell=f^\star-f(m_\ell)\in[0,\tau].\)
Assume that there are $r_0>0$, constants $\mu_1,\ldots,\mu_K>0$,
and an exterior gap $g_0>\tau$ such that:
\begin{enumerate}
 \item The closed balls $\mathcal B_\ell=\bbB(m_\ell,r_0)$ are
 mutually disjoint and contained in $\operatorname{int}(\X)$.
 \textcolor{black}{By the growth condition in item~2, $m_\ell$ is the unique maximizer of $f$ on $\mathcal B_\ell$}.
 \item Within each ball,
 \begin{equation}\label{eq:separated-optima-growth}
  f(m_\ell)-f(x)\ge\mu_\ell\|x-m_\ell\|_2^\alpha,
  x\in\mathcal B_\ell.
 \end{equation}
 \item For any $x\in\X\setminus\bigcup_{\ell=1}^{K}B(m_\ell,r_0)$,
 one has $f^\star-f(x)\ge g_0$, where $B(m_\ell,r_0)$ is the open ball.
\end{enumerate}
Set $\underline\mu=\min_{1\le\ell\le K}\mu_\ell$.
For any choice $0<u_0<g_0$ and all $0<u\le u_0$,
\begin{equation}\label{eq:separated-optima-cover}
 \mathcal S_{u+}\subseteq
 \bigcup_{\ell:\,\Delta_\ell\le u}
 \bbB\!\left(m_\ell,
       \left(\frac{u-\Delta_\ell}{\mu_\ell}\right)^{1/\alpha}\right).
\end{equation}
Consequently, Assumption~\ref{as:new-cover} holds with this $K$, the same exponent $\alpha$, and $c_\rho=\underline\mu^{-1/\alpha}$, i.e., 
\begin{equation}\label{eq:separated-optima-cover-count}
 \mathcal N\bigl(\mathcal S_{u+},c_\rho u^{1/\alpha}\bigr)
 \le\#\{\ell:\Delta_\ell\le u\}\le K,
 0<u\le u_0.
\end{equation}
In particular, $u_0=\tau$ is admissible when $\tau>0$.
\end{proposition}
\begin{proof}
If $x\in\mathcal S_{u+}$, then $f^\star-f(x)\le u<g_0$. The exterior-gap condition therefore places $x$ in some $B(m_\ell,r_0)$. By~\eqref{eq:separated-optima-growth},
\[
 u\ge f^\star-f(x) =\Delta_\ell+f(m_\ell)-f(x)\ge\Delta_\ell+\mu_\ell\|x-m_\ell\|_2^\alpha.
\]
Thus $\Delta_\ell\le u$ and $x$ belongs to the corresponding ball in~\eqref{eq:separated-optima-cover}, and a \textcolor{black}{ball of} radius zero denotes the singleton $\{m_\ell\}$. \textcolor{black}{The radius of each ball in~\eqref{eq:separated-optima-cover}} is at most $\underline\mu^{-1/\alpha}u^{1/\alpha}$, proving
\eqref{eq:separated-optima-cover-count}.
\end{proof}

% The proposition permits one covering ball per target peak. At a given
% level $u$, only peaks with $\Delta_\ell\le u$ contribute, and some
% may share a covering ball, so $K$ remains an upper bound on the number
% of required neighborhoods. Separation and uniqueness within each
% ball provide an intuitive sufficient model; the proof itself uses
% the exterior gap and the growth inequality. To obtain the same
% covering constants for a family with varying $K$, require a common
% lower bound $\mu_\ell\ge\mu_*>0$ and a common $u_0<g_0$.
% Then $c_\rho=\mu_*^{-1/\alpha}$ works uniformly.

% \textcolor{black}{The proposition assigns one covering ball to each target peak. At a given
% level $u$, only peaks with $\Delta_\ell\le u$ contribute, and several
% may share a covering ball, so $K$ is an upper bound on the number
% of neighborhoods required. Separation and uniqueness of the maximizer within each
% ball give an intuitive sufficient model, although the proof itself uses only
% the exterior gap and the growth inequality. To obtain the same
% covering constants across a family in which $K$ varies, it suffices to require a common
% lower bound $\mu_\ell\ge\mu_*>0$ and a common $u_0<g_0$;
% then $c_\rho=\mu_*^{-1/\alpha}$ works uniformly.}

\subsection{Evaluation Complexity for Optimization and Region Learning}\label{subsec:cap-driven-variant}\label{sec:new-region}

We now bound the number of simulation evaluations needed to reach a prescribed \textcolor{black}{accuracy level}. The result covers both pure optimization ($T=0$) and simultaneous region learning ($T>0$). In the positive-tolerance range of Section~\ref{sec:new-lower-region}, it yields rates in the accuracy, tolerance, and \textcolor{black}{covering number bound} that match the lower bound of Theorem~\ref{thm:region-minimax-lower}.

We first present the following lemma that establishes $\mathbb P(\Econf)\ge1-\delta$ for the shared confidence event defined in~\eqref{eq:shared-confidence-event}.

\begin{lemma}\label{lem:new-confidence}
Let $\X$ be a nonempty compact subset of $\R^d$, and suppose Assumptions~\ref{as:new-noise}--\ref{as:new-rkhs} hold. Run Algorithm~\ref{alg:continuous-ccucb} with $N\ge2$ and $\delta\in(0,1)$, using the confidence parameters in~\eqref{eq:new-confidence-parameters}. With probability at least $1-\delta$, simultaneously over all phases, \textcolor{black}{all sample sizes within a phase by time} $N$, and all $x\in\X$,
\begin{equation}\label{eq:new-band}
 |f(x)-\mu_n(x)|\le\beta_Ns_n(x)+\zeta_N.
\end{equation}
\end{lemma}

\begingroup\color{black}
\paragraph{Phase accuracy and the full-budget allocation.}
Fix a phase $h$ of one track, with tolerance $T$ and accuracy level $\eps_h=2B\vartheta^{-h}$. 
% The \textcolor{black}{screening function fixed at the start of the phase} $G_h$ and retained width $W_{h-1}$ are defined in~\eqref{eq:new-frozen-cap}. 
Its acquisition function and maximum are
\begin{equation}\label{eq:cap-driven-width}
 a_n(x)=\min\{G_h(x),W_{h-1}(x),2\beta_Ns_n(x)+2\zeta_N\},\qquad
 a_n^\star=\max_{x\in\X}a_n(x).
\end{equation}
The phase samples a maximizer when $a_n^\star>2\eps_h$ and its track has budget remaining. A sampled point $x$ therefore requires both $G_h(x)>2\eps_h$ and $W_{h-1}(x)>2\eps_h$, that is, the point remains relevant to the tolerance and its maintained interval is still too wide. If $a_n^\star\le2\eps_h$, the phase completes and updates the shared bands. If instead the track's budget is exhausted first, the phase ends and its incomplete batch is discarded.
\par\endgroup

\begingroup\color{black}
The following theorem establishes the evaluations needed to obtain an \textcolor{black}{\textcolor{black}{estimated optimal point} with optimality gap} at most $2\eps$ and simultaneous region bounds with \textcolor{black}{region estimation error} at most $4\eps$. It applies to the full-budget allocation at both zero and positive tolerances.
\par\endgroup

\begingroup\color{black}
\begin{theorem}[Evaluation complexity for optimization and region learning]
\label{thm:region-learning-budget}
Let $\X$ be a nonempty compact subset of $\R^d$, and suppose Assumptions~\ref{as:new-noise}--\ref{as:new-rkhs} hold. Run Algorithm~\ref{alg:continuous-ccucb} with fixed $\vartheta>1$, $N\ge2$, $\delta\in(0,1)$, and $T\ge0$, using $m=0$ for $T=0$ and $m=N$ for $T>0$. On $\Econf$, which has probability at least $1-\delta$, the following claims hold simultaneously for all targets $\eps=2B\vartheta^{-H}\ge2\zeta_N$. When phase $H$ completes, its saved output satisfies
\begin{equation}\label{eq:target-phase-accuracy}
\begin{aligned}
 \hi_H-\lo_H&\le2\eps,\qquad f^\star-f(\widehat x_H)\le2\eps,\\
 I_{H,\tau}&\subseteq\mathcal S_\tau\subseteq O_{H,\tau},\qquad 0\le\tau\le T,\\
 O_{H,\tau}\setminus I_{H,\tau}
 &\subseteq\{x:|f^\star-f(x)-\tau|\le4\eps\},\qquad 0\le\tau\le T.
\end{aligned}
\end{equation}
A sufficient number of evaluations for phase $H$ to complete is
\begin{equation}\label{eq:region-global-cost}
 C\log^{2+d/\nu}(2N)\log^{1+d/(2\nu)}(2N/\delta)
 \bigl(1+\eps^{-(2+d/\nu)}\bigr).
\end{equation}
Under Assumption~\ref{as:new-cover} with $\alpha>\nu$, uniformly for $0\le T\le u_0/2$, the bound improves to
\begin{equation}\label{eq:region-local-cost}
\begin{split}
 &C\log^{2+d/\nu}(2N)\log^{1+d/(2\nu)}(2N/\delta)\left[1+\min\left\{\eps^{-(2+d/\nu)},\,
 K\eps^{-2}+K(T+\eps)^{d/\alpha}\eps^{-(2+d/\nu)}\right\}\right].
\end{split}
\end{equation}
% Each bound guarantees completion only when its right-hand side is at most $N$. Otherwise, the theorem makes no assertion that the target phase completes, and the algorithm returns its last saved output.
The constants depend only on the fixed model parameters and, for the localized bound, the fixed covering parameters other than $K$, that is, they are independent of $K,T,N,\delta,\eps$.
\end{theorem}
\par\endgroup

% \paragraph{Proof idea.}
% Proposition~\ref{prop:phase-accuracy} gives~\eqref{eq:target-phase-accuracy}. It also bounds each phase-$h$ query's performance gap by $T+L_\vartheta\eps_h$. The information-gain estimate in Lemma~\ref{lem:new-mig} and the capped phase-cost bound in Lemma~\ref{lem:cap-driven-phase-cost}, both in \textcolor{black}{the appendices}, control the evaluations at each resolution. Summing over the geometric phases gives the displayed budgets. \textcolor{black}{Appendix}~\ref{ec:region-budget} gives the complete proof.\hfill$\square$

Here $N$ is the maximum number of evaluations declared before the run, whereas the theorem bounds the number actually needed to reach the requested accuracy. The dependence on $N$ comes from calibrating $\beta_N$ and $\zeta_N$, which provides simultaneous confidence through the declared horizon. A larger horizon changes these confidence parameters, and hence the sufficient evaluation count, even if the target is reached early. To plan a run for a specified $\eps$, one can choose $N$ so that the right-hand side of~\eqref{eq:region-global-cost}, or of~\eqref{eq:region-local-cost}, is at most $N$ and $2\zeta_N\le\eps$. The corollary below makes such a choice explicitly and eliminates $N$ from the final accuracy-based bound.

\begingroup\color{black}
At $T=0$, the \textcolor{black}{\textcolor{black}{estimated optimal point}} satisfies $f^\star-f(\widehat x_H)\le2\eps$, while $\mathcal S_0=\varnothing$. For $\alpha>\nu$ and $\eps\le1$, the localized evaluation bound reduces, apart from the logarithmic factors in~\eqref{eq:region-local-cost} and \textcolor{black}{fixed costs of the initial phases}, to $\min\{\eps^{-(2+d/\nu)},\ K\eps^{-(2+d/\nu-d/\alpha)}\}.$
The second exponent is smaller than the first by $d/\alpha$, which is because as the target error decreases, the covering condition confines the relevant points to smaller neighborhoods of the maximizers, reducing the evaluations needed for optimization. For fixed $T>0$, the term $KT^{d/\alpha}\eps^{-(2+d/\nu)}$ instead accounts for refining an entire positive-tolerance region, and the same completed output also provides the \textcolor{black}{\textcolor{black}{estimated optimal point}}.
\par\endgroup

\begingroup\color{black}
Applying Theorem~\ref{thm:region-learning-budget} with a target accuracy level $\eta/(8\vartheta)<\eps\le\eta/8$, and choosing the horizon $N$ accordingly, gives the following upper bound on the minimax evaluation count, which matches Theorem~\ref{thm:region-minimax-lower}.
\par\endgroup

\begin{corollary}\label{cor:region-minimax-matching}
Under the fixed model and class of Theorem~\ref{thm:region-minimax-lower}, uniformly over its stated $K,T,\eta$ range and $0<\delta\le1/10$,
\begin{equation}\label{eq:region-minimax-matching}
\begin{aligned}
 % n_{\rm all}(\eta,T,\delta;\mathcal F_K)
 % &\ge c\sigma^2\operatorname{kl}(\delta,1-\delta)
       % \eta^{-(2+d/\nu)}\min\{1,KT^{d/\alpha}\},\\
 n_{\rm all}(\eta,T,\delta;\mathcal F_K)
 &\le C\eta^{-(2+d/\nu)}\min\{1,KT^{d/\alpha}\}
       \left[\log\frac{2}{\eta\delta}\right]^{3+3d/(2\nu)}.
\end{aligned}
\end{equation}
The constants depend only on the fixed model and class parameters and are independent of $K,T,\eta,\delta$. 

\end{corollary}

% \paragraph{Proof idea.}
% Since $\eta\le T/128$ and $\alpha>\nu$, the term $K\eta^{-2}$ in the evaluation bound is absorbed by $KT^{d/\alpha}\eta^{-(2+d/\nu)}$. Choose the declared budget $N$ to be the ceiling of a sufficiently large constant times the displayed upper-bound expression. This ensures both phase completion and $2\zeta_N\le\eps$. \textcolor{black}{Appendix}~\ref{ec:region-inversion} verifies these statements uniformly in $\delta$. The lower bound is Theorem~\ref{thm:region-minimax-lower}.\hfill$\square$

Theorem~\ref{thm:region-minimax-lower} and Corollary~\ref{cor:region-minimax-matching} show that, for any fixed $0<\delta\le1/10$, the minimax evaluation count has matching rates
\[
 n_{\rm all}(\eta,T,\delta;\mathcal F_K)
 =\widetilde\Theta\!\left(\eta^{-(2+d/\nu)}\min\{1,KT^{d/\alpha}\}\right).
\]
Here the tilde suppresses logarithms in the accuracy, with the confidence factors given explicitly in~\eqref{eq:region-minimax-lower} and~\eqref{eq:region-minimax-matching}.
% : the lower bound contains $\operatorname{kl}(\delta,1-\delta)\asymp\log(1/\delta)$, and the upper bound contains $[\log(2/(\eta\delta))]^{3+3d/(2\nu)}$. Thus the matching claim concerns \textcolor{black}{the orders in $\eta,K,T$}; the displayed bounds retain their respective confidence dependence.

% The simultaneous lower-bound construction uses different tolerances
% $\tau_i\in(0,T)$ to detect perturbations throughout a shell. In contrast,
% Theorem~\ref{thm:fixed-tolerance-lower} places the perturbation centers on
% a single level surface, so all tests use the same $\tau$, and its lower rate is different. The matching statement~\eqref{eq:region-minimax-matching} continues to concern the entire family of regions. 

% \textcolor{black}{Theorem~\ref{thm:fixed-tolerance-lower} places the perturbation centers on
% a single level surface, so that all tests use the same $\tau$, and its lower rate is
% correspondingly different. The matching statement~\eqref{eq:region-minimax-matching} concerns the entire family of regions.} 

% At $T=0$, the strict region is already known
% to be empty; the positive-tolerance and accuracy conditions in
% Corollary~\ref{cor:region-minimax-matching} exclude that trivial problem.

If $T=c_T\eta$ for a fixed $c_T\ge128$, the two accuracy terms in the evaluation bound are
\[
 \min\{\eta^{-(2+d/\nu)},K\eta^{-(2+d/\nu-d/\alpha)}\}.
\]
These are also the terms governing the evaluations needed \textcolor{black}{to achieve an optimality gap} of order $\eta$ in the $T=0$ specialization, where the first is the global bound, and the second uses the covering condition to reduce its exponent by $d/\alpha$. For fixed positive $T\le \textcolor{black}{T_{\mathrm{reg}}}$, the inverse-accuracy exponent is $2+d/\nu$, and the geometric multiplier is $\min\{1,KT^{d/\alpha}\}$. When $K<T^{-d/\alpha}$, the bound is smaller than the global rate by the factor $KT^{d/\alpha}$, and when $K\ge T^{-d/\alpha}$, the factor equals one.
% This explains the transition at $K\asymp T^{-d/\alpha}$. 
% For a fixed requested tolerance $T$, changing the simulation budget changes the achievable accuracy, but does not change this threshold in $K$.

% \paragraph{Evaluation budgets and decision precision.}
From a decision-making perspective, Theorem~\ref{thm:region-minimax-lower} and Corollary~\ref{cor:region-minimax-matching} quantify the evaluation cost of \textcolor{black}{keeping} the post-sampling choice of tolerance described in Section~\ref{subsec:region-decisions}. For $\alpha>\nu$ and $KT^{d/\alpha}<1$, the minimax budget grows as $KT^{d/\alpha}\eta^{-(2+d/\nu)}$, up to logarithmic factors.

% For $\alpha>\nu$, fixed confidence, $K\ge \textcolor{black}{K_{\mathrm{reg}}}$, $0<T\le \textcolor{black}{T_{\mathrm{reg}}}$, $0<\eta\le T/128$, and $KT^{d/\alpha}<1$, the minimax budget grows as $KT^{d/\alpha}\eta^{-(2+d/\nu)}$, up to logarithmic factors.

% In this expression, halving the \textcolor{black}{target region estimation error}
% $\eta$ multiplies the term by $2^{2+d/\nu}$; doubling the largest
% permitted loss $T$ multiplies it by $2^{d/\alpha}$, provided
% $2T\le \textcolor{black}{T_{\mathrm{reg}}}$ and $K(2T)^{d/\alpha}<1$.

% \paragraph{Tolerance and quota choices.}
\begin{remark}
    Theorem~\ref{thm:region-learning-budget} concerns the full-budget allocation $m\in\{0,N\}$. For $T>0$ and $m\le N/2$, Section~\ref{subsec:joint-tradeoff} quantifies the tradeoff between region accuracy and cumulative regret and identifies quotas that preserve the optimization rates. For $T=0$ and $m=0$, Section~\ref{sec:covering-minimax} establishes the pure-optimization regret bounds and their matching minimax ranges.
\end{remark}

\begingroup\color{black}
For $T=0$, the full-budget allocation is $m=0$, and its regret guarantees are those of Theorem~\ref{thm:new-regret}. The following proposition gives the optimization guarantees of the full-budget allocation $m=N$ for $T>0$, in which all evaluations are spent by the region track. This case is covered neither by Theorem~\ref{thm:new-regret}, which requires $T=0$, nor by Theorem~\ref{thm:budgeted-ccucb}, which requires $m\le\lfloor N/2\rfloor$.
\par\endgroup

\begingroup\color{black}
\begin{proposition}\label{prop:full-budget-global-regret}
Under the assumptions of Theorem~\ref{thm:region-learning-budget}, run Algorithm~\ref{alg:continuous-ccucb} with budget $N\ge2$, $T>0$, $m=N$, and $\delta=N^{-2}$. Then
\begin{equation}\label{eq:joint-global-regret}
\begin{aligned}
 \mathbb E r_N&\le C\log^{3/2}(2N)N^{-1/(2+d/\nu)},\\
 \mathbb E R_N&\le\min\{2BN,\ NT+C\log^{3/2}(2N)N^{1-1/(2+d/\nu)}\}.
\end{aligned}
\end{equation}
The constant depends only on the fixed model parameters and $\vartheta$, and is uniform in $T$ and $N$.
\end{proposition}
\par\endgroup

% \paragraph{Proof idea.}
% On $\Econf[N^{-2}]$, global phase costs and completed-phase accuracy bound the optimality gap. Each query contributes at most $T+L_\vartheta\eps_h$ to cumulative regret, so charging $T$ per evaluation and summing the remaining phase costs gives~\eqref{eq:joint-global-regret}. The terms involving $\zeta_N$ and the contributions from $\Econf[N^{-2}]^c$ are smaller than these bounds. Appendix~\ref{ec:full-budget-regret} gives the proof.\hfill$\square$

\begingroup\color{black}
Thus, devoting all evaluations to region learning keeps the global rate for simple regret. The designs that remain relevant to the region track always include the maximizers of $f$, so each completed region phase also gives $f^\star-f(\widehat x_h)\le2\eps_h$ by Proposition~\ref{prop:phase-accuracy}. In contrast, the cumulative regret bound contains the linear term $NT$, because learning the full positive-tolerance region may spend evaluations at designs whose performance gap is of order $T$. This term cannot be avoided at the fastest region accuracy. With $m=N$, Corollary~\ref{cor:region-minimax-matching} shows that SASORI achieves, up to logarithmic factors, the fastest region estimation error $(\min\{1,KT^{d/\alpha}\}/N)^{1/(2+d/\nu)}$. By~\eqref{eq:joint-regret-cost-lower}, as shown in Section~\ref{subsec:joint-tradeoff}, any valid procedure that achieves this accuracy, up to logarithmic factors, has worst-case cumulative regret of order at least $TN$, again up to logarithmic factors. For fixed $T>0$, the bound~\eqref{eq:joint-global-regret} is at most $2NT$ for all sufficiently large $N$. Hence, at the fastest region accuracy, the full-budget allocation attains the smallest possible order of cumulative regret, up to logarithmic factors. This complements Corollary~\ref{cor:joint-frontier}, which covers quotas $m_\eta\le\lfloor N/2\rfloor$. Section~\ref{subsec:joint-tradeoff} shows how a region quota $m<N$ reduces the cumulative regret to the optimization rate $N\rho_N$ at the price of a larger region estimation error.
\par\endgroup

\subsection{Optimization Regret Benchmarks}\label{subsec:optimization-benchmarks}
\label{sec:covering-minimax}

We now establish the pure-optimization benchmark used in the joint-learning analysis. Throughout this subsection, SASORI uses $T=0$ and $m=0$, so only the optimization track is active. The strict zero-tolerance region is already known to be empty. For $N\ge2$ and covering parameters $K\ge1$ and $\alpha>0$, define
\begin{equation}\label{eq:optimization-benchmark}
 \rho_N=\rho_N(K,\alpha):=\min\left\{(K/N)^{1/\max\{2,2+d/\nu-d/\alpha\}}, N^{-1/(2+d/\nu)}\right\}.
\end{equation}
The quantities $\rho_N$ and $N\rho_N$ describe the \textcolor{black}{simple regret and cumulative regret} rates, respectively. We give upper bounds for both, followed by matching lower bounds when $\alpha>\nu$. 

% The matching lower bounds hold beyond fixed class-dependent thresholds in $K$ and $N$.

% \textcolor{black}{The global rates below agree with those established in \citet{vakili2021optimal,li2022batches}. The theorem verifies these rates for the continuous capped policy and gives localized bounds uniformly in the covering number bound $K$. In~\eqref{eq:new-local-rates}, the \textcolor{black}{contributions involving $\zeta_N$} are $C\zeta_N$ for simple regret and $CN\zeta_N$ for cumulative regret; the corresponding \textcolor{black}{contributions from $\Econf[N^{-2}]^c$} are $2B/N^2$ and $2B/N$. These upper bounds are compared with minimax lower bounds on the same covering classes below. They also provide the optimization-only rates $\rho_N$ and $N\rho_N$ used to measure the additional regret cost of region learning in Section~\ref{subsec:joint-tradeoff}.}

\begingroup\color{black}
\begin{theorem}[Global and localized regret of batched SASORI ]
\label{thm:new-regret}
Under Assumptions~\ref{as:new-noise}--\ref{as:new-rkhs}, the policy in Algorithm~\ref{alg:continuous-ccucb} with $T=0$ and $\delta=N^{-2}$ satisfies
\begin{equation}\label{eq:new-global-rates}
 \mathbb E r_N=\widetilde O(N^{-\nu/(2\nu+d)}),\qquad
 \mathbb E R_N=\widetilde O(N^{(\nu+d)/(2\nu+d)}).
\end{equation}
Under Assumption~\ref{as:new-cover}, the same policy satisfies
\begin{equation}\label{eq:new-local-rates}
\begin{split}
 \mathbb E r_N &\le C{\log^{3+3d/(2\nu)}(2N)} \rho_N+C\zeta_N+2B/N^2,\\
 \mathbb E R_N &\le C{\log^{3+3d/(2\nu)}(2N)}\left[1+N\rho_N\right]+CN\zeta_N+2B/N.
\end{split}
\end{equation}
% \textcolor{black}{For the proof below, . The displayed logarithmic factors are independent of $K$.} 
\textcolor{black}{Constants in the localized statement may depend on $\vartheta$,} $c_\rho,u_0,\alpha$ as well as the fixed
model parameters. 
% Neither $K$ nor these geometric parameters is an input to the policy.
\end{theorem}
\par\endgroup

The global bounds in~\eqref{eq:new-global-rates} hold for arbitrarily large $K$ and without any localization assumption. Their \textcolor{black}{simple regret and cumulative regret} rates recover those of \citet{vakili2021optimal} and \citet{li2022batches}, respectively, and match the global lower rates of \citet{scarlett2017lower}. For $\alpha>\nu$, \textcolor{black}{the localized bound improves on the global bound} until $K$ reaches order $N^{d/(\alpha(2+d/\nu))}$, up to logarithmic factors. Theorem~\ref{thm:covering-lower-alpha-large} below shows that this transition is minimax sharp on the same fixed class.

\paragraph{Matching lower bounds when $\alpha>\nu$.}
\label{subsec:covering-lower-alpha-large}

For a fixed function class $\mathcal F$, define the minimax risks
\[
 \mathcal R_N^s(\mathcal F)
 =\inf_{\pi,\widehat x_N}\sup_{f\in\mathcal F}\mathbb E_f r_N,
 \qquad
 \mathcal R_N^c(\mathcal F)
 =\inf_\pi\sup_{f\in\mathcal F}\mathbb E_f R_N.
\]
Here $\pi$ ranges over sampling policies using $N$ evaluations, and the first infimum is also taken over the final \textcolor{black}{design selection rule}. The supremum is over the fixed class, so the resulting lower bounds apply to procedures that are required to work uniformly on that class. These risks give the smallest worst-case expected regret achievable with $N$ evaluations. We now derive lower bounds on the same classes $\mathcal F_K$ of Section~\ref{sec:new-lower-region}, which are also the classes appearing in Theorem~\ref{thm:new-regret}. These classes are indexed by their \textcolor{black}{covering number bound $K$}, while the norm upper bound and all other parameters are fixed as $K$ and $N$ vary. We prove matching minimax rates for the case $\alpha>\nu$ as shown in the following theorem, while other cases are discussed in Section~\ref{subsec:other-covering-regimes}.

% treat the critical endpoint and the remaining covering regimes. The unrestricted Sobolev rate does not require a covering assumption.

% Use the fixed covering class $\mathcal F_K(B,c_\rho,u_0,\alpha)$
% defined in Section~\ref{sec:new-lower-region}.
% The domain contains a fixed interior Euclidean ball, and the normalized
% Mat\'ern smoothness $\nu>0$ and length scale are fixed. The result
% below covers any $\alpha>\nu$, including $\nu<\alpha<2$.
% It uses independent Gaussian observation noise with fixed variance
% $\sigma^2>0$.

% The localized-power regularity used in the construction is stated in
% Lemma~\ref{lem:localized-power-native} \textcolor{black}{of the appendices}
% Section~\ref{ec:localized-power}.

% \paragraph{Proof idea.}
% Dyadic annuli have $H^{\nu+d/2}$ norms bounded by
% $C2^{-j(\alpha-\nu)}$, which are summable when $\alpha>\nu$.
% \textcolor{black}{Appendix}~\ref{ec:localized-power} supplies the
% cutoff and restriction argument.\hfill$\square$

\begin{theorem}[\textcolor{black}{Matching lower bounds for} $\alpha>\nu$]
\label{thm:covering-lower-alpha-large}
Fix $\alpha>\nu$, $c_\rho,u_0,B,\sigma>0$, the kernel, and the domain. Assume independent Gaussian observation errors with variance $\sigma^2$. There are a finite integer $\textcolor{black}{K_{\mathrm{opt}}}\ge1$ and a finite threshold $\textcolor{black}{N_{\mathrm{opt}}}\ge2$, depending only on these fixed parameters, such that, for any integer $K\ge \textcolor{black}{K_{\mathrm{opt}}}$ and any $N\ge \textcolor{black}{N_{\mathrm{opt}}}$,
\begin{align}
 \inf_{\pi,\widehat x_N}\sup_{f\in\mathcal F_K}
       \mathbb E_f r_N
 &\ge c\rho_N,\label{eq:covering-simple-lower}\\
 \inf_\pi\sup_{f\in\mathcal F_K}\mathbb E_f R_N
 &\ge cN\rho_N.\label{eq:covering-regret-lower}
\end{align}
The constants $c>0$, $\textcolor{black}{K_{\mathrm{opt}}}$, and $\textcolor{black}{N_{\mathrm{opt}}}$ are independent of $K,N$.
\end{theorem}

% \paragraph{Proof idea.}
% Use the common baseline $-\mu_K\|x-c\|^\alpha$, with
% $\mu_K$ proportional to $K^{-\alpha/d}$. Its norm decreases with
% $K$, leaving a fixed budget for a local perturbation. Inside its
% near-optimal basin, replace the objective on disjoint balls of radius
% proportional to $\eps^{1/\nu}$ by peaks with an exact
% $\alpha$-power top. This replacement preserves the covering condition
% at all smaller levels, rather than only at the testing resolution.
% The number of distinguishable alternatives is bounded below by a
% constant times
% \[
%  \min\{\eps^{-d/\nu},K\eps^{-d/\nu+d/\alpha}\}.
% \]
% Gaussian testing with $N\eps^2$ comparable to this count gives both
% \textcolor{black}{regret orders}. \textcolor{black}{Appendix}~\ref{ec:covering-lower}
% verifies the fixed norm budget, \textcolor{black}{covering condition at every near-optimality level}, and uniform
% choices of constants, \textcolor{black}{including the sufficient richness condition discussed below}.\hfill$\square$

% The theorem keeps the norm budget, noise variance, kernel, domain, and covering parameters $c_\rho,u_0,\alpha$ fixed while $K$ and $N$ vary. 
The lower bounds in Theorem \ref{thm:covering-lower-alpha-large} hold uniformly for $K\ge \textcolor{black}{K_{\mathrm{opt}}}$ and $N\ge \textcolor{black}{N_{\mathrm{opt}}}$. 
Combining Theorem \ref{thm:covering-lower-alpha-large} with Theorem~\ref{thm:new-regret}, and writing $a=2+d/\nu-d/\alpha>2$, gives
\begin{equation}\label{eq:covering-minimax-summary}
 \mathcal R_N^s(\mathcal F_K)
 =\widetilde\Theta(\rho_N),\qquad
 \mathcal R_N^c(\mathcal F_K)
 =\widetilde\Theta(N\rho_N).
\end{equation}
Recall that the \textcolor{black}{covering number bound} $K$ is the uniform upper bound in Assumption~\ref{as:new-cover} on the number of balls of radius $c_\rho u^{1/\alpha}$ needed to cover $\mathcal S_{u+}$ for $0<u\le u_0$. Increasing $K$ weakens this geometric restriction and enlarges the class. The minimax rate switches between the localized rate $(K/N)^{1/a}$ and the global rate $N^{-1/(2+d/\nu)}$. Equating these two rates yields
\[
 K_\star(N)=N^{1-a/(2+d/\nu)} =N^{d/[\alpha(2+d/\nu)]} =N^{\nu d/[\alpha(2\nu+d)]}.
\]
Thus localization improves the minimax rates when $K$ is below $K_\star(N)$, while for $K\gtrsim K_\star(N)$ the simple and cumulative regrets have the global Mat\'ern/Sobolev rates $N^{-\nu/(2\nu+d)}$ and $N^{(\nu+d)/(2\nu+d)}$, respectively, up to logarithmic factors \citep{scarlett2017lower,vakili2021optimal,li2022batches}. Note that the global upper bound remains valid without any covering assumption.

\subsection{The Regret Cost of Simultaneous Region Learning}
\label{sec:joint}\label{subsec:joint-tradeoff}

Region learning spends evaluations at points that may already be known to be suboptimal. We first quantify the cumulative regret that any valid region guarantee forces, and then choose a region quota that controls this additional cost. Throughout the matching results below, $\alpha>\nu$, $K\ge\max\{\textcolor{black}{K_{\mathrm{reg}}},\textcolor{black}{K_{\mathrm{opt}}}\}$, $0<T\le \textcolor{black}{T_{\mathrm{reg}}}$, and $0<\eta\le T/128$, and $N\ge \textcolor{black}{N_{\mathrm{opt}}}$, with the fixed class parameters of Theorem~\ref{thm:region-minimax-lower}. 

% \textcolor{black}{Throughout the matching results below, $\alpha>\nu$, $K\ge\max\{\textcolor{black}{K_{\mathrm{reg}}},\textcolor{black}{K_{\mathrm{opt}}}\}$, $0<T\le \textcolor{black}{T_{\mathrm{reg}}}$, $0<\eta\le T/128$, and $N\ge \textcolor{black}{N_{\mathrm{opt}}}$, with the fixed class parameters of Theorem~\ref{thm:region-minimax-lower}; the thresholds $\textcolor{black}{K_{\mathrm{reg}}},\textcolor{black}{T_{\mathrm{reg}}}$ come from Theorem~\ref{thm:region-minimax-lower}, and the richness and horizon thresholds $\textcolor{black}{K_{\mathrm{opt}}},\textcolor{black}{N_{\mathrm{opt}}}$ from Theorem~\ref{thm:covering-lower-alpha-large}. The optimization benchmark is $\rho_N$ from~\eqref{eq:optimization-benchmark}, with \textcolor{black}{cumulative regret} scale $N\rho_N$.}

At a fixed positive $T$, suitable region quotas preserve the optimization-only rates for both simple and cumulative regret while allowing consistent region estimation. However, keeping cumulative regret at the optimization rate requires a smaller region quota than devoting all $N$ evaluations to region learning, so \textcolor{black}{region estimation error} decreases more slowly. Theorem~\ref{thm:joint-regret-cost} gives a lower bound on the regret required by any prescribed \textcolor{black}{region} accuracy. Theorem~\ref{thm:budgeted-ccucb} then provides an allocation whose cumulative regret matches this lower bound up to the explicit logarithmic factors in Corollary~\ref{cor:joint-frontier}.

\begin{theorem}\label{thm:joint-regret-cost}
Under the fixed model and parameter range of Theorem~\ref{thm:region-minimax-lower}, any $(\eta,T,\delta)$-valid procedure using a fixed budget of $N$ observations satisfies
\begin{equation}\label{eq:joint-regret-cost-lower}
 \sup_{f\in\mathcal F_K}\mathbb E_f R_N
 \ge c\sigma^2 T\min\{1,KT^{d/\alpha}\}\eta^{-(2+d/\nu)}
            \operatorname{kl}(\delta,1-\delta).
\end{equation}
Here $K\ge \textcolor{black}{K_{\mathrm{reg}}}$, $0<T\le \textcolor{black}{T_{\mathrm{reg}}}$, $0<\eta\le T/128$, and $0<\delta\le1/10$, with constants uniform in $K,T,\eta,\delta,N$. The result also holds when the procedure knows the common optimum value and global maximizer of the lower-bound family.
\end{theorem}
% \paragraph{Proof idea.}
% Each informative perturbation in the region lower-bound construction has baseline performance gap at least $T/8$. The tests that require order $\eta^{-2}$ observations per perturbation therefore require regret of order $T\eta^{-2}$ there. Summing over the disjoint perturbations proves~\eqref{eq:joint-regret-cost-lower}; \textcolor{black}{Appendix}~\ref{ec:joint-tradeoff} gives the proof.\hfill$\square$

Recall that Algorithm~\ref{alg:continuous-ccucb} assigns quotas $N-m$ and $m$ to the optimization and region tracks, respectively, and alternates whole phases while sharing completed confidence bands. With $T=0$ and $m=0$, all $N$ evaluations use the optimization track. With $T>0$ and $m=N$, all evaluations use the region track analyzed in Theorem~\ref{thm:region-learning-budget}. The next theorem considers $m\le N/2$, which reserves at least half the evaluations for optimization.

\begingroup\color{black}
\begin{theorem}\label{thm:budgeted-ccucb}
Let $\X$ be a nonempty compact subset of $\R^d$, and suppose Assumptions~\ref{as:new-noise}--\ref{as:new-rkhs} hold. Let $N\ge2$, $T\ge0$, and $0\le m\le\lfloor N/2\rfloor$, with $m=0$ if $T=0$. For $0<\delta<1$, set $\bar\delta=\min\{\delta,N^{-2}\}$ and calibrate~\eqref{eq:new-confidence-parameters} at horizon $N$ and \textcolor{black}{error probability} $\bar\delta$.
Under Assumption~\ref{as:new-cover}, for any $\alpha>0$, the simple and cumulative regret satisfy
\begin{equation}\label{eq:budgeted-simple-regret}
\begin{split}
 \mathbb E_f r_N\le{}&C\log^{2+d/\nu}(2N)\log^{1+d/(2\nu)}(2N/\bar\delta)\rho_N
       +C\zeta_N+2B\bar\delta,
\end{split}
\end{equation}
\begin{equation}\label{eq:budgeted-cumulative-regret}
\begin{split}
 \mathbb E_f R_N\le{}&CN\rho_N\log^{2+d/\nu}(2N)\log^{1+d/(2\nu)}(2N/\bar\delta)+CTm+CN\zeta_N+2BN\bar\delta.
\end{split}
\end{equation}
Without Assumption~\ref{as:new-cover}, both inequalities hold with $\rho_N$ replaced by $N^{-1/(2+d/\nu)}$.

For $T>0$, on $\Econf[\bar\delta]$ from~\eqref{eq:shared-confidence-event}, if the region track completes a phase with accuracy level $\eps=\textcolor{black}{2B\vartheta^{-H}}\ge2\zeta_N$, the output satisfies, simultaneously for $0\le\tau\le T$,
\begin{equation}\label{eq:budgeted-output-accuracy}
\begin{aligned}
 f^\star-f(\widehat x_N)&\le2\eps,\qquad
 I_{N,\tau}\subseteq\mathcal S_\tau\subseteq O_{N,\tau},\\
 O_{N,\tau}\setminus I_{N,\tau}
 &\subseteq\{x:|f^\star-f(x)-\tau|\le4\eps\}.
\end{aligned}
\end{equation}
For any $T>0$, on $\Econf[\bar\delta]$, the region track completes phase $H$ (together with all preceding phases) within its quota of $m$ observations and saves the output in~\eqref{eq:budgeted-output-accuracy} if
\begin{equation}\label{eq:budgeted-region-global-cost}
 m\ge C\log^{2+d/\nu}(2N)\log^{1+d/(2\nu)}(2N/\bar\delta)
            (1+\eps^{-(2+d/\nu)}).
\end{equation}
Under Assumption~\ref{as:new-cover} with $\alpha>\nu$, for $0<T\le u_0/2$, the following smaller quota suffices for the same target-phase completion:
\begin{equation}\label{eq:budgeted-region-cost}
\begin{split}
 m\ge{}&C\log^{2+d/\nu}(2N)\log^{1+d/(2\nu)}(2N/\bar\delta)\left[1+\min\left\{\eps^{-(2+d/\nu)},
 K\eps^{-2}+K(T+\eps)^{d/\alpha}\eps^{-(2+d/\nu)}\right\}\right].
\end{split}
\end{equation}
% At $T=0$, $I_{N,0}=O_{N,0}=\varnothing$; a completed optimization phase supplies the \textcolor{black}{\textcolor{black}{estimated optimal point}} guarantee in~\eqref{eq:target-phase-accuracy}. 
The shared confidence event $\Econf[\bar\delta]$ in~\eqref{eq:shared-confidence-event} has probability at least $1-\bar\delta$. \textcolor{black}{Constants depend only on the fixed model parameters, $\vartheta$, and, for the covering-dependent} bounds~\eqref{eq:budgeted-simple-regret}, \eqref{eq:budgeted-cumulative-regret}, and~\eqref{eq:budgeted-region-cost}, the fixed covering parameters other than $K$; they are independent of $K,N,T,m,\delta$.
\end{theorem}
\par\endgroup
% \paragraph{Proof idea.}
% Monotone intersections preserve each track's width and \textcolor{black}{\textcolor{black}{estimated optimal point}} guarantees on $\Econf[\bar\delta]$. Optimization and region queries have performance gaps at most $L_\vartheta\eps$ and $T+L_\vartheta\eps$, respectively. Summing phase costs above a balancing resolution gives the optimization term, while the region observations contribute at most the same term plus $CTm$. Summing the region track's own phase costs gives~\eqref{eq:budgeted-region-cost}; \textcolor{black}{Appendix}~\ref{ec:joint-tradeoff} gives the details.\hfill$\square$

Combining the optimization lower bound~\eqref{eq:covering-regret-lower}, the region-induced regret lower bound~\eqref{eq:joint-regret-cost-lower}, and Theorem~\ref{thm:budgeted-ccucb} yields the following characterization.

\begingroup\color{black}
\begin{corollary}\label{cor:joint-frontier}
Fix $\alpha>\nu$ and $0<\delta\le1/10$. Let $K\ge\max\{\textcolor{black}{K_{\mathrm{reg}}},\textcolor{black}{K_{\mathrm{opt}}}\}$ and $N\ge \textcolor{black}{N_{\mathrm{opt}}}$, where $\textcolor{black}{K_{\mathrm{reg}}},\textcolor{black}{T_{\mathrm{reg}}}$ are from Theorem~\ref{thm:region-minimax-lower} and $\textcolor{black}{K_{\mathrm{opt}}},\textcolor{black}{N_{\mathrm{opt}}}$ are from Theorem~\ref{thm:covering-lower-alpha-large}. Let $0<T\le \textcolor{black}{T_{\mathrm{reg}}}$, $0<\eta\le T/128$, and $\bar\delta=\min\{\delta,N^{-2}\}$. Suppose $\eta\ge\textcolor{black}{16\max\{2,\vartheta\}\zeta_N}$ and
\begin{equation}\label{eq:joint-frontier-feasibility}
\begin{split}
 m_\eta:=\Bigl\lceil{}&C\log^{2+d/\nu}(2N)\log^{1+d/(2\nu)}(2N/\bar\delta)\min\{1,KT^{d/\alpha}\}\eta^{-(2+d/\nu)}\Bigr\rceil\le\lfloor N/2\rfloor,
\end{split}
\end{equation}
with a sufficiently large fixed $C$. Define $\mathfrak R_N(\eta,T,\delta;\mathcal F_K)$ as the infimum of worst-case expected cumulative regret among $N$-observation policies that are $(\eta,T,\delta)$-valid on $\mathcal F_K$. Then
\begin{equation}\label{eq:joint-frontier-lower}
\begin{split}
 \mathfrak R_N(\eta,T,\delta;\mathcal F_K)\ge c\max\Bigl\{N\rho_N,\;T\min\{1,KT^{d/\alpha}\}\eta^{-(2+d/\nu)}\operatorname{kl}(\delta,1-\delta)\Bigr\},
\end{split}
\end{equation}
and
\begin{equation}\label{eq:joint-frontier}
\begin{split}
 \mathfrak R_N(\eta,T,\delta;\mathcal F_K)\le{}&C\log^{2+d/\nu}(2N)\log^{1+d/(2\nu)}(2N/\bar\delta)\\
 &{}\times\max\Bigl\{N\rho_N,\;T\min\{1,KT^{d/\alpha}\}\eta^{-(2+d/\nu)}\Bigr\}.
\end{split}
\end{equation}
\textcolor{black}{The constants may depend on fixed $\vartheta$ and are independent of $K,N,T,\eta,\delta$. SASORI} with $m=m_\eta$ attains the upper bound while also satisfying the \textcolor{black}{simple regret} bound~\eqref{eq:budgeted-simple-regret}. 
% \textcolor{black}{At fixed confidence, the lower bound~\eqref{eq:joint-frontier-lower} and upper bound~\eqref{eq:joint-frontier} have the same dependence on $N,K,T,\eta$, up to the displayed logarithmic factor.}
\end{corollary}
\par\endgroup
% \paragraph{Proof idea.}
% Every region-valid policy remains subject to the optimization lower bound, and also pays the regret required by~\eqref{eq:joint-regret-cost-lower}. Taking their maximum gives~\eqref{eq:joint-frontier-lower}. For the upper bound, use a target $\textcolor{black}{\eta/(8\vartheta)<\eps\le\eta/8}$ and the quota~\eqref{eq:joint-frontier-feasibility} in Theorem~\ref{thm:budgeted-ccucb}. The two nonnegative regret contributions add to at most twice their maximum. \textcolor{black}{Appendix}~\ref{ec:joint-tradeoff} verifies the constants and remainder terms.\hfill$\square$

% \paragraph{Quotas that preserve the optimization rates.}

For $T>0$, the choice
\begin{equation}\label{eq:regret-preserving-quota}
 m=\left\lfloor\min\left\{\frac N2,\frac{N\rho_N}{T}\right\}\right\rfloor
\end{equation}
keeps the extra term $Tm$ below the optimization contribution in~\eqref{eq:budgeted-cumulative-regret}. This tuning uses $K$ and $\alpha$. A choice independent of these parameters is
\begin{equation}\label{eq:geometry-free-quota}
 m=\left\lfloor\min\{N/2,\sqrt N/T\}\right\rfloor.
\end{equation}
Indeed, $K\ge1$ and $2+d/\nu-d/\alpha>2$ imply $\sqrt N\le N\rho_N$, so this choice also preserves the regret bounds.

\begingroup\color{black}
For either positive quota $m$, a sufficient condition on the target region estimation error is
\begin{equation}\label{eq:regret-preserving-region-accuracy}
\begin{aligned}
 \eta&\ge16\max\{2,\vartheta\}\zeta_N, \eta\le T/128,
 \eta&\ge C\Bigl[\frac{\log^{2+d/\nu}(2N)\log^{1+d/(2\nu)}(2N/\bar\delta)}m
             \min\{1,KT^{d/\alpha}\}\Bigr]^{1/(2+d/\nu)}.
\end{aligned}
\end{equation}
The upper restriction $\eta\le T/128$ is the accuracy range in which the positive-tolerance lower bound applies and the $K\eps^{-2}$ term is absorbed. The derivations behind the following rates and the feasibility conditions stated below are given in Section~\ref{ec:proof-quota}. To obtain explicit rates from~\eqref{eq:regret-preserving-region-accuracy}, fix $K$, $T>0$, and $\delta>0$, and let $N$ increase. Since $\alpha>\nu$, the localized rate eventually leads to $\rho_N=(K/N)^{1/(2+d/\nu-d/\alpha)}$. Thus the quotas in~\eqref{eq:regret-preserving-quota} and~\eqref{eq:geometry-free-quota} have respective orders
\[
 m\asymp N^{1-1/(2+d/\nu-d/\alpha)}\text{ and }m\asymp N^{1/2},
\]
where constants may depend on the fixed $K,T$. 
% The truncation at $N/2$ and integer rounding do not change these orders for sufficiently large $N$. 
Also, $\bar\delta=N^{-2}$ for all large $N$, so the logarithmic product in~\eqref{eq:regret-preserving-region-accuracy} has order $\log^{3+3d/(2\nu)}(2N)$. Substituting the two quota orders into~\eqref{eq:regret-preserving-region-accuracy} therefore gives the following sufficient conditions on the target region estimation error:
\begin{equation}\label{eq:tuned-region-rate}
 \eta\ge C\log^{3/2}(2N)N^{-\{1-1/(2+d/\nu-d/\alpha)\}/(2+d/\nu)}\text{ for~\eqref{eq:regret-preserving-quota}},
\end{equation}
\begin{equation}\label{eq:conservative-region-rate}
 \eta\ge C\log^{3/2}(2N)N^{-1/[2(2+d/\nu)]}\text{ for~\eqref{eq:geometry-free-quota}}.
\end{equation}
Both lower thresholds decrease to zero. For large $N$ they exceed $16\max\{2,\vartheta\}\zeta_N$ and fall below the fixed $T/128$, so the required accuracy interval is nonempty. Meanwhile, $Tm\le N\rho_N$ for both quotas, which preserves the regret rates in~\eqref{eq:budgeted-simple-regret}--\eqref{eq:budgeted-cumulative-regret}.
\par\endgroup

% \paragraph{The accuracy cost of preserving cumulative regret.}
The accuracy cost of preserving cumulative regret can now be made explicit. If all evaluations can be devoted to region learning, its fastest accuracy has rate
\[
 \left(\frac{\min\{1,KT^{d/\alpha}\}}N\right)^{1/(2+d/\nu)}.
\]
Fix $K$, $T>0$, and \textcolor{black}{$\delta$. If the target region estimation error} is at most a fixed multiple of this \textcolor{black}{fastest region-accuracy} rate, substituting it into~\eqref{eq:joint-regret-cost-lower} gives a \textcolor{black}{cumulative regret} lower bound $cTN$. Thus \textcolor{black}{achieving} that accuracy requires a positive average performance gap $\sup_f\mathbb E_fR_N/N\ge cT$. In contrast, the \textcolor{black}{optimal cumulative regret} is sublinear in $N$. Indeed, for fixed $K$,
\[
 N\rho_N\asymp N^{1-1/(2+d/\nu-d/\alpha)}, 1-\frac{1}{2+d/\nu-d/\alpha}<1.
\]
% Multiplying this benchmark by any fixed polynomial in $\log N$ still gives $o(N)$, so its average performance gap tends to zero. 
% Allowing a logarithmic factor in regional accuracy only divides this lower bound by a polynomial in $\log N$; the resulting bound still exceeds the optimization rate for sufficiently large $N$. 
Hence, the linear lower bound $cTN$ is incompatible with the \textcolor{black}{optimal cumulative regret} rate. Consequently, the fastest \textcolor{black}{region} accuracy and the \textcolor{black}{optimal cumulative regret} cannot both be \textcolor{black}{achieved}. More generally, a \textcolor{black}{cumulative regret} bound of order $N\rho_N$ implies, at fixed confidence,
\begin{equation}\label{eq:joint-necessary-accuracy}
\begin{split}
 \eta\ge c\max\Biggl\{&\left(\frac{\min\{1,KT^{d/\alpha}\}}N\right)^{1/(2+d/\nu)},\left(\frac{T\min\{1,KT^{d/\alpha}\}}{N\rho_N}\right)^{1/(2+d/\nu)}\Biggr\}.
\end{split}
\end{equation}
Simple regret, on the other hand, remains compatible with the fastest \textcolor{black}{region} accuracy by choosing $m=\lfloor N/2\rfloor$. 
% Allowing the explicit logarithmic losses in~\eqref{eq:budgeted-cumulative-regret} allows the corresponding logarithmic relaxation of the second term.

\begin{remark}[When the tolerance and accuracy ranges are feasible]
For a chosen $m\ge1$, the sufficient interval~\eqref{eq:regret-preserving-region-accuracy} is nonempty whenever $T\le \textcolor{black}{T_{\mathrm{reg}}}$ and
\begin{equation}\label{eq:joint-tolerance-feasibility}
\begin{split}
 T\ge{\color{black}2048\max\{2,\vartheta\}}\zeta_N,\textcolor{black}{T}\ge C\min\Biggl\{&\left[\frac{K\log^{2+d/\nu}(2N)\log^{1+d/(2\nu)}(2N/\bar\delta)}m\right]^{1/(2+d/\nu-d/\alpha)},\\
 &\left[\frac{\log^{2+d/\nu}(2N)\log^{1+d/(2\nu)}(2N/\bar\delta)}m\right]^{1/(2+d/\nu)}\Biggr\}.
\end{split}
\end{equation}
% These conditions follow by substituting the largest permitted error $\eta=T/128$ into the quota requirement. They are sufficient conditions, rather than impossibility thresholds.
% In particular, the current bound does not certify $T$ equal to a fixed multiple of the optimization rate without logarithmic inflation. 
For~\eqref{eq:regret-preserving-quota}, at fixed confidence a sufficient range \textcolor{black}{when the localized rate $(K/N)^{1/(2+d/\nu-d/\alpha)}$ in $\rho_N$ is smaller} is
\[
 T\ge C(K/N)^{1/(2+d/\nu-d/\alpha)}\log^{(3+3d/(2\nu))/(1+d/\nu-d/\alpha)}(2N),
\]
when $KT^{d/\alpha}\le1$ and $(K/N)^{1/(2+d/\nu-d/\alpha)}\le N^{-1/(2+d/\nu)}$. \textcolor{black}{When the global rate $N^{-1/(2+d/\nu)}$ in $\rho_N$ is smaller}, a sufficient range is
\[
 T\ge C N^{-1/(2+d/\nu)}\log^{(3+3d/(2\nu))/(1+d/\nu)}(2N),
\]
when $KT^{d/\alpha}\ge1$ and $N^{-1/(2+d/\nu)}\le(K/N)^{1/(2+d/\nu-d/\alpha)}$. Both also require \textcolor{black}{$T\ge2048\max\{2,\vartheta\}\zeta_N$} and $T\le \textcolor{black}{T_{\mathrm{reg}}}$. Thus the tolerance can have the same polynomial order as the optimization error, up to logarithmic factors.
\end{remark}

% END tmp/joint_tradeoff_results

% BEGIN region_boundary

\subsection{Region Identification at a Prespecified Tolerance}\label{subsec:fixed-tolerance-screening}

\textcolor{black}{When} fixing a single tolerance $\tau>0$ before sampling, the procedure estimates the unknown maximum and \textcolor{black}{determines} membership in the single region $\mathcal S_\tau$. We use Algorithm~\ref{alg:continuous-ccucb} with only the region track active ($T=\tau,m=N$), for a chosen budget $N\ge2$ and \textcolor{black}{error probability} $\delta\in(0,1)$. Only the \textcolor{black}{screening function for the phase} and the returned region change, and all other steps, including \textcolor{black}{the check for inconsistent lower and upper bounds}, follow Algorithm~\ref{alg:continuous-ccucb}.

\begingroup\color{black}
We keep the phase accuracy levels $\eps_h=2B\vartheta^{-h}$ with fixed $\vartheta>1$ and the notation for the confidence bounds from Algorithm~\ref{alg:continuous-ccucb} and~\eqref{eq:new-elimination}.
Immediately before phase $h$, abbreviate the current bands and extrema by $\ell=\ell_{h-1}$, $u=u_{h-1}$, $\lo=\lo_{h-1}$, and $\hi=\hi_{h-1}$, and replace the region screening function by
\begin{equation}\label{eq:boundary-query-set}
\begin{split}
 g_{h-1}(x)&=\max\{u(x)-\lo,\min\{\hi-\tau-\ell(x),u(x)-\lo+\tau\}\},\\
 G_h(x)&=[g_{h-1}(x)+4\eps_h]_+.
\end{split}
\end{equation}
The first term in the maximum in~\eqref{eq:boundary-query-set}, $u(x)-\lo$, keeps possible maximizers, and the second term keeps points whose membership at the chosen tolerance is undetermined. After a completed phase, the procedure returns
\begin{equation}\label{eq:boundary-output}
 I_{h,\tau}=\{x:\ell_h(x)>\hi_h-\tau\},O_{h,\tau}=\{x:u_h(x)>\lo_h-\tau\}.
\end{equation}
The following theorem gives accuracy and a sufficient budget for this single-tolerance output.
\par\endgroup

\begingroup\color{black}
\begin{theorem}
\label{thm:fixed-tolerance-boundary}

Let $\X$ be any nonempty compact subset of $\R^d$, and assume that
Assumptions~\ref{as:new-noise}--\ref{as:new-rkhs} hold. With probability
at least $1-\delta$, simultaneously over all completed phases,
\begin{equation}\label{eq:boundary-optimum-accuracy}
\begin{split}
 \lo_h\le f^\star\le\hi_h,\hi_h-\lo_h\le2\eps_h,f^\star-f(\widehat x_h)\le2\eps_h,\widehat x_h\in\argmax_{x\in\X}\ell_h(x),
\end{split}
\end{equation}
and
\begin{equation}\label{eq:boundary-certificate}
 I_{h,\tau}\subseteq\mathcal S_\tau\subseteq O_{h,\tau},
 O_{h,\tau}\setminus I_{h,\tau} \subseteq\{x:|f^\star-f(x)-\tau|\le4\eps_h\}.
\end{equation}
For any target phase $H$ with $\eps_H\ge2\zeta_N$, a sufficient evaluation budget for completion on the same event is
\begin{equation}\label{eq:boundary-budget}
 C\log^{2+d/\nu}(2N)\log^{1+d/(2\nu)}(2N/\delta) \bigl(1+\eps_H^{-(2+d/\nu)}\bigr),
\end{equation}
provided this expression is at most $N$. The constant $C$ depends only on the fixed model parameters and $\vartheta$, and is independent of $N,\delta,\tau,H$.
\end{theorem}
\par\endgroup

\begingroup\color{black}
Theorem~\ref{thm:fixed-tolerance-boundary} provides both an estimated optimal point and an inner/outer approximation to the prescribed region. Choosing an accuracy level $\eps_H$ of order $\eta$ with $4\eps_H\le\eta$ gives an optimality gap at most $2\eps_H$ and leaves membership undetermined only where $|f^\star-f(x)-\tau|\le\eta$, provided that $\eps_H\ge2\zeta_N$ and the sufficient evaluation bound in~\eqref{eq:boundary-budget} is at most $N$. The general sufficient budget is $\widetilde O(\eta^{-2-d/\nu})$ and requires neither Assumption~\ref{as:new-cover} nor regularity of the tolerance boundary. Its exponent of $\eta^{-1}$ is the same as in the global simultaneous-region bound in Theorem~\ref{thm:region-learning-budget}. Thus fixing one tolerance alone does not by itself improve this general upper bound, but can improve the rates under boundary geometry condition, as quantified below.
\par\endgroup

% The screening function keeps possible maximizers and unresolved classifications. At phase completion their confidence widths control both the optimum-value error and the region estimation error. 
% Geometric summation of the phase costs gives~\eqref{eq:boundary-budget}. Appendix~\ref{ec:boundary} proves these claims and gives a refined budget in terms of boundary covers.

The first corollary extracts the cumulative regret cost of the testing construction used for Theorem~\ref{thm:fixed-tolerance-lower}.

\begin{corollary}\label{cor:fixed-tolerance-regret-lower}
Under the model assumptions and parameter range of
Theorem~\ref{thm:fixed-tolerance-lower}, consider any procedure using a fixed budget of $N$ observations that, for any $f\in\mathcal F_K$, returns a region guarantee satisfying~\eqref{eq_region_require} at the prescribed tolerance $\tau$, with \textcolor{black}{region estimation error} at most $\eta$ and probability at least $1-\delta$. Then
\begin{equation}\label{eq:fixed-tolerance-regret-lower}
\begin{split}
 \sup_{f\in\mathcal F_K}\mathbb E_fR_N\ge{}&c\sigma^2\tau\operatorname{kl}(\delta,1-\delta)\eta^{-2-(d-1)/\nu}\bigl(\min\{1,K\tau^{d/\alpha}\}\bigr)^{(d-1)/d}.
\end{split}
\end{equation}
The constant is independent of $K,N,\tau,\eta,\delta$. The bound continues to hold when the common optimizer and optimum value of the lower-bound family are \textcolor{black}{provided} before sampling.
\end{corollary}

\begingroup\color{black}
The next corollary gives simple regret and cumulative regret upper bounds when the procedure refines the region through a prescribed target phase and then uses its saved estimated optimal point for the remaining evaluations. The bound separates the cost of learning the region from the regret accumulated after the target has been reached.
\par\endgroup

\begingroup\color{black}
\begin{corollary}\label{cor:fixed-tolerance-regret-upper}
Under the assumptions of Theorem~\ref{thm:fixed-tolerance-boundary},
fix $\tau>0$, $N\ge2$, and $\delta\in(0,1)$. Choose a target
phase $H$ with $\eps=\eps_H\ge2\zeta_N$, and suppose the
sufficient evaluation budget in~\eqref{eq:boundary-budget} is at most
$N$. Run the fixed-tolerance specialization through phase $H$. After its completion, use $\widehat x_H\in\argmax_{x\in\X}\ell_H(x)$
for the remaining evaluations and as the final estimated optimal point
$\widehat x_N$. Then
\begin{equation}\label{eq:fixed-tolerance-simple-regret}
 \mathbb E_f r_N\le2\eps+2B\delta,
\end{equation}
and
\begin{equation}\label{eq:fixed-tolerance-cumulative-regret}
\begin{split}
 \mathbb E_f R_N\le{} C\log^{2+d/\nu}(2N)\log^{1+d/(2\nu)}(2N/\delta)\bigl[1+(\tau+\eps)\eps^{-(2+d/\nu)}\bigr]+2N\eps+2BN\delta.
\end{split}
\end{equation}
The constant $C$ depends only on the fixed model parameters and $\vartheta$.
\end{corollary}
\par\endgroup

\begingroup\color{black}
To compare the two corollaries, choose $\eps_H$ of order $\eta$ with $4\eps_H\le\eta$. For fixed $K,\tau,\delta$ in the parameter range of Theorem~\ref{thm:fixed-tolerance-lower}, the necessary cumulative regret has order at least $\tau\eta^{-2-(d-1)/\nu}$, whereas the region-learning term in the general upper bound has order $\widetilde O(\tau\eta^{-2-d/\nu})$. Their exponents of $\eta^{-1}$ differ by $1/\nu$. The lower bound depends on $K$ and $\tau$ through the geometric factor $\bigl(\min\{1,K\tau^{d/\alpha}\}\bigr)^{(d-1)/d}$, whereas this factor is absent from the general upper bound. In addition, the upper bound contains $O(N\eta+BN\delta)$, which accounts for the evaluations at the saved estimated optimal point and for the event $\Econf^c$, whose probability is at most $\delta$.
% Thus the exponent comparison concerns the region-learning terms, rather than the complete cumulative regret bounds.
On the other hand, the simple regret upper bound is $O(\eta+B\delta)$, so the simple regret bound decreases as the target accuracy and error probability decrease.
\par\endgroup 

The budget in Theorem~\ref{thm:fixed-tolerance-boundary} is a general sufficient bound. A smaller budget suffices when the remaining maximizer candidates and the uncertain tolerance boundary have small covering numbers. To quantify this geometry, fix $c_r>0$ and define
\begin{equation}\label{eq:boundary-cover-number}
 J_{\tau,f}(u)=\mathcal N\!\left(\mathcal S_{(8u)+}\cup\{x:|f^\star-f(x)-\tau|\le8u\},c_r u^{1/\nu}\right).
\end{equation}
The first set accounts for estimating $f^\star$, and the second accounts for classifying the chosen boundary. 
% The refined budget~\eqref{eq:boundary-budget-refined} in Appendix~\ref{ec:boundary} uses covers of their union. 
% For simultaneous guarantees over $[0,T]$, the boundary moves through the entire positive-tolerance region, as reflected in Theorem~\ref{thm:region-learning-budget}. 
The following corollary describes the gain available for a single boundary when $J_{\tau,f}$ grows slowly as $u$ goes to zero.

\begin{corollary}\label{cor:boundary-cover-rate}
Fix \textcolor{black}{an error probability} $\delta\in(0,1)$ and a tolerance $\tau>0$. Suppose, in addition to the assumptions of Theorem~\ref{thm:fixed-tolerance-boundary}, that $J_{\tau,f}(u)\le C_Ju^{-\kappa_{\rm cov}}$ for all sufficiently small $u>0$, with fixed $C_J<\infty$ and covering exponent $\kappa_{\rm cov}\ge0$. Then the fixed-tolerance specialization of Algorithm~\ref{alg:continuous-ccucb} identifies the region $\mathcal S_\tau$ with \textcolor{black}{region estimation error} at most $\eta$ using a budget of $\widetilde O(\eta^{-2-\kappa_{\rm cov}})$ evaluations.
\end{corollary}

\begin{example}\label{ex:regular-tolerance-boundary}
In this example, we show how a regular tolerance boundary reduces the spatial covering exponent from $d$ to $d-1$, yielding a smaller sufficient budget for one prespecified region. Let $d\ge2$, $\nu>1$, $\X=\overline B(c,R)$, and $f(x)=-\mu\|x-c\|^2$ for fixed $\mu,R>0$. Fix $0<\tau<\mu R^2$. A smooth compactly supported extension of this quadratic belongs to the \textcolor{black}{Mat\'ern RKHS} for any fixed finite $\nu$. Thus the example satisfies the RKHS assumption whenever the fixed $B$ is at least the norm of its restriction to $\X$. The boundary is the sphere of radius $\sqrt{\tau/\mu}$, where $\|\nabla f\|=2\sqrt{\mu\tau}>0$. This nonvanishing gradient is the regularity that converts value uncertainty into spatial \textcolor{black}{uncertainty of order $u$ around the boundary}. For sufficiently small $u$, \textcolor{black}{all points $x$ of} the set $\{x:|f^\star-f(x)-\tau|\le8u\}$ \textcolor{black}{satisfy $|\|x-c\|-\sqrt{\tau/\mu}|\le C_0u$ with $C_0=8/\sqrt{\mu\tau}$}. Letting $r=c_ru^{1/\nu}$, a maximal $r$-separated \textcolor{black}{subset of this set yields disjoint balls of radius $r/2$ contained in $\{x:|\|x-c\|-\sqrt{\tau/\mu}|\le C_0u+r/2\}$, which} has volume $O(u+r)$. Hence \textcolor{black}{the number of balls of radius $r$ needed to cover $\{x:|f^\star-f(x)-\tau|\le8u\}$} is
\[
 O\!\left(r^{-(d-1)}(1+u/r)\right)=O\!\left(u^{-(d-1)/\nu}\right),
\]
where the equality uses $\nu>1$. Meanwhile $\mathcal S_{(8u)+}$ is a ball of radius $O(\sqrt u)$ and has covering number at most
\[
 C\left(1+u^{-\max\{d/\nu-d/2,0\}}\right) \le C' u^{-(d-1)/\nu}.
\]
The last inequality follows from $1/\nu\le d/2$. Corollary~\ref{cor:boundary-cover-rate} therefore gives the fixed-tolerance sufficient budget $\widetilde O(\eta^{-2-(d-1)/\nu})$.
This has a strictly smaller exponent than the whole-region bound $\widetilde O(\eta^{-2-d/\nu})$, since for the same quadratic and fixed $T=\tau>0$, $\mathcal S_{T+}$ contains a ball of fixed positive radius, whose radius-$r$ covering number is of order $r^{-d}$. At fixed confidence and budget $N$, \textcolor{black}{the bound on region estimation error thus improves} from $\widetilde O(N^{-\nu/(2\nu+d)})$, the general upper bound of Theorem~\ref{thm:fixed-tolerance-boundary}, to $\widetilde O(N^{-\nu/(2\nu+d-1)})$ on this example. 
% The gain comes from resolving a single thin boundary shell, while the shrinking set of possible maximizers does not dominate the covering cost.
\end{example}

% END region_boundary

% BEGIN covering_critical_lower
% General-dimensional critical lower bound; active manuscript source.
% Uses the class and testing/ball-packing argument in covering_minimax.tex.
\subsection{\texorpdfstring{Regret Bounds for $2\le\alpha\le\nu$}{Regret Bounds for 2 <= alpha <= nu}}
\label{subsec:other-covering-regimes}
\label{subsec:restored-critical}
\label{subsec:general-periodic-cover-lower}
\label{subsec:covering-saturation}

In this subsection, we provide the optimization-regret results for $2\le\alpha\le\nu$. All model and covering parameters other than $K$ and $N$ remain fixed. 

\paragraph{Critical endpoint $\alpha=\nu$.}

At the critical endpoint $\alpha=\nu$, the localized upper exponent in~\eqref{eq:optimization-benchmark} is $a=\max\{2,2+d/\nu-d/\alpha\}=2$, so that $\rho_N=\min\{\sqrt{K/N},N^{-1/(2+d/\nu)}\}$. The following result establishes matching rates at this endpoint.

\begin{theorem}
\label{thm:restored-critical-lower}
Fix $\alpha=\nu\ge2$, the normalized Mat\'ern kernel, positive $B,c_\rho,u_0,\sigma$, and a compact domain containing an interior
ball. There are fixed constants $\textcolor{black}{K_{\mathrm{crit}}},\textcolor{black}{N_{\mathrm{crit}}},\textcolor{black}{C_{\mathrm{crit}}},c>0$ such that, for all integers $K\ge \textcolor{black}{K_{\mathrm{crit}}}$ and $N\ge \textcolor{black}{N_{\mathrm{crit}}}$, 
\begin{align*}
 \mathcal R_N^s(\mathcal F_K)&\ge c[\log(2N)]^{-d/(2\nu)}\min\{\sqrt{K/N},N^{-1/(2+d/\nu)}\},\\
 \mathcal R_N^c(\mathcal F_K)&\ge c[\log(2N)]^{-d/(2\nu)}\min\{\sqrt{KN},N^{1-1/(2+d/\nu)}\}.
\end{align*}
The logarithmic factor can be omitted if $K\ge \textcolor{black}{C_{\mathrm{crit}}}[\log(2N)]^{d/\nu}$. If $\nu$ is an even integer, it can be omitted for any $K\ge \textcolor{black}{K_{\mathrm{crit}}}$. All constants are independent of $K,N$. 
\end{theorem}

Together with Theorem~\ref{thm:new-regret}, this proves
\[
 \mathcal R_N^s(\mathcal F_K)=\widetilde\Theta\!\left(\min\{\sqrt{K/N},N^{-1/(2+d/\nu)}\}\right),\mathcal R_N^c(\mathcal F_K)=\widetilde\Theta\!\left(\min\{\sqrt{KN},N^{1-1/(2+d/\nu)}\}\right).
\]
\textcolor{black}{The localized and global bounds are equal} at $K\asymp N^{d/(2\nu+d)}$. The lower bounds also apply to joint procedures on this class, whereas the matching SASORI upper bounds use $T=0$ and $m=0$. 

% END covering_critical_lower

% BEGIN covering_smooth_lower
% General-dimensional lower bounds and sufficient saturation range.
% No one-dimensional special cases and no experiments.
\paragraph{Lower bounds for $2\le\alpha\le\nu$.}

Fix $d\ge1$, $2\le\alpha\le\nu$, a normalized Mat\'ern kernel,
and positive $B,c_\rho,u_0,\sigma$. Let $\X$ be a fixed compact
domain with nonempty interior, and let $\mathcal F_K$ be the covering class of Section~\ref{sec:new-lower-region}. The following proposition also covers $\alpha=\nu$, while Theorem~\ref{thm:restored-critical-lower} gives the sharper lower bound.

\begin{proposition}\label{prop:general-periodic-cover-lower}
There are constants $\textcolor{black}{K_{\mathrm{per}}},\textcolor{black}{N_{\mathrm{per}}},c>0$, independent of $K,N$, such that for all integers $K\ge \textcolor{black}{K_{\mathrm{per}}}$ and $N\ge \textcolor{black}{N_{\mathrm{per}}}$,
\begin{align*}
 \mathcal R_N^s(\mathcal F_K) &\ge c\min\{K^{\alpha/(2\nu+d)}N^{-1/2}, N^{-\nu/(2\nu+d)}\},\\
 \mathcal R_N^c(\mathcal F_K)&\ge c\min\{K^{\alpha/(2\nu+d)}N^{1/2}, N^{(\nu+d)/(2\nu+d)}\}.
\end{align*}
\end{proposition}

When $N\ge C K^{2\alpha/d}$, the first term in each minimum is the smaller one. Proposition~\ref{prop:general-periodic-cover-lower} therefore gives \textcolor{black}{simple regret and cumulative regret} lower bounds of order $K^{\alpha/(2\nu+d)}N^{-1/2}$ and $K^{\alpha/(2\nu+d)}N^{1/2}$, respectively. The two terms in each minimum coincide at $K\asymp N^{d/(2\alpha)}$, and above this scale Proposition~\ref{prop:general-periodic-cover-lower} yields the global Mat\'ern lower rates. However, the next proposition shows that the global lower rates already hold at the smaller scale $K\ge C N^{\nu d/[\alpha(2\nu+d)]}$.

% \paragraph{Sufficient covering number bound for the global rate.}

\begin{proposition}\label{prop:covering-saturation}
Fix $2\le\alpha\le\nu$ and all model and covering parameters as
above. There are fixed $C,c>0$ and $\textcolor{black}{N_{\mathrm{sat}}}$ such that, whenever
$N\ge \textcolor{black}{N_{\mathrm{sat}}}$ and
\begin{equation}\label{eq:covering-saturation-threshold}
 K\ge C N^{d/[\alpha(2+d/\nu)]}=C N^{\nu d/[\alpha(2\nu+d)]},
\end{equation}
one has
\[
 \mathcal R_N^s(\mathcal F_K)\ge cN^{-1/(2+d/\nu)},
 \mathcal R_N^c(\mathcal F_K)\ge cN^{1-1/(2+d/\nu)}.
\]
\end{proposition}

Combining Proposition~\ref{prop:covering-saturation} with the global upper bounds in Theorem~\ref{thm:new-regret} gives
\[
 \mathcal R_N^s(\mathcal F_K)=\widetilde\Theta\!\left(N^{-\nu/(2\nu+d)}\right),\mathcal R_N^c(\mathcal F_K)=\widetilde\Theta\!\left(N^{(\nu+d)/(2\nu+d)}\right)
\]
whenever~\eqref{eq:covering-saturation-threshold} holds. Thus a sufficiently large \textcolor{black}{covering number bound makes the Mat\'ern regret rates without the covering condition} unavoidable. 
% This conclusion holds for every fixed $B>0$; the constants and the required thresholds may depend on $B$. The condition~\eqref{eq:covering-saturation-threshold} is sufficient and is not asserted to be necessary.

\begingroup\color{black}
Recall that Theorem~\ref{thm:new-regret} gives the upper bounds
\[
\begin{aligned}
 \mathcal R_N^s(\mathcal F_K)&=\widetilde O\!\left(\min\{\sqrt{K/N},N^{-\nu/(2\nu+d)}\}\right),\\
 \mathcal R_N^c(\mathcal F_K)&=\widetilde O\!\left(\min\{\sqrt{KN},N^{(\nu+d)/(2\nu+d)}\}\right).
\end{aligned}
\]
For $2\le\alpha<\nu$, their localized and global terms coincide at $K\asymp N^{d/(2\nu+d)}$. In contrast, Proposition~\ref{prop:covering-saturation} establishes matching global lower bounds only once $K\ge C N^{\nu d/[\alpha(2\nu+d)]}$. Because $\nu/\alpha>1$, this sufficient lower-bound threshold has a larger exponent of $N$. 
% Between these two scales, the upper bound is already global, but the available lower bounds do not establish that it is sharp.
\par\endgroup

\begingroup\color{black}
For $2\le\alpha<\nu$ below the upper-bound transition, the localized upper and lower bounds have the same orders $N^{-1/2}$ for simple regret and $N^{1/2}$ for cumulative regret. Their dependence on $K$ differs: the upper prefactor is $K^{1/2}$, whereas Proposition~\ref{prop:general-periodic-cover-lower} gives $K^{\alpha/(2\nu+d)}$. Their ratio is $K^{1/2-\alpha/(2\nu+d)}$, apart from logarithmic factors. For any fixed integer $K\ge K_{\mathrm{per}}$, both localized bounds apply once $N\ge\max\{2,N_{\mathrm{per}},C K^{(2\nu+d)/d}\}$, where $K_{\mathrm{per}},N_{\mathrm{per}}$ are the thresholds in Proposition~\ref{prop:general-periodic-cover-lower} and $C$ is a sufficiently large fixed constant. The ratio is then constant in $N$, so the rates match in their orders in $N$. 
% For growing $K$, the optimal dependence on $K$ and the exact transition to the global rate remain unresolved in this regime.
At $\alpha=\nu$, Theorem~\ref{thm:restored-critical-lower} gives matching rates for $K\ge K_{\mathrm{crit}}$ and $N\ge N_{\mathrm{crit}}$, with the distinct thresholds defined in Theorem~\ref{thm:restored-critical-lower}.
\par\endgroup

% \textcolor{black}{The periodic lower bound uses thresholds $K_{\mathrm{per}},N_{\mathrm{per}}$ from Proposition~\ref{prop:general-periodic-cover-lower}; the critical theorem uses its own thresholds $K_{\mathrm{crit}},N_{\mathrm{crit}}$.}

% END covering_smooth_lower

% Main numerical results; the former study scaffold is archived separately.

\begingroup

\clubpenalty=3000
\widowpenalty=3000

\makeatletter
\newenvironment{SASORINumBlock}[1]{%
  \par\addvspace{.4\baselineskip}%
  \noindent
  \begin{minipage}{\linewidth}%
  \def\@captype{#1}%
}{%
  \end{minipage}%
  \par\smallskip%
}
\makeatother

\section{Numerical Experiments}\label{sec:numerical}

In this section, we first describe the experimental design and evaluation metrics, then present numerical results for Branin-2 and Hartmann-3.

\subsection{Experimental Design}\label{subsec:num-design}

We test three variants of Algorithm~\ref{alg:continuous-ccucb}: SASORI-Opt uses $m=0$, SASORI-Reg uses $m=N$, and SASORI-Joint-Free uses $m=\lfloor\min\{N/2,\sqrt{N}/T\}\rfloor$. Two additional baselines, GP-UCB and Sobol-GP, provide comparisons based on optimization and uniform exploration, respectively. Each method uses $N=3500$ evaluations with $30$ macro-replications. We consider Branin-2 and Hartmann-3, two standard benchmarks for simulation optimization. We set $T=0.1$ for Branin-2 and $T=0.3$ for Hartmann-3, with SASORI multiplier $\beta=2.5$, phase ratio $\vartheta=3$, and GP-UCB multiplier $\kappa=2.5$. Function definitions, kernel parameters, and other implementation details appear in Sections~\ref{ec:num-benchmarks}--\ref{ec:num-implementation}.

Following Remark~\ref{rem:threshold-query}, we query a design with acquisition value above $2\eps_h$ and accept phase completion only after an additional numerical search finds no such design. Section~\ref{ec:num-implementation} gives the implementation details.

Following Algorithm~\ref{alg:continuous-ccucb}, we update the retained bands using the GP confidence bands from completed phases. Let $L,U$ denote the resulting lower and upper functions, and let $b_{\rm num},c_{\rm num}$ denote their maxima. We then compute $I_\tau=\{x:L(x)>c_{\rm num}-\tau\}$ and $O_\tau=\{x:U(x)>b_{\rm num}-\tau\}$ as in~\eqref{eq:ccucb-joint-output}. For each benchmark, we use an independent set $Q$ of $32768$ scrambled Sobol points for evaluation, shared by all methods and macro-replications. The main text reports the original algorithm outputs, while the E-Companion additionally presents region estimates obtained by fitting a single GP to all observations, including those from incomplete phases.

We report three metrics for region identification. First, the empirical region error is $\eta_Q(\tau)=\max_{x\in(O_\tau\setminus I_\tau)\cap Q}|f^\star-f(x)-\tau|$, with value zero when the maximization set is empty. It measures the largest absolute difference between an unclassified evaluation point's performance and the acceptance threshold $f^\star-\tau$. Second, inner recovery is $\operatorname{Rec}_Q(\tau)=|I_\tau\cap\mathcal S_\tau\cap Q|/|\mathcal S_\tau\cap Q|$, reported as a percentage. It measures the fraction of truly acceptable evaluation points included in the inner set. Third, the unclassified percentage is $100|(O_\tau\setminus I_\tau)\cap Q|/|Q|$, measuring the fraction of the evaluation set whose membership remains unresolved. These three metrics describe how closely the unresolved designs lie to the acceptance threshold, how much of the acceptable region is recovered, and how much of the domain remains unclassified, respectively.

We also assess optimization performance using simple regret $r_N$ and cumulative regret $R_N$ from Section~\ref{sec:relaxed-model}. Further reporting details appear in Section~\ref{ec:num-metrics}.

\subsection{Branin-2: Region Identification and Optimization}\label{subsec:num-branin}

We first show three metrics of region identification at $\tau=T=0.1$.
\vspace{-1em}
\begin{SASORINumBlock}{table}
\begingroup\centering\small\singlespacing
\setlength{\abovecaptionskip}{4pt}
\setlength{\belowcaptionskip}{4pt}
\setlength{\tabcolsep}{4pt}
\renewcommand{\arraystretch}{1.12}
\caption{Branin-2 algorithm-output region estimates at $\tau=T=0.1$ (mean $\pm$ standard error).}
\label{tab:num-branin-summary}
\begin{tabular*}{\linewidth}{@{\extracolsep{\fill}}lccc@{}}
\toprule
Method & $\eta_Q(\tau)$ & Recovery (\%) & Unclassified (\%)\\
\midrule
SASORI-Opt & $0.0621\pm0.0011$ & $59.97 \pm 0.64$ & $23.49 \pm 0.17$\\
SASORI-Joint-Free & $0.0671\pm0.0016$ & $\boldsymbol{60.89 \pm 0.50}$ & $23.48 \pm 0.14$\\
SASORI-Reg & $\boldsymbol{0.0518\pm0.0007}$ & $57.25 \pm 0.55$ & $\boldsymbol{20.58 \pm 0.17}$\\
GP-UCB & $0.7024\pm0.0716$ & $20.12 \pm 3.94$ & $62.01 \pm 2.51$\\
Sobol-GP & $0.0661\pm0.0010$ & $53.20 \pm 0.74$ & $22.91 \pm 0.27$\\
\bottomrule
\end{tabular*}
\par\endgroup
\end{SASORINumBlock}

\noindent
Table~\ref{tab:num-branin-summary} shows that all three SASORI variants achieve higher inner recovery than either baseline. Reg attains the smallest empirical region error and unclassified fraction, indicating that fewer designs remain unresolved and that their performance values lie closer to the acceptance threshold. This is consistent with its objective of improving region estimation accuracy. A smaller region error does not necessarily imply higher inner recovery or a smaller unclassified fraction, as these metrics capture different aspects of region identification. Joint-Free achieves the highest mean inner recovery while allocating only a limited portion of its evaluation budget to region learning. This illustrates how its two tracks can identify acceptable alternatives while reserving most evaluations for optimization, consistent with its theoretical objective of preserving the optimization regret rates. Opt also recovers a substantial portion of the acceptable region without dedicated region evaluations, since reducing uncertainty around promising designs improves the confidence bounds used to construct the region estimates.

To illustrate the different allocations, we first visualize the queried designs from one macro-replication.

\begin{SASORINumBlock}{figure}
\begingroup\centering\singlespacing
\setlength{\abovecaptionskip}{4pt}
\setlength{\belowcaptionskip}{4pt}

\begin{minipage}{.90\linewidth}
\centering
\includegraphics[width=.325\linewidth]{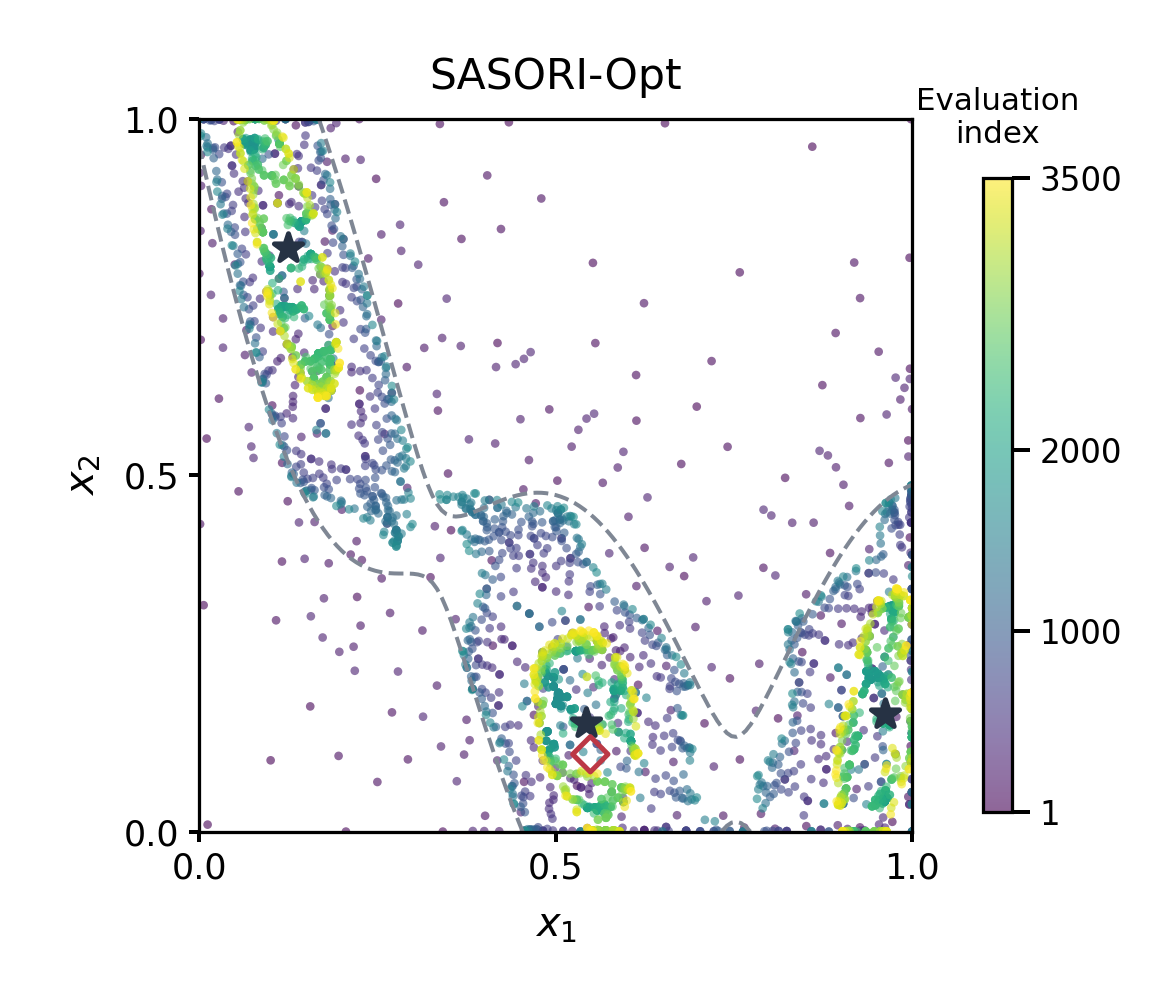}\hfill
\includegraphics[width=.325\linewidth]{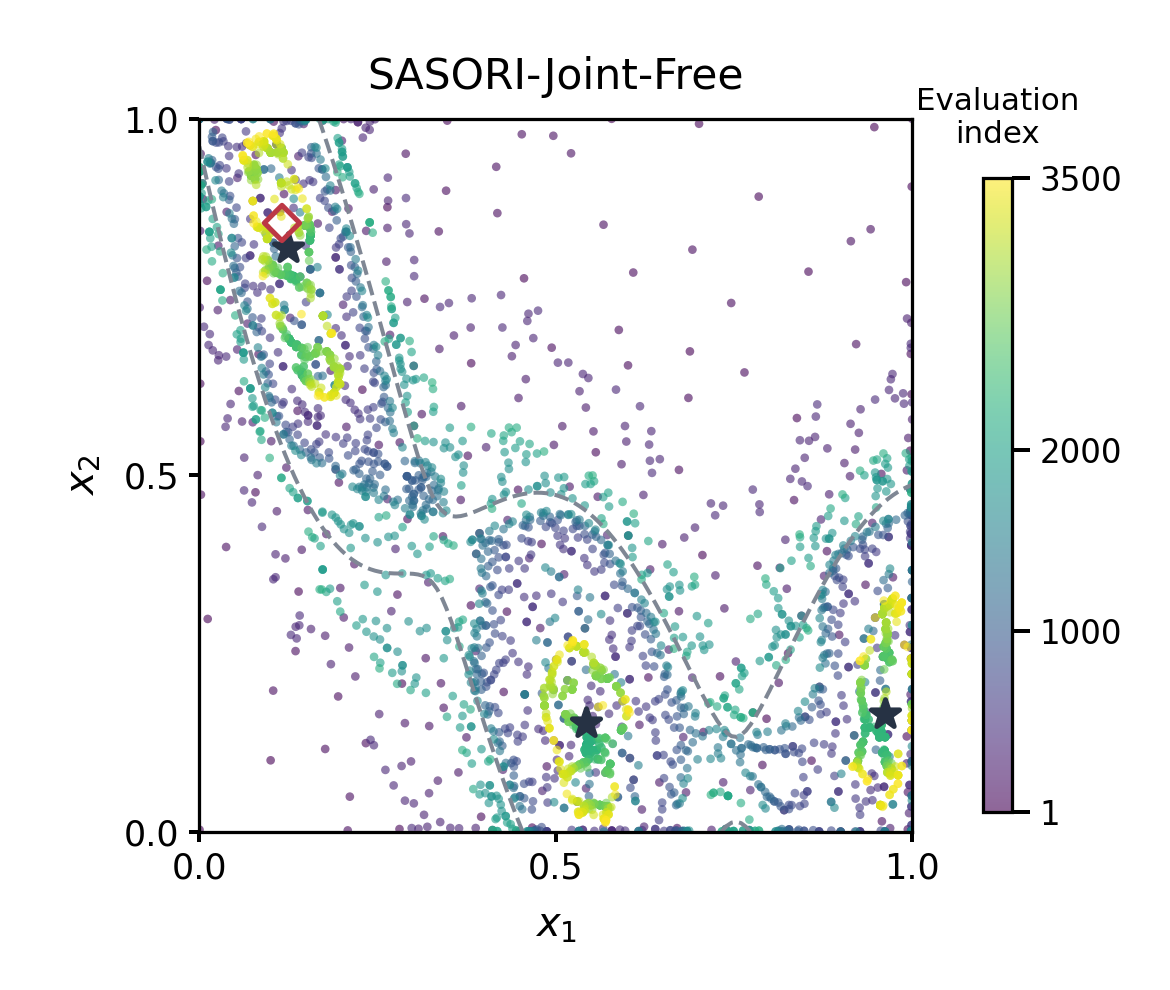}\hfill
\includegraphics[width=.325\linewidth]{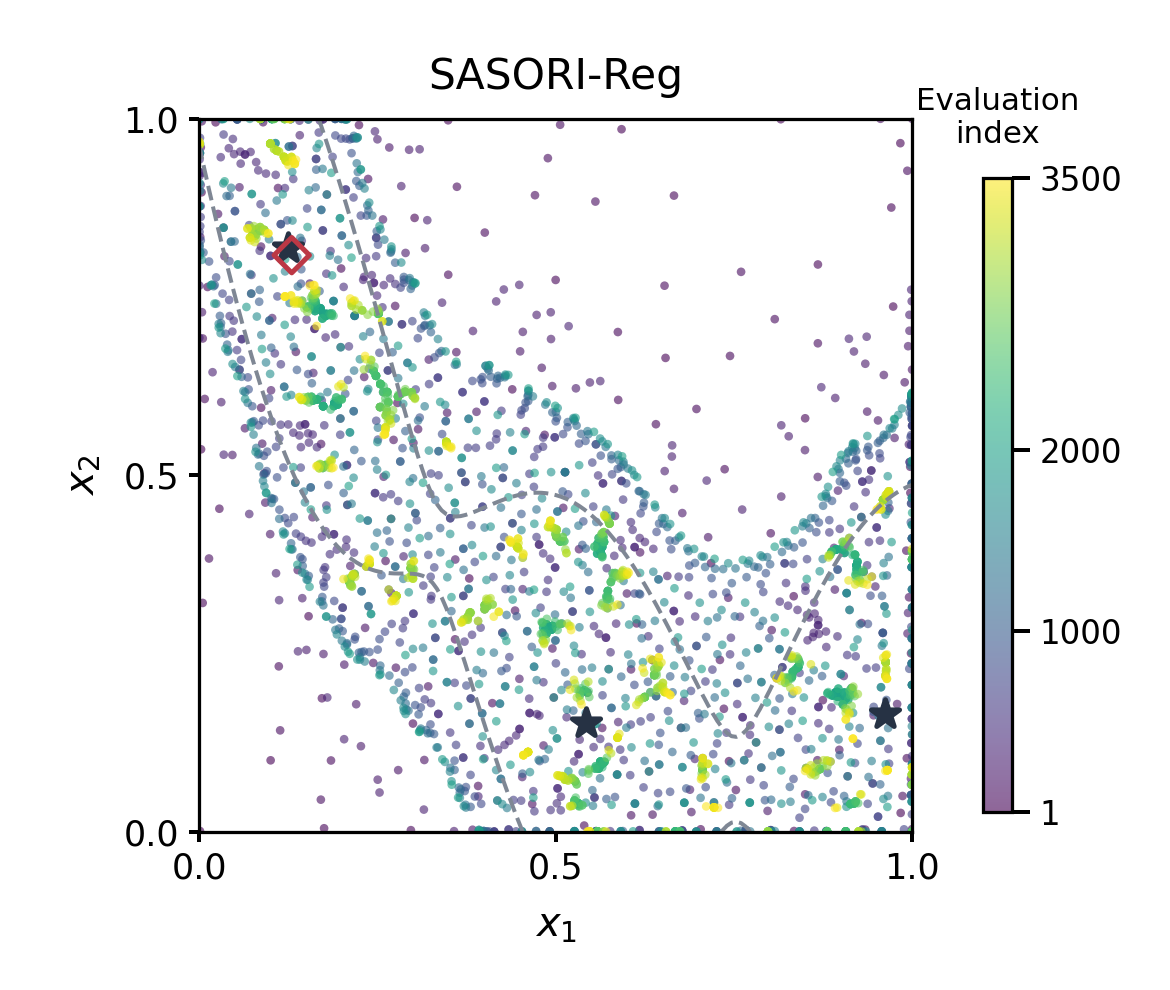}
\end{minipage}

\caption{Branin-2 evaluations for the three SASORI variants from one replication. Point colors indicate evaluation order. Stars and diamonds denote the true maxima and original recommendations, respectively.}
\label{fig:num-branin-sampling}
\par\endgroup
\end{SASORINumBlock}

\noindent
Figure~\ref{fig:num-branin-sampling} illustrates the different sampling behaviors of the three SASORI variants. Opt concentrates many evaluations around the three maxima, but also explores uncertain designs that remain competitive at the current phase accuracy. Joint-Free exhibits an intermediate pattern, combining evaluations near the maxima with additional exploration for region learning. Reg samples more broadly across the acceptable region and its surroundings, consistent with its emphasis on region identification.

We next visualize the corresponding inner and outer region estimates at $\tau=0.1$.

\begin{SASORINumBlock}{figure}
\begingroup\centering\singlespacing
\setlength{\abovecaptionskip}{4pt}
\setlength{\belowcaptionskip}{4pt}

\begin{minipage}{.90\linewidth}
\centering
\includegraphics[width=.325\linewidth]{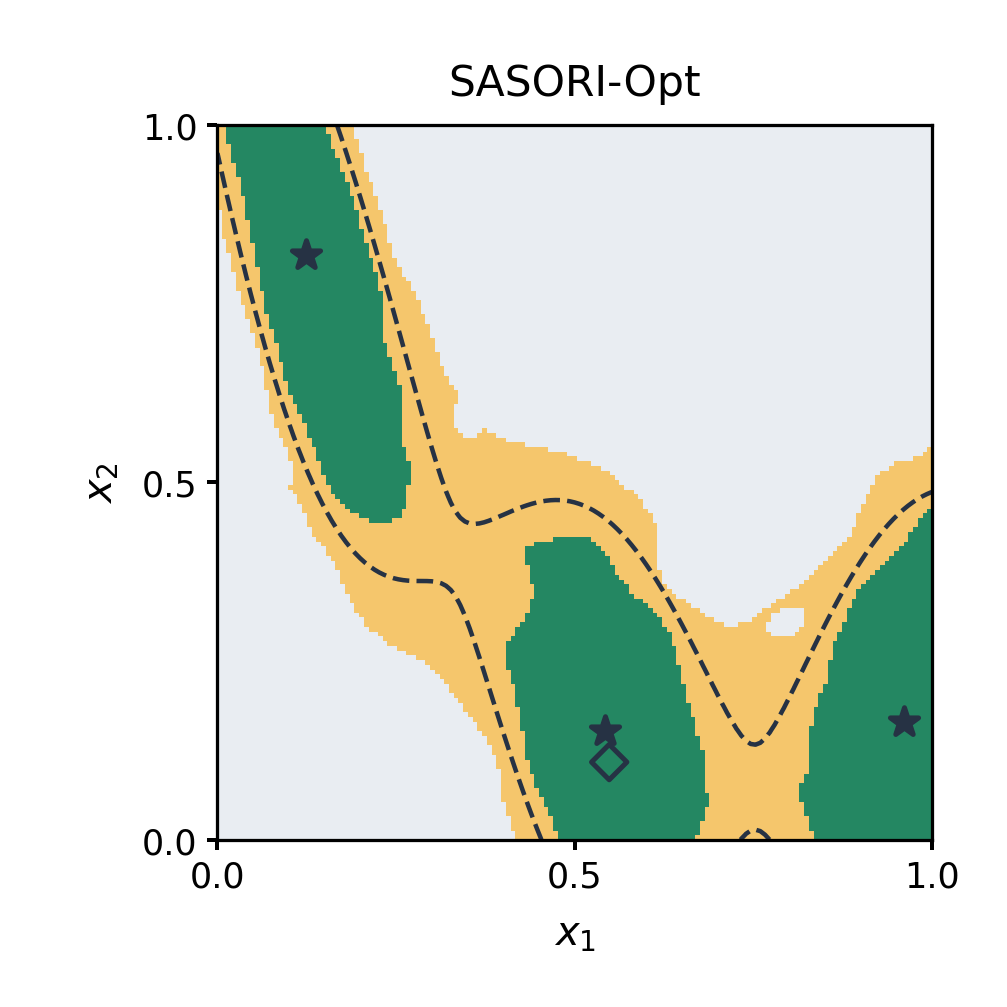}\hfill
\includegraphics[width=.325\linewidth]{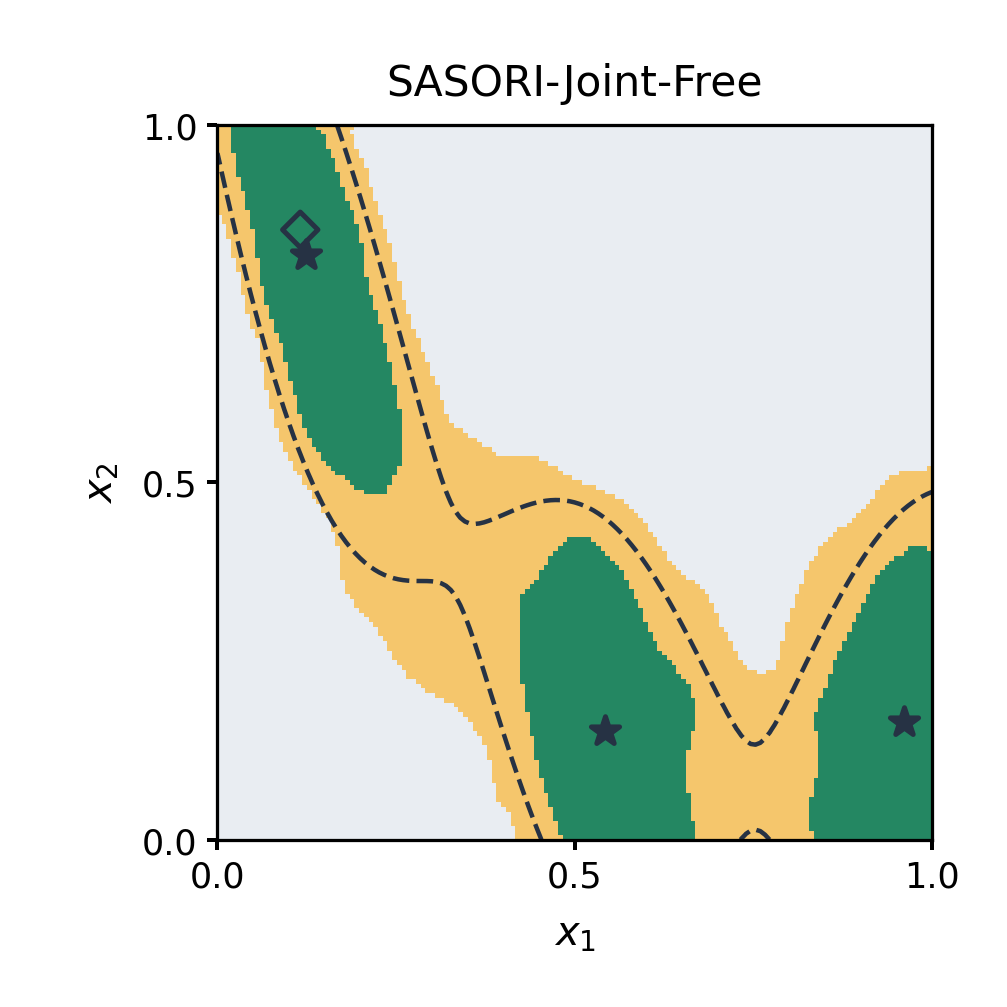}\hfill
\includegraphics[width=.325\linewidth]{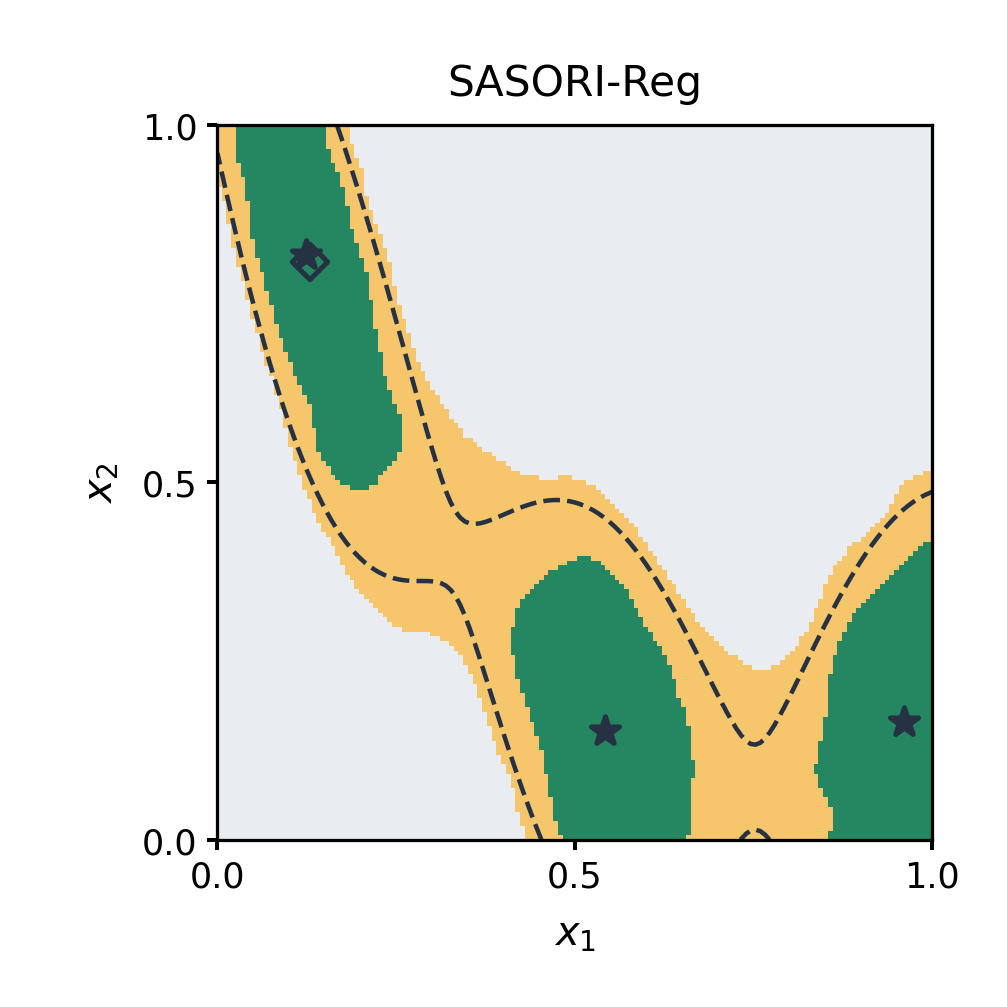}
\par\smallskip
\includegraphics[width=.97\linewidth]{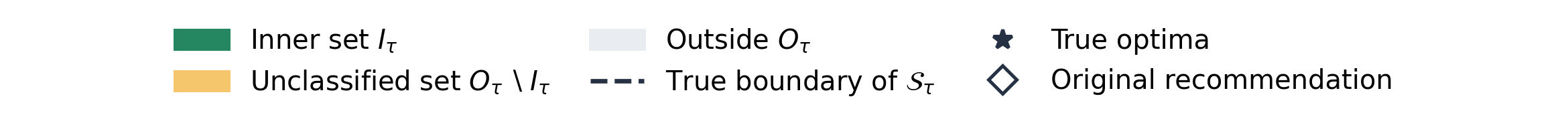}
\end{minipage}

\caption{Branin-2 algorithm regions at $\tau=0.1$ from the same replication.}
\label{fig:num-branin-regions}
\par\endgroup
\end{SASORINumBlock}

\noindent
Figure~\ref{fig:num-branin-regions} shows the resulting differences in region identification. All three variants recover substantial portions of the acceptable regions around the maxima. In particular, Reg produces a narrower and smoother unclassified region around the true boundary, suggesting that its broader exploration helps resolve uncertainty across the region. This observation agrees with its smaller empirical region error and unclassified fraction in this example. Results at other tolerances appear in Section~\ref{ec:num-branin-tolerances}.

We next examine optimization performance through the regret curves.

\begin{SASORINumBlock}{figure}
\begingroup\centering\singlespacing
\setlength{\abovecaptionskip}{4pt}
\setlength{\belowcaptionskip}{4pt}
\includegraphics[width=.9\linewidth]{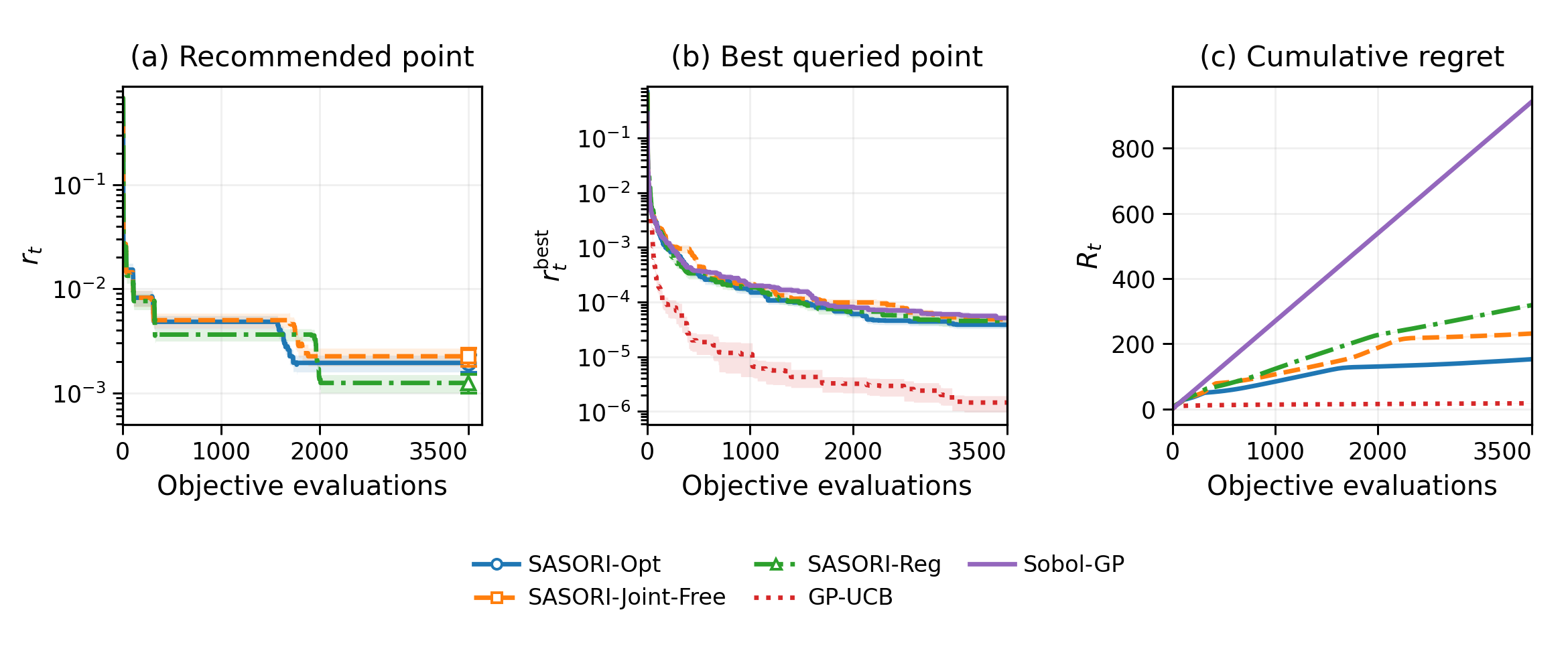}
\caption{Branin-2 simple and cumulative regret. The middle panel uses the best evaluated point; shading indicates one standard error. Recommendation curves show SASORI.}
\label{fig:num-branin-regret}
\par\endgroup
\end{SASORINumBlock}

\noindent
Figure~\ref{fig:num-branin-regret} illustrates how the allocation of evaluations affects optimization performance. Among the SASORI variants, Opt achieves the lowest cumulative regret and slightly lower best query regret, while Joint-Free exhibits intermediate cumulative regret. This is consistent with their respective allocations, as Opt concentrates its evaluations on optimization and Joint-Free reserves a smaller portion for region learning. Interestingly, Reg attains the lowest mean regret at the final recommendation, despite devoting all its evaluations to region learning. One possible explanation is that its broader exploration may also help refine the confidence bounds near the maximizers, thereby improving the final recommendation. Indeed, both tracks use the same lower confidence bound rule to select the recommended point, so region learning can also contribute to recommendation quality. This is reflected in the difference between the best query and recommendation results, where Opt visits slightly better designs but Reg produces a more accurate final recommendation. As we will see in Section~\ref{subsec:num-hartmann}, this ordering is different for Hartmann-3, where Opt achieves the lowest recommendation regret among the SASORI variants. Finally, GP-UCB achieves the lowest best query and cumulative regret overall, but its region estimates are considerably less accurate, as shown in Table~\ref{tab:num-branin-summary}. These results illustrate the distinction between optimization efficiency, recommendation quality, and region identification accuracy.

\subsection{Hartmann-3}\label{subsec:num-hartmann}

We next present the region identification and optimization results for Hartmann-3.
\vspace{-1em}
\begin{SASORINumBlock}{table}
\begingroup\centering\small\singlespacing
\setlength{\abovecaptionskip}{4pt}
\setlength{\belowcaptionskip}{4pt}
\setlength{\tabcolsep}{4pt}
\renewcommand{\arraystretch}{1.12}
\caption{Hartmann-3 algorithm-output region estimates at $\tau=T=0.3$ (mean $\pm$ standard error).}
\label{tab:num-hartmann-summary}
\begin{tabular*}{\linewidth}{@{\extracolsep{\fill}}lccc@{}}
\toprule
Method & $\eta_Q(\tau)$ & Recovery (\%) & Unclassified (\%)\\
\midrule
SASORI-Opt & $0.1672\pm0.0023$ & $55.59 \pm 0.44$ & $9.11 \pm 0.05$\\
SASORI-Joint-Free & $0.1747\pm0.0028$ & $\boldsymbol{55.61 \pm 0.54}$ & $9.19 \pm 0.06$\\
SASORI-Reg & $\boldsymbol{0.1206\pm0.0020}$ & $53.66 \pm 0.65$ & $\boldsymbol{6.79 \pm 0.06}$\\
GP-UCB & $0.665667\pm0.000004$ & $12.53 \pm 1.36$ & $81.53 \pm 0.57$\\
Sobol-GP & $0.2036\pm0.0044$ & $37.75 \pm 1.14$ & $9.60 \pm 0.12$\\
\bottomrule
\end{tabular*}
\par\endgroup
\end{SASORINumBlock}

\noindent
Table~\ref{tab:num-hartmann-summary} shows that the regional advantages of SASORI extend to three dimensions, with all three variants achieving higher inner recovery and smaller empirical region error than either baseline. As in Branin-2, Reg attains the smallest empirical region error and unclassified fraction, consistent with its emphasis on region estimation accuracy. Opt and Joint-Free achieve similar mean inner recovery, while Joint-Free reserves only $197$ evaluations for region learning, leaving most of its budget for optimization. Results at other tolerances appear in Section~\ref{ec:num-hartmann-tolerances}.

We finally compare the regret curves for Hartmann-3.
\begin{SASORINumBlock}{figure}
\begingroup\centering\singlespacing
\setlength{\abovecaptionskip}{4pt}
\setlength{\belowcaptionskip}{4pt}
\includegraphics[width=.9\linewidth]{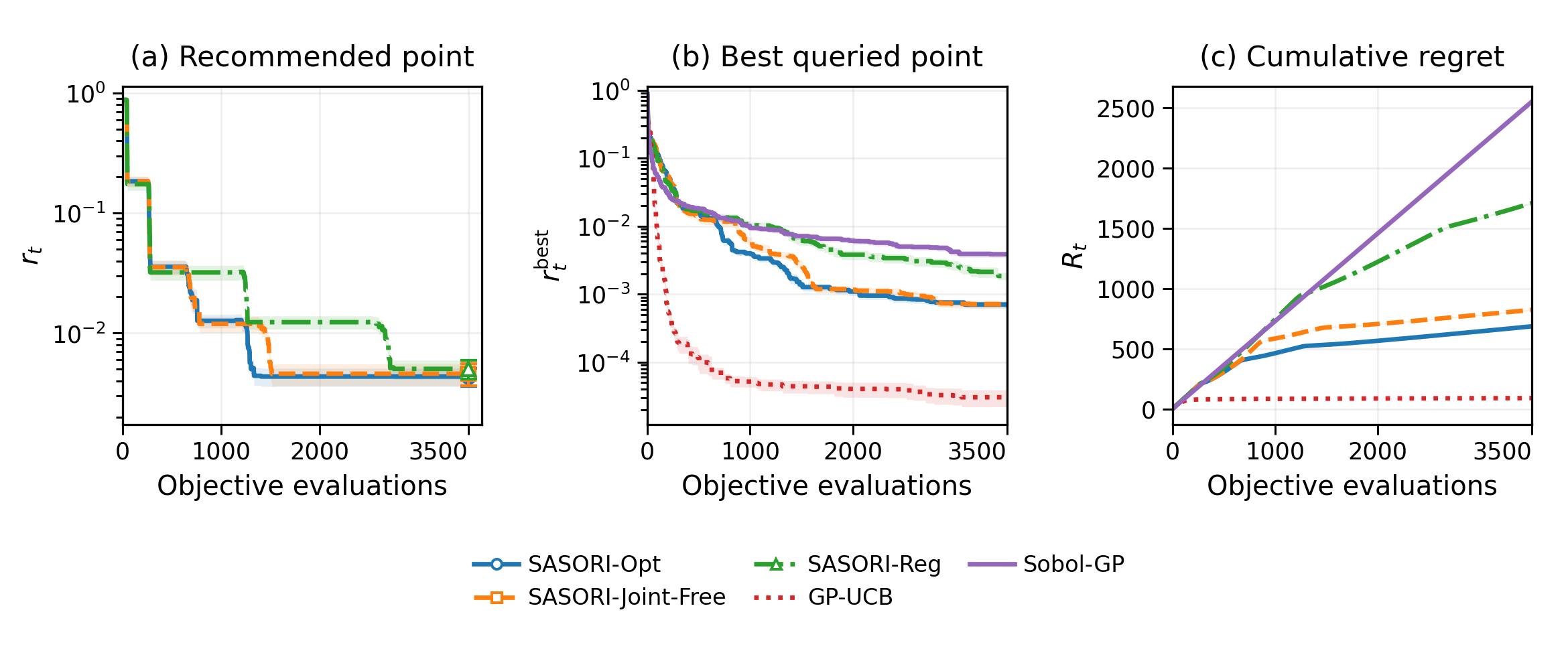}
\caption{Hartmann-3 simple and cumulative regret. The middle panel uses the best evaluated point; shading indicates one standard error. Recommendation curves show SASORI.}
\label{fig:num-hartmann-regret}
\par\endgroup
\end{SASORINumBlock}

\noindent
Figure~\ref{fig:num-hartmann-regret} shows that Opt achieves the lowest recommendation, best query, and cumulative regret among the SASORI variants. Unlike Branin-2, where Reg produces the most accurate final recommendation, Opt also achieves the lowest recommendation regret on Hartmann-3. This illustrates that the relative recommendation quality can vary across benchmarks, while Opt consistently incurs the lowest cumulative regret among the three variants. Joint-Free maintains recommendation quality close to Opt and incurs substantially less cumulative regret than Reg, while allocating part of its budget to region learning. Reg incurs greater cumulative regret, accompanied by the smallest empirical region error and unclassified fraction reported in Table~\ref{tab:num-hartmann-summary}. GP-UCB again achieves the lowest best query and cumulative regret overall, while leaving considerably greater uncertainty in its region estimates. Together with the Branin-2 results, these observations illustrate how different evaluation allocations affect optimization performance and region identification accuracy.

\endgroup

\section{Conclusions and Discussion}\label{sec:conclusions}

This paper studies the identification of near-optimal regions in continuous simulation optimization, where a decision maker needs the set of designs within a tolerance of the optimum and want to choose the tolerance after seeing the results. We show that region identification is not a by-product of optimization. Specifically, when $\alpha>\nu$, the minimax number of simulation evaluations needed to identify the acceptable regions for all tolerances in $[0,T]$ with region estimation error at most $\eta$ is $\widetilde\Theta(\eta^{-(2+d/\nu)}\min\{1,KT^{d/\alpha}\})$, and the lower bound holds even when the optimal design and its value are known in advance. SASORI achieves this rate without knowing the covering parameters, delivers the estimated optimal point and the regions from the same evaluations, and achieves the optimal tradeoff between region estimation error and cumulative regret. Minimax simple regret is compatible with the fastest region identification rate, whereas minimax cumulative regret is not. At zero tolerance, SASORI is a simulation optimization procedure with minimax simple and cumulative regret on the covering class. For practice, only the largest tolerance $T$ must be fixed before sampling, and the tolerance used in the decision can be chosen afterwards with the same probability guarantee.

Several directions remain open. The current guarantees assume that the smoothness class, its norm bound, and the noise scale are known. Future work could develop methods for calibrating these quantities from pilot evaluations and quantifying the resulting effect on coverage. A variance-adaptive version that estimates the simulation variance from replications at each design point would address the heteroscedasticity common in queueing and inventory models, and exploiting common random numbers could further reduce the evaluation cost. On the theoretical side, matching bounds for a single prespecified tolerance under general boundary geometry, and the sharp dependence on the covering number bound when $2\le\alpha<\nu$, remain to be established.

\fancyhead[L]{\footnotesize Simulation Optimization and Near-Optimal Regions}
\bibliographystyle{abbrvnat}
\bibliography{ref}

\clearpage
\appendix
\singlespacing
% Standard appendix numbering, with continuous manuscript page numbers.
\numberwithin{equation}{section}
\numberwithin{figure}{section}
\numberwithin{table}{section}
% Keep appendix hyperlink destinations distinct from the main text.
\def\theHsection{appendix.\Alph{section}}
\def\theHequation{\theHsection.\arabic{equation}}
\def\theHfigure{\theHsection.\arabic{figure}}
\def\theHtable{\theHsection.\arabic{table}}
\def\theHtheorem{\theHsection.\arabic{theorem}}
\def\theHlemma{\theHsection.\arabic{lemma}}
\def\theHproposition{\theHsection.\arabic{proposition}}
\def\theHcorollary{\theHsection.\arabic{corollary}}
\def\theHassumption{\theHsection.\arabic{assumption}}
\def\theHdefinition{\theHsection.\arabic{definition}}
\def\theHremark{\theHsection.\arabic{remark}}
\def\theHexample{\theHsection.\arabic{example}}
\fancyhead[L]{\footnotesize Appendix: Identifying Near-Optimal Regions}

\section*{Appendices}
The appendices first present additional confidence bounds and algorithmic guarantees (Appendix~\ref{ec:confidence-algorithm}), then give the complete proofs (Appendices~\ref{ec:proofs} and~\ref{ec:lemma-proofs}), and finally provide additional numerical details and results (Appendix~\ref{ec:numerical}).
% Completed editorial note, retained for provenance only:
% \ww{To Yilin: add additional numerical experiments results if needed.}

\section{Confidence Bounds and Algorithmic Guarantees}\label{ec:confidence-algorithm}

\subsection{Confidence Bounds for Near-Optimal Regions}\label{ec:region-envelopes}
The region argument in this subsection does not require a Mat\'ern kernel, a mode count, or Euclidean geometry. Let $\X$ be a compact metric space and let $f:\X\to\R$ be continuous. Proposition~\ref{thm:general-region-certificate} below is deterministic. It only requires pointwise bounds $L(x)\le f(x)\le U(x)$ for all $x\in\X$ and bounds $\lo\le f^\star\le\hi$ on the optimal value. In SASORI, after a completed phase $h$, these bounds are given by the shared confidence bounds $L=\ell_h$ and $U=u_h$ and by their maxima $\lo=\lo_h$ and $\hi=\hi_h$ in~\eqref{eq:new-elimination}. They hold on the shared confidence event $\Econf$ in~\eqref{eq:shared-confidence-event}, on which all confidence bands computed by either track contain $f$, so that $\ell_h\le f\le u_h$ and hence $\lo_h=\max_\X\ell_h\le f^\star\le\max_\X u_h=\hi_h$. By Lemma~\ref{lem:new-confidence}, $\mathbb P(\Econf)\ge1-\delta$. Therefore, on $\Econf$, Proposition~\ref{thm:general-region-certificate} applies simultaneously to all completed phases and to the final output.

\begin{proposition}[Simultaneous region bounds with separate error budgets]
\label{thm:general-region-certificate}
Suppose $L(x)\le f(x)\le U(x)$ on $\X$ and
$\lo\le f^\star\le\hi$, where $\Gamma=\hi-\lo\ge0$. Define, for any $\tau\ge0$,
\begin{equation}\label{eq:general-region-sets}
 I_\tau=\{x:L(x)>\hi-\tau\},
 O_\tau=\{x:U(x)>\lo-\tau\}.
\end{equation}
Then, simultaneously for all $\tau\ge0$,
\begin{equation}\label{eq:general-region-inclusions}
 I_\tau\subseteq\mathcal S_\tau\subseteq O_\tau.
\end{equation}
If $U(x)-L(x)\le w_\tau$ on $O_\tau$ for a nonnegative bound $w_\tau$, let $\eta_\tau=\Gamma+w_\tau$.
Then
\begin{equation}\label{eq:general-region-ambiguity}
 O_\tau\setminus I_\tau
 \subseteq\{x:\tau-\eta_\tau\le f^\star-f(x)<\tau+\eta_\tau\}.
\end{equation}
More precisely, at an unclassified design point of width $w(x)=U(x)-L(x)$,
\begin{equation}\label{eq:general-region-pointwise}
 \tau-(\hi-f^\star)-w(x)\le f^\star-f(x)
       <\tau+(f^\star-\lo)+w(x).
\end{equation}
With $\mathcal S_v=\varnothing$ for $v\le0$, one also has
\begin{equation}\label{eq:general-region-value-sandwich}
 \mathcal S_{\tau-\eta_\tau}\subseteq I_\tau
 \subseteq\mathcal S_\tau\subseteq O_\tau
 \subseteq\mathcal S_{\tau+\eta_\tau}.
\end{equation}
The optimization uncertainty $\Gamma$ and local function uncertainty
$w_\tau$ need not be equal. If $w_T$ bounds the width on $O_T$,
the same error $\Gamma+w_T$ works simultaneously for all $0\le\tau\le T$.
\end{proposition}

\subsection{Phase Guarantees of SASORI}\label{ec:sasori-analysis}\label{ec:shared-bands}\label{ec:cap-transfer}
This subsection discusses two guarantees for the phases of Algorithm~\ref{alg:continuous-ccucb} that are used in the analysis of Section~\ref{sec:new-regret}. The first, Proposition~\ref{lem:shared-invariant} in Section~\ref{sec:new-algorithm}, shows that the accuracy guarantees established by a track remain valid after any later update of the shared confidence bounds, and the second gives the guarantees of a single phase of either track. Their proofs are given in Sections~\ref{ec:proof-shared-bands} and~\ref{ec:proof-phase-accuracy}.

The following proposition provides the phase guarantees used by Theorem~\ref{thm:region-learning-budget} and the subsequent regret analysis. It applies to either track, including when both tracks share confidence bounds.

\begin{proposition}\label{prop:phase-accuracy}
Let $\X$ be a nonempty compact subset of $\R^d$, and suppose Assumptions~\ref{as:new-noise}--\ref{as:new-rkhs} hold. Consider a track of Algorithm~\ref{alg:continuous-ccucb}, letting $T\ge0$ for its tolerance and $h$ for its local phase index. Use the event $\Econf$ in~\eqref{eq:shared-confidence-event}, whose probability is at least $1-\delta$ by Lemma~\ref{lem:new-confidence}.
On $\Econf$, any completed phase with $\eps_h\ge2\zeta_N$ satisfies
\[
 u_h(x)-\ell_h(x)\le2\eps_h
 \text{ whenever }u_h(x)>\lo_h-T-2\eps_h,
 \hi_h-\lo_h\le2\eps_h,
 f^\star-f(\widehat x_h)\le2\eps_h.
\]
Its region outputs satisfy, simultaneously for every $0\le\tau\le T$,
\begin{equation}\label{eq:cap-driven-region-accuracy}
\begin{split}
 I_{h,\tau}\subseteq\mathcal S_\tau\subseteq O_{h,\tau},
 O_{h,\tau}\setminus I_{h,\tau}
 \subseteq\{x:|f^\star-f(x)-\tau|\le4\eps_h\}.
\end{split}
\end{equation}
At $\tau=0$, both output sets are empty by convention.
Moreover, any point sampled in a phase with $\eps_h\ge2\zeta_N$, including a phase truncated by its track's budget, satisfies
\begin{equation}\label{eq:region-active-quality}
 x_{h,j}\in\mathcal S_{(T+L_\vartheta\eps_h)+}.
\end{equation}
\end{proposition}

Proposition~\ref{lem:shared-invariant} explains why the two tracks of SASORI can share their confidence bands without interfering with each other. After track $j$ completes a phase with accuracy level $\bar\eps$, its guarantees consist of a bound $2\bar\eps$ on the width $u^{\rm old}-\ell^{\rm old}$ over the designs that are still relevant at its tolerance $T_j$, namely those with $u^{\rm old}(x)>\lo^{\rm old}-T_j-2\bar\eps$, and a bound $2\bar\eps$ on the optimal-value error $f^\star-\lo^{\rm old}$. Any later band, whether it is provided by the same track or by the other track, is intersected with the current confidence bounds, so the upper confidence bound can only decrease, while the lower confidence bound and its maximum can only increase. Hence the set of relevant designs can only shrink, the widths on this set can only decrease, and $\lo$ can only move closer to $f^\star$. In particular, evaluations spent by the region track never invalidate the guarantee of the optimization track, and evaluations spent by the optimization track never invalidate an earlier region guarantee. Because all bands are valid on the single event $\Econf$, this sharing requires neither a separate data set nor an additional union bound.

Proposition~\ref{prop:phase-accuracy} gives the two properties of a single phase on which the complexity and regret analyses rely. The first concerns accuracy. When a phase with accuracy level $\eps_h$ completes, the stopping rule~\eqref{eq:new-phase-stop} forces the width $u_h-\ell_h$ to be at most $2\eps_h$ at any design that may still matter at tolerance $T$. Applied at a maximizer of $u_h$, this gives $\hi_h-\lo_h\le2\eps_h$, so the estimated optimal point has optimality gap at most $2\eps_h$. Combined with Proposition~\ref{thm:general-region-certificate}, it gives the region guarantee with region estimation error $4\eps_h$ simultaneously for all $\tau\in[0,T]$. The second concerns where the evaluations are spent. By~\eqref{eq:region-active-quality}, any point sampled in phase $h$, including one sampled in a phase cut off by the budget, has performance gap at most $T+L_\vartheta\eps_h$. This has two consequences. First, each evaluation contributes at most $T+L_\vartheta\eps_h$ to the cumulative regret, which yields the regret bounds in Theorem~\ref{thm:budgeted-ccucb} and Proposition~\ref{prop:full-budget-global-regret}. Second, under Assumption~\ref{as:new-cover} and when $T+L_\vartheta\eps_h\le u_0$, the points sampled in phase $h$ lie in a set covered by $K$ balls of radius $c_\rho(T+L_\vartheta\eps_h)^{1/\alpha}$, which yields the localized evaluation bound~\eqref{eq:region-local-cost}. For the optimization track ($T=0$), the sampled points therefore concentrate near the maximizers as $\eps_h$ decreases. For the region track ($T>0$), they may remain at designs whose performance gap is of order $T$, which is the source of the regret cost of region learning quantified in Section~\ref{subsec:joint-tradeoff}.

\section{Proofs of the Main Results}\label{ec:proofs}

This section proves the theorems, propositions, and corollaries stated in the main text and in Section~\ref{ec:confidence-algorithm}. The proofs of all lemmas used in this section are provided in Section~\ref{ec:lemma-proofs}.

We use the following notation throughout Sections~\ref{ec:proofs} and~\ref{ec:lemma-proofs}. With $s=\nu+d/2$ as in Section~\ref{subsec:kernel-smoothness}, let $p=2+d/\nu$ and $a=\max\{2,2+d/\nu-d/\alpha\}$, so that $\rho_N=\min\{(K/N)^{1/a},N^{-1/p}\}$ in~\eqref{eq:optimization-benchmark}. For the logarithmic factors, let
\begin{equation}\label{eq:ec-log-factors}
\Lambda_{N,\delta}=\log^{2+d/\nu}(2N)\log^{1+d/(2\nu)}(2N/\delta),
L_N=\log^{3+3d/(2\nu)}(2N),
\mathcal V_N=\beta_N^2\log^2(2N),
\end{equation}
where $\beta_N$ is calibrated at error probability $\delta$ in~\eqref{eq:new-confidence-parameters}. By the definition of $\beta_N$ and $M_N\le(1+2D_{\X})^dN^{4d/\theta}$, we have $\beta_N^2\le2B^2+(4\sigma^2/\lambda)\log(4M_NN^2/\delta)\le C\log(2N/\delta)$. Since also $\beta_N\ge B$, $N\ge2$, and $\log(2N^3)\le3\log(2N)$,
\begin{equation}\label{eq:ec-log-bounds}
B^2\log^24\le\mathcal V_N\le C\mathcal V_N^{1+d/(2\nu)}\le C\Lambda_{N,\delta},
1\le\Lambda_{N,\delta},
\Lambda_{N,N^{-2}}\le3^{1+d/(2\nu)}L_N.
\end{equation}
For $\varphi\in H^s(\mathbb R^d)$, $z\in\mathbb R^d$, and $t\ge1$, the Fourier representation of the $H^s(\mathbb R^d)$ norm gives the scaling inequality
\begin{equation}\label{eq:ec-scaling}
\|\varphi(t(\cdot-z))\|_{H^s(\mathbb R^d)}\le t^{s-d/2}\|\varphi\|_{H^s(\mathbb R^d)}=t^\nu\|\varphi\|_{H^s(\mathbb R^d)}.
\end{equation}
Moreover, the Mat\'ern RKHS on $\mathbb R^d$ coincides with $H^s(\mathbb R^d)$ with an equivalent norm, and the RKHS of the restriction of $k$ to $\X\times\X$ consists of the restrictions to $\X$ of the functions in the RKHS on $\mathbb R^d$, where the RKHS norm on $\X$ of a function is the smallest RKHS norm on $\mathbb R^d$ among its extensions \citep[Chapter~10]{wendland2004scattered}. Hence, for any $\X\subseteq\mathbb R^d$ and without any regularity condition on its boundary, the RKHS norm on $\X$ of a function is at most a fixed constant times the $H^s(\mathbb R^d)$ norm of any extension of it to $\mathbb R^d$. Unless stated otherwise, $C$ denotes a positive constant depending only on the fixed parameters, whose value may change between occurrences.
% , each in its own subsection. Following the order of the paper, Subsections~\ref{ec:region-lower}--\ref{ec:proof-joint-frontier} prove the results of the main text, and Subsections~\ref{ec:proof-separated-optima}--\ref{ec:proof-full-budget} prove the additional results presented before this section.

% Each subsection first presents the lemmas used in the proof and then gives the proof; a lemma already presented in an earlier subsection is cited rather than restated. 

\subsection{Proof of Theorem~\ref{thm:region-minimax-lower}}\label{ec:region-lower}
We first present a lemma used in this proof, which is also used in the proof of Theorem~\ref{thm:covering-lower-alpha-large}, and its proof is given in Section~\ref{ec:proof-localized-power}.

\begin{lemma}\label{lem:localized-power-native}
If $\alpha>\nu$ and $\varphi\in C_c^\infty(\mathbb R^d)$, then the function $x\mapsto\varphi(x)\|x\|^\alpha$ belongs to $H^{\nu+d/2}(\mathbb R^d)$. Consequently, for any fixed $c\in\mathbb R^d$, the restriction of the function $x\mapsto\|x-c\|^\alpha$ to a fixed bounded domain belongs to the Mat\'ern RKHS and has finite RKHS norm.
\end{lemma}

\begin{proof}[Proof of Theorem~\ref{thm:region-minimax-lower}]
% \noindent\textbf{A common baseline with covering slack.}
Choose $x^\star\in\X$ and $r_*>0$ such that $\overline B(x^\star,2r_*)\subset\operatorname{int}(\X)$. By Lemma~\ref{lem:localized-power-native}, $C_g=\bigl\|\|\cdot-x^\star\|^\alpha\bigr\|_{\mathcal H_k(\X)}<\infty$.
Set $\mu=2(4/c_\rho)^\alpha,\kappa_*=\mu^{d/\alpha}r_*^d, k_T=\min\{K,\lfloor\kappa_*T^{-d/\alpha}\rfloor\}$.
Choose the fixed $K_{\mathrm{reg}}$ large enough that
$\mu C_gK_{\mathrm{reg}}^{-\alpha/d}\le B/2$. Choose $T_{\mathrm{reg}}$ so that $T_{\mathrm{reg}}\le\min\{1,u_0/2,2048B\}$ and $\kappa_*T_{\mathrm{reg}}^{-d/\alpha}\ge2K_{\mathrm{reg}}$. This proof does not use the constraint $T_{\mathrm{reg}}\le\min\{1,u_0/2,2048B\}$; it is imposed so that the same $T_{\mathrm{reg}}$ can be used in the upper bounds of Corollaries~\ref{cor:region-minimax-matching} and~\ref{cor:joint-frontier}. Then $k_T\ge K_{\mathrm{reg}}$ for all
parameters in Theorem~\ref{thm:region-minimax-lower}. Define $g(x)=-\mu {k_T}^{-\alpha/d}\|x-x^\star\|^\alpha$.
The baseline has RKHS norm at most $B/2$, unique global maximum $g(x^\star)=0$, and a norm bound independent of $K,T$.
Letting $R=(T/\mu)^{1/\alpha}{k_T}^{1/d}$,
we have $R\le r_*$ and
\begin{equation}\label{eq:region-effective-volume}
 R^d\asymp\min\{1,K T^{d/\alpha}\}.
\end{equation}
The constants in this comparison are fixed. 
% The floor is at least half its argument, so $kT^{d/\alpha}\asymp\min\{K T^{d/\alpha},\kappa_*\}$.

% \noindent\textbf{Many suboptimal perturbations.}

Let $\psi\in C_c^\infty(B(0,1))$ satisfy
$0\le\psi\le1$ and $\psi(0)=1$. Fourier scaling gives a fixed
$C_\psi$ such that $\|\psi((\cdot-z)/h)\|_{\mathcal H_k(\X)} \le C_\psi h^{-\nu}$ for $0<h\le1$
uniformly for the centers considered below. Set $L=(16C_\psi/B)^{1/\nu}$ and $h=L\eta^{1/\nu}$.

Let $\mathcal A_R=\{x:4^{-1/\alpha}R\le\|x-x^\star\|\le2^{-1/\alpha}R\}$. Choose a fixed $\rho_0>0$ small enough that any point within distance $\rho_0R$ of $\mathcal A_R$ has baseline gap in $[T/8,3T/4]$ and lies in $B(x^\star,R)$. Such a choice exists because the baseline gap is $-g(x)=T(\|x-x^\star\|/R)^\alpha$, which lies in $[T/8,3T/4]$ exactly when $8^{-1/\alpha}R\le\|x-x^\star\|\le(3/4)^{1/\alpha}R$, and $8^{-1/\alpha}<4^{-1/\alpha}<2^{-1/\alpha}<(3/4)^{1/\alpha}<1$.
Since $\alpha>\nu$, decreasing the fixed $T_{\mathrm{reg}}$ further ensures
\begin{equation}\label{eq:region-lower-h}
 \frac hR
 =L\mu^{1/\alpha}{k_T}^{-1/d}
             \eta^{1/\nu}T^{-1/\alpha}
 \le L\mu^{1/\alpha}K_{\mathrm{reg}}^{-1/d}
          128^{-1/\nu}T^{1/\nu-1/\alpha}
 \le\rho_0,
 h\le1.
\end{equation}

Take a maximal $2h$-separated set of centers $z_i$ in
$\mathcal A_R$. The support balls $A_i=B(z_i,h)$ are disjoint. By maximality, any point of $\mathcal A_R$ lies within distance $2h$ of some center, so the balls $B(z_i,2h)$ cover $\mathcal A_R$. Since the volume of $\mathcal A_R$ is a constant multiple of $R^d$, comparing volumes and using~\eqref{eq:region-effective-volume} gives
\begin{equation}\label{eq:region-perturbation-count}
 M\ge c(R/h)^d
   \ge c\eta^{-d/\nu}\min\{1,K T^{d/\alpha}\}.
\end{equation}
The fixed $\rho_0$ can also be decreased to ensure $M\ge2$. At each center, $(-g(z_i))\in[T/4,T/2]$, and throughout each support, $(-g(x))\in[T/8,3T/4]$.

Define the alternatives $f_i(x)=g(x)+8\eta\psi((x-z_i)/h)$.
Their RKHS norms are at most
$B/2+8\eta C_\psi h^{-\nu}=B$.
Because $8\eta\le T/16$ and $-g\ge T/8$ on the
supports, each perturbation stays strictly below zero. Any
$f_i$ therefore keeps the same unique global maximizer $x^\star$
and the same value $f_i^\star=0$. In particular,
\begin{equation}\label{eq:region-gap-comparison}
 \tfrac12(-g(x))\le\bigl(f_i^\star-f_i(x)\bigr)
   =(-g(x))-8\eta\psi((x-z_i)/h)
   \le(-g(x))\text{ for }x\in\X.
\end{equation}

For the baseline and any alternative,
\eqref{eq:region-gap-comparison} implies
\[
 \mathcal S_{u+}\subseteq
 \overline B\left(x^\star,(2u/\mu)^{1/\alpha}{k_T}^{1/d}\right)
 =\overline B\left(x^\star,\tfrac14c_\rho {k_T}^{1/d}u^{1/\alpha}\right),
\]
for any $u>0$. This ball is coverable by at most $k_T$ balls of radius $c_\rho u^{1/\alpha}$. For $k_T<2^d$, one ball suffices; for $k_T\ge2^d$, a maximal separated packing gives the upper bound $(1+{k_T}^{1/d}/2)^d\le k_T$. Thus $g,f_1,\ldots,f_M$ all belong to the same prescribed $\mathcal F_K$, since $k_T\le K$. 
% This verification holds for any $u>0$.

% and not merely for levels comparable to $T$ or $\eta$.

Consider any $(\eta,T,\delta)$-valid procedure with fixed budget $n$. For a response surface $f\in\mathcal F_K$, let $\mathcal E_f$ be the event that the output $(I_\tau,O_\tau)_{0\le\tau\le T}$ satisfies~\eqref{eq:region-minimax-loss} for all $\tau\in[0,T]$, which we call the validity event under $f$. By the definition of an $(\eta,T,\delta)$-valid procedure, the probability of $\mathcal E_f$ is at least $1-\delta$ when the response surface is $f$. We use this property for $f=g$ and for $f=f_i$, $1\le i\le M$, all of which belong to $\mathcal F_K$ by the covering argument above.

For each $i$, set $\tau_i=(-g(z_i))-4\eta$, which lies in $(0,T)$ because $-g(z_i)\in[T/4,T/2]$ and $4\eta\le T/32$, and let $E_i=\{z_i\in I_{\tau_i}\}$. Since $z_i$ and $\tau_i$ are deterministic, $E_i$ is an event by the measurability of membership in the output sets. Because $g^\star=f_i^\star=0$ and $\psi(0)=1$, the performance gaps of $z_i$ are $g^\star-g(z_i)=\tau_i+4\eta$ under $g$ and $f_i^\star-f_i(z_i)=-g(z_i)-8\eta=\tau_i-4\eta$ under $f_i$. We now show that $E_i\cap\mathcal E_g=\varnothing$ and $\mathcal E_{f_i}\subseteq E_i$.
\begin{itemize}
\item On $\mathcal E_g$, since $g^\star-g(z_i)>\tau_i$, we have $z_i\notin\mathcal S_{\tau_i}(g)$, and the inclusion $I_{\tau_i}\subseteq\mathcal S_{\tau_i}(g)$ gives $z_i\notin I_{\tau_i}$.
\item On $\mathcal E_{f_i}$, since $f_i^\star-f_i(z_i)<\tau_i$, we have $z_i\in\mathcal S_{\tau_i}(f_i)\subseteq O_{\tau_i}$. Moreover, $|f_i^\star-f_i(z_i)-\tau_i|=4\eta>\eta$, so the second condition in~\eqref{eq:region-minimax-loss} gives $z_i\notin O_{\tau_i}\setminus I_{\tau_i}$. Hence $z_i\in I_{\tau_i}$.
\end{itemize}

Let $P_0$ and $P_i$ denote the joint distributions of the sampling history and the output, including the internal randomization of the procedure, when the response surface is $g$ and $f_i$, respectively. The two inclusions give $P_0(E_i)\le P_0(\mathcal E_g^c)\le\delta$ and $P_i(E_i)\ge P_i(\mathcal E_{f_i})\ge1-\delta$. The first inequality uses the validity of the procedure under the baseline $g$.
% which is why we verified above that $g$ itself belongs to $\mathcal F_K$.
% This differs from the proofs of the optimization lower bounds in Theorem~\ref{thm:covering-lower-alpha-large} and Proposition~\ref{prop:covering-saturation}, where the reference distribution only serves as a common point of comparison and its response surface need not belong to the class.

Next, we bound $\KL(P_0,P_i)$ from below. For $p,q\in[0,1]$, $\operatorname{kl}(p,q)$ is the relative entropy between the Bernoulli distributions with means $p$ and $q$. Since $\mathbf 1_{E_i}$ is a measurable function of the history and the output, the data-processing inequality for relative entropy gives $\KL(P_0,P_i)\ge\operatorname{kl}(P_0(E_i),P_i(E_i))$; see, for example, \citet[Lemma~1]{kaufmann2016complexity}. For fixed $q$, the function $p\mapsto\operatorname{kl}(p,q)$ is convex with minimum value zero at $p=q$, so it is nonincreasing for $p\le q$. Similarly, for fixed $p$, the function $q\mapsto\operatorname{kl}(p,q)$ is nondecreasing for $q\ge p$. Because $\delta\le1/10$, we have $P_0(E_i)\le\delta<1-\delta\le P_i(E_i)$, and hence
\begin{equation}\label{eq:region-lower-dp}
\KL(P_0,P_i)\ge\operatorname{kl}\bigl(P_0(E_i),P_i(E_i)\bigr)\ge\operatorname{kl}\bigl(\delta,P_i(E_i)\bigr)\ge\operatorname{kl}(\delta,1-\delta).
\end{equation}

We now compute $\KL(P_0,P_i)$. Let $U$ denote the internal randomization of the procedure, with distribution $P_U$, let $\mathcal H_0=U$, and let $\mathcal H_t=(U,x_1,y_1,\ldots,x_t,y_t)$ be the history after $t$ evaluations. Let $o=(I_\tau,O_\tau)_{0\le\tau\le T}$ denote the output. The procedure is specified by the following decision rules:
\begin{itemize}
\item for each $t=1,\ldots,n$, a conditional distribution $\pi_t(\cdot\mid\mathcal H_{t-1})$ on $\X$, according to which $x_t$ is chosen given $\mathcal H_{t-1}$;
\item a conditional distribution $\pi_{\rm out}(\cdot\mid\mathcal H_n)$, according to which the output $o$ is chosen given $\mathcal H_n$.
\end{itemize}
% A deterministic rule corresponds to a point mass. These decision rules are determined by the procedure alone. 
The procedure does not observe the response surface, which affects the data only through the responses $y_t$. Hence the same $P_U$, $\pi_1,\ldots,\pi_n$, and $\pi_{\rm out}$ apply under $g$ and under each $f_i$, although the distributions of the histories differ. 
% If the procedure is given the common maximizer and optimal value before sampling, its decision rules may also depend on this information, which is again the same under $g$ and every $f_i$.
Under a response surface $f$, given $\mathcal H_{t-1}$ and $x_t$, the response is $y_t=f(x_t)+\varepsilon_t$, where $\varepsilon_t\sim N(0,\sigma^2)$ is independent of $(\mathcal H_{t-1},x_t)$. Let $\phi_\sigma(z)=(2\pi\sigma^2)^{-1/2}\exp\{-z^2/(2\sigma^2)\}$ denote the $N(0,\sigma^2)$ density. The joint distribution $P_f$ of $(\mathcal H_n,o)$ therefore factorizes as
\[
\begin{aligned}
&P_f(du,dx_1,dy_1,\ldots,dx_n,dy_n,do)\\
&=P_U(du)\prod_{t=1}^n\Bigl\{\pi_t(dx_t\mid\mathcal H_{t-1})\,\phi_\sigma\bigl(y_t-f(x_t)\bigr)\,dy_t\Bigr\}\pi_{\rm out}(do\mid\mathcal H_n).
\end{aligned}
\]
% The factors $P_U$, $\pi_t$, and $\pi_{\rm out}$ are the same for $f=g$ and $f=f_i$. 
Hence, with respect to the common measure obtained by replacing each factor $\phi_\sigma(y_t-f(x_t))\,dy_t$ in this factorization by $dy_t$, the distributions $P_0$ and $P_i$ have the positive densities $\prod_{t=1}^n\phi_\sigma(y_t-g(x_t))$ and $\prod_{t=1}^n\phi_\sigma(y_t-f_i(x_t))$, respectively. In particular, $P_0$ and $P_i$ are mutually absolutely continuous, and the decision rules cancel in the likelihood ratio as
\[
\log\frac{dP_0}{dP_i}=\sum_{t=1}^n\log\frac{\phi_\sigma(y_t-g(x_t))}{\phi_\sigma(y_t-f_i(x_t))}=\sum_{t=1}^n\frac{\{y_t-f_i(x_t)\}^2-\{y_t-g(x_t)\}^2}{2\sigma^2}.
\]
Under $P_0$, we have $y_t=g(x_t)+\varepsilon_t$, so the $t$-th summand equals
\[
\frac{\{g(x_t)-f_i(x_t)+\varepsilon_t\}^2-\varepsilon_t^2}{2\sigma^2}=\frac{\{g(x_t)-f_i(x_t)\}^2+2\varepsilon_t\{g(x_t)-f_i(x_t)\}}{2\sigma^2}.
\]
Since $\mathbb E_0[\varepsilon_t\mid\mathcal H_{t-1},x_t]=0$ and $|g-f_i|\le8\eta$, the conditional expectation of the $t$th summand given $(\mathcal H_{t-1},x_t)$ is $\{g(x_t)-f_i(x_t)\}^2/(2\sigma^2)$. By the tower property and the definition of relative entropy, we have
\begin{equation}\label{eq:region-lower-kl}
\KL(P_0,P_i)=\mathbb E_0\Bigl[\log\frac{dP_0}{dP_i}\Bigr]=\frac1{2\sigma^2}\mathbb E_0\sum_{t=1}^n\{g(x_t)-f_i(x_t)\}^2.
\end{equation}
Let $N_i=\sum_{t=1}^n\mathbf 1\{x_t\in A_i\}$ be the number of evaluations allocated to $A_i$. Since $0\le f_i-g=8\eta\psi((\cdot-z_i)/h)\le8\eta\,\mathbf 1_{A_i}$, we have $\{g(x_t)-f_i(x_t)\}^2\le64\eta^2\,\mathbf 1\{x_t\in A_i\}$, and therefore
\begin{equation}\label{eq:region-lower-kl-bound}
\KL(P_0,P_i)\le\frac{64\eta^2}{2\sigma^2}\mathbb E_0N_i=\frac{32\eta^2}{\sigma^2}\mathbb E_0N_i.
\end{equation}

Combining the two bounds on $\KL(P_0,P_i)$ gives $\mathbb E_0N_i\ge\sigma^2\operatorname{kl}(\delta,1-\delta)/(32\eta^2)$ for each $i$. Since the supports $A_i$ are disjoint, $\sum_{i=1}^MN_i\le n$, and hence
\[
n\ge\sum_{i=1}^M\mathbb E_0N_i\ge\frac{\sigma^2 M}{32\eta^2}\operatorname{kl}(\delta,1-\delta)\ge c\sigma^2\operatorname{kl}(\delta,1-\delta)\,\eta^{-2-d/\nu}\min\{1,KT^{d/\alpha}\},
\]
where the last step substitutes~\eqref{eq:region-perturbation-count}. Since the valid procedure was arbitrary, this proves~\eqref{eq:region-minimax-lower}. The same argument applies when the procedure is given the common maximizer and the common optimal value before sampling, because this information is the same under $P_0$ and every $P_i$.
\end{proof}

\subsection{Proof of Theorem~\ref{thm:region-learning-budget}}\label{ec:region-budget}

In addition to Lemma~\ref{lem:new-confidence}, which gives the simultaneous confidence bounds used throughout Section~\ref{sec:new-regret}, this proof uses the two lemmas presented below. The proofs of these three lemmas are given in Sections~\ref{ec:proof-confidence}--\ref{ec:proof-capped-phase}. 
% The proof also uses Proposition~\ref{prop:phase-accuracy}, stated in \textcolor{black}{Subsection}~\ref{ec:cap-transfer} and proved in Section~\ref{ec:proof-phase-accuracy}. 

Lemma~\ref{lem:new-mig} controls the information gain, which underlies the bounds \eqref{eq:cap-driven-gamma}, \eqref{eq:cap-driven-cover-cost}, \eqref{eq:new-phase-cost}, and~\eqref{eq:new-phase-cost-local} on the number of evaluations in a phase in Lemmas~\ref{lem:cap-driven-phase-cost} and~\ref{lem:new-phase-cost}.
For $A\subseteq\X$, define the information gain $\gamma_n(A;\lambda)=\sup_{x_{1:n}\in A^n}\tfrac12\log\det(I+\lambda^{-1}K_n)$.
Conditioning on additional observations can only reduce posterior variance. Consequently, an arbitrary subsequence of $n$ sampled points in $A$ has total variance at most $C_\lambda\gamma_n(A;\lambda)$, where $C_\lambda=2/\log(1+\lambda^{-1})$. 

The following lemma bounds information gain on a union of covering balls, without a separation requirement, for any $\nu>0$.

\begin{lemma}\label{lem:new-mig}
For any $\nu>0$ and any fixed $\bar r>0$, there is a constant $C$, depending only on the kernel, $\lambda$, $\bar r$, and the fixed domain, such that, for $1\le n\le N$ and $0<r\le\bar r$,
\begin{equation}\label{eq:new-mig}
\begin{aligned}
 \gamma_n(A;\lambda)\le {}&C\log^2(2N)\min\Bigl\{
       1+(n/\lambda)^{d/(2\nu+d)},K+K^{2\nu/(2\nu+d)}r^{2\nu d/(2\nu+d)}
                       (n/\lambda)^{d/(2\nu+d)}\Bigr\}
\end{aligned}
\end{equation}
whenever $A$ is covered by $K$ Euclidean radius-$r$ balls.
\end{lemma}

Lemma~\ref{lem:cap-driven-phase-cost} studies a single phase whose predictor uses only observations from that phase with any nonnegative continuous screening function $G$ fixed at the phase start. The continuous maintained bands $\ell,u$ are obtained by intersections starting from $[-B,B]$, so $W=u-\ell\le2B$.

Define $a_n(x)=\min\{G(x),W(x),2\beta_Ns_n(x)+2\zeta_N\}$ and $a_n^\star=\max_{x\in\X}a_n(x)$. The phase samples a maximizer of $a_n$ while $a_n^\star>2\eps$ and budget remains, and completes when $a_n^\star\le2\eps$.

This general statement applies both to the screening function in Algorithm~\ref{alg:continuous-ccucb} and to the screening function for the chosen boundary in Section~\ref{subsec:fixed-tolerance-screening}.

\begin{lemma}\label{lem:cap-driven-phase-cost}
Any point sampled in a phase with $\eps\ge2\zeta_N$ satisfies
\[
 G(x_j)>2\eps,W(x_j)>2\eps,
 s_{j-1}(x_j)>\frac{\eps-\zeta_N}{\beta_N}
                       \ge\frac{\eps}{2\beta_N}.
\]
Conditional on the history available at the start of the phase, define the fixed compact set $D=\{x\in\X:G(x)\ge2\eps,\ W(x)\ge2\eps\}$.
For its number $n$ of observations, including a truncated phase,
\begin{equation}\label{eq:cap-driven-gamma}
 n\eps^2\le4\beta_N^2C_\lambda\gamma_n(D;\lambda).
\end{equation}
A phase with no evaluations has cost zero and needs no new confidence
event. On $\Econf$, a completed phase has
maintained width at most $2\eps$ wherever the function $G$, held fixed throughout the phase, satisfies $G(x)>2\eps$.
If $D$ is covered by $J\ge1$ balls of radius $r$ in a fixed bounded
range, then
\begin{equation}\label{eq:cap-driven-cover-cost}
\begin{aligned}
 n\le C\left[1+J\mathcal V_N\eps^{-2}
       +J\mathcal V_N^{1+d/(2\nu)}r^d\eps^{-(2+d/\nu)}\right].
\end{aligned}
\end{equation}
% The definition of $\mathcal V_N$ displays the confidence and logarithmic contributions explicitly.
\end{lemma}

\begin{proof}[Proof of Theorem~\ref{thm:region-learning-budget}]
Throughout the proof, we work on $\Econf$, which has probability at least $1-\delta$ by Lemma~\ref{lem:new-confidence}. The choice $m=0$ for $T=0$ and $m=N$ for $T>0$ gives the track under analysis all $N$ evaluations. The other track has quota zero and therefore never makes an evaluation. Each of its phases either completes without any evaluation, which leaves the confidence bounds unchanged and hence, by the fixed rule for breaking ties, also the estimated optimal point and the saved regions, or finishes that track by line~6 or line~9 of Algorithm~\ref{alg:continuous-ccucb}. Hence the zero-quota track does not affect the analysis below. We first show that each completed phase satisfies the accuracy bounds in~\eqref{eq:target-phase-accuracy}. We then bound the number of evaluations in each phase, sum these bounds over the phases, and show by contradiction that the target phase $H$ completes within the budgets~\eqref{eq:region-global-cost} and~\eqref{eq:region-local-cost}.

By Proposition~\ref{prop:phase-accuracy}, any completed phase $h$ with $\eps_h\ge2\zeta_N$ satisfies $\hi_h-\lo_h\le2\eps_h$ and $f^\star-f(\widehat x_h)\le2\eps_h$, and its region outputs satisfy $I_{h,\tau}\subseteq\mathcal S_\tau\subseteq O_{h,\tau}$ and $O_{h,\tau}\setminus I_{h,\tau}\subseteq\{x:|f^\star-f(x)-\tau|\le4\eps_h\}$ simultaneously for all $0\le\tau\le T$. Algorithm~\ref{alg:band-sharing} saves these quantities when phase $h$ completes. For $h=H$, where $\eps_H=\eps\ge2\zeta_N$, these are \eqref{eq:target-phase-accuracy}. It remains to show that phase $H$ completes within the stated budgets.

We next locate the points sampled in a phase. Fix a phase $h\le H$, so that $\eps_h\ge\eps\ge2\zeta_N$, and let $G_h$ and $W_{h-1}$ be the screening function and the maintained width in~\eqref{eq:new-frozen-cap}, both of which are fixed at the start of phase $h$. Define $D_h=\{x\in\X:G_h(x)\ge2\eps_h,\ W_{h-1}(x)\ge2\eps_h\}$. By Lemma~\ref{lem:cap-driven-phase-cost}, any point $x$ sampled in phase $h$ satisfies $G_h(x)>2\eps_h$ and $W_{h-1}(x)>2\eps_h$, and hence lies in $D_h$. Because $D_h$ depends only on quantities fixed at the start of the phase, this also holds when phase $h$ is cut off by the budget. We now show that
\begin{equation}\label{eq:region-budget-active}
D_h\subseteq\mathcal S_{(T+L_\vartheta\eps_h)+},
\end{equation}
that is, all points of $D_h$ have performance gap at most $T+L_\vartheta\eps_h$. For $h=0$, this holds because $f^\star-f(x)\le2B=\eps_0$ for all $x\in\X$. For $h\ge1$, phase $h-1$ has completed, because a phase starts only after the previous phase of the same track completes. Let $x\in D_h$. Since $G_h(x)\ge2\eps_h>0$, the definition of $G_h$ in~\eqref{eq:new-frozen-cap} gives $u_{h-1}(x)\ge\lo_{h-1}-T-2\eps_h>\lo_{h-1}-T-2\eps_{h-1}$. Proposition~\ref{prop:phase-accuracy}, applied to the completed phase $h-1$, therefore gives $u_{h-1}(x)-\ell_{h-1}(x)\le2\eps_{h-1}=2\vartheta\eps_h$ and $f^\star-\lo_{h-1}\le\hi_{h-1}-\lo_{h-1}\le2\vartheta\eps_h$, where $f^\star\le\hi_{h-1}$ because $u_{h-1}\ge f$ on $\Econf$. Since also $\ell_{h-1}\le f$ on $\Econf$,
\[
f^\star-f(x)\le(f^\star-\lo_{h-1})+(\lo_{h-1}-u_{h-1}(x))+(u_{h-1}(x)-\ell_{h-1}(x))\le2\vartheta\eps_h+(T+2\eps_h)+2\vartheta\eps_h,
\]
which is at most $T+L_\vartheta\eps_h$ because $L_\vartheta=4(\vartheta+1)$. This proves~\eqref{eq:region-budget-active}.

We now bound the number $n_h$ of evaluations in phase $h$, including a phase cut off by the budget. By~\eqref{eq:ec-log-bounds}, $\mathcal V_N\le C\mathcal V_N^{1+d/(2\nu)}\le C\Lambda_{N,\delta}$. We also use $1\le\Lambda_{N,\delta}$ and $\eps_h^{-2}\le1+\eps_h^{-p}$, which holds because $p>2$.

The global phase bound uses no geometric condition. The set $D_h\subseteq\X$ is covered by one ball of radius $D_{\X}$, the diameter of $\X$. Applying~\eqref{eq:cap-driven-cover-cost} with $J=1$ and $r=D_{\X}$ gives $n_h\le C[1+\mathcal V_N\eps_h^{-2}+\mathcal V_N^{1+d/(2\nu)}D_{\X}^d\eps_h^{-p}]$, and thus
\begin{equation}\label{eq:region-budget-phase-global}
n_h\le C\Lambda_{N,\delta}(1+\eps_h^{-p}).
\end{equation}

The localized phase bound uses Assumption~\ref{as:new-cover}. Suppose that this assumption holds with $\alpha>\nu$ and that $0\le T\le u_0/2$, and first consider a phase with $T+L_\vartheta\eps_h\le u_0$. By~\eqref{eq:region-budget-active} and~\eqref{eq:new-cover} with $u=T+L_\vartheta\eps_h$, the set $D_h$ is covered by $K$ balls of radius $r_h=c_\rho(T+L_\vartheta\eps_h)^{1/\alpha}\le c_\rho u_0^{1/\alpha}$. Applying~\eqref{eq:cap-driven-cover-cost} with $J=K$ and $r=r_h$ gives $n_h\le C[1+K\mathcal V_N\eps_h^{-2}+K\mathcal V_N^{1+d/(2\nu)}r_h^d\eps_h^{-p}]$. Since $L_\vartheta\ge1$, we have $r_h^d\le c_\rho^dL_\vartheta^{d/\alpha}(T+\eps_h)^{d/\alpha}$, and therefore
\begin{equation}\label{eq:region-budget-phase-local}
n_h\le C\Lambda_{N,\delta}\bigl[1+K\eps_h^{-2}+K(T+\eps_h)^{d/\alpha}\eps_h^{-p}\bigr].
\end{equation}
Because $T\le u_0/2$, the condition $T+L_\vartheta\eps_h\le u_0$ holds whenever $\eps_h\le u_0/(2L_\vartheta)$. The remaining phases have $\eps_h>u_0/(2L_\vartheta)$. There is at most a fixed number of them, and by~\eqref{eq:region-budget-phase-global} each satisfies $n_h\le C\Lambda_{N,\delta}\{1+(2L_\vartheta/u_0)^p\}$, which is at most the right-hand side of~\eqref{eq:region-budget-phase-local} after increasing $C$. Hence~\eqref{eq:region-budget-phase-local} holds for every phase $h\le H$.

We now sum the phase bounds. Write $\eps=\eps_H$. Since $\eps_h=\vartheta^{H-h}\eps$ for $h\le H$, we have $\sum_{h=0}^{H}\eps_h^{-b}=\eps^{-b}\sum_{j=0}^{H}\vartheta^{-jb}\le\eps^{-b}/(1-\vartheta^{-b})$ for any $b>0$, and the number of phases satisfies $H+1=1+\log_\vartheta(2B/\eps)\le C(1+\eps^{-2})$. Summing the global phase bound~\eqref{eq:region-budget-phase-global} over $h\le H$, with $b=p$, gives
\begin{equation}\label{eq:region-budget-sum-global}
\sum_{h=0}^{H}n_h\le C_1\Lambda_{N,\delta}\bigl(1+\eps^{-p}\bigr).
\end{equation}
To sum the localized phase bound~\eqref{eq:region-budget-phase-local}, we use $(T+\eps_h)^{d/\alpha}\le C_{d,\alpha}(T^{d/\alpha}+\eps_h^{d/\alpha})$, so that $(T+\eps_h)^{d/\alpha}\eps_h^{-p}\le C_{d,\alpha}(T^{d/\alpha}\eps_h^{-p}+\eps_h^{-(p-d/\alpha)})$, where $p-d/\alpha=2+d/\nu-d/\alpha>2$ because $\alpha>\nu$. Applying the geometric sum with $b=2$, $b=p$, and $b=p-d/\alpha$, and then using $T^{d/\alpha}+\eps^{d/\alpha}\le2(T+\eps)^{d/\alpha}$, gives
\begin{equation}\label{eq:region-budget-sum-local}
\sum_{h=0}^{H}n_h\le C_1\Lambda_{N,\delta}\bigl[1+K\eps^{-2}+K(T+\eps)^{d/\alpha}\eps^{-p}\bigr].
\end{equation}
Here $C_1$ denotes a constant for which both~\eqref{eq:region-budget-sum-global} and~\eqref{eq:region-budget-sum-local} hold, and both bounds also apply at $T=0$. Only the compactness of $\X$ is used in these upper bounds. The right-hand side of~\eqref{eq:region-budget-sum-global} is the expression in~\eqref{eq:region-global-cost} with $C_1$ in place of $C$. Under Assumption~\ref{as:new-cover} with $\alpha>\nu$ and $0\le T\le u_0/2$, taking the smaller of~\eqref{eq:region-budget-sum-global} and~\eqref{eq:region-budget-sum-local} gives the expression in~\eqref{eq:region-local-cost} with $C_1$ in place of $C$.

Finally, we show that phase $H$ completes within the sufficient budgets~\eqref{eq:region-global-cost} and~\eqref{eq:region-local-cost} of Theorem~\ref{thm:region-learning-budget}, whose constant $C$ we take larger than $C_1$. On $\Econf$, the safeguard for inconsistent bounds in Algorithm~\ref{alg:continuous-ccucb} is never triggered, and the refinement stop of Algorithm~\ref{alg:continuous-ccucb}, which applies only once the accuracy level falls below $2\zeta_N$, does not affect phases with $\eps_h\ge2\zeta_N$. Hence phase $H$ can fail to complete only if the budget is exhausted first. Suppose that $N$ is at least the expression in~\eqref{eq:region-global-cost}, or at least the expression in~\eqref{eq:region-local-cost} under its assumptions, but phase $H$ does not complete. Since phases are run in order, the track then uses all $N$ evaluations in phases $h\le H$, and each of these phases, including the one cut off by the budget, satisfies \eqref{eq:region-budget-phase-global} and, under the assumptions of~\eqref{eq:region-local-cost}, \eqref{eq:region-budget-phase-local}. Thus $N=\sum_{h=0}^{H}n_h$, where $n_h=0$ for phases that are not reached. By \eqref{eq:region-budget-sum-global} and~\eqref{eq:region-budget-sum-local}, $N$ is at most the corresponding expression with $C_1$ in place of $C$. Since $C>C_1$, this contradicts the assumption on $N$. Therefore phase $H$ completes.

% The order of the two tests in Algorithm~\ref{alg:sample-allocation} ensures that this argument also covers the boundary cases. The stopping test is checked before the budget test, so if the last available evaluation completes phase $H$, the phase is recorded as completed and its output is saved. Moreover, a phase whose stopping test already holds at its start completes without any new evaluation, even if no budget remains.

% Combining the two parts, 
Therefore, on $\Econf$, phase $H$ completes whenever $N$ is at least the sufficient budget, and the output saved at its completion satisfies~\eqref{eq:target-phase-accuracy}. Since $\Econf$ does not depend on $H$, these conclusions hold simultaneously for all targets $\eps=2B\vartheta^{-H}\ge2\zeta_N$, which proves the theorem.
\end{proof}

\subsection{Proof of Corollary~\ref{cor:region-minimax-matching}}\label{ec:region-inversion}
\begin{proof}[Proof of Corollary~\ref{cor:region-minimax-matching}]

% The lower bound is already established. for the upper bound we choose a declared horizon for which Theorem~\ref{thm:region-learning-budget} guarantees completion of the target phase, and then verify the condition on $\zeta_N$. 

% The lower bound is Theorem~\ref{thm:region-minimax-lower}. For the upper bound, c
Recall from Section~\ref{ec:region-lower} that $T_{\mathrm{reg}}\le\min\{1,u_0/2,2048B\}$. Hence $T\le1$; $T\le u_0/2$, so that the localized bound~\eqref{eq:region-local-cost} applies; and $\eta\le T/128\le16B$, so that $\eta/8\le2B=\eps_0$ and there is a phase target $\eps=\eps_H$ with $\eta/(8\vartheta)<\eps\le\eta/8$. In the stated range, $T+\eps\asymp T$. Since $\eps\le\eta/8\le T\le1$ and $\alpha>\nu$, we have $\eps^{1/\nu}\le\eps^{1/\alpha}\le T^{1/\alpha}$, and hence $K\eps^{-2}\le KT^{d/\alpha}\eps^{-(2+d/\nu)}$.

Consequently, the sufficient evaluation count in Theorem~\ref{thm:region-learning-budget} is at most
\[
 C F(\eta,T,K)\Lambda_{N,\delta},
 F(\eta,T,K)=\eta^{-(2+d/\nu)}\min\{1,KT^{d/\alpha}\}.
\]
Since $T\ge128\eta$, $K\ge K_{\mathrm{reg}}\ge1$, and $\eta\le T/128<1$, we have
\begin{equation}\label{eq:region-inversion-F}
c\eta^{-(2+d/\nu-d/\alpha)}\le F(\eta,T,K) \le\eta^{-(2+d/\nu)}.
\end{equation}

Choose the declared horizon
\begin{equation}\label{eq:region-inversion-horizon}
 N=\left\lceil A F(\eta,T,K)
       \left[\log\frac{2}{\eta\delta}\right]^{3+3d/(2\nu)}\right\rceil
\end{equation}
for a sufficiently large fixed constant $A$, which may depend on $\vartheta$. By~\eqref{eq:region-inversion-F} and~\eqref{eq:region-inversion-horizon}, we have
\[
 \log(2N)\le\log(2N/\delta)
 \le C(1+\log A)\log\frac{2}{\eta\delta}.
\]
The sufficient evaluation count is therefore at most
\[
 C(1+\log A)^{3+3d/(2\nu)}F(\eta,T,K)
       \left[\log\frac{2}{\eta\delta}\right]^{3+3d/(2\nu)}.
\]
Choose $A$ so that $C(1+\log A)^{3+3d/(2\nu)}\le A$. This makes the sufficient count at most $N$, uniformly in $K,T,\eta,\delta$.

The confidence parameters in~\eqref{eq:new-confidence-parameters} satisfy
$\zeta_N\le C N^{-7/2}\sqrt{\log(2N/\delta)}$. Hence
\begin{equation}\label{eq:region-inversion-zeta}
\begin{split}
 \zeta_N\le{}&C A^{-7/2}(1+\log A)^{1/2}
 \eta^{\frac72(2+d/\nu-d/\alpha)}\left[\log\frac{2}{\eta\delta}\right]^{
       \frac12-\frac72(3+3d/(2\nu))}
 \le\frac{\eta}{16\max\{2,\vartheta\}}.
\end{split}
\end{equation}
The last inequality in~\eqref{eq:region-inversion-zeta} holds after increasing the fixed $A$, because $2+d/\nu-d/\alpha>2$, $\eta<1$, and the exponent of the logarithm is negative. Thus $2\zeta_N\le\eps$, and phase completion gives the accuracy~\eqref{eq:target-phase-accuracy}, with region estimation error at most $4\eps\le\eta$.

The ceiling in $N$ is absorbed into the constant $C$ in~\eqref{eq:region-minimax-matching} because $F(\eta,T,K)$ is bounded below by a positive constant in the stated range. All constants may depend on the fixed phase ratio $\vartheta$, but are independent of $\delta,K,T,\eta$.

\end{proof}

\subsection{Proof of Theorem~\ref{thm:new-regret}}\label{ec:regret}

% The theorem concerns the $T=0$ setting of Algorithm~\ref{alg:continuous-ccucb}.
We first present two lemmas used in this proof, whose proofs are given in Sections~\ref{ec:proof-completed-bands} and~\ref{ec:proof-phase-cost}. Lemma~\ref{lem:new-active} bounds the band widths after a completed phase and the optimality gaps of the designs sampled in the following phase.

\begin{lemma}\label{lem:new-active}
On $\Econf$, after phase $h$ completes,
\[
 W_h(x)\le2\eps_h\text{ if }u_h(x)>\lo_h-T-2\eps_h,
 \hi_h-\lo_h\le2\eps_h,f^\star-f(\widehat x_h)\le2\eps_h.
\]
Any point sampled in phase $h$ has gap at most $T+L_\vartheta\eps_h$ and at most $2B$.
\end{lemma}

For the $T=0$ setting of Algorithm~\ref{alg:continuous-ccucb}, Lemma~\ref{lem:new-phase-cost} converts the information-gain bound~\eqref{eq:new-mig} into a bound on the number of evaluations in each phase.

\begin{lemma}\label{lem:new-phase-cost}
Under Assumptions~\ref{as:new-noise}--\ref{as:new-rkhs}, consider Algorithm~\ref{alg:continuous-ccucb} with $T=0$, $N\ge2$, and $\delta\in(0,1)$. Let $n_h$ count observations collected by the sampling loop in phase $h$, including repetitions and an unfinished phase truncated at the horizon. If $\eps_h\ge2\zeta_N$, then deterministically
\begin{equation}\label{eq:new-phase-cost}
 n_h\le C\Lambda_{N,\delta}\eps_h^{-p}.
\end{equation}
For the localized bound, work on $\Econf$, defined in~\eqref{eq:shared-confidence-event}, with $\mathbb P(\Econf)\ge1-\delta$ by Lemma~\ref{lem:new-confidence}. On $\Econf$, under Assumption~\ref{as:new-cover} and $L_\vartheta\eps_h\le u_0$,
\begin{equation}\label{eq:new-phase-cost-local}
 n_h\le C\Lambda_{N,\delta}K\eps_h^{-a}.
\end{equation}
The constants are independent of $K,N,\delta$.
\end{lemma}

\begin{proof}[Proof of Theorem~\ref{thm:new-regret}]

Fix $\vartheta>1$. All upper-bound constants in this proof will depend on $\vartheta$. The region track has quota $m=0$ and, as noted in the proof of Theorem~\ref{thm:region-learning-budget}, does not affect the analysis. For $\delta=N^{-2}$, the factor $\Lambda_{N,\delta}$ in the phase bounds of Lemma~\ref{lem:new-phase-cost} is at most $CL_N$ by~\eqref{eq:ec-log-bounds}.

If $\zeta_N>B$, the deterministic bounds $r_N\le2B$ and
$R_N\le2BN$ are absorbed by the $\zeta_N$ term.
Hence assume $\zeta_N\le B$, and then any nonempty phase, that is, any phase in which the sampling loop of Algorithm~\ref{alg:sample-allocation} makes at least one evaluation, has
$\eps_h\ge2\zeta_N$, because Algorithm~\ref{alg:continuous-ccucb} allocates all remaining evaluations to the saved estimated optimal point once the accuracy level falls below $2\zeta_N$. Work on $\Econf[N^{-2}]$. By Lemma~\ref{lem:new-phase-cost}, all nonempty phases, including truncated ones, satisfy
\begin{equation}\label{eq:regret-phase-costs}
 n_h\le CL_N\eps_h^{-p}\text{ for all }h,
 n_h\le CL_NK\eps_h^{-a}\text{ if }L_\vartheta\eps_h\le u_0,
\end{equation}
and by Lemma~\ref{lem:new-active} with $T=0$, all points sampled in phase $h$ have gap at most $L_\vartheta\eps_h$.

The finitely many coarse scales with $L_\vartheta\eps_h>u_0$ have $\eps_h\ge u_0/L_\vartheta$ fixed, so~\eqref{eq:regret-phase-costs} bounds their total weighted cost by
$\sum_{h:\,L_\vartheta\eps_h>u_0}L_\vartheta\eps_hn_h\le CL_N$.

Split the nonempty phases at any scale $\eps\in[2\zeta_N,2B]$. We use the following three facts, where $R_N=\sum_{t=1}^N\{f^\star-f(x_t)\}$.
\begin{itemize}
\item \emph{Gap of each sampled point.} If $x_t$ is sampled in phase $h\ge1$, then $f^\star-f(x_t)\le L_\vartheta\eps_h$, by Lemma~\ref{lem:new-active} with $T=0$. The reason is that $x_t$ is sampled only if its acquisition value exceeds $2\eps_h$, which implies $u_{h-1}(x_t)>\lo_{h-1}-2\eps_h$. Since phase $h-1$ has completed, Lemma~\ref{lem:new-active} also gives $u_{h-1}(x_t)-f(x_t)\le W_{h-1}(x_t)\le2\eps_{h-1}=2\vartheta\eps_h$ and $f^\star-\lo_{h-1}\le2\vartheta\eps_h$. Hence
\[
f^\star-f(x_t)=(f^\star-\lo_{h-1})+(\lo_{h-1}-u_{h-1}(x_t))+(u_{h-1}(x_t)-f(x_t))<(4\vartheta+2)\eps_h\le L_\vartheta\eps_h.
\]
In phase $0$, $f^\star-f(x_t)\le2B=\eps_0\le L_\vartheta\eps_0$.
\item \emph{Fine phases.} The phases with $\eps_h<\eps$ contain at most $N$ evaluations in total. By the previous fact, each of these evaluations has gap at most $L_\vartheta\eps_h<L_\vartheta\eps$, so together they contribute at most $L_\vartheta N\eps$ to $R_N$.
\item \emph{Evaluations at the saved estimated optimal point.} Suppose that a phase with accuracy level below $2\zeta_N$ is reached, and let $h'$ be the first such phase. Since $\eps_0=2B\ge2\zeta_N$, we have $h'\ge1$, and phase $h'-1$ has completed with $\eps_{h'-1}=\vartheta\eps_{h'}<2\vartheta\zeta_N$. By Lemma~\ref{lem:new-active}, the saved estimated optimal point $\widehat x_{h'-1}$ satisfies $f^\star-f(\widehat x_{h'-1})\le2\eps_{h'-1}<4\vartheta\zeta_N$. All remaining evaluations are made at this point, and there are at most $N$ of them, so they contribute at most $4\vartheta N\zeta_N$ to $R_N$.
\end{itemize}
On $\Econf[N^{-2}]$, the safeguard for inconsistent bounds is never triggered. Hence all evaluations are made in a phase with $L_\vartheta\eps_h>u_0$, in a phase with $\eps\le\eps_h\le u_0/L_\vartheta$, in a phase with $\eps_h<\eps$, or at the saved estimated optimal point after the accuracy level falls below $2\zeta_N$. The first group contributes at most $CL_N$ by the preceding paragraph, the second group contributes at most $L_\vartheta\sum_{h:\,\eps\le\eps_h\le u_0/L_\vartheta}\eps_hn_h$ by the first fact, and the last two groups are bounded by the last two facts. Together they give
\[
 R_N
 \le CL_N
       +L_\vartheta\sum_{h:\,\eps\le\eps_h\le u_0/L_\vartheta}\eps_hn_h
       +L_\vartheta N\eps+4\vartheta N\zeta_N.
\]
For any $b>1$,
\begin{equation}\label{eq:regret-geometric-sum}
\sum_{h:\,\eps_h\ge\eps}\eps_h^{1-b}\le\frac{\eps^{1-b}}{1-\vartheta^{1-b}}.
\end{equation}
Inserting~\eqref{eq:regret-phase-costs} and applying \eqref{eq:regret-geometric-sum} with $b=p$ and $b=a$ gives
\[
 \sum_{h:\,\eps\le\eps_h\le u_0/L_\vartheta}\eps_hn_h
 \le CL_N\min\Bigl\{\sum_{h:\eps_h\ge\eps}\eps_h^{1-p},\ K\sum_{h:\eps_h\ge\eps}\eps_h^{1-a}\Bigr\}
 \le CL_N\min\{\eps^{-1-d/\nu},K\eps^{1-a}\},
\]
so that
\begin{equation}\label{eq:regret-split-bound}
 R_N\le CL_N+CL_N\min\{K\eps^{1-a},\eps^{-1-d/\nu}\}+CN\eps+CN\zeta_N.
\end{equation}
The minimum in~\eqref{eq:regret-split-bound} has a localized term $K\eps^{1-a}$ and a global term $\eps^{-1-d/\nu}=\eps^{1-p}$. We balance each of these two terms against the term $N\eps$ separately, which leads to the choices $\eps_{\rm loc}=(L_NK/N)^{1/a}$ and $\eps_{\rm glob}=(L_N/N)^{1/p}$,
each clipped to $[2\zeta_N,2B]$. When no clipping is active, these choices give
\[
 L_NK\eps_{\rm loc}^{1-a}=N\eps_{\rm loc}=L_N^{1/a}K^{1/a}N^{1-1/a},L_N\eps_{\rm glob}^{1-p}=N\eps_{\rm glob}=L_N^{1/p}N^{1-1/p},
\]
and since $a\ge2$ and $p>2$, the terms $L_N^{1/a}$ and $L_N^{1/p}$ are at most $L_N$.

If the clipping at the lower endpoint $2\zeta_N$ is active for one of these choices, say $(L_NK/N)^{1/a}<2\zeta_N$, then $L_NK<N(2\zeta_N)^a$, so the second and third terms of~\eqref{eq:regret-split-bound} evaluated at $\eps=2\zeta_N$ are $O(N\zeta_N)$ (similarly $L_N<N(2\zeta_N)^p$ when the global term is smaller); if the clipping at the upper endpoint $2B$ is active, use $R_N\le2BN$. In all cases,
\[
 R_N\le CL_N\bigl[1+N\min\{(K/N)^{1/a},N^{-1/p}\}\bigr]+CN\zeta_N
 =CL_N[1+N\rho_N]+CN\zeta_N
\]
on $\Econf[N^{-2}]$.

Next, we consider simple regret. If the stopping condition $\eps_h<2\zeta_N$ has not been met, let $h$ be the last nonempty phase. Phase $h-1$ has completed (possibly without observations) unless $h=0$, so the saved estimated optimal point has gap at most $2\eps_{h-1}=2\vartheta\eps_h$ by Lemma~\ref{lem:new-active}. If phase $h$ itself has completed, the bound is smaller, and if no phase completed, $r_N\le2B=\eps_0$. Additional empty phases after budget exhaustion leave the bands and the estimated optimal point unchanged.

Since $h$ is the last nonempty phase, all $N$ evaluations are made in phases $0,1,\ldots,h$, that is, $N=\sum_{j=0}^{h}n_j$. By~\eqref{eq:regret-phase-costs} and geometric summation, $N\le CL_N\eps_h^{-p}$ in the global analysis and $N\le CL_N\bigl[1+K\eps_h^{-a}\bigr]$ in the localized analysis. In the latter bound, the additive constant $1$ accounts for the finitely many phases with $L_\vartheta\eps_j>u_0$, to which the second bound in~\eqref{eq:regret-phase-costs} does not apply; each of these phases satisfies $n_j\le CL_N\eps_j^{-p}\le CL_N(u_0/L_\vartheta)^{-p}$ by the first bound in~\eqref{eq:regret-phase-costs}. For $N\ge2CL_N$, we then have
\[
 \eps_h\le(CL_N/N)^{1/p},\eps_h\le(2CL_NK/N)^{1/a},r_N\le2\vartheta\eps_h\le CL_N\rho_N,
\]
again bounding the logarithmic terms by $L_N$. For $N<2CL_N$ the bound $r_N\le2B\le CL_N\rho_N$ holds because $\rho_N\ge N^{-1/2}$.

Once $\eps_h<2\zeta_N$, the estimated optimal point has gap at most $4\vartheta\zeta_N$, by the third fact above.

Finally, $\Econf[N^{-2}]^c$ has probability at most $N^{-2}$ and adds at most $2BN\cdot N^{-2}=2B/N$ to expected cumulative regret and $2B\cdot N^{-2}$ to expected simple regret, which gives~\eqref{eq:new-local-rates}. The global rates~\eqref{eq:new-global-rates} follow from the global bound and $\zeta_N=\widetilde O(N^{-7/2})$. This finishes the proof.
\end{proof}

\subsection{Proof of Theorem~\ref{thm:covering-lower-alpha-large}}\label{ec:covering-lower}

% The minimax risks $\mathcal R_N^s$ and $\mathcal R_N^c$ are defined in Section~\ref{subsec:optimization-benchmarks}. The proof uses Lemma~\ref{lem:localized-power-native}, presented in Section~\ref{ec:region-lower}.

\begin{proof}[Proof of Theorem~\ref{thm:covering-lower-alpha-large}]

% \noindent\textbf{A common baseline response surface.}

Choose $c\in\X$ and $r_*>0$ with
$\overline B(c,2r_*)\subset\operatorname{int}(\X)$.
By Lemma~\ref{lem:localized-power-native},
$C_g=\|\|\cdot-c\|^\alpha\|_{\mathcal H_k(\X)}<\infty$.
Set $\mu_0=(4/c_\rho)^\alpha,\mu_K=\mu_0K^{-\alpha/d},g_K(x)=-\mu_K\|x-c\|^\alpha.$
Take
\[
 K_{\mathrm{opt}}=\max\left\{1,\left\lceil(2\mu_0C_g/B)^{d/\alpha}\right\rceil\right\}.
\]
Then $\|g_K\|_{\mathcal H_k(\X)}\le B/2$ for
$K\ge K_{\mathrm{opt}}$. Note that this norm decreases with $K$.

% ; the construction does not \textcolor{black}{place $K$ separate fixed-size peaks in the baseline}.

We will use the following elementary covering fact. A ball of radius $R_u=(u/\mu_K)^{1/\alpha} =(c_\rho/4)K^{1/d}u^{1/\alpha}$
is coverable by at most $K$ balls of radius
$r_u=c_\rho u^{1/\alpha}$. If $K<2^d$, then
$R_u<r_u/2$ and one ball suffices. If $K\ge2^d$, a maximal
$r_u$-separated packing provides a cover with at most $(1+2R_u/r_u)^d =\left(1+\tfrac12K^{1/d}\right)^d\le K$
balls. 
% This is a Euclidean ball packing argument, and uses no spatial partition.

% \noindent\textbf{Many indistinguishable modifications.}
Let $\chi\in C_c^\infty(\mathbb R^d)$ be a function taking values in $[0,1]$, with $\chi(v)=1$ for $\|v\|\le1/2$ and $\chi(v)=0$ for $\|v\|\ge1$. For a small height parameter $\Delta>0$, put
\(R=\min\{r_*,(\Delta/\mu_K)^{1/\alpha}\}, h=(2C_*\Delta/B)^{1/\nu},\)
where $C_*$ is a fixed RKHS norm bound specified below. For
sufficiently small $\Delta$, uniformly over $K\ge1$,
\begin{equation}\label{eq:cover-hard-smallness}
 h\le R/32,h\le1,
 h{\Delta}^{-1/\alpha}\le c_\rho.
\end{equation}
Indeed $\alpha>\nu$, and
$R\ge\min\{r_*,(\Delta/\mu_0)^{1/\alpha}\}$.

Take a maximal $2h$-separated set of centers $z_i$ in the set $\{x:R/2\le\|x-c\|\le3R/4\}$. The open balls
$A_i=B(z_i,h)$ are disjoint and lie in $B(c,R)$.
By maximality, the balls $B(z_i,2h)$ cover this set, so a volume comparison gives
\begin{equation}\label{eq:cover-hard-packing}
 M\ge c_d(R/h)^d
 \ge c_M\min\{
       {\Delta}^{-d/\nu},
       K{\Delta}^{-d/\nu+d/\alpha}\}.
\end{equation}
The constant $c_M>0$ is independent of $K,\Delta$.
In particular, the volume comparison gives~\eqref{eq:cover-hard-packing} with $c_d=(3/8)^d-(1/4)^d$, so $h\le R/32$ makes the first lower bound in~\eqref{eq:cover-hard-packing} at least
$12^d-8^d\ge4$, so $M\ge4$.

For $v=(x-z_i)/h$, define
\begin{equation}\label{eq:cover-hard-blend}
 f_i(x)=[1-\chi(v)]g_K(x)
       +\chi(v)[4\Delta-\Delta\|v\|^\alpha].
\end{equation}
Outside $A_i$ this equals $g_K$, and for $\|v\|\le1/2$ it equals
$4\Delta-\Delta\|v\|^\alpha$.
% This exact local replacement cancels the baseline gradient. It avoids an unintended quadratic local peak when $\alpha<2$.

% \noindent\textbf{Uniform bound on the RKHS norm.}
The modification is
\[
 f_i-g_K=\Delta\chi(v)\left[4-\|v\|^\alpha+\frac{\mu_K}{\Delta}\|z_i-c+hv\|^\alpha\right].
\]
The first two cutoff terms have finite $H^{\nu+d/2}$ norm by
Lemma~\ref{lem:localized-power-native}. For the last term write
$z_i-c=Ra_i$, $\bar h=h/R$, and $\bar\mu_K=\mu_KR^\alpha/\Delta\le1$. The normalized term is
\[
 \chi(v)\bar\mu_K\|a_i+\bar h v\|^\alpha,1/2\le\|a_i\|\le3/4,0<\bar h\le1/8.
\]
On the support of $\chi$, we have $\|v\|\le1$, and hence $3/8\le\|a_i\|-\bar h\le\|a_i+\bar h v\|\le\|a_i\|+\bar h\le7/8$. Since the Euclidean norm is smooth away from the origin, the map $v\mapsto\|a_i+\bar h v\|^\alpha$ is smooth on this support. All derivatives of this smooth compactly supported family
are uniformly bounded, and hence its $H^{\nu+d/2}$ norms are uniformly bounded.
Fourier scaling and restriction therefore give a fixed $C_*$ such
that $\|f_i-g_K\|_{\mathcal H_k(\X)} \le C_*\Delta h^{-\nu}=B/2$.
Thus any $f_i$ lies in the same ball of radius $B$, independently
of $K,N,\Delta$.

% \noindent\textbf{The covering condition at every near-optimality level.}
Since $g_K\le0$, the maximum of $f_i$ is exactly
$f_i^\star=4\Delta$, achieved only at $z_i$.
For $x\in A_i$, the gap satisfies
\[
 \bigl(f_i^\star-f_i(x)\bigr)
 =[1-\chi(v)](4\Delta+\mu_K\|x-c\|^\alpha)
       +\chi(v)\Delta\|v\|^\alpha
 \ge{\Delta}\|v\|^\alpha.
\]
Outside $A_i$, $\bigl(f_i^\star-f_i(x)\bigr)=4\Delta+\mu_K\|x-c\|^\alpha$.
For $0<u<\Delta$ it follows that
\[
 \mathcal S_{u+}(f_i)\subseteq
 \overline B(z_i,h(u/\Delta)^{1/\alpha})
 \subseteq\overline B(z_i,c_\rho u^{1/\alpha}),
\]
by~\eqref{eq:cover-hard-smallness}, and thus one ball suffices.

For $u\ge{\Delta}$, a point of $\mathcal S_{u+}(f_i)$ outside $A_i$ lies in the ball $B(c,(u/\mu_K)^{1/\alpha})$.
Points inside $A_i$ also lie in the same ball, because $A_i\subset B(c,R)$ and $R\le(\Delta/\mu_K)^{1/\alpha}\le(u/\mu_K)^{1/\alpha}$.
The elementary covering fact provides at most $K$ balls of the
required radius. Therefore any $f_i$ satisfies the prescribed
cover condition for any $u>0$, in particular for $0<u\le u_0$.

% \noindent\textbf{Testing lower bounds.}

On $A_i\subset B(c,R)$ the baseline satisfies $|g_K(x)|=\mu_K\|x-c\|^\alpha\le\mu_KR^\alpha\le{\Delta}$, so the modification is supported on $A_i$ and
\[
 0\le f_i(x)-g_K(x)=\chi(v)\bigl[4\Delta-\Delta\|v\|^\alpha-g_K(x)\bigr]
 \le4\Delta+\Delta=5\Delta\text{ for }x\in\X.
\]
Let $P_0$ denote the joint distribution of the sampling history and final output, including the procedure's randomization, under $g_K$, and let $P_i$ denote the corresponding distribution under $f_i$.

% the reference $g_K$ is common to all alternatives and need not itself belong to the subclass used to establish the lower bound.

Let $N_i=\sum_{t\le N}\mathbf1\{x_t\in A_i\}$ be the number of evaluations allocated to $A_i$. Similar to the proof of Theorem~\ref{thm:region-minimax-lower} in Section~\ref{ec:region-lower}, the decision rules of the procedure are the same under $g_K$ and under $f_i$ and therefore cancel in the likelihood ratio. Hence, factoring the conditional Gaussian densities over the $N$ evaluations and using $0\le f_i-g_K\le5\Delta\mathbf 1_{A_i}$ gives
\begin{equation}\label{eq:cover-hard-kl}
 \operatorname{KL}(P_0,P_i)
 =\frac1{2\sigma^2}\mathbb E_0\sum_{t=1}^N\{f_i(x_t)-g_K(x_t)\}^2
 \le\frac{25{\Delta}^2}{2\sigma^2}\mathbb E_0N_i.
\end{equation}

Let $\operatorname{TV}(P_0,P_i)=\sup_E|P_0(E)-P_i(E)|$ denote the total variation distance, where the supremum is over events $E$ determined by the sampling history and the final output. For any random variable $Z$ that is a function of the sampling history and the final output and takes values in $[0,1]$, Pinsker's inequality and \eqref{eq:cover-hard-kl} give
\begin{equation}\label{eq:cover-hard-tv}
|\mathbb E_iZ-\mathbb E_0Z|\le\operatorname{TV}(P_0,P_i)\le\sqrt{\tfrac12\operatorname{KL}(P_0,P_i)}\le\frac{5\Delta}{2\sigma}\sqrt{\mathbb E_0N_i}.
\end{equation}
Applying~\eqref{eq:cover-hard-tv} with $Z=\mathbf 1\{\widehat x_N\in A_i\}$, and using the disjointness of the $A_i$ (which gives $\sum_iP_0(\widehat x_N\in A_i)\le1$ and $\sum_i\mathbb E_0N_i\le N$) and the Cauchy--Schwarz inequality, we obtain
\begin{equation}\label{eq:cover-hard-success}
\begin{aligned}
 \frac1M\sum_{i=1}^M P_i(\widehat x_N\in A_i)
 &\le\frac1M\sum_{i=1}^M\Bigl[P_0(\widehat x_N\in A_i)+\sqrt{\tfrac12\operatorname{KL}(P_0,P_i)}\Bigr]\\
 &\le\frac1M+\frac{5\Delta}{2\sigma}\sqrt{\frac1M\sum_{i=1}^M\mathbb E_0N_i}
 \le\frac1M+\frac{5\Delta}{2\sigma}\sqrt{\frac NM}.
\end{aligned}
\end{equation}
The same upper bound holds for the average fraction of evaluations allocated to the correct support. Indeed, since $0\le N_i\le N$, the statistic $Z=N_i/N$ takes values in $[0,1]$, and~\eqref{eq:cover-hard-tv} gives $\mathbb E_iN_i/N\le\mathbb E_0N_i/N+(5\Delta/(2\sigma))\sqrt{\mathbb E_0N_i}$. Averaging over $i$ and using $\sum_i\mathbb E_0N_i\le N$ and the Cauchy--Schwarz inequality as in~\eqref{eq:cover-hard-success}, we obtain
\begin{equation}\label{eq:cover-hard-fraction}
\frac1M\sum_{i=1}^M\frac{\mathbb E_iN_i}{N}\le\frac1M+\frac{5\Delta}{2\sigma}\sqrt{\frac NM}.
\end{equation}

Set $\Delta=c_* \min\{(K/N)^{1/a},N^{-1/(2+d/\nu)}\}$.
By~\eqref{eq:cover-hard-packing},
\begin{equation}\label{eq:cover-hard-ratio}
 \frac{N{\Delta}^2}{M}
 \le c_M^{-1}\max\{N{\Delta}^{2+d/\nu},
                         (N/K){\Delta}^a\}
 \le c_M^{-1}\max\{c_*^{2+d/\nu},c_*^a\}.
\end{equation}

Choose $c_*>0$ small enough that the second term on the right-hand sides of~\eqref{eq:cover-hard-success} and~\eqref{eq:cover-hard-fraction}, namely $(5\Delta/(2\sigma))\sqrt{N/M}=(5/(2\sigma))\sqrt{N{\Delta}^2/M}$, is at most
$1/4$; this is possible by~\eqref{eq:cover-hard-ratio}. Since $\Delta\le c_*N^{-1/(2+d/\nu)}$ and the conditions in~\eqref{eq:cover-hard-smallness} hold for all sufficiently small $\Delta$ uniformly over $K\ge1$, there is an $N_{\mathrm{opt}}$, independent of $K$, such that~\eqref{eq:cover-hard-smallness} holds for all $N\ge N_{\mathrm{opt}}$. For such $N$, we have $M\ge4$, as shown after~\eqref{eq:cover-hard-packing}, so $1/M\le1/2$. Hence, by~\eqref{eq:cover-hard-success} and~\eqref{eq:cover-hard-fraction}, both the average success probability $M^{-1}\sum_{i=1}^MP_i(\widehat x_N\in A_i)$ and the average fraction $M^{-1}\sum_{i=1}^M\mathbb E_iN_i/N$ are at most $1/2+1/4=3/4$.
Outside $A_i$ the gap under $f_i$ is $f_i^\star-f_i(x)=4\Delta-g_K(x)\ge4\Delta$, hence
\[
 \frac1M\sum_{i=1}^M\mathbb E_ir_N
 \ge4\Delta\Bigl(1-\frac1M\sum_{i=1}^MP_i(\widehat x_N\in A_i)\Bigr)\ge{\Delta},
 \frac1M\sum_{i=1}^M\mathbb E_iR_N
 \ge4\Delta\Bigl(N-\frac1M\sum_{i=1}^M\mathbb E_iN_i\Bigr)\ge N\Delta.
\]

Taking the supremum separately gives $\sup_{f\in\mathcal F_K}\mathbb E_fr_N\ge{\Delta}=c_*\rho_N$ and $\sup_{f\in\mathcal F_K}\mathbb E_fR_N\ge N\Delta=c_*N\rho_N$. 
% The alternatives establishing the two inequalities need not coincide. 
Since the policy and the rule for the estimated optimal point were arbitrary, taking the corresponding infima proves the claimed minimax lower bounds.
\end{proof}

\subsection{Proof of Theorem~\ref{thm:joint-regret-cost}}\label{ec:joint-tradeoff}

\begin{proof}[Proof of Theorem~\ref{thm:joint-regret-cost}]
Use the baseline $g$, alternatives $f_i$, support balls
$A_i$, and tests $E_i$ constructed in Section~\ref{ec:region-lower}.
In particular, all objectives lie in the same fixed
$\mathcal F_K$, the common optimum and optimizer are unchanged,
and
\[
 g^\star-g(x)\ge T/8\text{ for }x\in A_i,
 M\ge c\eta^{-d/\nu}\min\{1,KT^{d/\alpha}\}.
\]

Let $N_i$ be the number of evaluations allocated to $A_i$ among the fixed total of
$N$ observations. As shown in the proof of Theorem~\ref{thm:region-minimax-lower}, the event
$E_i$ satisfies $P_g(E_i)\le\delta$ and
$P_{f_i}(E_i)\ge1-\delta$, where $P_g$ and $P_{f_i}$ denote the joint distributions of the sampling history and the output under $g$ and $f_i$. Similar to the proof of Theorem~\ref{thm:region-minimax-lower}, combining~\eqref{eq:region-lower-dp} and~\eqref{eq:region-lower-kl}, with $n$, $P_0$, and $P_i$ replaced by $N$, $P_g$, and $P_{f_i}$, and using $|g-f_i|\le8\eta\mathbf 1_{A_i}$, we have
% The binary data-processing inequality and the Gaussian likelihood-ratio identity under adaptive sampling give
\[
 \operatorname{kl}(\delta,1-\delta)\le\operatorname{KL}(P_g,P_{f_i}) =\frac1{2\sigma^2}\mathbb E_g \sum_{t=1}^{N}\{g(x_t)-f_i(x_t)\}^2 \le\frac{32\eta^2}{\sigma^2}\mathbb E_gN_i.
\]

% The identity follows by factoring the conditional observation
% densities over the $N$ evaluations; adaptive actions have the same
% conditional rule under both models and contribute no additional
% likelihood factors. 

The supports are disjoint, so
\[
 \mathbb E_g R_N
 \ge\frac T8\sum_{i=1}^M\mathbb E_gN_i
 \ge\frac{\sigma^2TM}{256\eta^2}
            \operatorname{kl}(\delta,1-\delta)
 \ge c\sigma^2T \min\{1,KT^{d/\alpha}\}\eta^{-(2+d/\nu)}
            \operatorname{kl}(\delta,1-\delta).
\]

Since $g\in\mathcal F_K$, taking a supremum proves the claim.
Knowing the common optimum and maximizer does not change any of
these comparisons between data distributions. All range conditions and constants are those verified in the proof of Theorem~\ref{thm:region-minimax-lower}.
\end{proof}

\subsection{Proof of Theorem~\ref{thm:budgeted-ccucb}}\label{ec:proof-budgeted}

% The proof uses Lemmas~\ref{lem:new-confidence} and~\ref{lem:new-phase-cost}, presented in Sections~\ref{ec:region-budget} and~\ref{ec:regret}, together with Propositions~\ref{thm:general-region-certificate} and~\ref{prop:phase-accuracy}.

\begin{proof}[Proof of Theorem~\ref{thm:budgeted-ccucb}]
As in Algorithm~\ref{alg:continuous-ccucb}, index the tracks by $j\in\{0,1\}$, with tolerances $T_0^{\rm alg}=0$ and $T_1^{\rm alg}=T$ and evaluation quotas $N_0=N-m\ge N/2$ and $N_1=m$. Fix $\vartheta>1$. Each track has its own phase counter $h_j$ and accuracy levels $\eps_{h_j}=2B\vartheta^{-h_j}$, and both tracks use the same maintained confidence bounds $(\ell,u)$. For each track $j$, let $\bar\eps_j$ denote the accuracy level of its last completed phase, as recorded by Algorithm~\ref{alg:band-sharing}, and let $n_{j,h}$ be the number of observations collected by track $j$ in phase $h$, including a phase truncated by the quota $N_j$. Upper-bound constants in this proof may depend on $\vartheta$. Most steps follow the proofs of Theorems~\ref{thm:region-learning-budget} and~\ref{thm:new-regret} for a single track, and we verify that they apply to each track when the bounds are shared.

We first collect the properties of the shared bounds. By Lemma~\ref{lem:new-confidence} with horizon $N$ and error probability $\bar\delta$, $\mathbb P(\Econf[\bar\delta])\ge1-\bar\delta$. Except where stated otherwise, we work on $\Econf[\bar\delta]$. Then all bands contain $f$ by~\eqref{eq:new-band}, and the maintained bounds are obtained from the initial bounds $\ell=-B\le f\le B=u$ by intersections with the bands of completed nonempty batches of either track. Hence, at all phase boundaries,
\begin{equation}\label{eq:budgeted-monotone}
\ell\le f\le u\text{ on }\X,
\end{equation}
$\ell$ can only increase, $u$ can only decrease, and the safeguard for inconsistent bounds is never triggered.

We next show that each track keeps the guarantees of a single track. Since the proof of Proposition~\ref{prop:phase-accuracy} accounts for the bands of intervening phases of the other track through Proposition~\ref{lem:shared-invariant}, Proposition~\ref{prop:phase-accuracy} applies to each track $j$ with $T=T_j^{\rm alg}$ and its own phase counter. Line~6 of Algorithm~\ref{alg:continuous-ccucb} starts a phase only if its accuracy level is at least $2\zeta_N$, so, by~\eqref{eq:region-active-quality}, any point $x$ sampled by track $j$ in phase $h$ satisfies
\begin{equation}\label{eq:budgeted-gap}
f^\star-f(x)\le T_j^{\rm alg}+L_\vartheta\eps_h.
\end{equation}
At the completion of the last completed phase of track $j$, Proposition~\ref{prop:phase-accuracy} gives the two bounds in~\eqref{eq:shared-invariant-old} with $\bar\eps=\bar\eps_j$ and $T_j=T_j^{\rm alg}$, where the second bound uses $f^\star\le\hi$. By Proposition~\ref{lem:shared-invariant}, at any later phase boundary, the bounds $(\ell,u)$ and $\lo=\max_\X\ell$ satisfy
\begin{equation}\label{eq:budgeted-new-accuracy}
u(x)-\ell(x)\le2\bar\eps_j\text{ whenever }u(x)>\lo-T_j^{\rm alg}-2\bar\eps_j,f^\star-\lo\le2\bar\eps_j.
\end{equation}
In particular, by~\eqref{eq:budgeted-monotone}, the saved estimated optimal point $\widehat x\in\argmax_{x\in\X}\ell(x)$ at any such time satisfies
\begin{equation}\label{eq:budgeted-final-point}
f^\star-f(\widehat x)\le f^\star-\lo\le2\bar\eps_j.
\end{equation}
Similarly, Lemma~\ref{lem:cap-driven-phase-cost} applies to each phase of track $j$ with $G=G_h$, $W=W_{h-1}$, and $D=D_{j,h}=\{x\in\X:G_h(x)\ge2\eps_h,\ W_{h-1}(x)\ge2\eps_h\}$, and~\eqref{eq:phase-accuracy-active-set} in the proof of Proposition~\ref{prop:phase-accuracy} gives $D_{j,h}\subseteq\mathcal S_{(T_j^{\rm alg}+L_\vartheta\eps_h)+}$. The derivations of~\eqref{eq:region-budget-phase-global}, and of~\eqref{eq:region-budget-phase-local} for a phase with $T+L_\vartheta\eps_h\le u_0$, use only these two facts, and the latter does not use $\alpha>\nu$. Hence these per-phase bounds hold for all phases of track $j$, with $n_h=n_{j,h}$, $T=T_j^{\rm alg}$, and $\delta=\bar\delta$, and they depend on the other track only through the bounds at the start of the phase. For track zero, whose tolerance is $T_0^{\rm alg}=0$, the derivation in the proof of Lemma~\ref{lem:new-phase-cost}, which uses only Lemma~\ref{lem:cap-driven-phase-cost} and~\eqref{eq:phase-accuracy-active-set} with $T=0$, gives the bounds~\eqref{eq:new-phase-cost} and~\eqref{eq:new-phase-cost-local} with $\delta$ replaced by $\bar\delta$.

We now prove~\eqref{eq:budgeted-output-accuracy}. The bounds change only when a phase completes, and Algorithm~\ref{alg:band-sharing} then recomputes and saves the output, so the final output is computed from the final bounds $(\ell,u)$ by~\eqref{eq:new-elimination} and~\eqref{eq:ccucb-joint-output}. Suppose that track one completes phase $H$, so that $\bar\eps_1\le\eps_H=\eps$. Then~\eqref{eq:budgeted-final-point} with $j=1$ gives the first bound in~\eqref{eq:budgeted-output-accuracy}. Moreover, \eqref{eq:budgeted-new-accuracy} with $j=1$ is the width bound~\eqref{eq:phase-accuracy-width-bound} in the proof of Proposition~\ref{prop:phase-accuracy}, with the final bounds in place of $(\ell_h,u_h)$ and $\bar\eps_1$ in place of $\eps_h$. The argument there for~\eqref{eq:phase-accuracy-value-bound} and for~\eqref{eq:cap-driven-region-accuracy} uses only this width bound, \eqref{eq:budgeted-monotone}, and Proposition~\ref{thm:general-region-certificate}. Hence it applies to the final bounds and gives the region bounds in~\eqref{eq:budgeted-output-accuracy}, with $4\bar\eps_1\le4\eps$ in place of $4\eps$.

We now bound the regret, first assuming $\zeta_N\le B$, and decompose $R_N=R^{(0)}+R^{(1)}$ into the cumulative regret of the observations of track zero and of track one. On $\Econf[\bar\delta]$, track zero finishes only when its quota is exhausted or its accuracy level falls below $2\zeta_N$. Hence the simple-regret argument in the proof of Theorem~\ref{thm:new-regret} applies to track zero, with $N$ replaced by $N_0\ge N/2$, $L_N$ by $\Lambda_{N,\bar\delta}$, \eqref{eq:regret-phase-costs} by these bounds for track zero, and Lemma~\ref{lem:new-active} by~\eqref{eq:budgeted-final-point} with $j=0$, which remains valid after later intersections by track one. It gives, on $\Econf[\bar\delta]$,
\begin{equation}\label{eq:budgeted-simple-bound}
r_N\le C\Lambda_{N,\bar\delta}\rho_N+C\zeta_N.
\end{equation}
With the same replacements, the three facts in the proof of Theorem~\ref{thm:new-regret} hold for the $N_0\le N$ evaluations of track zero, where~\eqref{eq:budgeted-gap} with $j=0$ gives the first fact. The argument leading to~\eqref{eq:regret-split-bound} and the same choices of $\eps_{\rm loc}$ and $\eps_{\rm glob}$ then give
\begin{equation}\label{eq:budgeted-track-zero-regret}
R^{(0)}\le C\Lambda_{N,\bar\delta}\bigl[1+N\rho_N\bigr]+CN\zeta_N\le CN\Lambda_{N,\bar\delta}\rho_N+CN\zeta_N,
\end{equation}
where the last step uses $N\rho_N\ge N^{1/2}\ge1$.

We next show that $R^{(1)}$ is at most the same term plus $CTm$. Fix a split scale $\tilde\eps\in[2\zeta_N,2B]$, and divide the evaluations of track one into three groups.
\begin{itemize}

\item \emph{Phases with $\eps_h>\max\{T,\tilde\eps\}$.} Since $T<\eps_h$, \eqref{eq:budgeted-gap} with $j=1$ bounds the gap of all points sampled in phase $h$ by $(L_\vartheta+1)\eps_h$. Moreover, $(T+\eps_h)^{d/\alpha}\le(2\eps_h)^{d/\alpha}$, so~\eqref{eq:region-budget-phase-global} and~\eqref{eq:region-budget-phase-local} for track one give the same bounds as for track zero, namely $n_{1,h}\le C\Lambda_{N,\bar\delta}\eps_h^{-p}$, and $n_{1,h}\le C\Lambda_{N,\bar\delta} K\eps_h^{-a}$ when $(L_\vartheta+1)\eps_h\le u_0$. Hence, as for the phases with $\eps_h\ge\eps$ in the proof of Theorem~\ref{thm:new-regret}, with $L_\vartheta$ replaced by $L_\vartheta+1$ and $\eps$ by $\tilde\eps$, these phases contribute at most $C\Lambda_{N,\bar\delta}[1+\min\{K\tilde\eps^{1-a},\tilde\eps^{1-p}\}]$ to $R^{(1)}$.

\item \emph{Phases with $\eps_h\le\max\{T,\tilde\eps\}$.} By~\eqref{eq:budgeted-gap} with $j=1$, all points sampled in such a phase have gap at most $T+L_\vartheta\max\{T,\tilde\eps\}\le(L_\vartheta+1)T+L_\vartheta\tilde\eps$. Since track one makes at most $m\le N$ evaluations, these phases contribute at most $CTm+CN\tilde\eps$.

\item \emph{Evaluations at the saved estimated optimal point.} As for track zero, \eqref{eq:budgeted-final-point} with $j=1$ bounds the gap of each of these at most $m\le N$ evaluations by $4\vartheta\zeta_N$, so they contribute at most $4\vartheta N\zeta_N$.

\end{itemize}
Summing the three contributions gives
\begin{equation}\label{eq:budgeted-track-one-split}
R^{(1)}\le C\Lambda_{N,\bar\delta}\bigl[1+\min\{K\tilde\eps^{1-a},\tilde\eps^{1-p}\}\bigr]+CTm+CN\tilde\eps+CN\zeta_N.
\end{equation}
Choosing $\tilde\eps$ in~\eqref{eq:budgeted-track-one-split} as $\eps_{\rm loc}$ or $\eps_{\rm glob}$ in the proof of Theorem~\ref{thm:new-regret}, with $L_N$ replaced by $\Lambda_{N,\bar\delta}$, gives
\begin{equation}\label{eq:budgeted-track-one-regret}
R^{(1)}\le C\Lambda_{N,\bar\delta}\bigl[1+N\rho_N\bigr]+CTm+CN\zeta_N\le CN\Lambda_{N,\bar\delta}\rho_N+CTm+CN\zeta_N.
\end{equation}

If $\zeta_N>B$, then $r_N\le2B<2\zeta_N$, $R^{(0)}\le2BN<2N\zeta_N$, and $R^{(1)}\le2Bm<2N\zeta_N$, so~\eqref{eq:budgeted-simple-bound}, \eqref{eq:budgeted-track-zero-regret}, and~\eqref{eq:budgeted-track-one-regret} hold trivially. Adding~\eqref{eq:budgeted-track-zero-regret} and~\eqref{eq:budgeted-track-one-regret} gives $R_N\le CN\Lambda_{N,\bar\delta}\rho_N+CTm+CN\zeta_N$ on $\Econf[\bar\delta]$. On the complement of $\Econf[\bar\delta]$, which has probability at most $\bar\delta$, we have $r_N\le2B$ and $R_N\le2BN$. Taking expectations gives~\eqref{eq:budgeted-simple-regret} and~\eqref{eq:budgeted-cumulative-regret}, whose logarithmic factor is $\Lambda_{N,\bar\delta}$. Without Assumption~\ref{as:new-cover}, the same arguments with only the global bounds give these bounds with $\rho_N$ replaced by $N^{-1/(2+d/\nu)}$.

It remains to show that the quotas in~\eqref{eq:budgeted-region-global-cost} and~\eqref{eq:budgeted-region-cost} suffice. Let $T>0$, and let $H$ be the target phase of track one, with accuracy level $\eps=\eps_H=2B\vartheta^{-H}\ge2\zeta_N$. The argument showing that phase $H$ completes in the proof of Theorem~\ref{thm:region-learning-budget} applies to track one, with the budget $N$ replaced by the quota $m$ and $n_h$ by $n_{1,h}$. Indeed, all phases $h\le H$ of track one, including a phase truncated by the quota, satisfy~\eqref{eq:region-budget-phase-global} and, under Assumption~\ref{as:new-cover} with $\alpha>\nu$ and $0<T\le u_0/2$, also~\eqref{eq:region-budget-phase-local}, with $\delta=\bar\delta$, and these bounds count only the observations of track one. Moreover, the safeguard is never triggered, line~6 of Algorithm~\ref{alg:continuous-ccucb} does not stop track one before phase $H$ completes, because $\eps_h\ge\eps_H\ge2\zeta_N$ for all $h\le H$, and Algorithm~\ref{alg:sample-allocation} tests the stopping condition before the quota. Hence, if phase $H$ of track one did not complete, all $m$ evaluations of track one would be made in phases $h\le H$, and summing as in~\eqref{eq:region-budget-sum-global} and~\eqref{eq:region-budget-sum-local} would give
\[
m\le C_1\Lambda_{N,\bar\delta}\bigl(1+\eps^{-p}\bigr),
m\le C_1\Lambda_{N,\bar\delta}\bigl[1+K\eps^{-2}+K(T+\eps)^{d/\alpha}\eps^{-p}\bigr],
\]
where $C_1$ is the constant in~\eqref{eq:region-budget-sum-global} and~\eqref{eq:region-budget-sum-local}, and the second bound requires the assumptions of~\eqref{eq:budgeted-region-cost}. Taking the constants $C$ in~\eqref{eq:budgeted-region-global-cost} and~\eqref{eq:budgeted-region-cost} larger than $C_1$, this contradicts the assumption on $m$. Therefore, phase $H$ of track one completes within its quota, and~\eqref{eq:budgeted-output-accuracy} holds by the output guarantees proved above.
\end{proof}

\subsection{Proof of Corollary~\ref{cor:joint-frontier}}\label{ec:proof-joint-frontier}

\begin{proof}[Proof of Corollary~\ref{cor:joint-frontier}]
Any procedure with valid region bounds is also an optimization policy. Thus Theorem~\ref{thm:covering-lower-alpha-large} gives an optimization lower bound, and~\eqref{eq:joint-regret-cost-lower} gives the additional lower bound from region identification. Hence, for any such policy $\pi$,
\[
 \sup_{f\in\mathcal F_K}\mathbb E_f^{\pi}R_N
 \ge c\max\Bigl\{N\rho_N,\ T\min\{1,KT^{d/\alpha}\}\eta^{-(2+d/\nu)}\operatorname{kl}(\delta,1-\delta)\Bigr\},
\]
with $\sigma^2$ absorbed into $c$, and taking the infimum over admissible policies gives~\eqref{eq:joint-frontier-lower}, with $\operatorname{kl}(\delta,1-\delta)$ kept explicitly. This finishes the proof of lower bound.

For the upper bound, as in the proof of Corollary~\ref{cor:region-minimax-matching}, $T\le T_{\mathrm{reg}}\le\min\{1,u_0/2,2048B\}$ gives $T\le u_0/2$, so that~\eqref{eq:budgeted-region-cost} applies, and $\eta\le T/128\le16B$, so that there is a phase target $\eps=\eps_H$ with $\eta/(8\vartheta)<\eps\le\eta/8$. The condition $\eta\ge16\max\{2,\vartheta\}\zeta_N$ ensures $\eps>2\zeta_N$. In this range, the proof of Corollary~\ref{cor:region-minimax-matching} absorbs $K\eps^{-2}$ into $K(T+\eps)^{d/\alpha}\eps^{-(2+d/\nu)}$. Thus the quota~\eqref{eq:joint-frontier-feasibility} completes that region phase, with region estimation error at most $4\eps\le\eta$.

This remains valid when phases are interleaved, since the other track only tightens the region track's starting bands. Theorem~\ref{thm:budgeted-ccucb} with $m=m_\eta$ then gives
\begin{equation}\label{eq:joint-frontier-upper-terms}
\mathbb E_fR_N\le C\Lambda_{N,\bar\delta}N\rho_N+CTm_\eta+CN\zeta_N+2BN\bar\delta.
\end{equation}
We bound each term on the right-hand side of~\eqref{eq:joint-frontier-upper-terms} by a constant multiple of
\[
\Lambda_{N,\bar\delta}\max\bigl\{N\rho_N,\,T\min\{1,KT^{d/\alpha}\}\eta^{-p}\bigr\}.
\]
By~\eqref{eq:joint-frontier-feasibility}, $m_\eta\le C\Lambda_{N,\bar\delta}\min\{1,KT^{d/\alpha}\}\eta^{-p}+1$. Since $T\le T_{\mathrm{reg}}\le1$ and $\eta\le T/128$, we have $\eta^{-p}\ge T^{-p}\ge1$ and $KT^{d/\alpha}\eta^{-p}\ge KT^{d/\alpha-p}\ge1$, because $K\ge1$ and $d/\alpha<d/\nu<p$. Hence $\min\{1,KT^{d/\alpha}\}\eta^{-p}\ge1$, and since $\Lambda_{N,\bar\delta}\ge1$, the additional term $1$ from rounding is absorbed:
\[
Tm_\eta\le T\bigl[C\Lambda_{N,\bar\delta}\min\{1,KT^{d/\alpha}\}\eta^{-p}+1\bigr]\le(C+1)\Lambda_{N,\bar\delta}T\min\{1,KT^{d/\alpha}\}\eta^{-p}.
\]
Moreover, $\zeta_N\le CN^{-7/2}\log^{1/2}(2N/\bar\delta)$ by~\eqref{eq:new-confidence-parameters}, so $CN\zeta_N\le C\Lambda_{N,\bar\delta}N\rho_N$, and $\bar\delta\le N^{-2}$ gives $2BN\bar\delta\le2B/N\le C\Lambda_{N,\bar\delta}N\rho_N$. Inserting these bounds into~\eqref{eq:joint-frontier-upper-terms} and using $x+y\le2\max\{x,y\}$ for $x,y\ge0$ proves~\eqref{eq:joint-frontier}.

The same argument applied to~\eqref{eq:budgeted-simple-regret} proves the simultaneous simple regret bound. All upper-bound constants may depend on the fixed $\vartheta$ and are uniform in $\delta$.
\end{proof}

\subsection{Supporting Calculations for the Accuracy--Regret Tradeoff}\label{ec:proof-quota}

This subsection provides the direct regret bound behind Corollary~\ref{cor:joint-frontier} and derives the rates and feasibility conditions stated in Section~\ref{subsec:joint-tradeoff}. Throughout, $m_{\rm opt}$ and $m_{\rm free}$ denote the quotas in~\eqref{eq:regret-preserving-quota} and~\eqref{eq:geometry-free-quota}, respectively.

\paragraph{Regret cost of a region run at a prescribed target.}
The sharper direct bound on cumulative regret underlying the optimal tradeoff can also be seen
without the two-track balancing split. Suppose $\alpha>\nu$,
$0<T\le T_{\mathrm{reg}}$, and consider the observations allocated to a region
track through completion of its first phase with accuracy level comparable
to $\eta\le T/128$. For coarse scales
$\eps_h\ge T$, each sampled point has gap at most $(L_\vartheta+1)\eps_h$ by~\eqref{eq:region-active-quality}, and \eqref{eq:region-budget-phase-global} and~\eqref{eq:region-budget-phase-local} give $n_h=\widetilde O(\min\{\eps_h^{-(2+d/\nu)},K\eps_h^{-a}\})$, where $a=2+d/\nu-d/\alpha$ because $\alpha>\nu$, so the summed weighted cost is
\begin{equation}\label{eq:quota-coarse-cost}
 \sum_{h:\,\eps_h\ge T}(L_\vartheta+1)\eps_hn_h
 =\widetilde O\bigl(\min\{T^{-1-d/\nu},KT^{1-a}\}\bigr)
 =\widetilde O\bigl(T\min\{1,KT^{d/\alpha}\}T^{-(2+d/\nu)}\bigr),
\end{equation}
because $KT^{1-a}=KT^{d/\alpha}\cdot T^{-1-d/\nu}$; the fixed initial scales are absorbed into~\eqref{eq:quota-coarse-cost}.

For fine
scales $\eta\lesssim\eps_h\le T$, \eqref{eq:region-budget-phase-local} gives
$n_h=\widetilde O(\min\{1,KT^{d/\alpha}\}\eps_h^{-(2+d/\nu)})$, including the absorbed
$K\eps_h^{-2}$ term, and each sampled point has gap at most $(L_\vartheta+1)T$ by~\eqref{eq:region-active-quality}, so
\begin{equation}\label{eq:quota-fine-cost}
 \sum_{h:\,\eta\lesssim\eps_h\le T}(L_\vartheta+1)T\,n_h
 =\widetilde O\bigl(T\min\{1,KT^{d/\alpha}\}\eta^{-(2+d/\nu)}\bigr),
\end{equation}
which also absorbs~\eqref{eq:quota-coarse-cost} because $\eta\le T$.

Together, \eqref{eq:quota-coarse-cost} and~\eqref{eq:quota-fine-cost} bound the regret contributed by the
observations needed to attain the prescribed regional accuracy. Later
observations are still collected within the fixed budget, and their
regret is controlled by the proof of Theorem~\ref{thm:budgeted-ccucb} in Section~\ref{ec:proof-budgeted}.

\paragraph{Quota tuning and tolerance feasibility.}

For $m_{\rm opt}$ and $m_{\rm free}$, respectively,
\[
 Tm\le T\cdot\frac{N\rho_N}{T}=N\rho_N,
 Tm\le T\cdot\frac{\sqrt N}{T}=\sqrt N\le N\rho_N,
\]
where the last inequality uses $K\ge1$ and $a\ge2$, so that $N\rho_N=\min\{K^{1/a}N^{1-1/a},N^{1-1/p}\}\ge N^{1/2}$. In both cases the term $CTm$ in~\eqref{eq:budgeted-cumulative-regret} is dominated by the optimization term, so Theorem~\ref{thm:budgeted-ccucb} preserves both regret orders.

In the range $\eta\le T/128$, where $K\eps^{-2}$ is absorbed as in Section~\ref{ec:region-inversion}, the sufficient region quota~\eqref{eq:budgeted-region-cost} with $\eps\asymp\eta$ reads
\[
 m\ge C\Lambda_{N,\bar\delta}\min\{1,KT^{d/\alpha}\}\eta^{-p}
 \Longleftrightarrow
 \eta\ge\Bigl[\frac{C\Lambda_{N,\bar\delta}\min\{1,KT^{d/\alpha}\}}{m}\Bigr]^{1/p},
\]
which, together with the condition $\eta\ge16\max\{2,\vartheta\}\zeta_N$, is~\eqref{eq:regret-preserving-region-accuracy}, and at fixed $K,T,\delta$ the two quota sizes yield~\eqref{eq:tuned-region-rate} and~\eqref{eq:conservative-region-rate}.

At fixed confidence, a cumulative regret bound $\sup_f\mathbb E_fR_N\le CN\rho_N$ combined with~\eqref{eq:joint-regret-cost-lower}, and the evaluation lower bound~\eqref{eq:region-minimax-lower} with $N$ evaluations available, give respectively
\[
\begin{aligned}
 cT\min\{1,KT^{d/\alpha}\}\eta^{-p}\le CN\rho_N
 &\ \Longrightarrow\
 \eta\ge c\Bigl[\frac{T\min\{1,KT^{d/\alpha}\}}{N\rho_N}\Bigr]^{1/p},\\
 c\min\{1,KT^{d/\alpha}\}\eta^{-p}\le N
 &\ \Longrightarrow\
 \eta\ge c\Bigl[\frac{\min\{1,KT^{d/\alpha}\}}{N}\Bigr]^{1/p},
\end{aligned}
\]
which are the two terms in~\eqref{eq:joint-necessary-accuracy}.

If $\eta\le C_1[\min\{1,KT^{d/\alpha}\}/N]^{1/p}$, substituting into~\eqref{eq:joint-regret-cost-lower} gives $\sup_f\mathbb E_fR_N\ge cC_1^{-p}TN$, a regret of order $TN$. If the accuracy is relaxed by a factor $\log^{3/2}(2N)$, that is,
\[
\eta\le C_1\log^{3/2}(2N)\bigl[\min\{1,KT^{d/\alpha}\}/N\bigr]^{1/p},
\]
then the same substitution gives $\sup_f\mathbb E_fR_N\ge cC_1^{-p}TN/L_N$, because $\log^{3p/2}(2N)=L_N$. For fixed $K$ and $T$, this lower bound still exceeds the optimal rate $N\rho_N$ for all sufficiently large $N$, because $N\rho_N$ grows polynomially slower than $N$. The quota $m=\lfloor N/2\rfloor$ keeps matching simple regret and region-estimation orders.

Finally, substituting $\eta=T/128$ into~\eqref{eq:regret-preserving-region-accuracy} gives $T\ge2048\max\{2,\vartheta\}\zeta_N$. Using
\[
\min\{1,KT^{d/\alpha}\}T^{-p}=\min\{T^{-p},KT^{-(p-d/\alpha)}\},
\]
the remaining quota condition becomes
\[
 C\Lambda_{N,\bar\delta}\min\{T^{-p},KT^{-(p-d/\alpha)}\}\le m
 \Longleftrightarrow
 T\ge C\min\Bigl\{\Bigl[\frac{K\Lambda_{N,\bar\delta}}{m}\Bigr]^{1/(p-d/\alpha)},
 \Bigl[\frac{\Lambda_{N,\bar\delta}}{m}\Bigr]^{1/p}\Bigr\},
\]
which is~\eqref{eq:joint-tolerance-feasibility}. Substituting $m_{\rm opt}$, $m\asymp N\rho_N/T$, into the localized bound, where $\rho_N=(K/N)^{1/a}$ with $a=p-d/\alpha$, gives
\[
 T^{a-1}\ge\frac{C\Lambda_{N,\bar\delta} K}{N\rho_N}=C\Lambda_{N,\bar\delta}\Bigl(\frac KN\Bigr)^{1-1/a}
 \Longleftrightarrow
 T\ge C\Bigl(\frac KN\Bigr)^{1/a}\Lambda_{N,\bar\delta}^{1/(a-1)},a-1=1+\frac d\nu-\frac d\alpha,
\]
and into the global bound, where $\rho_N=N^{-1/p}$,
\[
 T^{p-1}\ge\frac{C\Lambda_{N,\bar\delta}}{N\rho_N}=C\Lambda_{N,\bar\delta} N^{-1+1/p}
 \Longleftrightarrow
 T\ge CN^{-1/p}\Lambda_{N,\bar\delta}^{1/(p-1)},p-1=1+\frac d\nu;
\]
with $\bar\delta=N^{-2}$, $\Lambda_{N,\bar\delta}\asymp L_N$, which is the logarithmic inflation stated in the main text. 

\subsection{Proof of Theorem~\ref{thm:fixed-tolerance-lower}}\label{ec:fixed-tolerance-lower}
The proof reuses the construction and fixed constants from the proof of Theorem~\ref{thm:region-minimax-lower} in Section~\ref{ec:region-lower}, including the norm bound provided by Lemma~\ref{lem:localized-power-native}.
\begin{proof}[Proof of Theorem~\ref{thm:fixed-tolerance-lower}]
We use the fixed constants $x^\star,r_*,\mu,\kappa_*,C_\psi,L,\rho_0,K_{\mathrm{reg}}$, and $T_{\mathrm{reg}}$ from the proof of Theorem~\ref{thm:region-minimax-lower}. We apply that construction with $T=4\tau$, but place all perturbation centers on a single level set of the baseline, so that the single tolerance $\tau$ distinguishes all alternatives from the baseline.

Set $T=4\tau$ and
\[
k_T=\min\{K,\lfloor\kappa_*(4\tau)^{-d/\alpha}\rfloor\},
g(x)=-\mu {k_T}^{-\alpha/d}\|x-x^\star\|^\alpha,
R=(4\tau/\mu)^{1/\alpha}{k_T}^{1/d},
h=L\eta^{1/\nu}.
\]
These are the quantities $k_T$, $g$, $R$, and $h$ in the proof of Theorem~\ref{thm:region-minimax-lower} with $T=4\tau$. Take $K_1=K_{\mathrm{reg}}$ and $\tau_0\le T_{\mathrm{reg}}/4$, where $\tau_0$ may be decreased further below. Then $T\le T_{\mathrm{reg}}$, and the choices of $K_{\mathrm{reg}}$ and $T_{\mathrm{reg}}$ in the proof of Theorem~\ref{thm:region-minimax-lower} in Section~\ref{ec:region-lower} give the following bounds. Since $K\ge K_{\mathrm{reg}}$ and $\kappa_*(4\tau)^{-d/\alpha}\ge\kappa_*T_{\mathrm{reg}}^{-d/\alpha}\ge2K_{\mathrm{reg}}$, we have $k_T\ge K_{\mathrm{reg}}$. Hence, with $C_g$ as in that proof, $\|g\|_{\Hk(\X)}=\mu C_g{k_T}^{-\alpha/d}\le\mu C_gK_{\mathrm{reg}}^{-\alpha/d}\le B/2$. Moreover, $k_T\le\kappa_*(4\tau)^{-d/\alpha}$ and $\kappa_*=\mu^{d/\alpha}r_*^d$ give $R^d=(4\tau/\mu)^{d/\alpha}k_T\le r_*^d$, that is, $R\le r_*$. Finally, by~\eqref{eq:region-effective-volume} with $T=4\tau$,
\begin{equation}\label{eq:fixed-tau-volume}
R^d\asymp\min\{1,K\tau^{d/\alpha}\}
\end{equation}
uniformly over $K$ and $\tau$. The baseline has the unique maximizer $x^\star$ and the maximum value $g(x^\star)=0$.

The baseline gap is $-g(x)=4\tau(\|x-x^\star\|/R)^\alpha$. Hence it equals $\tau+4\eta$ exactly on the sphere
\[
\mathcal C_\tau=\{x:\|x-x^\star\|=r_\tau\},
r_\tau=R\Bigl(\frac{\tau+4\eta}{4\tau}\Bigr)^{1/\alpha}.
\]
Since $0<\eta\le\tau/32$, we have $(\tau+4\eta)/(4\tau)\in[1/4,9/32]\subset[1/4,1/2]$, so $4^{-1/\alpha}R\le r_\tau\le2^{-1/\alpha}R$. Thus $\mathcal C_\tau\subseteq\mathcal A_R$, where $\mathcal A_R$ is the set defined in the proof of Theorem~\ref{thm:region-minimax-lower} with the present value of $R$, and $r_\tau\asymp R$ uniformly. Since $\eta\le\tau/32=T/128$, \eqref{eq:region-lower-h} gives $h\le\rho_0R$ and $h\le1$. Because $\alpha>\nu$, decreasing $\tau_0$ also ensures $h\le r_\tau/4$.

We next place disjoint supports along $\mathcal C_\tau$. For $d\ge2$, take a maximal $2h$-separated set of centers $z_1,\ldots,z_M$ in $\mathcal C_\tau$. The open balls $A_i=B(z_i,h)$ are disjoint, and by maximality the closed balls $\overline B(z_i,2h)$ cover $\mathcal C_\tau$. Let $\omega_{d-1}$ denote the surface area of the unit sphere $\{x\in\R^d:\|x\|=1\}$, so that $\mathcal C_\tau$ has surface area $\omega_{d-1}r_\tau^{d-1}$. Since $2h\le r_\tau/2$, after a rotation about $x^\star$, the set $\overline B(z_i,2h)\cap\mathcal C_\tau$ is the graph of a function with uniformly bounded slope over a $(d-1)$-dimensional ball of radius $2h$, and hence its surface area is at most $C_dh^{d-1}$. Comparing surface areas gives
\[
\omega_{d-1}r_\tau^{d-1}
\le\sum_{i=1}^M\operatorname{area}\bigl(\overline B(z_i,2h)\cap\mathcal C_\tau\bigr)
\le MC_dh^{d-1},
\]
so that $M\ge c_d(r_\tau/h)^{d-1}\ge c(R/h)^{d-1}$. For $d=1$, take the two centers $z_1=x^\star-r_\tau$ and $z_2=x^\star+r_\tau$. They lie in $\X$ because $r_\tau\le R\le r_*$, and the intervals $A_i=(z_i-h,z_i+h)$ are disjoint because $h\le r_\tau/4$, so $M=2$. In both cases, $h=L\eta^{1/\nu}$ and~\eqref{eq:fixed-tau-volume} give
\begin{equation}\label{eq:fixed-tau-packing}
M\ge c\eta^{-(d-1)/\nu}
\bigl(\min\{1,K\tau^{d/\alpha}\}\bigr)^{(d-1)/d}.
\end{equation}

Define the alternatives $f_i(x)=g(x)+8\eta\psi((x-z_i)/h)$, $1\le i\le M$, where $\psi$ is the bump function in the proof of Theorem~\ref{thm:region-minimax-lower}. As in that proof, $\|f_i\|_{\Hk(\X)}\le B/2+8\eta C_\psi h^{-\nu}=B$. All points of $A_i$ lie within distance $h\le\rho_0R$ of $\mathcal C_\tau\subseteq\mathcal A_R$, so the choice of $\rho_0$ gives $-g(x)\in[T/8,3T/4]$ on $A_i$, whereas $8\eta\le T/16$. Hence, as in~\eqref{eq:region-gap-comparison}, every $f_i$ has the unique maximizer $x^\star$, the maximum value $f_i^\star=0$, and
\[
\tfrac12(-g(x))\le f_i^\star-f_i(x)\le-g(x)\text{ for }x\in\X.
\]
Consequently, for all $u>0$, the baseline and all alternatives satisfy $\mathcal S_{u+}\subseteq\overline B\bigl(x^\star,\tfrac14c_\rho {k_T}^{1/d}u^{1/\alpha}\bigr)$, and, as in the proof of Theorem~\ref{thm:region-minimax-lower}, this ball is covered by at most $k_T\le K$ balls of radius $c_\rho u^{1/\alpha}$. Hence $g,f_1,\ldots,f_M$ belong to the same class $\mathcal F_K$, which does not depend on $\eta$.

Consider any procedure with a fixed budget of $n$ evaluations that satisfies~\eqref{eq_region_require} at the prescribed tolerance $\tau$ with probability at least $1-\delta$ for $f\in\mathcal F_K$. For $f\in\{g,f_1,\ldots,f_M\}$, let $\mathcal E_f$ be the event that the output $(I_\tau,O_\tau)$ satisfies~\eqref{eq_region_require} at $\tau$, so that $\mathcal E_f$ has probability at least $1-\delta$ when the response surface is $f$. Let $E_i=\{z_i\in I_\tau\}$. Since $g^\star=f_i^\star=0$ and $\psi(0)=1$, the performance gap of $z_i$ is $g^\star-g(z_i)=\tau+4\eta$ under $g$ and $f_i^\star-f_i(z_i)=\tau-4\eta$ under $f_i$. Similar to the proof of Theorem~\ref{thm:region-minimax-lower}, we have $E_i\cap\mathcal E_g=\varnothing$ and $\mathcal E_{f_i}\subseteq E_i$:
\begin{itemize}
\item on $\mathcal E_g$, since $g^\star-g(z_i)>\tau$, we have $z_i\notin\mathcal S_\tau(g)$, and the inclusion $I_\tau\subseteq\mathcal S_\tau(g)$ gives $z_i\notin I_\tau$;
\item on $\mathcal E_{f_i}$, since $f_i^\star-f_i(z_i)<\tau$, we have $z_i\in\mathcal S_\tau(f_i)\subseteq O_\tau$, and since $|f_i^\star-f_i(z_i)-\tau|=4\eta>\eta$, the second condition in~\eqref{eq_region_require} gives $z_i\notin O_\tau\setminus I_\tau$, so that $z_i\in I_\tau$.
\end{itemize}
Let $P_0$ and $P_i$ be the joint distributions of the sampling history and the output, including the internal randomization of the procedure, when the response surface is $g$ and $f_i$, respectively. The two inclusions give $P_0(E_i)\le\delta$ and $P_i(E_i)\ge1-\delta$, so~\eqref{eq:region-lower-dp} gives $\KL(P_0,P_i)\ge\operatorname{kl}(\delta,1-\delta)$. Let $N_i=\sum_{t=1}^n\mathbf 1\{x_t\in A_i\}$ be the number of evaluations allocated to $A_i$. Since $0\le f_i-g\le8\eta\mathbf 1_{A_i}$, the derivation of~\eqref{eq:region-lower-kl} and~\eqref{eq:region-lower-kl-bound} applies without change and gives $\KL(P_0,P_i)\le(32\eta^2/\sigma^2)\mathbb E_0N_i$. Combining the two bounds on $\KL(P_0,P_i)$ gives
\begin{equation}\label{eq:fixed-tau-count}
\mathbb E_0N_i\ge\frac{\sigma^2}{32\eta^2}\operatorname{kl}(\delta,1-\delta),1\le i\le M.
\end{equation}

Since the supports $A_i$ are disjoint, $\sum_{i=1}^MN_i\le n$. Hence~\eqref{eq:fixed-tau-count} and~\eqref{eq:fixed-tau-packing} give
\[
n\ge\sum_{i=1}^M\mathbb E_0N_i
\ge\frac{\sigma^2M}{32\eta^2}\operatorname{kl}(\delta,1-\delta)
\ge c\sigma^2\operatorname{kl}(\delta,1-\delta)\,\eta^{-2-(d-1)/\nu}\bigl(\min\{1,K\tau^{d/\alpha}\}\bigr)^{(d-1)/d}.
\]
Since the procedure was arbitrary, this proves~\eqref{eq:fixed-tolerance-lower}. The same argument applies when the procedure is given $(x^\star,0)$ before sampling, because this information is the same under $P_0$ and $P_i$. The constants depend only on the fixed parameters and are independent of $K,\tau,\eta,\delta$ in the stated range.
\end{proof}

\subsection{Proof of Corollary~\ref{cor:fixed-tolerance-regret-lower}}\label{ec:proof-fixed-tolerance-regret-lower}
\begin{proof}[Proof of Corollary~\ref{cor:fixed-tolerance-regret-lower}]
Use the baseline $g$, the alternatives $f_i$, and the disjoint supports $A_i$ from the proof of Theorem~\ref{thm:fixed-tolerance-lower}, where $T=4\tau$. All points of $A_i$ have baseline gap $g^\star-g(x)\ge T/8=\tau/2$. Let $N_i=\sum_{t=1}^N\mathbf 1\{x_t\in A_i\}$. The procedure in the corollary satisfies~\eqref{eq_region_require} at $\tau$ with probability at least $1-\delta$ for $f\in\mathcal F_K$. Hence the argument leading to~\eqref{eq:fixed-tau-count}, with $n$ replaced by $N$, gives $\mathbb E_gN_i\ge\sigma^2\operatorname{kl}(\delta,1-\delta)/(32\eta^2)$ for each $i$. Since the supports are disjoint and all evaluations in $A_i$ contribute at least $\tau/2$ to the cumulative regret under $g$,
\[
\mathbb E_gR_N\ge\frac{\tau}{2}\sum_{i=1}^{M}\mathbb E_gN_i
\ge\frac{\sigma^2\tau M}{64\eta^2}
\operatorname{kl}(\delta,1-\delta).
\]
Since $g\in\mathcal F_K$, substituting~\eqref{eq:fixed-tau-packing} proves~\eqref{eq:fixed-tolerance-regret-lower}. As in the proof of Theorem~\ref{thm:fixed-tolerance-lower}, providing the common optimizer and optimal value before sampling does not change these distributions.
\end{proof}

\subsection{Proof of Theorem~\ref{thm:fixed-tolerance-boundary}}\label{ec:boundary}

% The proof uses Lemmas~\ref{lem:new-confidence}, \ref{lem:new-mig}, and~\ref{lem:cap-driven-phase-cost}, presented in Section~\ref{ec:region-budget}.
Define the compact sets $Q_h=\{x\in\X:g_{h-1}(x)\ge-2\eps_h\}$. 
% They are analysis sets, not algorithm inputs. 
We show in the proof that, on $\Econf$, these sets are nested, contain all global maximizers and all points sampled in phase $h$, and satisfy
\begin{equation}\label{eq:boundary-localization}
Q_h\subseteq\mathcal S_{(L_\vartheta\eps_h)+}\cup
\{x:|f^\star-f(x)-\tau|\le L_\vartheta\eps_h\}.
\end{equation}
In addition to the global budget~\eqref{eq:boundary-budget} in Theorem~\ref{thm:fixed-tolerance-boundary}, we prove the refined sufficient budget
\begin{equation}\label{eq:boundary-budget-refined}
C\Lambda_{N,\delta}\sum_{h=0}^{H}\xi_{\tau,f}(\eps_h),
\end{equation}
where $\xi_{\tau,f}(u)=\min\{u^{-(2+d/\nu)},u^{-2}J_{\tau,f}(u)\}$ and $J_{\tau,f}$ is defined in~\eqref{eq:boundary-cover-number}. That is, whenever~\eqref{eq:boundary-budget-refined} is at most $N$ and $\eps_H\ge2\zeta_N$, phase $H$ completes on $\Econf$. Here $C$ may also depend on the fixed phase ratio $\vartheta$ and the covering-radius constant $c_r$.

\begin{proof}[Proof of Theorem~\ref{thm:fixed-tolerance-boundary}]
Throughout the proof, we work on $\Econf$, which has probability at least $1-\delta$ by Lemma~\ref{lem:new-confidence}. The lemma applies because, as in Algorithm~\ref{alg:continuous-ccucb}, the screening function in~\eqref{eq:boundary-query-set} and the maintained width are fixed at the start of each phase, the sampling and stopping decisions within a phase do not use the responses of that phase, and empty phases need no confidence event.

We first show that the bounds and the screening functions are monotone over phases. The bands of completed phases are intersected, so
\begin{equation}\label{eq:boundary-monotone}
\ell_h\ge\ell_{h-1},u_h\le u_{h-1},
\lo_h\ge\lo_{h-1},\hi_h\le\hi_{h-1}.
\end{equation}
Hence $u_h-\lo_h\le u_{h-1}-\lo_{h-1}$, $\hi_h-\tau-\ell_h\le\hi_{h-1}-\tau-\ell_{h-1}$, and $u_h-\lo_h+\tau\le u_{h-1}-\lo_{h-1}+\tau$. Since $g_h$ is obtained from these three functions by the same minimum and maximum as in~\eqref{eq:boundary-query-set}, it follows that $g_h\le g_{h-1}$.

We next prove the accuracy claims~\eqref{eq:boundary-optimum-accuracy} and~\eqref{eq:boundary-certificate} for a completed phase $h$. At completion, the stopping rule~\eqref{eq:new-phase-stop} gives $a_{h,n}(x)\le2\eps_h$ for all $x\in\X$, where $a_{h,n}=\min\{G_h,W_{h-1},2\beta_Ns_n+2\zeta_N\}$. Hence, at any $x$ with $G_h(x)>2\eps_h$,
\begin{equation}\label{eq:boundary-completed-width}
u_h(x)-\ell_h(x)
\le\min\{u_{h-1}(x)-\ell_{h-1}(x),\,2\beta_Ns_n(x)+2\zeta_N\}
\le2\eps_h.
\end{equation}
For a nonempty batch, the first inequality holds because the band $[\mu_n-\beta_Ns_n-\zeta_N,\mu_n+\beta_Ns_n+\zeta_N]$ is intersected with $[\ell_{h-1},u_{h-1}]$. For an empty batch, $2\beta_Ns_0+2\zeta_N\ge2B\ge u_{h-1}-\ell_{h-1}$ because $s_0\equiv1$ and $\beta_N\ge B$, so the minimum equals the previous width. If $g_h(x)\ge0$, then $g_{h-1}(x)\ge g_h(x)\ge0$, so $G_h(x)=[g_{h-1}(x)+4\eps_h]_+\ge4\eps_h>2\eps_h$, and~\eqref{eq:boundary-completed-width} gives $u_h(x)-\ell_h(x)\le2\eps_h$.

By~\eqref{eq:boundary-query-set} with $(\ell,u,\lo,\hi)=(\ell_h,u_h,\lo_h,\hi_h)$, we have $\{g_h\ge0\}=V_h\cup C_h$, where
\[
V_h=\{x:u_h(x)\ge\lo_h\},
C_h=\{x:\ell_h(x)\le\hi_h-\tau,\ u_h(x)\ge\lo_h-\tau\}.
\]
Any global maximizer $x^\star$ satisfies $u_h(x^\star)\ge f^\star\ge\lo_h$ and hence belongs to $V_h$. If $x_c\in\argmax_\X u_h$, then $u_h(x_c)=\hi_h\ge\lo_h$, so $x_c\in V_h$, and the width bound at $x_c$ gives
\begin{equation}\label{eq:boundary-opt-width}
\lo_h\le f^\star\le \hi_h,
\hi_h\le\ell_h(x_c)+2\eps_h\le \lo_h+2\eps_h.
\end{equation}
Moreover, $\widehat x_h\in\argmax_\X\ell_h$ satisfies $f(\widehat x_h)\ge\ell_h(\widehat x_h)=\lo_h\ge f^\star-2\eps_h$. This proves~\eqref{eq:boundary-optimum-accuracy}. For~\eqref{eq:boundary-certificate}, Proposition~\ref{thm:general-region-certificate}, applied with $L=\ell_h$, $U=u_h$, $\lo=\lo_h$, and $\hi=\hi_h$, gives $I_{h,\tau}\subseteq\mathcal S_\tau\subseteq O_{h,\tau}$ by~\eqref{eq:general-region-inclusions}. If $x\in O_{h,\tau}\setminus I_{h,\tau}$, then $u_h(x)>\lo_h-\tau$ and $\ell_h(x)\le\hi_h-\tau$, so $x\in C_h$ and $u_h(x)-\ell_h(x)\le2\eps_h$. Hence
\[
\lo_h-\tau-2\eps_h<u_h(x)-2\eps_h\le\ell_h(x)\le f(x)\le u_h(x)\le\ell_h(x)+2\eps_h\le\hi_h-\tau+2\eps_h,
\]
and~\eqref{eq:boundary-opt-width} gives $f^\star-\tau-4\eps_h<f(x)\le f^\star-\tau+4\eps_h$. This proves the second inclusion in~\eqref{eq:boundary-certificate}. Only the points of $O_{h,\tau}\setminus I_{h,\tau}$, rather than the whole outer set, require this width control.

We now prove the properties of the sets $Q_h$ stated before the proof. A point $x$ is sampled in phase $h$ only if $a_{h,n}(x)>2\eps_h$, so $G_h(x)>2\eps_h$, which gives $g_{h-1}(x)>-2\eps_h$ and $x\in Q_h$. Since $g_h\le g_{h-1}$ and $\eps_{h+1}<\eps_h$, we have $Q_{h+1}=\{g_h\ge-2\eps_{h+1}\}\subseteq\{g_{h-1}\ge-2\eps_h\}=Q_h$. Moreover, if $x\notin Q_h$, then $g_j(x)\le g_{h-1}(x)<-2\eps_h<0$ for all $j\ge h-1$, so $x$ does not belong to any later set $\{g_j\ge0\}$. Any global maximizer $x^\star$ satisfies $u_{h-1}(x^\star)-\lo_{h-1}\ge f^\star-\lo_{h-1}\ge0$, so $g_{h-1}(x^\star)\ge0$ and $x^\star\in Q_h$. For $h=0$, the initial bounds $u=B$ and $\lo=-B$ give $Q_0=\X$, and $f^\star-f(x)\le2B=\eps_0\le L_\vartheta\eps_0$ for all $x\in\X$.

For $h\ge1$, let $x\in Q_h$. Since $\eps_{h-1}=\vartheta\eps_h$ and $g_{h-2}\ge g_{h-1}$, the screening function of phase $h-1$ satisfies
\[
G_{h-1}(x)\ge g_{h-2}(x)+4\eps_{h-1}
\ge g_{h-1}(x)+4\vartheta\eps_h
\ge(4\vartheta-2)\eps_h>2\eps_{h-1}.
\]
Hence~\eqref{eq:boundary-completed-width} for the completed phase $h-1$ gives $u_{h-1}(x)-\ell_{h-1}(x)\le2\eps_{h-1}=2\vartheta\eps_h$, and~\eqref{eq:boundary-opt-width} for phase $h-1$ gives $f^\star-\lo_{h-1}\le2\vartheta\eps_h$ and $\hi_{h-1}-f^\star\le2\vartheta\eps_h$. Since $g_{h-1}(x)\ge-2\eps_h$, one of the two terms in the maximum in~\eqref{eq:boundary-query-set} is at least $-2\eps_h$.
\begin{itemize}
\item If $u_{h-1}(x)-\lo_{h-1}\ge-2\eps_h$, then
\[
f^\star-f(x)\le(f^\star-\lo_{h-1})
+(\lo_{h-1}-u_{h-1}(x))
+(u_{h-1}(x)-f(x))\le(4\vartheta+2)\eps_h.
\]
\item Otherwise, $\min\{\hi_{h-1}-\tau-\ell_{h-1}(x),u_{h-1}(x)-\lo_{h-1}+\tau\}\ge-2\eps_h$, which gives
\begin{equation}\label{eq:boundary-second-term}
\lo_{h-1}-\tau-2\eps_h\le u_{h-1}(x),
\ell_{h-1}(x)\le \hi_{h-1}-\tau+2\eps_h.
\end{equation}
Combining~\eqref{eq:boundary-second-term} with $u_{h-1}(x)-\ell_{h-1}(x)\le2\vartheta\eps_h$ and with the bounds on $f^\star-\lo_{h-1}$ and $\hi_{h-1}-f^\star$ gives
\[
\begin{aligned}
f(x)&\ge u_{h-1}(x)-2\vartheta\eps_h
\ge\lo_{h-1}-\tau-(2\vartheta+2)\eps_h
\ge f^\star-\tau-(4\vartheta+2)\eps_h,\\
f(x)&\le\ell_{h-1}(x)+2\vartheta\eps_h
\le\hi_{h-1}-\tau+(2\vartheta+2)\eps_h
\le f^\star-\tau+(4\vartheta+2)\eps_h,
\end{aligned}
\]
that is, $|f^\star-f(x)-\tau|\le(4\vartheta+2)\eps_h$.
\end{itemize}
Since $4\vartheta+2\le L_\vartheta=4(\vartheta+1)$, this proves~\eqref{eq:boundary-localization}.

We next bound the number $n_h$ of evaluations in phase $h$, including an unfinished phase. Lemma~\ref{lem:cap-driven-phase-cost}, applied with $G=G_h$ and $W=W_{h-1}$, shows that all points sampled in phase $h$ lie in $D_h=\{x\in\X:G_h(x)\ge2\eps_h,\ W_{h-1}(x)\ge2\eps_h\}$, and $D_h\subseteq Q_h$ because $G_h(x)\ge2\eps_h$ implies $g_{h-1}(x)\ge-2\eps_h$. Hence~\eqref{eq:cap-driven-gamma} and the monotonicity of the information gain in the set give $n_h\eps_h^2\le4\beta_N^2C_\lambda\gamma_{n_h}(Q_h;\lambda)$, and~\eqref{eq:cap-driven-cover-cost} applies to any cover of $Q_h$. The case $n_h=0$ is trivial. For $h\ge1$, put $v=\eps_{h-1}=\vartheta\eps_h$. Then $L_\vartheta\eps_h=4(1+\vartheta^{-1})v<8v$, so~\eqref{eq:boundary-localization} gives
\[
Q_h\subseteq\mathcal S_{(8v)+}\cup\{x:|f^\star-f(x)-\tau|\le8v\},
\]
and, by~\eqref{eq:boundary-cover-number}, this set is covered by $J=J_{\tau,f}(v)$ balls of radius $r=c_rv^{1/\nu}$. Applying~\eqref{eq:cap-driven-cover-cost} with this cover, as in the derivation of~\eqref{eq:region-budget-phase-local}, and using $r^d=c_r^dv^{d/\nu}$, $v=\vartheta\eps_h$, $J\ge1$, and $v\le2B$, we obtain
\[
n_h\le C\Lambda_{N,\delta}
\bigl[1+J\eps_h^{-2}+Jr^d\eps_h^{-(2+d/\nu)}\bigr]
=C\Lambda_{N,\delta}
\bigl[1+(\vartheta^2+c_r^d\vartheta^{2+d/\nu})J\,v^{-2}\bigr]
\le C\Lambda_{N,\delta}J_{\tau,f}(v)v^{-2}.
\]
In addition, \eqref{eq:region-budget-phase-global} gives $n_h\le C\Lambda_{N,\delta}(1+\eps_h^{-(2+d/\nu)})\le C\Lambda_{N,\delta}v^{-(2+d/\nu)}$. Hence
\begin{equation}\label{eq:boundary-phase-cost}
n_h\le C\Lambda_{N,\delta}\min\{v^{-(2+d/\nu)},v^{-2}J_{\tau,f}(v)\}
=C\Lambda_{N,\delta}\,\xi_{\tau,f}(\eps_{h-1}),h\ge1.
\end{equation}
The initial phase has $n_0=0$, because $a_{0,0}\le W_{-1}=2B=\eps_0\le2\eps_0$, so the stopping rule holds at its start. 
% The shift from $\eps_h$ to $v=\eps_{h-1}$ explains why~\eqref{eq:boundary-cover-number} uses the level $8u$ for every fixed $\vartheta>1$.

Summing~\eqref{eq:boundary-phase-cost} over $h\le H$ and adding the nonnegative term $\xi_{\tau,f}(\eps_H)$ gives
\begin{equation}\label{eq:boundary-phase-sum}
\sum_{h=0}^{H}n_h
\le C\Lambda_{N,\delta}\sum_{h=1}^{H}\xi_{\tau,f}(\eps_{h-1})
\le C\Lambda_{N,\delta}\sum_{h=0}^{H}\xi_{\tau,f}(\eps_h).
\end{equation}
Since $\xi_{\tau,f}(u)\le u^{-(2+d/\nu)}$ and $\eps_h\ge\eps_H$ for $h\le H$, \eqref{eq:regret-geometric-sum} with $b=3+d/\nu$ gives
\[
\sum_{h=0}^{H}\xi_{\tau,f}(\eps_h)
\le\sum_{h=0}^{H}\eps_h^{-(2+d/\nu)}
\le\frac{\eps_H^{-(2+d/\nu)}}{1-\vartheta^{-(2+d/\nu)}}
\le C\eps_H^{-(2+d/\nu)},
\]
so~\eqref{eq:boundary-budget-refined} is at most a constant multiple of~\eqref{eq:boundary-budget}.

Finally, suppose that~\eqref{eq:boundary-budget-refined}, with a sufficiently large constant $C$, is at most $N$, but phase $H$ does not complete on $\Econf$. As in the proof of Theorem~\ref{thm:region-learning-budget}, on $\Econf$ the safeguard for inconsistent bounds is never triggered, and the refinement stop in Algorithm~\ref{alg:continuous-ccucb} does not apply to phases with $\eps_h\ge\eps_H\ge2\zeta_N$. Hence the budget is exhausted during some phase $h'\le H$, and all $N$ evaluations are made in phases $0,1,\ldots,h'$. Then~\eqref{eq:boundary-phase-sum} gives $N\le C_1\Lambda_{N,\delta}\sum_{h=0}^{H}\xi_{\tau,f}(\eps_h)$ for a constant $C_1<C$, which contradicts the assumption. Therefore phase $H$ completes. By the comparison with~\eqref{eq:boundary-budget} above, this also proves the sufficient budget~\eqref{eq:boundary-budget}. Therefore, we finish the proof. 
% An incomplete batch provides no new confidence band and hence no new accuracy claim.
\end{proof}

% ---- Commented out: approximate-maximization variant of the fixed-tolerance specialization (not used). ----
% For an optimization routine with a fixed approximation factor and guaranteed global upper bounds, assume $\kappa\in(0,1]$ and $\vartheta>2-\kappa$, and use
% $\eps_h\ge2\zeta_N/\kappa$ and replace the analysis set by
% \[
%  Q_{h,\kappa}=\{x:g_{h-1}(x)\ge-(4-2\kappa)\eps_h\}.
% \]
% It remains nested and contains any query. At its points the
% previous screening function is at least $(4\vartheta-4+2\kappa)\eps_h>2\eps_{h-1}$,
% so the preceding width bound is unchanged. The localization
% constant becomes $L_\vartheta-2\kappa\le L_\vartheta$, and the query-variance
% bound gains a factor $\kappa$. Thus the same region guarantee and
% budget hold with constants depending on fixed $\vartheta$ and $\kappa$, using
% the stopping rule with a global upper bound. The initial phase, if nonempty under
% this optimization routine, has cost at most a fixed multiple of $\log^{2+d/\nu}(2N)\log^{1+d/(2\nu)}(2N/\delta)$
% by the global phase bound, absorbed by the initial summand; no current-batch responses may
% enter either output of the optimization routine.

\subsection{Proof of Corollary~\ref{cor:fixed-tolerance-regret-upper}}\label{ec:proof-fixed-tolerance-regret-upper}
\begin{proof}[Proof of Corollary~\ref{cor:fixed-tolerance-regret-upper}]
Work on $\Econf$, which has probability at least $1-\delta$ by Lemma~\ref{lem:new-confidence}. Since~\eqref{eq:boundary-budget} is at most $N$ and $\eps_H\ge2\zeta_N$, Theorem~\ref{thm:fixed-tolerance-boundary} shows that phase $H$ completes on $\Econf$. By~\eqref{eq:boundary-optimum-accuracy} with $h=H$, the final estimated optimal point $\widehat x_N=\widehat x_H$ satisfies $r_N=f^\star-f(\widehat x_H)\le2\eps_H=2\eps$.

For the cumulative regret, let $n_h$ be the number of evaluations in phase $h$, and let $m_H=\sum_{h=0}^{H}n_h$. All points $x$ sampled in phase $h$ lie in $Q_h$, so~\eqref{eq:boundary-localization} gives $f^\star-f(x)\le\tau+L_\vartheta\eps_h$. The remaining $N-m_H$ evaluations are made at $\widehat x_H$, whose gap is at most $2\eps$. Hence
\begin{equation}\label{eq:fixed-tolerance-regret-charge}
R_N\le\sum_{h=0}^{H}(\tau+L_\vartheta\eps_h)n_h
+2\eps(N-m_H).
\end{equation}
By~\eqref{eq:boundary-phase-cost}, we have $n_0=0$ and $n_h\le C\Lambda_{N,\delta}\xi_{\tau,f}(\eps_{h-1})$ for $h\ge1$. Since $\tau+L_\vartheta\eps_h\le L_\vartheta(\tau+\eps_{h-1})$, substituting these bounds into~\eqref{eq:fixed-tolerance-regret-charge}, shifting the index by one, and adding the nonnegative term with $h=H$ give, on $\Econf$,
\begin{equation}\label{eq:fixed-tolerance-regret-reindexed}
R_N\le C\Lambda_{N,\delta}
\sum_{h=0}^{H}(\tau+\eps_h)\xi_{\tau,f}(\eps_h)+2N\eps.
\end{equation}
Since $\xi_{\tau,f}(u)\le u^{-(2+d/\nu)}$ and $\eps_h\ge\eps$ for $h\le H$, \eqref{eq:regret-geometric-sum} with $b=3+d/\nu$ and $b=2+d/\nu$ gives
\begin{equation}\label{eq:fixed-tolerance-regret-sum}
\sum_{h=0}^{H}(\tau+\eps_h)\xi_{\tau,f}(\eps_h)
\le C\bigl[\tau\eps^{-(2+d/\nu)}+\eps^{-(1+d/\nu)}\bigr]
=C(\tau+\eps)\eps^{-(2+d/\nu)}.
\end{equation}
Inserting~\eqref{eq:fixed-tolerance-regret-sum} into~\eqref{eq:fixed-tolerance-regret-reindexed} shows that, on $\Econf$, $R_N\le C\Lambda_{N,\delta}(\tau+\eps)\eps^{-(2+d/\nu)}+2N\eps$, which is at most the right-hand side of~\eqref{eq:fixed-tolerance-cumulative-regret} without the term $2BN\delta$.

On $\Econf^c$, we have $r_N\le2B$ and $R_N\le2BN$. Since $\mathbb P(\Econf^c)\le\delta$, taking expectations gives~\eqref{eq:fixed-tolerance-simple-regret} and~\eqref{eq:fixed-tolerance-cumulative-regret}.
\end{proof}

\subsection{Proof of Corollary~\ref{cor:boundary-cover-rate}}\label{ec:boundary-rate}
\begin{proof}[Proof of Corollary~\ref{cor:boundary-cover-rate}]
Choose the target phase $H$ with $\eta/(4\vartheta)<\eps_H\le\eta/4$, so that the region estimation error bound $4\eps_H$ in~\eqref{eq:boundary-certificate} is at most $\eta$. Let $u_1>0$ be such that $J_{\tau,f}(u)\le C_Ju^{-\kappa_{\rm cov}}$ for $0<u\le u_1$.

We bound the refined budget~\eqref{eq:boundary-budget-refined}. The phases with $\eps_h>u_1$ are finitely many, and each satisfies $\xi_{\tau,f}(\eps_h)\le\eps_h^{-(2+d/\nu)}\le u_1^{-(2+d/\nu)}$, so together they contribute at most $C\Lambda_{N,\delta}$ to~\eqref{eq:boundary-budget-refined}. For $\eps_h\le u_1$, we have $\xi_{\tau,f}(\eps_h)\le\eps_h^{-2}J_{\tau,f}(\eps_h)\le C_J\eps_h^{-2-\kappa_{\rm cov}}$, and~\eqref{eq:regret-geometric-sum} with $b=3+\kappa_{\rm cov}$ gives
\[
\sum_{h\le H:\,\eps_h\le u_1}\eps_h^{-2-\kappa_{\rm cov}}
\le\frac{\eps_H^{-2-\kappa_{\rm cov}}}{1-\vartheta^{-2-\kappa_{\rm cov}}}.
\]
Hence~\eqref{eq:boundary-budget-refined} is at most $C\Lambda_{N,\delta}(1+\eps_H^{-2-\kappa_{\rm cov}})\le C\Lambda_{N,\delta}\eta^{-2-\kappa_{\rm cov}}$, where the last step uses $\eps_H>\eta/(4\vartheta)$.

Take $N$ to be a sufficiently large fixed multiple of $\eta^{-2-\kappa_{\rm cov}}$ times a suitable logarithmic factor, so that $C\Lambda_{N,\delta}\eta^{-2-\kappa_{\rm cov}}\le N$ and $\eps_H\ge2\zeta_N$; the latter is possible because $\zeta_N=\widetilde O(N^{-7/2})$ for fixed $\delta$. Then phase $H$ completes on $\Econf$, as shown in the proof of Theorem~\ref{thm:fixed-tolerance-boundary}, and~\eqref{eq:boundary-certificate} gives region estimation error at most $4\eps_H\le\eta$ with $N=\widetilde O(\eta^{-2-\kappa_{\rm cov}})$.
\end{proof}

% Supporting covering proofs in the electronic companion.
% Final audit: critical small-level containment uses closed balls.

\subsection{Proof of Theorem~\ref{thm:restored-critical-lower}}\label{app:critical-proof}
\begin{proof}[Proof of Theorem~\ref{thm:restored-critical-lower}]
The proof modifies the construction in the proof of Theorem~\ref{thm:covering-lower-alpha-large} in Section~\ref{ec:covering-lower} with $\alpha=\nu$. Two changes are needed. First, Lemma~\ref{lem:localized-power-native} requires $\alpha>\nu$, so the baseline $-\mu_K\|x-c\|^\nu$ is smoothed near $c$, and the RKHS norm of the smoothed baseline grows logarithmically. Second, the local peaks of the alternatives are quadratic, which is compatible with the covering condition because $\nu\ge2$. We first consider $K\ge C_{\mathrm{crit}}[\log(2N)]^{d/\nu}$, where $C_{\mathrm{crit}}$ is specified in~\eqref{eq:restored-critical-baseline-norm} below.

Let $\phi_0$ and $w(y)=\{\phi_0(y)-\phi_0(2y)\}\|y\|^\nu$ be the functions in the proof of Lemma~\ref{lem:localized-power-native} in Section~\ref{ec:proof-localized-power}, with $\alpha=\nu$. For $0<h<1/16$, define
\begin{equation}\label{eq:restored-critical-baseline}
q_h(x)=[1-\phi_0(x/(2h))]\|x\|^\nu+\phi_0(x/(2h))h^{\nu-2}\|x\|^2,x\in\mathbb R^d.
\end{equation}
The function $q_h$ is smooth, and $q_h(x)=\|x\|^\nu$ for $\|x\|\ge2h$. Moreover, $q_h(x)\ge c_\nu\|x\|^\nu$ for all $x$, where $c_\nu=2^{2-\nu}\le1$. Indeed, $q_h(x)$ is a convex combination of $\|x\|^\nu$ and $h^{\nu-2}\|x\|^2$, and if $\|x\|<2h$, then $h^{\nu-2}\|x\|^2=(h/\|x\|)^{\nu-2}\|x\|^\nu\ge2^{2-\nu}\|x\|^\nu$ because $\nu\ge2$.

Choose $c\in\X$ and $r_*>0$ with $\overline B(c,2r_*)\subset\operatorname{int}(\X)$, and choose $\varphi_{\X}\in C_c^\infty(\mathbb R^d)$ with $\varphi_{\X}=1$ on a neighborhood of the compact set $\{x-c:x\in\X\}\cup\overline B(0,1)$. We next show that
\begin{equation}\label{eq:restored-critical-native}
\|\varphi_{\X} q_h\|_{H^s(\mathbb R^d)}\le C[1+\log(1/h)],0<h<1/16.
\end{equation}
Rewrite $\varphi_{\X} q_h=\phi_0q_h+(\varphi_{\X}-\phi_0)q_h$. Since $\varphi_{\X}-\phi_0$ vanishes on $\overline B(0,1/2)$ and $q_h(x)=\|x\|^\nu$ for $\|x\|\ge2h$, the second term equals $(\varphi_{\X}-\phi_0)\|\cdot\|^\nu$, a fixed function in $C_c^\infty(\mathbb R^d)$. For the first term, let $j_h$ be the integer with $r_{j_h}=2^{-j_h}\in[4h,8h)$. Telescoping as in~\eqref{eq:localized-power-series}, but only up to level $j_h$, gives
\begin{equation}\label{eq:restored-critical-telescope}
\phi_0q_h=\sum_{j=0}^{j_h-1}\{\phi_0(2^j\cdot)-\phi_0(2^{j+1}\cdot)\}q_h+\phi_0(2^{j_h}\cdot)\,q_h
=\sum_{j=0}^{j_h-1}2^{-j\nu}w(2^j\cdot)+\phi_0(2^{j_h}\cdot)\,q_h.
\end{equation}
The second equality holds because the $j$th summand vanishes on $\overline B(0,2^{-j-2})\supseteq\overline B(0,r_{j_h}/2)$, while $q_h(x)=\|x\|^\nu$ for $\|x\|\ge2h$ and $r_{j_h}/2\ge2h$. By~\eqref{eq:localized-power-term} with $\alpha=\nu$, each term of the last sum has $H^s$ norm at most $\|w\|_{H^s(\mathbb R^d)}$. Unlike in the proof of Lemma~\ref{lem:localized-power-native}, these bounds are not summable, which is why the decomposition is truncated at level $j_h$. For the last term in~\eqref{eq:restored-critical-telescope}, let $b=h/r_{j_h}\in(1/8,1/4]$. A change of variables in~\eqref{eq:restored-critical-baseline} gives $\phi_0(2^{j_h}x)q_h(x)={r_{j_h}}^\nu\Phi_b(x/r_{j_h})$, where
\[
\Phi_b(y)=\phi_0(y)\bigl\{[1-\phi_0(y/(2b))]\|y\|^\nu+\phi_0(y/(2b))b^{\nu-2}\|y\|^2\bigr\}.
\]
Since $1-\phi_0(y/(2b))$ vanishes for $\|y\|\le b$ and $b>1/8$, the functions $\Phi_b$ with $b\in(1/8,1/4]$ are smooth, are supported in $\overline B(0,1)$, and have derivatives of all orders bounded uniformly in $b$. Hence their $H^s$ norms are uniformly bounded, and~\eqref{eq:ec-scaling} with $t=1/r_{j_h}$ bounds the $H^s$ norm of the last term by a constant. Summing these bounds and using $j_h\le\log_2(1/(4h))\le2\log(1/h)$ proves~\eqref{eq:restored-critical-native}.

Since $\varphi_{\X}=1$ on $\{x-c:x\in\X\}$, the function $(\varphi_{\X} q_h)(\cdot-c)$ extends $x\mapsto q_h(x-c)$ from $\X$. Hence~\eqref{eq:restored-critical-native}, the translation invariance of the $H^s(\mathbb R^d)$ norm, and the norm comparison following~\eqref{eq:ec-scaling} give $\|q_h(\cdot-c)\|_{\mathcal H_k(\X)}\le C[1+\log(1/h)]$. Set
\[
\mu_0=c_\nu^{-1}(4/c_\rho)^\nu,\mu_K=\mu_0K^{-\nu/d},g_{K,h}(x)=-\mu_Kq_h(x-c).
\]
Then $g_{K,h}\le0$, $g_{K,h}(c)=0$, and $\|g_{K,h}\|_{\mathcal H_k(\X)}\le C\mu_0K^{-\nu/d}[1+\log(1/h)]$. Since $c_\nu\mu_0=(4/c_\rho)^\nu$, the ball $\overline B(c,(u/(c_\nu\mu_K))^{1/\nu})$ has radius $(c_\rho/4)K^{1/d}u^{1/\nu}$, so the elementary covering fact in the proof of Theorem~\ref{thm:covering-lower-alpha-large} with $\alpha=\nu$ covers it by at most $K$ balls of radius $c_\rho u^{1/\nu}$.

We now use the construction in the proof of Theorem~\ref{thm:covering-lower-alpha-large} with $\alpha=\nu$ and $g_{K,h}$ in place of $g_K$. For a small height parameter $\Delta>0$, let $R$, $h=(2C_*\Delta/B)^{1/\nu}$, $z_1,\dots,z_M$, and $A_i=B(z_i,h)$ be as there, where the fixed constant $C_*$ is specified in~\eqref{eq:restored-critical-modification-norm} below, and write $h=L{\Delta}^{1/\nu}$ with $L=(2C_*/B)^{1/\nu}$. In place of~\eqref{eq:cover-hard-smallness}, we require
\begin{equation}\label{eq:restored-critical-smallness}
h\le R/32,h<1/16.
\end{equation}
Since $(\Delta/\mu_K)^{1/\nu}=(\Delta/\mu_0)^{1/\nu}K^{1/d}$ has the same order in $\Delta$ as $h$, the condition $h\le R/32$ now requires $K$ to be large. Indeed, \eqref{eq:restored-critical-smallness} holds for any $K\ge K_1=\max\{1,(32L\mu_0^{1/\nu})^d\}$ and $0<\Delta\le{\Delta}_1$, where ${\Delta}_1>0$ is a fixed constant such that $L{\Delta}_1^{1/\nu}\le\min\{r_*/32,1/17\}$. The volume comparison leading to~\eqref{eq:cover-hard-packing} gives $M\ge c_d(R/h)^d$, and $M\ge4$ because $h\le R/32$. Since $R/h=\min\{r_*L^{-1}{\Delta}^{-1/\nu},L^{-1}\mu_0^{-1/\nu}K^{1/d}\}$,
\begin{equation}\label{eq:restored-critical-packing}
M\ge c_M\min\{K,{\Delta}^{-d/\nu}\},M\ge4,
\end{equation}
where $c_M=c_dL^{-d}\min\{r_*^d,\mu_0^{-d/\nu}\}$ is independent of $K$ and $\Delta$. Moreover, any $x$ in the closure of $A_i$ satisfies $\|x-c\|\ge R/2-h\ge15h>2h$, so $g_{K,h}(x)=-\mu_K\|x-c\|^\nu$.

For $v=(x-z_i)/h$, we replace the local peak $4\Delta-\Delta\|v\|^\alpha$ in~\eqref{eq:cover-hard-blend} by a quadratic one and define
\begin{equation}\label{eq:restored-critical-blend}
f_i(x)=[1-\chi(v)]g_{K,h}(x)+\chi(v)[4\Delta-\Delta\|v\|^2].
\end{equation}
With $z_i-c=Ra_i$, $\bar h=h/R\le1/32$, and $\bar\mu_K=\mu_KR^\nu/\Delta\le1$, the modification is
\[
f_i(x)-g_{K,h}(x)=\Delta\chi(v)\bigl[4-\|v\|^2+\bar\mu_K\|a_i+\bar h v\|^\nu\bigr],1/2\le\|a_i\|\le3/4.
\]
This is the modification in the proof of Theorem~\ref{thm:covering-lower-alpha-large} with $\alpha=\nu$, except that $\|v\|^\alpha$ is replaced by the smooth function $\|v\|^2$. Hence the argument in the proof of Theorem~\ref{thm:covering-lower-alpha-large}, which now does not need Lemma~\ref{lem:localized-power-native}, together with~\eqref{eq:ec-scaling} for $t=1/h$, gives a fixed $C_*$ such that
\begin{equation}\label{eq:restored-critical-modification-norm}
\|f_i-g_{K,h}\|_{\mathcal H_k(\X)}\le C_*\Delta h^{-\nu}=C_*L^{-\nu}=B/2.
\end{equation}

Since $g_{K,h}\le0$, the maximum of $f_i$ is $f_i^\star=4\Delta$, attained only at $z_i$, and, as in the proof of Theorem~\ref{thm:covering-lower-alpha-large},
\begin{equation}\label{eq:restored-critical-gap}
f_i^\star-f_i(x)=[1-\chi(v)]\bigl(4\Delta-g_{K,h}(x)\bigr)+\chi(v)\Delta\|v\|^2
\ge\begin{cases}\Delta\|v\|^2,&x\in A_i,\\ 4\Delta+c_\nu\mu_K\|x-c\|^\nu,&x\notin A_i,\end{cases}
\end{equation}
where the second case uses $q_h(x-c)\ge c_\nu\|x-c\|^\nu$. We now verify the covering condition. For $u\ge{\Delta}$, the argument in the proof of Theorem~\ref{thm:covering-lower-alpha-large}, with $(u/\mu_K)^{1/\alpha}$ replaced by $(u/(c_\nu\mu_K))^{1/\nu}\ge R$, gives $\mathcal S_{u+}(f_i)\subseteq\overline B(c,(u/(c_\nu\mu_K))^{1/\nu})$, which is covered by at most $K$ balls of radius $c_\rho u^{1/\nu}$. For $0<u<\Delta$, \eqref{eq:restored-critical-gap} shows that any point of $\mathcal S_{u+}(f_i)$ lies in $A_i$ and satisfies $\Delta\|v\|^2\le u$. Hence
\[
\mathcal S_{u+}(f_i)\subseteq\overline B\bigl(z_i,h(u/\Delta)^{1/2}\bigr)
=\overline B\bigl(z_i,Lu^{1/\nu}(u/\Delta)^{1/2-1/\nu}\bigr)
\subseteq\overline B(z_i,Lu^{1/\nu}),
\]
where the last inclusion uses $\nu\ge2$ and $u<\Delta$. The condition $h{\Delta}^{-1/\alpha}\le c_\rho$ in~\eqref{eq:cover-hard-smallness} is not available here, because $h{\Delta}^{-1/\nu}=L$ is fixed; instead, the maximal packing argument in the elementary covering fact covers $\overline B(z_i,Lu^{1/\nu})$ by at most $(1+2L/c_\rho)^d$ balls of radius $c_\rho u^{1/\nu}$. Therefore, if $K_{\mathrm{crit}}\ge\max\{K_1,(1+2L/c_\rho)^d\}$, then $\mathcal N(\mathcal S_{u+}(f_i),c_\rho u^{1/\nu})\le K$ for all $u>0$ and all $K\ge K_{\mathrm{crit}}$.

Set
\begin{equation}\label{eq:restored-critical-eps}
\Delta=c_*\min\{\sqrt{K/N},N^{-1/(2+d/\nu)}\},
\end{equation}
where $c_*\in(0,1]$ is a fixed constant specified below. Since $K\ge1$, we have $c_*N^{-1/2}\le{\Delta}\le c_*N^{-1/(2+d/\nu)}$. Hence there is a fixed $N_1\ge2$ such that $\Delta\le{\Delta}_1$ for all $N\ge N_1$, uniformly over $K\ge1$. Moreover, $h=L{\Delta}^{1/\nu}\ge Lc_*^{1/\nu}N^{-1/(2\nu)}$, so $1+\log(1/h)\le C\log(2N)$. Combining this with the bound on $\|g_{K,h}\|_{\mathcal H_k(\X)}$ gives a fixed $C_q$ such that
\begin{equation}\label{eq:restored-critical-baseline-norm}
\|g_{K,h}\|_{\mathcal H_k(\X)}\le C_q\mu_0K^{-\nu/d}\log(2N)\le B/2
\text{ if }K\ge C_{\mathrm{crit}}[\log(2N)]^{d/\nu},C_{\mathrm{crit}}=(2C_q\mu_0/B)^{d/\nu}.
\end{equation}
By~\eqref{eq:restored-critical-modification-norm} and~\eqref{eq:restored-critical-baseline-norm}, $\|f_i\|_{\mathcal H_k(\X)}\le B$. Together with the covering condition, this shows that $f_i\in\mathcal F_K$ for $i=1,\dots,M$.

On $A_i\subset B(c,R)$, the baseline satisfies $|g_{K,h}(x)|=\mu_K\|x-c\|^\nu\le\mu_KR^\nu\le{\Delta}$, so $0\le f_i-g_{K,h}\le5\Delta\mathbf 1_{A_i}$ by~\eqref{eq:restored-critical-blend}. This is the only property of the alternatives used to derive~\eqref{eq:cover-hard-kl}--\eqref{eq:cover-hard-fraction}, which therefore hold with $g_K$ replaced by $g_{K,h}$, where $P_0$ and $P_i$ are the joint distributions of the sampling history and final output under $g_{K,h}$ and $f_i$, and $N_i$ is the number of evaluations allocated to $A_i$. In particular, the average success probability $M^{-1}\sum_{i=1}^MP_i(\widehat x_N\in A_i)$ and the average fraction $M^{-1}\sum_{i=1}^M\mathbb E_iN_i/N$ are both at most $1/M+(5/(2\sigma))\sqrt{N{\Delta}^2/M}$. By~\eqref{eq:restored-critical-packing} and~\eqref{eq:restored-critical-eps},
\begin{equation}\label{eq:restored-critical-ratio}
\frac{N{\Delta}^2}{M}
\le c_M^{-1}\max\{(N/K){\Delta}^2,N{\Delta}^{2+d/\nu}\}
\le c_M^{-1}\max\{c_*^2,c_*^{2+d/\nu}\}=c_M^{-1}c_*^2.
\end{equation}
Choose $c_*=\min\{1,\sigma c_M^{1/2}/10\}$, so that $(5/(2\sigma))\sqrt{N{\Delta}^2/M}\le1/4$ by~\eqref{eq:restored-critical-ratio}. Since $M\ge4$ by~\eqref{eq:restored-critical-packing}, both averages are at most $3/4$. By~\eqref{eq:restored-critical-gap}, the gap under $f_i$ outside $A_i$ is at least $4\Delta$. Hence, as at the end of the proof of Theorem~\ref{thm:covering-lower-alpha-large},
\[
\frac1M\sum_{i=1}^M\mathbb E_ir_N\ge4\Delta\Bigl(1-\frac34\Bigr)=\Delta,
\frac1M\sum_{i=1}^M\mathbb E_iR_N\ge4\Delta\Bigl(N-\frac34N\Bigr)=N\Delta.
\]
Let $N_{\mathrm{crit}}\ge N_1$. Since the policy and the rule for the estimated optimal point were arbitrary, for any $K\ge\max\{K_{\mathrm{crit}},C_{\mathrm{crit}}[\log(2N)]^{d/\nu}\}$ and $N\ge N_{\mathrm{crit}}$,
\begin{equation}\label{eq:restored-critical-large-K}
\mathcal R_N^s(\mathcal F_K)\ge c_*\min\{\sqrt{K/N},N^{-1/(2+d/\nu)}\},
\mathcal R_N^c(\mathcal F_K)\ge c_*\min\{\sqrt{KN},N^{1-1/(2+d/\nu)}\}.
\end{equation}
This proves the claimed lower bounds without the logarithmic factor when $K\ge C_{\mathrm{crit}}[\log(2N)]^{d/\nu}$, and hence also with it, since $\log(2N)\ge1$.

It remains to consider $K_{\mathrm{crit}}\le K<C_{\mathrm{crit}}[\log(2N)]^{d/\nu}$, for which we use a two-point construction that does not depend on $K$. Take $\psi_1,\psi_2\in C_c^\infty(\mathbb R^d)$ taking values in $[0,1]$, with disjoint supports $S_1,S_2\subset\operatorname{int}(\X)$, such that, for $j=1,2$, $\psi_j$ attains its maximum value one only at a point $y_j$ and has a negative definite Hessian at $y_j$. Choose fixed $a_0>0$ and ${\Delta}_0>0$ small enough that the functions $\tilde f_j=a_0(\psi_1+\psi_2)+\Delta\psi_j$, $j=1,2$, have RKHS norm at most $B$ for all $0<\Delta\le{\Delta}_0$. Let $\Psi=\psi_1+\psi_2$. The maximum of $\tilde f_j$ is $\tilde f_j^\star=a_0+\Delta$, attained only at $y_j$, and since $S_1\cap S_2=\varnothing$,
\begin{equation}\label{eq:restored-critical-two-gap}
\tilde f_j^\star-\tilde f_j(x)=a_0\{1-\Psi(x)\}+\Delta\{1-\psi_j(x)\}
\ge\max\bigl\{a_0\{1-\Psi(x)\},\,\Delta\mathbf 1\{x\notin S_j\}\bigr\}.
\end{equation}

We verify that $\tilde f_1,\tilde f_2\in\mathcal F_K$ after enlarging $K_{\mathrm{crit}}$. By Taylor's theorem and the negative definite Hessians, there are fixed $r_0,\kappa>0$ such that $1-\Psi(x)\ge\kappa\|x-y_j\|^2$ for $\|x-y_j\|\le r_0$, $j=1,2$. By continuity and compactness, there is a fixed $\kappa'>0$ such that $1-\Psi(x)\ge\kappa'$ for $x\in\X$ outside $B(y_1,r_0)\cup B(y_2,r_0)$. Let $u_1=\min\{1,a_0\kappa'\}$ and $L_1=(a_0\kappa)^{-1/2}$. For $0<u<u_1$, \eqref{eq:restored-critical-two-gap} gives
\[
\mathcal S_{u+}(\tilde f_j)\subseteq\overline B(y_1,L_1\sqrt u)\cup\overline B(y_2,L_1\sqrt u),
\]
and $L_1\sqrt u\le L_1u^{1/\nu}$ since $\nu\ge2$ and $u<1$. Hence at most $2(1+2L_1/c_\rho)^d$ balls of radius $c_\rho u^{1/\nu}$ cover $\mathcal S_{u+}(\tilde f_j)$. For $u\ge u_1$, the domain $\X$ is covered by $\mathcal N(\X,c_\rho u_1^{1/\nu})$ balls of radius $c_\rho u_1^{1/\nu}\le c_\rho u^{1/\nu}$, and this number is fixed. Enlarging $K_{\mathrm{crit}}$ to at least $\max\{2(1+2L_1/c_\rho)^d,\mathcal N(\X,c_\rho u_1^{1/\nu})\}$, we obtain $\tilde f_1,\tilde f_2\in\mathcal F_K$ for all $K\ge K_{\mathrm{crit}}$ and $0<\Delta\le{\Delta}_0$.

Set $\Delta=\sigma N^{-1/2}$, and enlarge $N_{\mathrm{crit}}$ so that $\Delta\le{\Delta}_0$ for $N\ge N_{\mathrm{crit}}$. Let $P_1,P_2$ denote the joint distributions of the sampling history and final output under $\tilde f_1,\tilde f_2$, and let $N_j'$ be the number of evaluations allocated to $S_j$. Since $|\tilde f_1-\tilde f_2|=\Delta|\psi_1-\psi_2|\le{\Delta}$, the computation leading to~\eqref{eq:cover-hard-kl} gives $\operatorname{KL}(P_1,P_2)\le N{\Delta}^2/(2\sigma^2)=1/2$, and hence $\operatorname{TV}(P_1,P_2)\le1/2$ by Pinsker's inequality. Since $S_1\cap S_2=\varnothing$, applying the first inequality in~\eqref{eq:cover-hard-tv} to $P_1,P_2$ with $Z=\mathbf 1\{\widehat x_N\in S_2\}$ and with $Z=N_2'/N$ gives
\[
\frac12\sum_{j=1}^2P_j(\widehat x_N\in S_j)\le\frac12\bigl[1+\operatorname{TV}(P_1,P_2)\bigr]\le\frac34,
\frac12\sum_{j=1}^2\frac{\mathbb E_jN_j'}N\le\frac12\bigl[1+\operatorname{TV}(P_1,P_2)\bigr]\le\frac34.
\]
By~\eqref{eq:restored-critical-two-gap}, the gap under $\tilde f_j$ is at least $\Delta$ outside $S_j$. Hence, as at the end of the proof of Theorem~\ref{thm:covering-lower-alpha-large},
\[
\frac12\sum_{j=1}^2\mathbb E_jr_N\ge{\Delta}\Bigl(1-\frac34\Bigr)=\frac{\sigma}{4\sqrt N},
\frac12\sum_{j=1}^2\mathbb E_jR_N\ge{\Delta}\Bigl(N-\frac34N\Bigr)=\frac{\sigma\sqrt N}4.
\]
If $K<C_{\mathrm{crit}}[\log(2N)]^{d/\nu}$, then $[\log(2N)]^{-d/(2\nu)}\sqrt{K/N}<\sqrt{C_{\mathrm{crit}}/N}$ and $[\log(2N)]^{-d/(2\nu)}\sqrt{KN}<\sqrt{C_{\mathrm{crit}}N}$. Since the policy and the rule for the estimated optimal point were arbitrary, the claimed lower bounds therefore hold with the constant $\min\{c_*,\sigma/(4\sqrt{C_{\mathrm{crit}}})\}$ for all $K\ge K_{\mathrm{crit}}$ and $N\ge N_{\mathrm{crit}}$.

Finally, suppose that $\nu$ is an even integer. Then $\|x\|^\nu=(x_1^2+\dots+x_d^2)^{\nu/2}$ is a polynomial, and we replace $q_h$ by $q(x)=\|x\|^\nu$ throughout. The function $\varphi_{\X} q$ is a fixed element of $C_c^\infty(\mathbb R^d)$, so $\|q(\cdot-c)\|_{\mathcal H_k(\X)}\le C$ independently of $h$, and~\eqref{eq:restored-critical-baseline-norm} is replaced by $\|\mu_Kq(\cdot-c)\|_{\mathcal H_k(\X)}\le C\mu_0K^{-\nu/d}\le B/2$, which holds for any $K\ge K_{\mathrm{crit}}$ after enlarging $K_{\mathrm{crit}}$. The other steps use only $q(x)\ge c_\nu\|x\|^\nu$ and $q(x)=\|x\|^\nu$ for $\|x\|\ge2h$, which hold since $c_\nu\le1$. Hence~\eqref{eq:restored-critical-large-K} holds for any $K\ge K_{\mathrm{crit}}$ and $N\ge N_{\mathrm{crit}}$, which proves the claimed lower bounds without the logarithmic factor.
\end{proof}

\subsection{Proof of Proposition~\ref{prop:general-periodic-cover-lower}}\label{app:periodic-proof}
\begin{proof}[Proof of Proposition~\ref{prop:general-periodic-cover-lower}]
The proof uses a periodic baseline whose near-optimal sets are unions of small balls around the points of a lattice, and it places the alternatives at these points. In addition to $C$, we let $c$ denote a positive constant depending only on the fixed parameters, whose value may change between occurrences.

For an integer $m\ge1$, put
\begin{equation}\label{eq:periodic-baseline}
A_m=A_0m^{-s},g_m(x)=-A_m\sum_{r=1}^d\sin^2(\pi mx_r),x\in\X,
\end{equation}
where $A_0>0$ is a fixed constant specified below. We first bound the RKHS norm of $g_m$. Extend $g_m$ to $\mathbb R^d$ by the same formula, and let $\varphi_{\X}\in C_c^\infty(\mathbb R^d)$ be a fixed function with $\varphi_{\X}=1$ on a neighborhood of $\X$. For any $\omega\in\mathbb R^d$, the Fourier transform of $x\mapsto{\varphi_{\X}}(x)e^{i\omega^\top x}$ is $\widehat{\varphi_{\X}}(\cdot-\omega)$. Hence, by the inequality $1+\|v+\omega\|^2\le2(1+\|v\|^2)(1+\|\omega\|^2)$,
\begin{equation}\label{eq:periodic-harmonic}
\|\varphi_{\X} e^{i\omega^\top\cdot}\|_{H^s(\mathbb R^d)}\le2^{s/2}(1+\|\omega\|^2)^{s/2}\|\varphi_{\X}\|_{H^s(\mathbb R^d)}\le C(1+\|\omega\|)^s.
\end{equation}
Since $\sin^2(\pi mt)=\{2-e^{2\pi imt}-e^{-2\pi imt}\}/4$, the function $g_m$ is the constant $-dA_m/2$ plus $2d$ harmonics with frequencies $\pm2\pi me_r$, $r=1,\dots,d$, each with coefficient $A_m/4$. By~\eqref{eq:periodic-harmonic}, $\|\varphi_{\X} g_m\|_{H^s(\mathbb R^d)}\le CA_m(1+2\pi m)^s\le CA_mm^s=CA_0$. Since $\varphi_{\X} g_m$ extends $g_m$ from $\X$, the norm comparison following~\eqref{eq:ec-scaling} gives a fixed $C_0$ such that $\|g_m\|_{\mathcal H_k(\X)}\le C_0A_0$. We set $A_0=B/(2C_0)$, so that $\|g_m\|_{\mathcal H_k(\X)}\le B/2$ for any $m\ge1$.

We next show that the near-optimal sets of $g_m$ satisfy the covering condition with $K$ replaced by a fixed multiple of $m^{sd/\alpha}$. Since $\X$ has nonempty interior, there is a fixed $m_0$ such that $\X$ contains a point of the lattice $m^{-1}\mathbb Z^d$ for $m\ge m_0$, and hence $g_m^\star=0$. Let $0<u<A_m$ and $x\in\mathcal S_{u+}(g_m)$. The inequality $\sin^2(\pi t)\ge4\operatorname{dist}(t,\mathbb Z)^2$ gives
\[
u\ge-g_m(x)=A_m\sum_{r=1}^d\sin^2(\pi mx_r)
\ge4A_m\sum_{r=1}^d\operatorname{dist}(mx_r,\mathbb Z)^2
=4A_mm^2\operatorname{dist}(x,m^{-1}\mathbb Z^d)^2.
\]
Hence $x$ lies within distance $R_u=\sqrt u/(2m\sqrt{A_m})<1/(2m)$ of a point of $m^{-1}\mathbb Z^d$. Since $\X$ is bounded, at most $C_\X m^d$ lattice points lie within distance $1/(2m)$ of $\X$. By the maximal packing argument in the elementary covering fact in the proof of Theorem~\ref{thm:covering-lower-alpha-large}, each ball of radius $R_u$ is covered by at most $(1+2R_u/(c_\rho u^{1/\alpha}))^d\le C[1+(R_u/(c_\rho u^{1/\alpha}))^d]$ balls of radius $c_\rho u^{1/\alpha}$. Therefore,
\begin{equation}\label{eq:periodic-cover-count}
\begin{aligned}
\mathcal N(\mathcal S_{u+}(g_m),c_\rho u^{1/\alpha})
&\le CC_\X m^d\left[1+\frac{u^{d/2}}{2^dm^dA_m^{d/2}c_\rho^du^{d/\alpha}}\right]
\le C\left[m^d+A_m^{-d/2}u^{d/2-d/\alpha}\right]\\
&\le C\left[m^d+A_m^{-d/\alpha}\right]
=C\left[m^d+A_0^{-d/\alpha}m^{sd/\alpha}\right],
\end{aligned}
\end{equation}
where the second line uses $u^{d/2-d/\alpha}\le A_m^{d/2-d/\alpha}$, which holds since $\alpha\ge2$ and $u<A_m$, and then $A_m=A_0m^{-s}$. For $u\ge A_m$, the bounded domain $\X$ is covered by at most $C(1+A_m^{-d/\alpha})$ balls of radius $c_\rho A_m^{1/\alpha}\le c_\rho u^{1/\alpha}$, which is bounded by the last expression in~\eqref{eq:periodic-cover-count} after increasing $C$. Since $s>\nu\ge\alpha$, we have $m^d\le m^{sd/\alpha}$. Hence there is a fixed $C_{\rm cov}\ge1$ such that, for any $m\ge m_0$,
\begin{equation}\label{eq:general-periodic-cover}
\mathcal N(\mathcal S_{u+}(g_m),c_\rho u^{1/\alpha})\le C_{\rm cov}m^{sd/\alpha},u>0.
\end{equation}

We now construct the alternatives. Fix a closed ball $\overline B(z_0,r_*)\subset\operatorname{int}(\X)$ and a constant $0<r_1<1/4$. After enlarging $m_0$, for any $m\ge m_0$, the ball $\overline B(z_0,r_*)$ contains $M$ points $z_1,\dots,z_M$ of $m^{-1}\mathbb Z^d$ with
\begin{equation}\label{eq:periodic-count}
c_\X m^d\le M\le C_\X m^d,
\end{equation}
and the balls $\mathcal B_i=B(z_i,r_1/m)$ are contained in $\X$. These balls are disjoint, because distinct lattice points are at distance at least $1/m>2r_1/m$. Let $\psi\in C_c^\infty(\mathbb R^d)$ be supported in the unit ball, with $0\le\psi\le1$, $\psi(0)=1$, and $\psi(v)<1$ for $v\ne0$. For a height parameter $\Delta>0$, define
\begin{equation}\label{eq:periodic-alternative}
f_i(x)=g_m(x)+\Delta\psi(m(x-z_i)/r_1).
\end{equation}
Since $g_m\le0=g_m(z_i)$, the unique maximizer of $f_i$ is $z_i$, with $f_i^\star=\Delta$, and
\begin{equation}\label{eq:periodic-gap}
f_i^\star-f_i(x)=-g_m(x)+\Delta[1-\psi(m(x-z_i)/r_1)]\ge\max\bigl\{-g_m(x),\,\Delta\mathbf 1\{x\notin \mathcal B_i\}\bigr\}.
\end{equation}
Since $g_m^\star=0$, \eqref{eq:periodic-gap} gives $\mathcal S_{u+}(f_i)\subseteq\mathcal S_{u+}(g_m)$ for any $u>0$. Hence~\eqref{eq:general-periodic-cover} holds with $g_m$ replaced by $f_i$, regardless of the ratio $\Delta/A_m$.

The scaling inequality~\eqref{eq:ec-scaling} with $t=m/r_1\ge1$ and the norm comparison following it give a fixed $C_\psi$ such that $\|f_i-g_m\|_{\mathcal H_k(\X)}\le C_\psi{\Delta} m^\nu$. Set
\begin{equation}\label{eq:periodic-eps}
\Delta=\frac\sigma2\sqrt{\frac MN}.
\end{equation}
By~\eqref{eq:periodic-count}, $\|f_i-g_m\|_{\mathcal H_k(\X)}\le C_\psi\sigma C_\X^{1/2}m^s/(2\sqrt N)$, which is at most $B/2$ whenever
\begin{equation}\label{eq:periodic-norm-condition}
m^{2s}\le c_BN,c_B=\bigl\{B/(C_\psi\sigma C_\X^{1/2})\bigr\}^2.
\end{equation}
Under~\eqref{eq:periodic-norm-condition}, $\|f_i\|_{\mathcal H_k(\X)}\le B$. Together with~\eqref{eq:general-periodic-cover} for $f_i$, this shows that $f_i\in\mathcal F_K$ whenever $C_{\rm cov}m^{sd/\alpha}\le K$.

Let $P_0$ and $P_i$ denote the joint distributions of the sampling history and final output under $g_m$ and $f_i$, respectively, and let $N_i$ be the number of evaluations allocated to $\mathcal B_i$. By~\eqref{eq:periodic-alternative}, $0\le f_i-g_m\le{\Delta}\mathbf 1_{\mathcal B_i}$. Hence the computation leading to~\eqref{eq:cover-hard-kl} gives
\[
\operatorname{KL}(P_0,P_i)\le\frac{{\Delta}^2}{2\sigma^2}\mathbb E_0N_i.
\]
Since the $\mathcal B_i$ are disjoint, $\sum_iP_0(\widehat x_N\in \mathcal B_i)\le1$ and $\sum_i\mathbb E_0N_i\le N$. Therefore, the arguments leading to~\eqref{eq:cover-hard-success} and~\eqref{eq:cover-hard-fraction}, with $5\Delta$ replaced by $\Delta$, give
\begin{equation}\label{eq:periodic-success}
\frac1M\sum_{i=1}^MP_i(\widehat x_N\in \mathcal B_i)\le\frac1M+\frac{\Delta}{2\sigma}\sqrt{\frac NM}=\frac1M+\frac14,
\frac1M\sum_{i=1}^M\frac{\mathbb E_iN_i}N\le\frac1M+\frac14,
\end{equation}
where the equality uses~\eqref{eq:periodic-eps}. If $M\ge4$, both averages in~\eqref{eq:periodic-success} are at most $1/2$. By~\eqref{eq:periodic-gap}, any evaluated point or estimated optimal point outside $\mathcal B_i$ has gap at least $\Delta$ under $f_i$. Hence
\[
\frac1M\sum_{i=1}^M\mathbb E_ir_N\ge{\Delta}\Bigl(1-\frac12\Bigr)=\frac{\Delta}2,
\frac1M\sum_{i=1}^M\mathbb E_iR_N\ge{\Delta}\Bigl(N-\frac N2\Bigr)=\frac{N\Delta}2.
\]

It remains to choose $m$. Let $c_{\mathrm{per}}>0$ be a fixed constant small enough that $C_{\rm cov}c_{\mathrm{per}}^{sd/\alpha}\le1$ and $c_{\mathrm{per}}^{2s}\le c_B$, and take
\begin{equation}\label{eq:periodic-m}
m=\left\lfloor c_{\mathrm{per}}\min\left\{K^{\alpha/(sd)},N^{1/(2s)}\right\}\right\rfloor.
\end{equation}
Then $C_{\rm cov}m^{sd/\alpha}\le K$ and $m^{2s}\le c_BN$, so $f_i\in\mathcal F_K$ for $i=1,\dots,M$. Choose fixed $K_{\mathrm{per}}$ and $N_{\mathrm{per}}$ such that $K\ge K_{\mathrm{per}}$ and $N\ge N_{\mathrm{per}}$ imply $m\ge m_0$ and $c_\X m^d\ge4$, so that $M\ge4$ by~\eqref{eq:periodic-count}. Since $\lfloor y\rfloor\ge y/2$ for $y\ge1$, \eqref{eq:periodic-count} and~\eqref{eq:periodic-m} give
\[
\sqrt{\frac MN}\ge\sqrt{\frac{c_\X m^d}N}\ge c\min\left\{K^{\alpha/(2s)}N^{-1/2},N^{d/(4s)-1/2}\right\}.
\]
Since $2s=2\nu+d$ and $d/(4s)-1/2=-\nu/(2\nu+d)$, \eqref{eq:periodic-eps} gives
\[
\frac{\Delta}2\ge c\min\{K^{\alpha/(2\nu+d)}N^{-1/2},N^{-\nu/(2\nu+d)}\},
\frac{N\Delta}2\ge c\min\{K^{\alpha/(2\nu+d)}N^{1/2},N^{(\nu+d)/(2\nu+d)}\}.
\]
Since the policy and the rule for the estimated optimal point were arbitrary, taking the corresponding infima proves the claimed lower bounds. All class parameters and constants are fixed as $K$ and $N$ vary.
\end{proof}

\subsection{Proof of Proposition~\ref{prop:covering-saturation}}\label{app:saturation-proof}
\begin{proof}[Proof of Proposition~\ref{prop:covering-saturation}]
We construct a family of compactly supported bumps of height $\Delta\asymp N^{-1/(2+d/\nu)}$ and width $h\asymp{\Delta}^{1/\nu}$, show that they all belong to $\mathcal F_K$ once $K$ exceeds the threshold in~\eqref{eq:covering-saturation-threshold}, and then apply the testing argument in the proof of Theorem~\ref{thm:covering-lower-alpha-large} with the zero function as the reference.

Let
\[
\psi(v)=
\begin{cases}
\exp\{-\|v\|^2/(1-\|v\|^2)\},&\|v\|<1,\\
0,&\|v\|\ge1.
\end{cases}
\]
This function belongs to $C_c^\infty(\mathbb R^d)$ and satisfies $\psi(0)=1$ and $0\le\psi\le1$. For $\|v\|<1$, we have $\psi(v)\le\exp(-\|v\|^2)$ and $1-e^{-t}\ge(1-e^{-1})t$ for $t\in[0,1]$ by the concavity of $t\mapsto1-e^{-t}$, while $1-\psi(v)=1$ for $\|v\|\ge1$. Hence,
\begin{equation}\label{eq:saturation-bump-gap}
1-\psi(v)\ge c_\psi\min\{\|v\|^2,1\},c_\psi=1-e^{-1}.
\end{equation}

The inequality~\eqref{eq:ec-scaling} and the norm comparison following it give a fixed constant $C_\psi$ such that, for any center $z\in\mathbb R^d$, $\Delta>0$, and $0<h\le1$,
\begin{equation}\label{eq:saturation-norm}
\|\Delta\psi((\cdot-z)/h)\|_{\mathcal H_k(\X)}\le C_\psi{\Delta} h^{-\nu}.
\end{equation}
Set
\begin{equation}\label{eq:saturation-eps}
\Delta=c_*N^{-1/(2+d/\nu)},h=(2C_\psi{\Delta}/B)^{1/\nu},
\end{equation}
where $c_*\in(0,1]$ is a fixed constant specified below, so that $C_\psi{\Delta} h^{-\nu}=B/2$. Fix a ball $B(c_0,r_*)$ with $\overline B(c_0,r_*)\subset\X$, and take a maximal $2h$-separated set of centers $c_1,\dots,c_M$ in $B(c_0,r_*/2)$. For $h<r_*/2$, the open balls $A_i=B(c_i,h)$ are disjoint and lie in $\X$. By maximality, the balls $B(c_i,2h)$ cover $B(c_0,r_*/2)$, so a volume comparison gives
\begin{equation}\label{eq:saturation-packing}
M\ge(r_*/(4h))^d=(r_*/4)^d(2C_\psi/B)^{-d/\nu}{\Delta}^{-d/\nu}.
\end{equation}
Hence there is a fixed $N_{\mathrm{sat}}$ such that $h\le\min\{1,r_*/4\}$ and $M\ge4$ for all $N\ge N_{\mathrm{sat}}$. Define
\[
f_i(x)=\Delta\psi((x-c_i)/h),i=1,\dots,M.
\]
By~\eqref{eq:saturation-norm}, $\|f_i\|_{\mathcal H_k(\X)}\le B/2$. The maximum of $f_i$ is $f_i^\star=\Delta$, attained only at $c_i$, and by~\eqref{eq:saturation-bump-gap},
\begin{equation}\label{eq:saturation-gap}
f_i^\star-f_i(x)=\Delta\{1-\psi((x-c_i)/h)\}\ge c_\psi{\Delta}\min\{\|x-c_i\|^2/h^2,1\},
\end{equation}
where the left-hand side equals $\Delta$ for $x\notin A_i$.

We next verify the covering condition for any $u>0$. Let $0<u<\Delta$. Since the gap equals $\Delta$ outside $A_i$, we have $\mathcal S_{u+}(f_i)\subseteq A_i$, and
\[
\mathcal S_{u+}(f_i)\subseteq\overline B(c_i,R_u),R_u=h\sqrt{u/(c_\psi{\Delta})}.
\]
Indeed, if $u<c_\psi{\Delta}$, then~\eqref{eq:saturation-gap} forces $\|x-c_i\|^2/h^2\le u/(c_\psi{\Delta})$ for any $x\in\mathcal S_{u+}(f_i)$, and if $c_\psi{\Delta}\le u<\Delta$, then $A_i\subseteq\overline B(c_i,R_u)$ since $R_u\ge h$. By the maximal packing argument in the elementary covering fact in the proof of Theorem~\ref{thm:covering-lower-alpha-large}, a ball of radius $R$ is covered by at most $(1+2R/r)^d\le C[1+(R/r)^d]$ balls of radius $r$. Since $\alpha\ge2$, the exponent $d/2-d/\alpha$ is nonnegative, and $u<\Delta$ gives $u^{d/2-d/\alpha}\le{\Delta}^{d/2-d/\alpha}$. Therefore, for $0<u<\Delta$,
\begin{equation}\label{eq:saturation-small-u}
\mathcal N(\mathcal S_{u+}(f_i),c_\rho u^{1/\alpha})
\le C\left[1+\frac{h^du^{d/2-d/\alpha}}{c_\psi^{d/2}c_\rho^d{\Delta}^{d/2}}\right]
\le C\bigl[1+h^d{\Delta}^{-d/\alpha}\bigr]
\le C\bigl[1+{\Delta}^{d/\nu-d/\alpha}\bigr],
\end{equation}
where the last step uses $h^d=(2C_\psi/B)^{d/\nu}{\Delta}^{d/\nu}$ by~\eqref{eq:saturation-eps}. For $u\ge{\Delta}$, the boundedness of $\X$ gives
\begin{equation}\label{eq:saturation-large-u}
\mathcal N(\mathcal S_{u+}(f_i),c_\rho u^{1/\alpha})
\le\mathcal N(\X,c_\rho{\Delta}^{1/\alpha})
\le C\bigl[1+{\Delta}^{-d/\alpha}\bigr].
\end{equation}
Since $\Delta\le1$, the right-hand side of~\eqref{eq:saturation-small-u} is at most that of~\eqref{eq:saturation-large-u}, which is at most $2C{\Delta}^{-d/\alpha}$. Hence there is a fixed $C_{\mathrm{sat}}$ such that
\begin{equation}\label{eq:saturation-class}
f_i\in\mathcal F_K\text{ for }i=1,\dots,M,\text{ whenever }K\ge C_{\mathrm{sat}}{\Delta}^{-d/\alpha}.
\end{equation}

Fix any policy using $N$ evaluations and any rule for the estimated optimal point. Let $P_i$ denote the joint distribution of the sampling history and final output under $f_i$, and let $P_0$ denote the corresponding distribution under the zero function, which serves only as a reference and need not belong to $\mathcal F_K$. Let $N_i=\sum_{t=1}^N\mathbf 1\{x_t\in A_i\}$. Since $0\le f_i\le{\Delta}\mathbf 1_{A_i}$, the computation leading to~\eqref{eq:cover-hard-kl} gives
\[
\operatorname{KL}(P_0,P_i)=\frac1{2\sigma^2}\mathbb E_0\sum_{t=1}^Nf_i(x_t)^2\le\frac{{\Delta}^2}{2\sigma^2}\mathbb E_0N_i.
\]
Since the $A_i$ are disjoint, $\sum_iP_0(\widehat x_N\in A_i)\le1$ and $\sum_i\mathbb E_0N_i\le N$. Hence the arguments leading to~\eqref{eq:cover-hard-success} and~\eqref{eq:cover-hard-fraction}, with $5\Delta$ replaced by $\Delta$, give
\begin{equation}\label{eq:saturation-success}
\frac1M\sum_{i=1}^MP_i(\widehat x_N\in A_i)\le\frac1M+\frac{\Delta}{2\sigma}\sqrt{\frac NM},
\frac1M\sum_{i=1}^M\frac{\mathbb E_iN_i}N\le\frac1M+\frac{\Delta}{2\sigma}\sqrt{\frac NM}.
\end{equation}
By~\eqref{eq:saturation-packing} and~\eqref{eq:saturation-eps},
\[
\Delta\sqrt{\frac NM}\le C\sqrt{N{\Delta}^{2+d/\nu}}=Cc_*^{1+d/(2\nu)}.
\]
Choose $c_*\in(0,1]$ small enough that $Cc_*^{1+d/(2\nu)}/(2\sigma)\le1/4$. Since $M\ge4$, both averages in~\eqref{eq:saturation-success} are then at most $1/2$. Since the gap under $f_i$ equals $\Delta$ outside $A_i$,
\[
\frac1M\sum_{i=1}^M\mathbb E_ir_N\ge{\Delta}\Bigl(1-\frac12\Bigr)=\frac{\Delta}2,
\frac1M\sum_{i=1}^M\mathbb E_iR_N\ge{\Delta}\Bigl(N-\frac N2\Bigr)=\frac{N\Delta}2.
\]

Finally, by~\eqref{eq:saturation-eps}, the condition $K\ge C_{\mathrm{sat}}{\Delta}^{-d/\alpha}$ in~\eqref{eq:saturation-class} is $K\ge C_{\mathrm{sat}}c_*^{-d/\alpha}N^{d/[\alpha(2+d/\nu)]}$, which is~\eqref{eq:covering-saturation-threshold} with $C=C_{\mathrm{sat}}c_*^{-d/\alpha}$. Under this condition, $\Delta/2=(c_*/2)N^{-1/(2+d/\nu)}$ and $N\Delta/2=(c_*/2)N^{1-1/(2+d/\nu)}$. Since the policy and the rule for the estimated optimal point were arbitrary, taking the corresponding infima proves the claimed lower bounds with $c=c_*/2$. All constants and $N_{\mathrm{sat}}$ depend only on the fixed model and covering parameters.
\end{proof}

\subsection{Proof of Proposition~\ref{thm:general-region-certificate}}\label{ec:proof-region-certificate}
\begin{proof}[Proof of Proposition~\ref{thm:general-region-certificate}]
The argument is deterministic and uses only the bounds $L\le f\le U$ on $\X$ and $\lo\le f^\star\le\hi$. Recall that $\mathcal S_v=\{x:f^\star-f(x)<v\}$.

We first prove~\eqref{eq:general-region-inclusions}. Let $\tau\ge0$. If $x\in I_\tau$, then, by~\eqref{eq:general-region-sets} and $\hi\ge f^\star$,
\[
f(x)\ge L(x)>\hi-\tau\ge f^\star-\tau,
\]
so $f^\star-f(x)<\tau$ and $x\in\mathcal S_\tau$. If $x\in\mathcal S_\tau$, then, by $f^\star\ge\lo$,
\[
U(x)\ge f(x)>f^\star-\tau\ge\lo-\tau,
\]
so $x\in O_\tau$ by~\eqref{eq:general-region-sets}.

We next prove~\eqref{eq:general-region-pointwise}, which implies~\eqref{eq:general-region-ambiguity}. Let $x\in O_\tau\setminus I_\tau$, so that $L(x)\le\hi-\tau$ and $U(x)>\lo-\tau$ by~\eqref{eq:general-region-sets}. Since $w(x)=U(x)-L(x)$,
\[
f(x)\le U(x)=L(x)+w(x)\le\hi-\tau+w(x),
f(x)\ge L(x)=U(x)-w(x)>\lo-\tau-w(x).
\]
Subtracting these bounds from $f^\star$ gives~\eqref{eq:general-region-pointwise}. Since $\hi-f^\star\le\hi-\lo=\Gamma$, $f^\star-\lo\le\Gamma$, and $w(x)\le w_\tau$ on $O_\tau$, \eqref{eq:general-region-pointwise} gives
\[
\tau-\eta_\tau\le f^\star-f(x)<\tau+\eta_\tau,
\]
which is~\eqref{eq:general-region-ambiguity}.

It remains to prove the first and last inclusions in~\eqref{eq:general-region-value-sandwich}, since the middle ones are~\eqref{eq:general-region-inclusions}. For the first inclusion, we may assume $\tau-\eta_\tau>0$, since otherwise $\mathcal S_{\tau-\eta_\tau}=\varnothing$. Let $x\in\mathcal S_{\tau-\eta_\tau}$. Then $x\in\mathcal S_\tau\subseteq O_\tau$ by~\eqref{eq:general-region-inclusions}, so $w(x)\le w_\tau$, and
\[
L(x)=U(x)-w(x)\ge f(x)-w_\tau>f^\star-\tau+\eta_\tau-w_\tau=f^\star-\tau+\Gamma\ge\hi-\tau,
\]
where the last step uses $f^\star\ge\lo$. Hence $x\in I_\tau$. For the last inclusion, let $x\in O_\tau$. Then $U(x)>\lo-\tau$ and $w(x)\le w_\tau$, so
\[
f(x)\ge L(x)=U(x)-w(x)>\lo-\tau-w_\tau.
\]
Hence $f^\star-f(x)<(f^\star-\lo)+\tau+w_\tau\le\tau+\eta_\tau$, that is, $x\in\mathcal S_{\tau+\eta_\tau}$.

Finally, if $0\le\tau\le T$, then $\lo-\tau\ge\lo-T$, so $O_\tau\subseteq O_T$ by~\eqref{eq:general-region-sets}. Hence, if $w_T$ bounds $U-L$ on $O_T$, then $w_\tau=w_T$ is admissible for all $0\le\tau\le T$, and~\eqref{eq:general-region-ambiguity} and~\eqref{eq:general-region-value-sandwich} hold with the same error $\eta_\tau=\Gamma+w_T$ for all these tolerances. We also note that, at $\tau=0$, both $I_0$ and $\mathcal S_0$ are empty, since $L\le f\le f^\star\le\hi$. Thus $O_0$ may be replaced by $\varnothing$ without affecting~\eqref{eq:general-region-inclusions}--\eqref{eq:general-region-value-sandwich}.
\end{proof}

\subsection{Proof of Proposition~\ref{lem:shared-invariant}}\label{ec:proof-shared-bands}
\begin{proof}[Proof of Proposition~\ref{lem:shared-invariant}]
When a nonempty phase of either track completes, its band is intersected with the maintained confidence bounds, and an empty phase leaves them unchanged. Hence the confidence bounds can only shrink, and
\begin{equation}\label{eq:shared-invariant-monotone}
u(x)\le u^{\rm old}(x),\ell(x)\ge\ell^{\rm old}(x)\text{ for }x\in\X,
\lo=\max_\X\ell\ge\max_\X\ell^{\rm old}=\lo^{\rm old}.
\end{equation}
Let $x$ satisfy $u(x)>\lo-T_j-2\bar\eps$. By~\eqref{eq:shared-invariant-monotone},
\[
u^{\rm old}(x)\ge u(x)>\lo-T_j-2\bar\eps\ge\lo^{\rm old}-T_j-2\bar\eps,
\]
so the first bound in~\eqref{eq:shared-invariant-old} applies at $x$. Using~\eqref{eq:shared-invariant-monotone} again,
\[
u(x)-\ell(x)\le u^{\rm old}(x)-\ell^{\rm old}(x)\le2\bar\eps.
\]
For the optimal-value bound, \eqref{eq:shared-invariant-monotone} and the second bound in~\eqref{eq:shared-invariant-old} give $f^\star-\lo\le f^\star-\lo^{\rm old}\le2\bar\eps$.
\end{proof}

\subsection{Proof of Proposition~\ref{prop:phase-accuracy}}\label{ec:proof-phase-accuracy}

\begin{proof}[Proof of Proposition~\ref{prop:phase-accuracy}]
Throughout the proof, we work on $\Econf$, on which $\ell\le f\le u$ for the confidence bounds $(\ell,u)$ at any time. For a phase $h$, let $\ell_{h-1},u_{h-1}$, $\lo_{h-1}$, $W_{h-1}$, and $G_h$ be the quantities in~\eqref{eq:new-frozen-cap}, which are fixed at the start of the phase. If the phase completes, let $\ell_h,u_h$ be the confidence bounds immediately after its completion, with $\lo_h$ and $\hi_h$ as in~\eqref{eq:new-elimination}. Since the bands are only intersected, $u_h\le u_{h-1}$, $\ell_h\ge\ell_{h-1}$, and $\lo_h\ge\lo_{h-1}$.

We first prove the width bound for a completed phase $h$ with $\eps_h\ge2\zeta_N$. Let $x$ satisfy $u_h(x)>\lo_h-T-2\eps_h$. Since $u_h\le u_{h-1}$ and $\lo_h\ge\lo_{h-1}$,
\begin{equation}\label{eq:phase-accuracy-width}
\begin{aligned}
&u_{h-1}(x)\ge u_h(x)>\lo_h-T-2\eps_h\ge\lo_{h-1}-T-2\eps_h,\\
&G_h(x)=[u_{h-1}(x)-\lo_{h-1}+T+4\eps_h]_+>2\eps_h,
\end{aligned}
\end{equation}
where the second line follows from the first and~\eqref{eq:new-frozen-cap}.
By Lemma~\ref{lem:cap-driven-phase-cost}, the width after the completed phase is at most $2\eps_h$ wherever $G_h>2\eps_h$. Hence we obtain the width bound
\begin{equation}\label{eq:phase-accuracy-width-bound}
u_h(x)-\ell_h(x)\le2\eps_h\text{ whenever }u_h(x)>\lo_h-T-2\eps_h.
\end{equation}
This also covers a phase completed without any evaluation, which leaves the bounds unchanged.

We next prove the bounds on the optimal value and on the region outputs. Let $x_c$ be a maximizer of $u_h$ over $\X$, so that $\hi_h=u_h(x_c)$. Since $\ell_h\le f\le u_h$, we have $\lo_h\le f^\star\le\hi_h$. In particular, $u_h(x_c)\ge\lo_h>\lo_h-T-2\eps_h$, so~\eqref{eq:phase-accuracy-width-bound} applies at $x_c$ and gives
\[
\hi_h=u_h(x_c)\le\ell_h(x_c)+2\eps_h\le\lo_h+2\eps_h.
\]
Hence we obtain the optimal-value bound
\begin{equation}\label{eq:phase-accuracy-value-bound}
\hi_h-\lo_h\le2\eps_h.
\end{equation}
Since $\widehat x_h$ maximizes $\ell_h$, we also have $f(\widehat x_h)\ge\ell_h(\widehat x_h)=\lo_h\ge\hi_h-2\eps_h\ge f^\star-2\eps_h$. For $0<\tau\le T$, the sets $I_{h,\tau}$ and $O_{h,\tau}$ in~\eqref{eq:ccucb-joint-output} are the sets in~\eqref{eq:general-region-sets} with $L=\ell_h$, $U=u_h$, $\lo=\lo_h$, and $\hi=\hi_h$. Any $x\in O_{h,\tau}$ satisfies $u_h(x)>\lo_h-\tau\ge\lo_h-T-2\eps_h$, so~\eqref{eq:phase-accuracy-width-bound} gives $u_h-\ell_h\le w_\tau=2\eps_h$ on $O_{h,\tau}$. Therefore, Proposition~\ref{thm:general-region-certificate} applies with $\Gamma=\hi_h-\lo_h\le2\eps_h$ by~\eqref{eq:phase-accuracy-value-bound} and $\eta_\tau=\Gamma+w_\tau\le4\eps_h$, and~\eqref{eq:general-region-inclusions} and~\eqref{eq:general-region-ambiguity} give~\eqref{eq:cap-driven-region-accuracy}. At $\tau=0$, both output sets and $\mathcal S_0$ are empty, so~\eqref{eq:cap-driven-region-accuracy} holds trivially.

Finally, we prove~\eqref{eq:region-active-quality}. Let $x$ be a point sampled in phase $h$ with $\eps_h\ge2\zeta_N$, possibly in a phase truncated by the budget. If $h=0$, then $f^\star-f(x)\le2\sup_\X|f|\le2B=\eps_0\le T+L_\vartheta\eps_0$, since $|f(x)|\le\|f\|_{\Hk(\X)}\sqrt{k(x,x)}\le B$. Let $h\ge1$. Phase $h-1$ has completed, because a phase starts only after the previous phase of the same track completes, and $\eps_{h-1}=\vartheta\eps_h\ge2\zeta_N$. Let $(\ell^{\rm end},u^{\rm end})$ denote the confidence bounds immediately after the completion of phase $h-1$, with $\lo^{\rm end}=\max_\X\ell^{\rm end}$ and $\hi^{\rm end}=\max_\X u^{\rm end}$. By~\eqref{eq:phase-accuracy-width-bound} and~\eqref{eq:phase-accuracy-value-bound}, applied to phase $h-1$, these bounds satisfy~\eqref{eq:shared-invariant-old} with $(\ell^{\rm old},u^{\rm old})=(\ell^{\rm end},u^{\rm end})$, $\bar\eps=\eps_{h-1}$, and $T_j=T$, where we use $f^\star-\lo^{\rm end}\le\hi^{\rm end}-\lo^{\rm end}$. The bounds $(\ell_{h-1},u_{h-1})$ fixed at the start of phase $h$ in~\eqref{eq:new-frozen-cap} are obtained from $(\ell^{\rm end},u^{\rm end})$ by intersecting the bands of the completed phases of the other track that intervene, if any. Hence, by Proposition~\ref{lem:shared-invariant}, the two bounds in~\eqref{eq:shared-invariant-old} also hold for $(\ell_{h-1},u_{h-1},\lo_{h-1})$:
\begin{equation}\label{eq:phase-accuracy-start}
u_{h-1}(z)-\ell_{h-1}(z)\le2\vartheta\eps_h\text{ whenever }u_{h-1}(z)>\lo_{h-1}-T-2\vartheta\eps_h,
f^\star-\lo_{h-1}\le2\vartheta\eps_h.
\end{equation}
By Lemma~\ref{lem:cap-driven-phase-cost}, $G_h(x)>2\eps_h$, and hence, by~\eqref{eq:new-frozen-cap}, $u_{h-1}(x)>\lo_{h-1}-T-2\eps_h>\lo_{h-1}-T-2\vartheta\eps_h$. Thus the first bound in~\eqref{eq:phase-accuracy-start} applies at $x$, and since $f\ge\ell_{h-1}$,
\[
\begin{aligned}
f^\star-f(x)
&=(f^\star-\lo_{h-1})+(\lo_{h-1}-u_{h-1}(x))+(u_{h-1}(x)-f(x))\\
&<2\vartheta\eps_h+(T+2\eps_h)+2\vartheta\eps_h=T+(4\vartheta+2)\eps_h.
\end{aligned}
\]
Since $4\vartheta+2<L_\vartheta=4(\vartheta+1)$, this proves~\eqref{eq:region-active-quality}. The same argument, with $G_h(x)\ge2\eps_h$ in place of $G_h(x)>2\eps_h$, shows that the closed set $D_h$ in Lemma~\ref{lem:cap-driven-phase-cost} satisfies 
\begin{equation}\label{eq:phase-accuracy-active-set}
D_h=\{x\in\X:G_h(x)\ge2\eps_h,\ W_{h-1}(x)\ge2\eps_h\}\subseteq\mathcal S_{(T+L_\vartheta\eps_h)+}.
\end{equation}
This is~\eqref{eq:region-budget-active}, now also when phases of the other track intervene.
\end{proof}

\subsection{Proof of Proposition~\ref{prop:full-budget-global-regret}}\label{ec:proof-full-budget}
\begin{proof}[Proof of Proposition~\ref{prop:full-budget-global-regret}]
Fix $\vartheta>1$. All upper-bound constants in this proof will depend on $\vartheta$. Since $m=N$, all evaluations are made by the region track, whose tolerance is $T$. The optimization track has quota zero and, as noted in the proof of Theorem~\ref{thm:region-learning-budget}, does not affect the analysis. The proof follows the proof of Theorem~\ref{thm:new-regret} in Section~\ref{ec:regret}, with the gap bound $L_\vartheta\eps_h$ of each sampled point replaced by $T+L_\vartheta\eps_h$.

With $\delta=N^{-2}$, the confidence parameters in~\eqref{eq:new-confidence-parameters} give $\zeta_N\le CN^{-7/2}\log^{1/2}(2N)$. If $\zeta_N>B$, then $r_N\le2B<2\zeta_N$ and $R_N\le2BN<2N\zeta_N$, and these bounds are covered by the final step below. Hence assume $\zeta_N\le B$. Then any nonempty phase has $\eps_h\ge2\zeta_N$, because Algorithm~\ref{alg:continuous-ccucb} allocates all remaining evaluations to the saved estimated optimal point once the accuracy level falls below $2\zeta_N$. Work on $\Econf[N^{-2}]$, on which the safeguard for inconsistent bounds is never triggered. With $\delta=N^{-2}$, the factor $\Lambda_{N,\delta}$ in~\eqref{eq:region-budget-phase-global} is at most $CL_N$ by~\eqref{eq:ec-log-bounds}, and $1+\eps_h^{-p}\le C\eps_h^{-p}$ because $\eps_h\le2B$. Hence the argument leading to~\eqref{eq:region-budget-phase-global} in the proof of Theorem~\ref{thm:region-learning-budget}, which applies to any phase with $\eps_h\ge2\zeta_N$, gives
\begin{equation}\label{eq:full-budget-phase-cost}
n_h\le CL_N\eps_h^{-p}
\end{equation}
for all nonempty phases, including a phase truncated by the budget. Moreover, by Proposition~\ref{prop:phase-accuracy}, any completed phase $h$ satisfies $f^\star-f(\widehat x_h)\le2\eps_h$, and, by~\eqref{eq:region-active-quality}, any point sampled in phase $h$ has gap at most $T+L_\vartheta\eps_h$.

We first consider cumulative regret. Split the nonempty phases at any scale $\eps\in[2\zeta_N,2B]$. As in the proof of Theorem~\ref{thm:new-regret}, all evaluations are made in a phase with $\eps_h\ge\eps$, in a phase with $2\zeta_N\le\eps_h<\eps$, or at the saved estimated optimal point after the accuracy level falls below $2\zeta_N$. By~\eqref{eq:region-active-quality}, an evaluation in phase $h$ contributes at most $T+L_\vartheta\eps_h$ to $R_N$, which is less than $T+L_\vartheta\eps$ if $\eps_h<\eps$. By the third fact in the proof of Theorem~\ref{thm:new-regret}, with Proposition~\ref{prop:phase-accuracy} in place of Lemma~\ref{lem:new-active}, each evaluation at the saved estimated optimal point contributes less than $4\vartheta\zeta_N$. Since there are at most $N$ evaluations in total,
\begin{equation}\label{eq:full-budget-split}
R_N\le NT+L_\vartheta\sum_{h:\,\eps_h\ge\eps}\eps_hn_h+L_\vartheta N\eps+4\vartheta N\zeta_N.
\end{equation}
By~\eqref{eq:full-budget-phase-cost} and~\eqref{eq:regret-geometric-sum} with $b=p$,
\[
\sum_{h:\,\eps_h\ge\eps}\eps_hn_h\le CL_N\sum_{h:\,\eps_h\ge\eps}\eps_h^{1-p}\le CL_N\eps^{1-p},
\]
so that
\begin{equation}\label{eq:full-budget-split-bound}
R_N\le NT+CL_N\eps^{-(1+d/\nu)}+CN\eps+CN\zeta_N.
\end{equation}
Choose $\eps=(L_N/N)^{1/p}$, clipped to $[2\zeta_N,2B]$. Since $(3+3d/(2\nu))/(2+d/\nu)=3/2$, when no clipping is active,
\[
\eps=\log^{3/2}(2N)N^{-1/(2+d/\nu)},L_N\eps^{1-p}=N\eps=\log^{3/2}(2N)N^{1-1/(2+d/\nu)}.
\]
If the clipping at the lower endpoint $2\zeta_N$ is active, then $L_N<N(2\zeta_N)^p$, so the second and third terms of~\eqref{eq:full-budget-split-bound} evaluated at $\eps=2\zeta_N$ are $O(N\zeta_N)$. If the clipping at the upper endpoint $2B$ is active, then $N(L_N/N)^{1/p}>2BN\ge R_N$. In all cases, on $\Econf[N^{-2}]$,
\begin{equation}\label{eq:full-budget-cumulative}
R_N\le NT+C\log^{3/2}(2N)N^{1-1/(2+d/\nu)}+CN\zeta_N.
\end{equation}

We next consider simple regret. Suppose first that no phase with accuracy level below $2\zeta_N$ is reached, and let $h$ be the last nonempty phase. Phase $h-1$ has completed unless $h=0$, possibly without evaluations, so by Proposition~\ref{prop:phase-accuracy}, the saved estimated optimal point has gap at most $2\eps_{h-1}=2\vartheta\eps_h$. If phase $h$ itself has completed, the bound is smaller, and if no phase has completed, $r_N\le2B=\eps_0$. Additional empty phases after the budget is exhausted leave the bands and the estimated optimal point unchanged. Hence $r_N\le2\vartheta\eps_h$ in all cases. Since $h$ is the last nonempty phase, all $N$ evaluations are made in phases $0,1,\ldots,h$, so~\eqref{eq:full-budget-phase-cost} and~\eqref{eq:regret-geometric-sum} with $b=p+1$ give
\[
N=\sum_{j=0}^hn_j\le CL_N\sum_{j=0}^h\eps_j^{-p}\le CL_N\eps_h^{-p}.
\]
Hence $\eps_h\le(CL_N/N)^{1/p}$, and
\begin{equation}\label{eq:full-budget-simple}
r_N\le2\vartheta\eps_h\le C(L_N/N)^{1/p}=C\log^{3/2}(2N)N^{-1/(2+d/\nu)}.
\end{equation}
If a phase with accuracy level below $2\zeta_N$ is reached, then, by the third fact in the proof of Theorem~\ref{thm:new-regret}, with Proposition~\ref{prop:phase-accuracy} in place of Lemma~\ref{lem:new-active}, the saved estimated optimal point has gap less than $4\vartheta\zeta_N$.

% Local line-breaking adjustment within this proof; the text is unchanged.
\emergencystretch=2em
Finally, $\Econf[N^{-2}]^c$ has probability at most $N^{-2}$ by Lemma~\ref{lem:new-confidence}, and $r_N\le2B$ and $R_N\le2BN$ always hold. Hence this event adds at most $2BN^{-2}$ to $\mathbb Er_N$ and at most $2B/N$ to $\mathbb ER_N$. Since $\zeta_N\le CN^{-7/2}\log^{1/2}(2N)$, the terms $4\vartheta\zeta_N$ and $2BN^{-2}$ are at most constant multiples of $\log^{3/2}(2N)N^{-1/(2+d/\nu)}$, and the terms $CN\zeta_N$ and $2B/N$ are at most constant multiples of $\log^{3/2}(2N)N^{1-1/(2+d/\nu)}$. Combining~\eqref{eq:full-budget-simple} and~\eqref{eq:full-budget-cumulative} with the deterministic bound $R_N\le2BN$ gives~\eqref{eq:joint-global-regret}.
\end{proof}

\section{Proofs of Lemmas}\label{ec:lemma-proofs}

This section proves the lemmas stated in the main text and in the subsections of Section~\ref{ec:proofs}.

\subsection{Proof of Lemma~\ref{lem:region-clipped-gap}}\label{ec:proof-clipped-gap}
\begin{proof}[Proof of Lemma~\ref{lem:region-clipped-gap}]
We first prove the measurability of $\widehat d(x)$ and then the two inequalities in~\eqref{eq:region-clipped-gap}. Fix $x\in\X$. For $q\in\mathbb Q\cap[0,T]$, let $Z_q=q\mathbf 1\{x\in I_q\}+T\mathbf 1\{x\notin I_q\}$, which is a random variable because membership in $I_q$ is measurable. Then
\[
\widehat d(x)=\min\Bigl\{T,\inf_{q\in\mathbb Q\cap[0,T]}Z_q\Bigr\}
\]
is the infimum of countably many random variables and is therefore measurable.

Now work on $\mathcal E_f$, and let $d(x)=f^\star-f(x)$. For the lower bound, if $x\in I_q$ for some $q\in\mathbb Q\cap[0,T]$, then $x\in\mathcal S_q$ by the first inclusion in~\eqref{eq:region-minimax-loss}, that is, $q>d(x)$. Hence all elements of the set in the definition of $\widehat d(x)$ are at least $\min\{d(x),T\}$, and so is its infimum.

For the upper bound, we always have $\widehat d(x)\le T$. If $d(x)+\eta\ge T$, then $\widehat d(x)\le T=\min\{d(x)+\eta,T\}$. Suppose now that $d(x)+\eta<T$. For $j\ge1$, the interval $(d(x)+\eta,\min\{T,d(x)+\eta+1/j\})$ has positive length, so it contains a rational number $q_j$. Since $q_j>d(x)$, we have $x\in\mathcal S_{q_j}\subseteq O_{q_j}$ by~\eqref{eq:region-minimax-loss}. Since $|f^\star-f(x)-q_j|=q_j-d(x)>\eta$, the second condition in~\eqref{eq:region-minimax-loss} gives $x\notin O_{q_j}\setminus I_{q_j}$. Hence $x\in I_{q_j}$, and
\[
\widehat d(x)\le q_j<d(x)+\eta+1/j.
\]
Letting $j\to\infty$ gives $\widehat d(x)\le d(x)+\eta=\min\{d(x)+\eta,T\}$. Since $x\in\X$ was arbitrary, \eqref{eq:region-clipped-gap} holds simultaneously for all $x\in\X$ on $\mathcal E_f$. 
% The argument does not require the output sets to be nested in $\tau$.
Finally, for an $(\eta,T,\delta)$-valid procedure, \eqref{eq:region-minimax-loss} holds for all $\tau\in[0,T]$ on an event of probability at least $1-\delta$, which can serve as $\mathcal E_f$.
\end{proof}

\subsection{Proof of Lemma~\ref{lem:localized-power-native}}\label{ec:proof-localized-power}
\begin{proof}[Proof of Lemma~\ref{lem:localized-power-native}]
Fix a smooth radial function $\phi_0$ on $\R^d$ with $0\le\phi_0\le1$, $\phi_0(x)=1$ for $\|x\|\le1/2$, and $\phi_0(x)=0$ for $\|x\|\ge1$. We decompose
\[
\varphi(x)\|x\|^\alpha=\varphi(x)\phi_0(x)\|x\|^\alpha+\varphi(x)\{1-\phi_0(x)\}\|x\|^\alpha.
\]
The second term is smooth and compactly supported, because $1-\phi_0$ vanishes when $\|x\|\le1/2$ and $\|x\|^\alpha$ is smooth away from the origin. Hence it belongs to $H^s(\R^d)$. Since multiplication by a smooth compactly supported function preserves $H^s(\R^d)$, it remains to show that $\phi_0(x)\|x\|^\alpha$ belongs to $H^s(\R^d)$.

Let $w(x)=\{\phi_0(x)-\phi_0(2x)\}\|x\|^\alpha$. Since $\phi_0(x)-\phi_0(2x)=0$ when $\|x\|\le1/4$ or $\|x\|\ge1$, the function $w$ is smooth, vanishes unless $1/4\le\|x\|\le1$, and belongs to $H^s(\R^d)$. For $x\ne0$, telescoping and $\phi_0(2^Jx)\to0$ as $J\to\infty$ give
\begin{equation}\label{eq:localized-power-series}
\phi_0(x)\|x\|^\alpha=\sum_{j=0}^{\infty}\{\phi_0(2^jx)-\phi_0(2^{j+1}x)\}\|x\|^\alpha=\sum_{j=0}^{\infty}2^{-j\alpha}w(2^jx).
\end{equation}
For $j\ge0$, the scaling inequality~\eqref{eq:ec-scaling} with $t=2^j$ gives
\begin{equation}\label{eq:localized-power-term}
\|2^{-j\alpha}w(2^j\cdot)\|_{H^s(\R^d)}\le2^{-j\alpha}\,2^{j\nu}\|w\|_{H^s(\R^d)}=2^{-j(\alpha-\nu)}\|w\|_{H^s(\R^d)}.
\end{equation}
Since $\alpha>\nu$, the norms in~\eqref{eq:localized-power-term} are summable over $j$, so the series in~\eqref{eq:localized-power-series} converges in $H^s(\R^d)$. Its limit agrees almost everywhere with the pointwise limit $\phi_0(x)\|x\|^\alpha$, which proves the first claim.

For the second claim, fix $c\in\R^d$ and a bounded domain $\mathcal D$, and choose $\varphi\in C_c^\infty(\R^d)$ equal to one on $\{x-c:x\in\mathcal D\}$. By the first claim and the translation invariance of the $H^s(\R^d)$ norm, the function $x\mapsto\varphi(x-c)\|x-c\|^\alpha$ belongs to $H^s(\R^d)$, and it agrees with $\|x-c\|^\alpha$ on $\mathcal D$. By the norm comparison following~\eqref{eq:ec-scaling}, its restriction to $\mathcal D$ therefore belongs to the Mat\'ern RKHS on $\mathcal D$ and has finite RKHS norm.
\end{proof}

\subsection{Proof of Lemma~\ref{lem:new-confidence}}\label{ec:proof-confidence}
\begin{proof}[Proof of Lemma~\ref{lem:new-confidence}]
We first establish the bound at the points of a finite net, using that the sampling locations of a phase do not depend on its responses, and then extend it to all of $\X$ through the H\"older bound $d_k(x,z)\le L_\theta\|x-z\|^\theta$ in Section~\ref{subsec:kernel-smoothness}.

Set $r_N^{\rm mesh}=N^{-4/\theta}$, and fix a deterministic maximal Euclidean $r_N^{\rm mesh}$-separated set $V_N\subset\X$. By maximality, $V_N$ is an $r_N^{\rm mesh}$-net of $\X$. The Euclidean balls of radius $r_N^{\rm mesh}/2$ centered at its points have disjoint interiors and lie in a ball of radius $D_{\X}+r_N^{\rm mesh}/2$. Comparing their volumes gives
\begin{equation}\label{eq:confidence-net-size}
|V_N|\le\left(1+\frac{2D_{\X}}{r_N^{\rm mesh}}\right)^d=M_N.
\end{equation}

Consider a nonempty phase of either track, and let $\mathcal G_h$ denote the history of both tracks available at its start. Since the two tracks alternate whole phases, all observations made during the phase belong to its batch. Within the phase, the sampling locations, the stopping test, and the quota check depend only on $\mathcal G_h$ and, through $s_n$, on the sampled locations, but not on the current responses. Hence, conditional on $\mathcal G_h$, the sampling locations $x_1,\dots,x_n$ of the phase are fixed at sample size $n$. Let $\boldsymbol\varepsilon=(\varepsilon_1,\ldots,\varepsilon_n)^\top$ be the corresponding observation noise, and $a_x=(K_n+\lambda I)^{-1}k_n(x)$, so that
\begin{equation}\label{eq:confidence-decomposition}
f(x)-\mu_n(x)=\Bigl\{f(x)-\sum_{i=1}^n(a_x)_if(x_i)\Bigr\}-a_x^\top\boldsymbol\varepsilon.
\end{equation}
The kernel ridge identity gives $s_n^2(x)=\|k_x-\sum_i(a_x)_ik_{x_i}\|_{\Hk}^2+\lambda\|a_x\|_2^2$. Hence, by the reproducing property and $\|f\|_{\Hk}\le B$,
\begin{equation}\label{eq:confidence-bias}
\Bigl|f(x)-\sum_{i=1}^n(a_x)_if(x_i)\Bigr|
\le\|f\|_{\Hk}\Bigl\|k_x-\sum_{i=1}^n(a_x)_ik_{x_i}\Bigr\|_{\Hk}
\le Bs_n(x),
\|a_x\|_2\le\frac{s_n(x)}{\sqrt\lambda}.
\end{equation}
Since $a_x$ is fixed given $\mathcal G_h$, iterating the conditional moment bound in Assumption~\ref{as:new-noise} over the evaluations of the phase gives
\[
\mathbb E[\exp(u a_x^\top\boldsymbol{\varepsilon})\mid\mathcal G_h]
\le\exp(u^2\sigma^2\|a_x\|_2^2/2),u\in\R.
\]
Note that this uses no independence assumption on the observation errors.

Fix a nonempty phase, a sample size $n\ge1$ within it, and a net point $v\in V_N$, and put
\[
t_v=\frac{\sigma}{\sqrt\lambda}\sqrt{2\log(4M_NN^2/\delta)}\,s_n(v)
\ \ge\ \sigma\|a_v\|_2\sqrt{2\log(4M_NN^2/\delta)},
\]
where the inequality uses~\eqref{eq:confidence-bias}. The Chernoff bound gives
\[
\mathbb P\bigl(|a_v^\top\boldsymbol\varepsilon|>t_v\,\big|\,\mathcal G_h\bigr)
\le2\exp\Bigl(-\frac{t_v^2}{2\sigma^2\|a_v\|_2^2}\Bigr)
\le\frac{\delta}{2M_NN^2}.
\]
On the complementary event, \eqref{eq:confidence-decomposition}, \eqref{eq:confidence-bias}, and the definition of $\beta_N$ in~\eqref{eq:new-confidence-parameters} give
\begin{equation}\label{eq:confidence-net-bound}
|f(v)-\mu_n(v)|\le Bs_n(v)+t_v=\beta_Ns_n(v).
\end{equation}
We next take a union bound over a deterministic index set. Since each nonempty phase contains at least one of the $N$ evaluations, there are at most $N$ nonempty phases of the two tracks. Index them by $i=1,\ldots,N$ in their order of occurrence, and, for $i\le N$, $n\le N$, and $v\in V_N$, let $F_{i,n,v}$ be the event that the $i$th nonempty phase exists, reaches sample size $n$, and violates~\eqref{eq:confidence-net-bound} at $v$ with sample size $n$. The start of this phase is a stopping time, and, given the history $\mathcal G_h$ at its start, both whether it reaches sample size $n$ and its first $n$ sampling locations are fixed. Hence the conditional bound above, integrated over $\mathcal G_h$, gives $\mathbb P(F_{i,n,v})\le\delta/(2M_NN^2)$. Therefore~\eqref{eq:confidence-net-bound} holds simultaneously for all nonempty phases, all sample sizes within them, and all net points, except on the union of the at most $N\cdot N\cdot M_N$ events $F_{i,n,v}$, whose probability is at most $N\cdot N\cdot M_N\cdot\delta/(2M_NN^2)=\delta/2$. Only phases with observations need to be indexed, since a phase completed without any evaluation leaves the bands unchanged and requires no new event. The case $n=0$ satisfies~\eqref{eq:new-band} deterministically, because $\mu_0\equiv0$, $s_0\equiv1$, and $\beta_N\ge B\ge|f|$.

We next bound the observations and the RKHS norm of $\mu_n$. Since $|y_t|\le|f(x_t)|+|\varepsilon_t|\le B+|\varepsilon_t|$ and, by Assumption~\ref{as:new-noise},
\[
\mathbb P\bigl(|\varepsilon_t|>\sigma\sqrt{2\log(4N/\delta)}\bigr)\le2\exp\{-\log(4N/\delta)\}=\frac\delta{2N},
\]
a union bound over $t\le N$ gives $|y_t|\le Y_N$ for all $t\le N$, except on an event of probability at most $\delta/2$. On this event, writing $\mu_n=\sum_i\alpha_ik_{x_i}$ with $\alpha=(K_n+\lambda I)^{-1}y_{1:n}$, the spectral decomposition of $K_n$ gives
\begin{equation}\label{eq:confidence-mean-norm}
\|\mu_n\|_{\Hk}^2
=y_{1:n}^\top(K_n+\lambda I)^{-1}K_n(K_n+\lambda I)^{-1}y_{1:n}
\le\sup_{q\ge0}\frac{q}{(q+\lambda)^2}\,\|y_{1:n}\|_2^2
=\frac{\|y_{1:n}\|_2^2}{4\lambda}
\le\frac{NY_N^2}{4\lambda}.
\end{equation}

We also need the continuity of $s_n$. Define
\[
\|g\|_{n,\lambda}^2=\min_{a\in\R^n}\Bigl\{\Bigl\|g-\sum_{i=1}^na_ik_{x_i}\Bigr\|_{\Hk}^2+\lambda\|a\|_2^2\Bigr\},g\in\Hk.
\]
This is the squared distance from $(g,0)$ to the subspace $\{(\sum_ia_ik_{x_i},-a):a\in\R^n\}$ of $\Hk\oplus\R^n$, equipped with the norm given by $\|(g,a)\|^2=\|g\|_{\Hk}^2+\lambda\|a\|_2^2$. Hence $\|\cdot\|_{n,\lambda}$ is a Hilbert seminorm on $\Hk$. Moreover, $\|g\|_{n,\lambda}\le\|g\|_{\Hk}$ by taking $a=0$, and $s_n(x)=\|k_x\|_{n,\lambda}$ by the kernel ridge identity. Therefore,
\begin{equation}\label{eq:confidence-sd-lipschitz}
|s_n(x)-s_n(v)|\le\|k_x-k_v\|_{n,\lambda}\le\|k_x-k_v\|_{\Hk}=d_k(x,v).
\end{equation}

Finally, the two exceptional events above have total probability at most $\delta$. Outside them, fix a phase with observations, a sample size $n\ge1$, and $x\in\X$, and choose $v\in V_N$ with $\|x-v\|\le r_N^{\rm mesh}$. Since $|g(x)-g(v)|=|\langle g,k_x-k_v\rangle_{\Hk}|\le\|g\|_{\Hk}d_k(x,v)$ for any $g\in\Hk$, as in~\eqref{eq:new-modulus}, the bounds~\eqref{eq:confidence-net-bound}, \eqref{eq:confidence-sd-lipschitz}, and~\eqref{eq:confidence-mean-norm} give
\[
\begin{aligned}
|f(x)-\mu_n(x)|
&\le |f(v)-\mu_n(v)|+(B+\|\mu_n\|_{\Hk})d_k(x,v)\\
&\le\beta_Ns_n(x)+(B+\|\mu_n\|_{\Hk}+\beta_N)d_k(x,v)\\
&\le\beta_Ns_n(x)+\left(B+\frac{\sqrt N Y_N}{2\sqrt\lambda}+\beta_N\right)L_\theta N^{-4}
=\beta_Ns_n(x)+\zeta_N,
\end{aligned}
\]
where the last line uses $d_k(x,v)\le L_\theta(r_N^{\rm mesh})^\theta=L_\theta N^{-4}$ and the definition of $\zeta_N$ in~\eqref{eq:new-confidence-parameters}. This proves~\eqref{eq:new-band}.
\end{proof}

\subsection{Proof of Lemma~\ref{lem:new-mig}}\label{ec:proof-mig}
\begin{proof}[Proof of Lemma~\ref{lem:new-mig}]
We first bound the information gain on a single ball, separately for $\nu>1/2$ and $0<\nu\le1/2$, and then combine the balls of a cover. For $\nu>1/2$, the radius-dependent deterministic bound in \citet[Corollary 8]{iwazaki2025improved} gives, for $0<r\le\bar r$,
\begin{equation}\label{eq:mig-single-ball}
\gamma_n(B(z,r);\lambda)\le C \log^2(2N)
[1+r^{2\nu d/(2\nu+d)}(n/\lambda)^{d/(2\nu+d)}].
\end{equation}
To verify~\eqref{eq:mig-single-ball}, note first that the Mat\'ern kernel is translation invariant, so that $\gamma_n(B(z,r);\lambda)=\gamma_n(B(0,r);\lambda)$. We use Theorem~7 and Corollary~8 of \citet{iwazaki2025improved}, in which the information gain is defined with a noise variance, denoted by $\lambda^2$ there, that corresponds to our $\lambda$. In our notation, their bound for the ball $B(0,r)$ is
\begin{equation}\label{eq:mig-iwazaki}
\begin{aligned}
\gamma_n(B(0,r);\lambda)&\le C_n\,\bar\gamma_n(r)+C_0,\text{ where}\\
\bar\gamma_n(r)&=C_1\log\Bigl(1+\frac{2n}{\lambda}\Bigr)+C_2\,r^{\frac{2\nu d}{2\nu+d}}\Bigl(\frac{n}{\lambda\ell^{2\nu}}\Bigr)^{\frac{d}{2\nu+d}}\log^{\frac{2\nu}{2\nu+d}}\Bigl(1+\frac{2n}{\lambda}\Bigr),
\end{aligned}
\end{equation}
$C_0$ is an absolute constant, $C_1$ and $C_2$ depend only on $d$ and $\nu$, and
\[
C_n=\max\Bigl\{1,\ \log_2\Bigl(1+\frac{\Gamma(\nu)}{C_\nu}\log\frac{n^2}{\lambda}\Bigr)+\frac1\nu\log_2\Bigl(\frac{n^2}{\nu\Gamma(\nu)\lambda}\Bigr)+1\Bigr\}
\]
for a constant $C_\nu>0$ depending only on $\nu$. Let $n_0=\lceil\{\lambda\max\{1,\nu\Gamma(\nu)\}\}^{1/2}\rceil$, so that the logarithms in $C_n$ are nonnegative for $n\ge n_0$. For fixed $\lambda$ and $n_0\le n\le N$, each of these logarithms is at most $C\log(2N)$, and hence $C_n\le C\log(2N)$. Moreover, $\log(1+2n/\lambda)\le C\log(2N)$ and $\log^{2\nu/(2\nu+d)}(1+2n/\lambda)\le\max\{1,\log(1+2n/\lambda)\}\le C\log(2N)$. Hence, with the fixed length scale $\ell$ absorbed into the constant,
\[
\bar\gamma_n(r)\le C\log(2N)\bigl[1+r^{2\nu d/(2\nu+d)}(n/\lambda)^{d/(2\nu+d)}\bigr].
\]
Inserting these two bounds into~\eqref{eq:mig-iwazaki} and using $\log^2(2N)\ge\log^24>1$ gives~\eqref{eq:mig-single-ball} for $n_0\le n\le N$. For $1\le n<n_0$, Hadamard's inequality and $k(x,x)=1$ give $\det(I+K_n/\lambda)\le\prod_{i=1}^n\{1+k(x_i,x_i)/\lambda\}=(1+\lambda^{-1})^n$ for any $x_1,\dots,x_n$, so that $\gamma_n(B(z,r);\lambda)\le(n_0/2)\log(1+\lambda^{-1})$. Since $n_0$ is fixed, this is at most a constant times the right-hand side of~\eqref{eq:mig-single-ball}, uniformly in $r$.

For $0<\nu\le1/2$, we prove~\eqref{eq:mig-single-ball} by a projection argument. As in Section~\ref{subsec:kernel-smoothness}, the Mat\'ern spectral density and $|1-e^{it}|^2\le C\min\{1,t^2\}$ give
\begin{equation}\label{eq:mig-modulus}
\begin{aligned}
d_k(x,z)^2
&=\int_{\R^d}|e^{i\omega^\top(x-z)}-1|^2\widehat k(\omega)\,d\omega\\
&\le C\int_{\R^d}\min\{1,\|\omega\|^2\|x-z\|^2\}
\bigl(2\nu/\ell^2+\|\omega\|^2\bigr)^{-(\nu+d/2)}\,d\omega
\le C\|x-z\|^{2\nu},
\end{aligned}
\end{equation}
where the last step splits the integral at $\|\omega\|=1/\|x-z\|$ and uses $0<\nu<1$. Fix $B(z,r)$ with $0<r\le\bar r$ and an integer $m\ge1$. Cover $B(z,r)$ by $m$ balls $B(z_q,r_m)$, $q=1,\dots,m$, with $r_m=Crm^{-1/d}$, and assign each design point $x_i\in B(z,r)$ to a center $z_{q(i)}$ with $\|x_i-z_{q(i)}\|\le r_m$. Let $\Pi$ be the orthogonal projection of $\Hk$ onto $\operatorname{span}\{k_{z_1},\ldots,k_{z_m}\}$. The ambient Mat\'ern features may be used because the kernel is defined on $\R^d$. Decomposing $k_{x_i}=\Pi k_{x_i}+(I-\Pi)k_{x_i}$ splits the Gram matrix as $K_n=K_n^{\parallel}+K_n^{\perp}$, where
\[
(K_n^{\parallel})_{ij}=\langle\Pi k_{x_i},\Pi k_{x_j}\rangle_{\Hk},
(K_n^{\perp})_{ij}=\langle(I-\Pi)k_{x_i},(I-\Pi)k_{x_j}\rangle_{\Hk}
\]
are both positive semidefinite. Since $k(x,x)=1$ and $\|(I-\Pi)k_{x_i}\|_{\Hk}\le\|k_{x_i}-k_{z_{q(i)}}\|_{\Hk}=d_k(x_i,z_{q(i)})$, \eqref{eq:mig-modulus} gives
\begin{equation}\label{eq:mig-projection}
\operatorname{rank}K_n^{\parallel}\le m,
\tr K_n^{\parallel}\le\tr K_n=n,
\tr K_n^{\perp}=\sum_{i=1}^n\|(I-\Pi)k_{x_i}\|_{\Hk}^2
\le Cnr^{2\nu}m^{-2\nu/d}.
\end{equation}
For positive semidefinite $A$ and $D$, we have $\log\det(I+A+D)\le\log\det(I+A)+\tr D$, and $\log\det(I+A)\le\operatorname{rank}(A)\log\bigl(1+\tr A/\operatorname{rank}(A)\bigr)$ by the concavity of the logarithm. Applying these with $A=K_n^{\parallel}/\lambda$ and $D=K_n^{\perp}/\lambda$, and using~\eqref{eq:mig-projection}, gives
\[
2\gamma_n(B(z,r);\lambda)
\le m\log\Bigl(1+\frac{n}{\lambda m}\Bigr)
+C\frac{n}{\lambda}r^{2\nu}m^{-2\nu/d}.
\]
Choosing $m\asymp\max\{1,[(n/\lambda)r^{2\nu}]^{d/(2\nu+d)}\}$ makes both terms at most a constant times
\[
\log(2N)\,\bigl[1+r^{2\nu d/(2\nu+d)}(n/\lambda)^{d/(2\nu+d)}\bigr],
\]
which is at most a constant times the right-hand side of~\eqref{eq:mig-single-ball}.
% the rounding of $m$ and the bound $r\le\bar r$ are absorbed into constants. 
Hence~\eqref{eq:mig-single-ball} also holds for $0<\nu\le1/2$.

Now let $A$ be covered by $K$ balls $B(z_1,r),\dots,B(z_K,r)$. Given design points $x_1,\dots,x_n\in A$, assign each of them to one covering ball that contains it, and let $n_i$ be the number of points assigned to $B(z_i,r)$, so that $\sum_in_i=n$. Relabel the design points so that the points assigned to the same ball are consecutive, which permutes the rows and columns of $K_n$ simultaneously and does not change $\det(I+K_n/\lambda)$. Let $K^{(i)}\in\R^{n_i\times n_i}$ be the Gram matrix of the points assigned to $B(z_i,r)$, so that the matrices $I+K^{(i)}/\lambda$ with $n_i>0$ are the diagonal blocks of $I+K_n/\lambda$. Since $I+K_n/\lambda$ is positive definite, Fischer's determinant inequality gives
\[
\tfrac12\log\det(I+K_n/\lambda)\le\sum_{i:\,n_i>0}\tfrac12\log\det\bigl(I+K^{(i)}/\lambda\bigr)\le\sum_{i:\,n_i>0}\gamma_{n_i}(B(z_i,r);\lambda),
\]
where the last step uses the definition of $\gamma_{n_i}$ and the fact that the points assigned to $B(z_i,r)$ lie in this ball. Hence, by~\eqref{eq:mig-single-ball} and the concavity of $t\mapsto t^{d/(2\nu+d)}$, which gives $\sum_in_i^{d/(2\nu+d)}\le K^{2\nu/(2\nu+d)}n^{d/(2\nu+d)}$,
\[
\begin{aligned}
\sum_{i:\,n_i>0}\gamma_{n_i}(B(z_i,r);\lambda)
&\le C\log^2(2N)\left[K+r^{2\nu d/(2\nu+d)}\lambda^{-d/(2\nu+d)}\sum_{i=1}^Kn_i^{d/(2\nu+d)}\right]\\
&\le C\log^2(2N)\left[K+K^{2\nu/(2\nu+d)}r^{2\nu d/(2\nu+d)}(n/\lambda)^{d/(2\nu+d)}\right].
\end{aligned}
\]
Since the right-hand side does not depend on the design points, taking the supremum over $x_1,\dots,x_n\in A$ gives the same bound for $\gamma_n(A;\lambda)$. This is the second term in the minimum in~\eqref{eq:new-mig}. For the first term, $\X$ is contained in one ball of the fixed radius $D_\X$, and~\eqref{eq:mig-single-ball} with $\bar r$ replaced by $\max\{\bar r,D_\X\}$ gives $\gamma_n(A;\lambda)\le\gamma_n(\X;\lambda)\le C\log^2(2N)[1+(n/\lambda)^{d/(2\nu+d)}]$, with a constant depending on the fixed domain.

Finally, we verify the bound on the total variance stated before Lemma~\ref{lem:new-mig}. Since $0\le s_{j-1}^2\le1$ and $t\mapsto\log(1+t/\lambda)$ is concave and vanishes at zero, $s_{j-1}^2(x_j)\le\log(1+s_{j-1}^2(x_j)/\lambda)/\log(1+\lambda^{-1})$. Hence, for design points $x_1,\dots,x_n\in A$,
\[
\sum_{j=1}^ns_{j-1}^2(x_j)
\le\frac1{\log(1+\lambda^{-1})}\sum_{j=1}^n\log\bigl(1+s_{j-1}^2(x_j)/\lambda\bigr)
=\frac{\log\det(I+K_n/\lambda)}{\log(1+\lambda^{-1})}
\le C_\lambda\gamma_n(A;\lambda).
\]
Since conditioning on additional observations can only reduce each posterior variance, the same bound holds for an arbitrary subsequence of $n$ sampled points in $A$.
\end{proof}

\subsection{Proof of Lemma~\ref{lem:cap-driven-phase-cost}}\label{ec:proof-capped-phase}
\begin{proof}[Proof of Lemma~\ref{lem:cap-driven-phase-cost}]
We first show that any sampled point has a large posterior standard deviation and lies in $D$. Summing the posterior variances then gives~\eqref{eq:cap-driven-gamma}, and solving the resulting inequality under a cover gives~\eqref{eq:cap-driven-cover-cost}.

A point $x_j$ is sampled only if the stopping test fails, that is, $a_{j-1}^\star=a_{j-1}(x_j)>2\eps$. By the definition of $a_{j-1}$, all three terms in the minimum exceed $2\eps$:
\[
G(x_j)>2\eps,W(x_j)>2\eps,2\beta_Ns_{j-1}(x_j)+2\zeta_N>2\eps.
\]
The last inequality and $\zeta_N\le\eps/2$ give $s_{j-1}(x_j)>(\eps-\zeta_N)/\beta_N\ge\eps/(2\beta_N)$. In particular, all sampled point lie in $D$, and if $D$ is empty, then no point is sampled. Since $D$ is fixed given the history at the start of the phase, the bound on the total variance stated before Lemma~\ref{lem:new-mig} gives, for $n>0$,
\begin{equation}\label{eq:capped-phase-variance}
\frac{n\eps^2}{4\beta_N^2}
<\sum_{j=1}^n s_{j-1}^2(x_j)
\le C_\lambda\gamma_n(D;\lambda).
\end{equation}
This proves~\eqref{eq:cap-driven-gamma}, which holds trivially for $n=0$. The argument also applies to a phase truncated by the budget, since it uses only the points sampled in the phase.

We next consider the maintained width after completion, working on $\Econf$. At completion, $a_n^\star\le2\eps$. Hence, wherever $G(x)>2\eps$,
\[
\min\{W(x),\,2\beta_Ns_n(x)+2\zeta_N\}\le2\eps.
\]
If $n>0$, the new band $[\mu_n-w_n,\mu_n+w_n]$ with $w_n=\beta_Ns_n+\zeta_N$ is intersected with the maintained bounds $(\ell,u)$, and both contain $f$ on $\Econf$. Hence the updated bounds $\ell'=\max\{\ell,\mu_n-w_n\}$ and $u'=\min\{u,\mu_n+w_n\}$ satisfy $\ell'\le f\le u'$. Since $u'\le u$, $\ell'\ge\ell$, $u'\le\mu_n+w_n$, and $\ell'\ge\mu_n-w_n$, wherever $G(x)>2\eps$,
\[
u'(x)-\ell'(x)\le\min\bigl\{u(x)-\ell(x),\,2w_n(x)\bigr\}=\min\{W(x),\,2\beta_Ns_n(x)+2\zeta_N\}\le2\eps.
\] If $n=0$, then $2\beta_Ns_0(x)+2\zeta_N\ge2\beta_N\ge2B\ge W(x)$, so the maintained width $W(x)$ itself is at most $2\eps$. In this case, the phase has cost zero and requires no new confidence event.

Finally, suppose that $D$ is covered by $J\ge1$ balls of radius $r$ in a fixed bounded range, and let $n\ge1$. By Lemma~\ref{lem:new-mig} with $A=D$ and $K=J$,
\[
\gamma_n(D;\lambda)\le C\log^2(2N)\bigl[J+J^{2\nu/(2\nu+d)}r^{2\nu d/(2\nu+d)} n^{d/(2\nu+d)}\bigr].
\]
Inserting this bound into~\eqref{eq:cap-driven-gamma} gives
\begin{equation}\label{eq:capped-phase-scalar}
n\le b_0+b_1n^{q},
b_0=CJ\mathcal V_N\eps^{-2},
b_1=C\mathcal V_NJ^{2\nu/(2\nu+d)}r^{2\nu d/(2\nu+d)}\eps^{-2},q=\frac{d}{2\nu+d}.
\end{equation}
If $b_0\ge b_1n^q$, then~\eqref{eq:capped-phase-scalar} gives $n\le2b_0$. Otherwise, $n<2b_1n^q$, and hence $n<(2b_1)^{1/(1-q)}$. Since $1/(1-q)=1+d/(2\nu)$,
\[
b_1^{1+d/(2\nu)}
=C\,\mathcal V_N^{1+d/(2\nu)}
J^{\frac{2\nu}{2\nu+d}\cdot\frac{2\nu+d}{2\nu}}
r^{\frac{2\nu d}{2\nu+d}\cdot\frac{2\nu+d}{2\nu}}
\eps^{-2(1+d/(2\nu))}
=C\,\mathcal V_N^{1+d/(2\nu)}J\,r^{d}\eps^{-(2+d/\nu)}.
\]
Combining the two cases, $n\le2b_0+(2b_1)^{1+d/(2\nu)}$, which yields~\eqref{eq:cap-driven-cover-cost}.
\end{proof}

\subsection{Proof of Lemma~\ref{lem:new-active}}\label{ec:proof-completed-bands}
\begin{proof}[Proof of Lemma~\ref{lem:new-active}]
Work on $\Econf$. Algorithm~\ref{alg:continuous-ccucb} starts a phase only if its accuracy level is at least $2\zeta_N$, and otherwise allocates the remaining evaluations of the track to the saved estimated optimal point. Hence all phase $h$ that completes or contains a sampled point has $\eps_h\ge2\zeta_N$, and Proposition~\ref{prop:phase-accuracy} applies. If phase $h$ completes, the first three bounds in Proposition~\ref{prop:phase-accuracy}, with $W_h=u_h-\ell_h$, are the three bounds claimed in the lemma.
% they are proved in Section~\ref{ec:proof-phase-accuracy}. 
Any point $x$ sampled in phase $h$ satisfies $f^\star-f(x)\le T+L_\vartheta\eps_h$ by~\eqref{eq:region-active-quality}, and $f^\star-f(x)\le2B$ because $|f|\le B$ on $\X$ by~\eqref{eq:new-modulus}.
\end{proof}

\subsection{Proof of Lemma~\ref{lem:new-phase-cost}}\label{ec:proof-phase-cost}
\begin{proof}[Proof of Lemma~\ref{lem:new-phase-cost}]
Both bounds follow from~\eqref{eq:cap-driven-cover-cost} in Lemma~\ref{lem:cap-driven-phase-cost}, applied to the set $D_h=\{x\in\X:G_h(x)\ge2\eps_h,\ W_{h-1}(x)\ge2\eps_h\}$ of phase $h$, with a single ball covering $\X$ for the global bound and with the cover in Assumption~\ref{as:new-cover} for the localized bound. Let $\eps_h\ge2\zeta_N$. Since $\beta_N\ge B$, $\eps_h\le2B$, and $N\ge2$,
\begin{equation}\label{eq:phase-cost-floor}
\mathcal V_N\eps_h^{-2}\ge\tfrac14\log^2(2N)\ge\tfrac14\log^24.
\end{equation}
Hence $1$ and $\mathcal V_N\eps_h^{-2}$ are both at most a fixed constant times $\mathcal V_N^{1+d/(2\nu)}\eps_h^{-p}=(\mathcal V_N\eps_h^{-2})^{1+d/(2\nu)}$.

For the global bound, $D_h\subseteq\X$ is covered by one ball of radius $D_\X$. Applying~\eqref{eq:cap-driven-cover-cost} with $J=1$ and $r=D_\X$, and then~\eqref{eq:phase-cost-floor}, gives
\begin{equation}\label{eq:phase-cost-global}
n_h\le C\bigl[1+\mathcal V_N\eps_h^{-2}+\mathcal V_N^{1+d/(2\nu)}D_\X^d\eps_h^{-p}\bigr]\le C\mathcal V_N^{1+d/(2\nu)}\eps_h^{-p}.
\end{equation}
This bound holds deterministically, because~\eqref{eq:cap-driven-cover-cost} does.

For the localized bound, work on $\Econf$. With $T=0$, \eqref{eq:phase-accuracy-active-set} in the proof of Proposition~\ref{prop:phase-accuracy} gives $D_h\subseteq\mathcal S_{(L_\vartheta\eps_h)+}$. If $L_\vartheta\eps_h\le u_0$, then~\eqref{eq:new-cover} with $u=L_\vartheta\eps_h$ shows that $D_h$ is covered by $K$ balls of radius $r_h=c_\rho(L_\vartheta\eps_h)^{1/\alpha}\le c_\rho u_0^{1/\alpha}$. Applying~\eqref{eq:cap-driven-cover-cost} with $J=K$ and $r=r_h$, and using $r_h^d=c_\rho^dL_\vartheta^{d/\alpha}\eps_h^{d/\alpha}$, gives
\begin{equation}\label{eq:phase-cost-local-raw}
n_h\le C\bigl[1+K\mathcal V_N\eps_h^{-2}+K\mathcal V_N^{1+d/(2\nu)}\eps_h^{-(2+d/\nu-d/\alpha)}\bigr]\le CK\mathcal V_N^{1+d/(2\nu)}\eps_h^{-a},
\end{equation}
where the last step uses~\eqref{eq:ec-log-bounds}, \eqref{eq:phase-cost-floor}, $K\ge1$, the definition of $a$, and $\eps_h\le2B$.

Finally, inserting the bound $\mathcal V_N^{1+d/(2\nu)}\le C\Lambda_{N,\delta}$ from~\eqref{eq:ec-log-bounds} into~\eqref{eq:phase-cost-global} and~\eqref{eq:phase-cost-local-raw} gives~\eqref{eq:new-phase-cost} and~\eqref{eq:new-phase-cost-local}.
\end{proof}

\clearpage
\begingroup
\clubpenalty=10000\widowpenalty=10000
\makeatletter\@clubpenalty=10000\makeatother
\makeatletter
\newenvironment{SASORIECblock}[1]{%
  \def\SASORIECtype{#1}%
  \par\addvspace{.6\baselineskip}\noindent
  \begin{minipage}{\linewidth}\def\@captype{#1}%
}{%
  \end{minipage}\par\nopagebreak[4]%
  \def\SASORIECtable{table}%
  \ifx\SASORIECtype\SASORIECtable
    \vspace{8pt}%
  \else
    \smallskip
  \fi
  \nopagebreak[4]%
}
\makeatother

\section{Additional Numerical Details and Results}\label{ec:numerical}

We first provide basic experimental details, including benchmark definitions, experimental settings, query point search and the evaluation metrics. We then present the additional Branin-2 results followed by the Hartmann-3 results. All experiments run $30$ independent macro-replications. Tables and curves report means $\pm$ sample standard errors over completed runs. Opt, Joint-Free, and Reg abbreviate the corresponding SASORIs. 

\subsection{Benchmark Definitions and Settings}\label{ec:num-benchmarks}

\textbf{Branin-2.}
For $z=(z_1,z_2)\in[-5,10]\times[0,15]$, define
\begin{equation}\label{eq:ec-num-branin}
 B_{\rm Br}(z)=\left(z_2-\frac{5.1z_1^2}{4\pi^2}
 +\frac{5z_1}{\pi}-6\right)^2
 +10\left(1-\frac{1}{8\pi}\right)\cos z_1+10.
\end{equation}
We maximize $f(x)=-B_{\rm Br}(15x_1-5,15x_2)/200$ on $[0,1]^2$.
Its reference maximum is $f^\star=-5/(800\pi)$, attained at the three transformed Branin minimizers. Multiplying a reported gap or tolerance by $200$ converts it to a difference in the original Branin response.

\textbf{Hartmann-3.}
On $[0,1]^3$, define
\begin{equation}\label{eq:ec-num-hartmann}
 H_3(x)=-\sum_{i=1}^{4}a_i
 \exp\left\{-\sum_{j=1}^{3}A_{ij}(x_j-P_{ij})^2\right\},
 \qquad f(x)=-H_3(x)/4,
\end{equation}
where
\[
 a=(1,1.2,3,3.2),\qquad
 A=\begin{pmatrix}
 3&10&30\\0.1&10&35\\3&10&30\\0.1&10&35
 \end{pmatrix},\qquad
 P=10^{-4}\begin{pmatrix}
 3689&1170&2673\\4699&4387&7470\\1091&8732&5547\\381&5743&8828
 \end{pmatrix}.
\]
Numerical refinement gives the fixed evaluation reference
\[
 \begin{aligned}
 x^\star&\approx(0.114588877,0.555648895,0.852546985),\\
 f^\star&\approx0.9656949468.
 \end{aligned}
\]
This reference is numerical, whereas the Branin reference is analytic. Neither reference is supplied to the sampling procedure. Both benchmarks use observations $y_t=f(x_t)+\varepsilon_t$ with independent $\varepsilon_t\sim N(0,0.02^2)$.

\begin{SASORIECblock}{table}
\begingroup\centering\singlespacing
\setlength{\abovecaptionskip}{4pt}\setlength{\belowcaptionskip}{4pt}
\small
\setlength{\tabcolsep}{3pt}
\caption{Settings for the two benchmarks.}
\label{tab:ec-num-settings}
\begin{tabular*}{\linewidth}{@{\extracolsep{\fill}}lcc@{}}
\toprule
Setting & Branin-2 & Hartmann-3\\
\midrule
Domain & $[0,1]^2$ & $[0,1]^3$\\
Objective & $-B_{\rm Br}/200$ & $-H_3/4$\\
Largest tolerance $T$ & $0.10$ & $0.30$\\
Evaluation tolerances & $.04,.06,.08,.10$ & $.12,.18,.24,.30$\\
Fixed length scale & $0.3\sqrt{2}$ & $0.2\sqrt{3}$\\
Joint-Free region quota & $591$ & $197$\\
\bottomrule
\end{tabular*}
\par\endgroup
\end{SASORIECblock}

Table~\ref{tab:ec-num-settings} gives the settings that differ between benchmarks. Both tasks use $N=3500$, a fixed Mat\'ern-$5/2$ kernel with zero prior mean and unit marginal variance, and regularization $0.0004$. Joint-Free's region quota is $m=\lfloor\min\{N/2,\sqrt{N}/T\}\rfloor$.

The runs use the fresh GP fits and band intersections in Algorithm~\ref{alg:continuous-ccucb}. Each track has its own phase index and schedule $\varepsilon_h=2B\vartheta^{-h}$. Noisy observations from the current phase do not affect its queries. A completed nonempty phase updates the shared bands. An incomplete phase contributes observations to the budget and regret but does not update the lower and upper bound, therefore not contribute to the region identifications.

We use the practical calibration $\beta=\max\{\beta_{\rm floor},B\}$, with $\beta_{\rm floor}=2.5$. The correction $\zeta_N$ is computed from~\eqref{eq:new-confidence-parameters} using this multiplier, $\delta=N^{-2}$, and $\theta=1/2$. The value $B=2$ is a working bound, not necessarily the valid RKHS norm bound of the benchmarks. 

SASORI starts with $x_0=(0.1,\ldots,0.1)$ and no separate initialization batch. GP-UCB begins with $\max\{8,2d\}=8$ Sobol observations, counted in $N$, then maximizes $\mu_t+\kappa s_t$ numerically. Sobol-GP uses a scrambled Sobol sequence throughout.

\subsection{Numerical Search for the Next Query Point}\label{ec:num-implementation}

Following Remark~\ref{rem:threshold-query}, we seek a point whose acquisition value exceeds $2\varepsilon_h$. We first evaluate all $1024$ scrambled Sobol points. If any point satisfies this condition, we query the point with the largest acquisition value among these candidates.

Otherwise, we apply L-BFGS-B from up to $10$ starting points, with at most $100$ iterations per restart. Up to two Powell restarts provide a fallback when gradient evaluation or local optimization fails, or when the searches produce no improvement. During either optimizer, the threshold condition is checked at every acquisition evaluation, including evaluations within line searches. Once a qualifying point is found, optimization terminates without waiting for convergence, and we query the point with the largest acquisition value among all candidates evaluated so far.

If this search finds no qualifying point, we perform an additional search review over the full domain before accepting phase completion. This review uses a fresh scramble of $2048$ Sobol points, retains the best previous candidates as anchors, and allows up to $20$ L-BFGS-B restarts with the same Powell fallback. Restart locations include points from different parts of the domain, with the remaining starts selected by acquisition value. The larger sample and more widely distributed restarts reduce the risk of overlooking a qualifying point. The same threshold rule applies during this review: finding such a point resumes querying. Otherwise, the phase satisfies the numerical stopping condition. 

\subsection{Evaluation Metrics}\label{ec:num-metrics}

Let $L,U$ denote the lower and upper functions used to compute regions, and let $b_{\rm num},c_{\rm num}$ be their numerical maxima. For $\tau>0$, we use
\begin{equation}\label{eq:ec-num-output}
 I_\tau=\{x:L(x)>c_{\rm num}-\tau\},\qquad
 O_\tau=\{x:U(x)>b_{\rm num}-\tau\}.
\end{equation}
For the \emph{algorithm output}, SASORI constructs its regions from the retained bounds of completed phases, using its own $\beta$ and $\zeta_N$. The baseline outputs use $L=\mu_N-3s_N$ and $U=\mu_N+3s_N$, without the initial range constraint or $\zeta_N$ correction. These are the region outputs reported in the main text.

Under this construction, SASORI's observations from incomplete phases do not contribute to its reported regions, whereas GP-UCB and Sobol-GP use all collected observations. To provide a fair comparison of region identification using all available observations, we therefore consider an additional report in which we fit \emph{one GP to all observations} from each method. This includes repeated observations and those collected during incomplete phases, so all $N$ observations contribute to the GP fit. Using the same kernel and regularization, we construct the bounds as
\begin{equation}\label{eq:ec-num-pooled}
 \begin{aligned}
 L^{\rm all}(x)&=\max\{-B,\mu_N(x)-3s_N(x)-\zeta_N^{\rm rep}\},\\
 U^{\rm all}(x)&=\min\{B,\mu_N(x)+3s_N(x)+\zeta_N^{\rm rep}\}.
 \end{aligned}
\end{equation}
Here $\zeta_N^{\rm rep}$ uses $N=3500$ and multiplier $3$. Numerical extrema are recomputed, without intersecting these bounds with earlier fresh GP bands. For SASORI, this changes both the observations used and the construction of the bounds. Baseline GPs already use all observations. Note that this additional report is no more than a different way to report the outputs of SASORIs, and it does not change the implementation procedures of SASORIs. 

We briefly recall the metrics of region identification and optimization for reference. Write $\Delta(x)=f^\star-f(x)$ and $A_\tau=O_\tau\setminus I_\tau$. On the independent evaluation set $Q$, inner recovery is
\begin{equation}\label{eq:num-recovery}
 \operatorname{Rec}_Q(\tau)
 =\frac{|I_\tau\cap\mathcal S_\tau\cap Q|}{|\mathcal S_\tau\cap Q|}.
\end{equation}
It is the fraction of truly acceptable evaluation points included in the inner set. The unclassified fraction and empirical region error are
\begin{align}
 \operatorname{Unc}_Q(\tau)&=\frac{|A_\tau\cap Q|}{|Q|},\label{eq:ec-num-unc}\\
 \eta_Q(\tau)&=\max_{x\in A_\tau\cap Q}|\Delta(x)-\tau|,\label{eq:ec-num-eta}
\end{align}
with $\eta_Q(\tau)=0$ when $A_\tau\cap Q$ is empty. Recovery and Unclassified are reported as percentages. The error $\eta_Q(\tau)$ is in the objective's units and describes how far the unresolved designs lie from the acceptance threshold. These metrics describe the reported sets on $Q$.

The main tables evaluate all three metrics at $\tau=T$. For the additional error summary, we compute
\[
 \eta_{Q,\max}=\max_{\tau\in\{0.4T,0.6T,0.8T,T\}}\eta_Q(\tau)
\]
within each run, then report the cross-run mean and standard error of $\eta_{Q,\max}/T$. The maximum is therefore taken before averaging. Each benchmark uses $32768$ fixed scrambled Sobol points as $Q$, shared by all methods and runs and generated independently of sampling.

Simple regret $r_t$ and cumulative regret $R_t$ have the definitions in Section~\ref{sec:relaxed-model}. For SASORI, $\widehat x_t$ is its latest recommendation, held fixed between completed phases. We also evaluate the best-query simple regret,
\begin{equation}\label{eq:num-best-query}
 r_t^{\rm best}=\min_{1\le s\le t}\{f^\star-f(x_s)\}.
\end{equation}
Both $r_t^{\rm best}$ and $R_t$ include observations in incomplete phases.

\newpage
\subsection{Additional Branin-2 Results}\label{ec:num-regions}\label{ec:num-branin}

\subsubsection{Other Tolerances}\label{ec:num-branin-tolerances}

We first compare Branin-2 empirical region error across $\tau\in\{0.04,0.06,0.08,0.10\}$ for the algorithm output and the GP fitted to all observations.
\par\nopagebreak[4]

\begin{SASORIECblock}{figure}
\begingroup\centering\singlespacing
\setlength{\abovecaptionskip}{4pt}\setlength{\belowcaptionskip}{4pt}
\includegraphics[width=.45\linewidth]{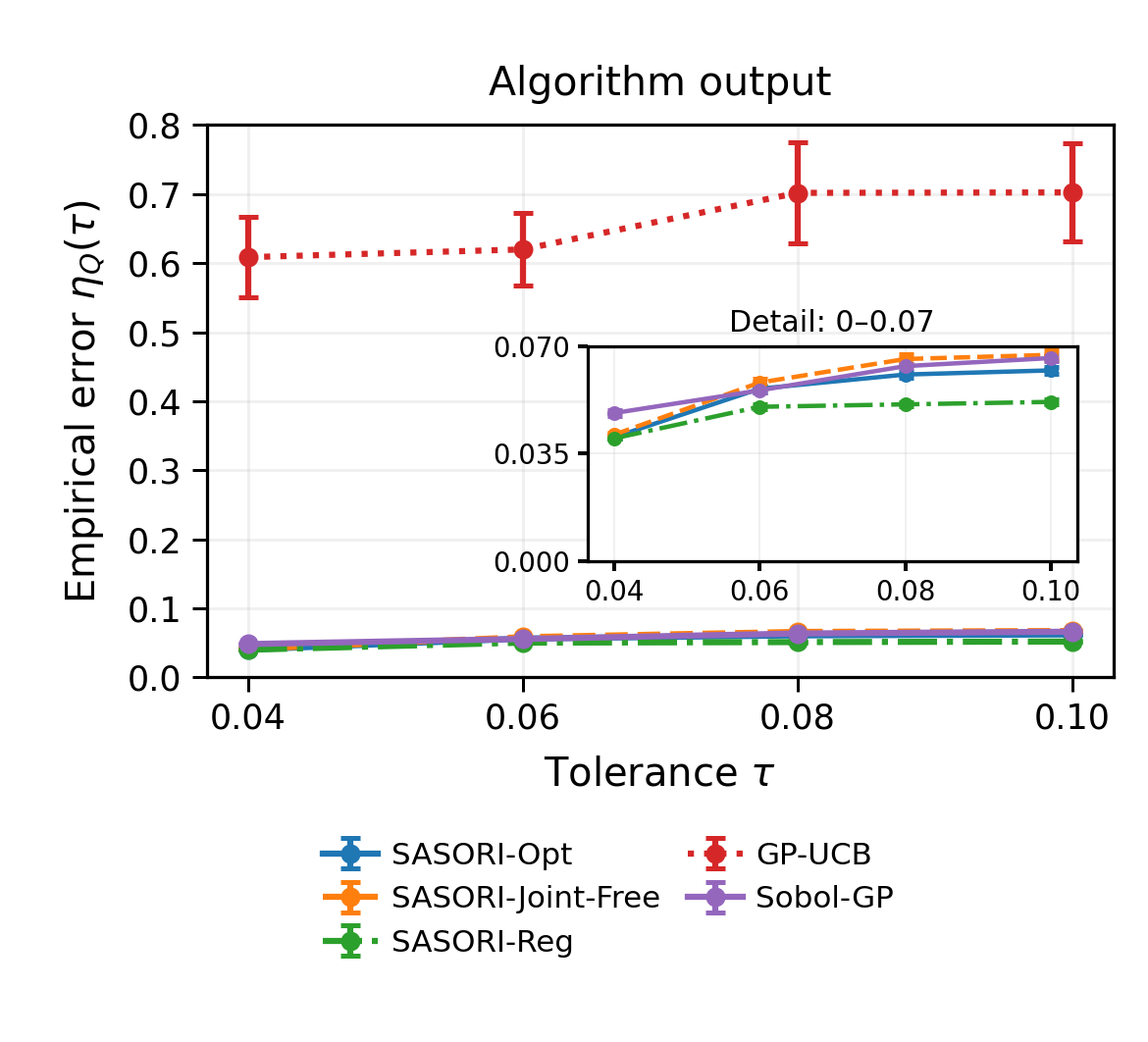}\hfill
\includegraphics[width=.45\linewidth]{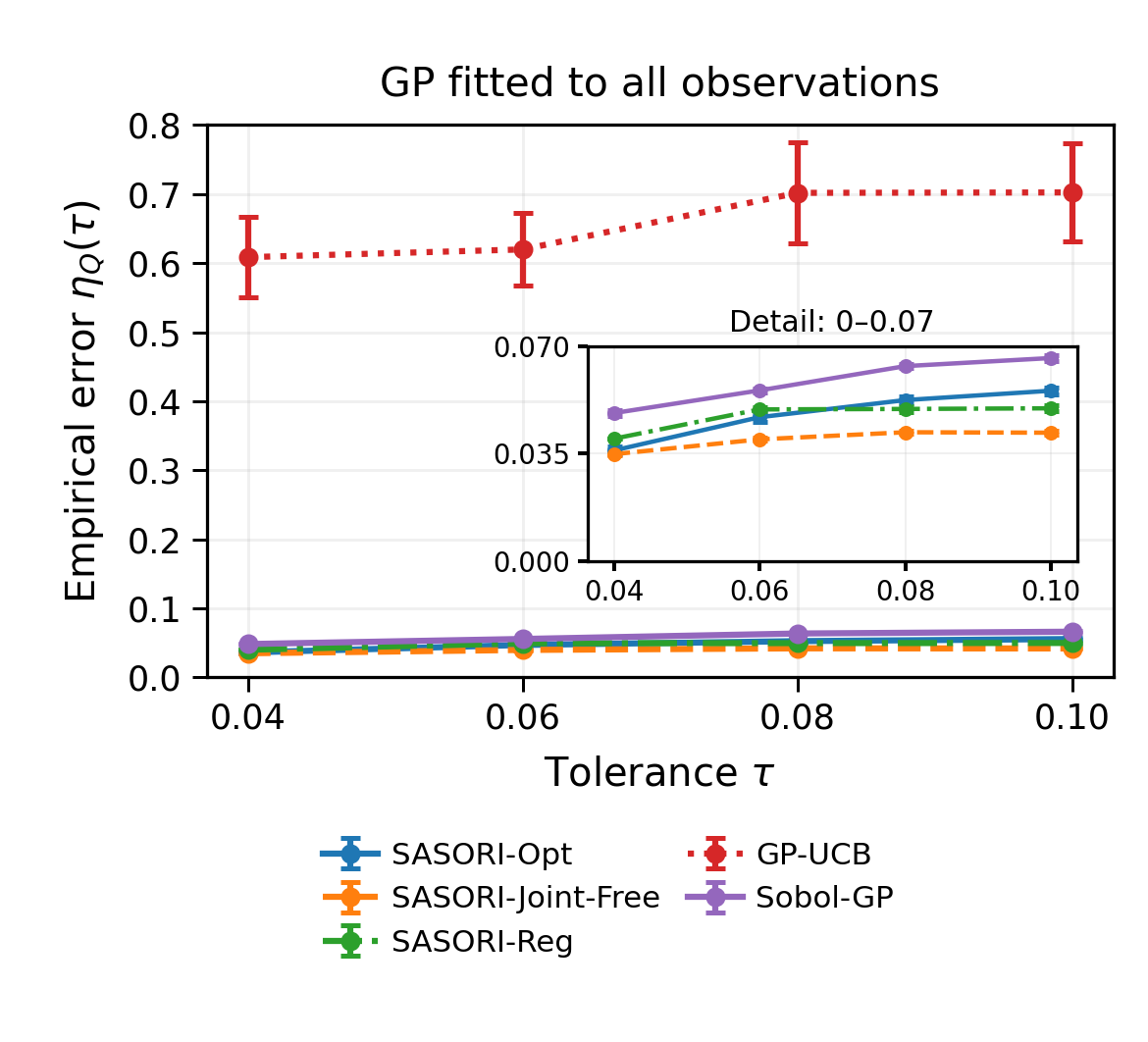}
\caption{Branin-2 empirical region error $\eta_Q(\tau)$: algorithm output (left) and GP fitted to all observations (right). Error bars show one standard error.}\label{fig:ec-num-branin-eta}
\par\endgroup
\end{SASORIECblock}

\noindent
Figure~\ref{fig:ec-num-branin-eta} shows that Reg has the smallest mean empirical region error at all four tolerances for the algorithm output, consistent with its emphasis on region estimation accuracy. With all observations fitted to one GP, Joint-Free has the smallest mean error. All three SASORI variants have much smaller errors than GP-UCB under both reporting methods.

We next summarize the maximum error across these tolerances as $\eta_{Q,\max}/T$.
\par\nopagebreak[4]

\begin{SASORIECblock}{table}
\begingroup\centering\small\singlespacing
\setlength{\abovecaptionskip}{4pt}\setlength{\belowcaptionskip}{4pt}
\setlength{\tabcolsep}{4pt}\renewcommand{\arraystretch}{1.12}
\caption{Branin-2 maximum empirical region error across the four tested tolerances (mean $\pm$ standard error).}\label{tab:ec-num-regions}
\begin{tabular*}{\linewidth}{@{\extracolsep{\fill}}lcc@{}}
\toprule
Method & Algorithm output: $\eta_{Q,\max}/T$ & All observations: $\eta_{Q,\max}/T$\\
\midrule
Opt & $0.638\pm0.010$ & $0.567\pm0.010$\\
Joint-Free & $0.688\pm0.017$ & $0.435\pm0.006$\\
Reg & $0.528\pm0.007$ & $0.514\pm0.008$\\
GP-UCB & $7.186\pm0.741$ & $7.186\pm0.741$\\
Sobol-GP & $0.678\pm0.008$ & $0.678\pm0.008$\\
\bottomrule
\end{tabular*}
\par\endgroup
\end{SASORIECblock}

\noindent
Table~\ref{tab:ec-num-regions} shows that Reg has the smallest mean maximum error for the algorithm output, while Joint-Free has the smallest value when all observations are used. These summaries are consistent with the error curves in Figure~\ref{fig:ec-num-branin-eta}; the maximum is taken within each run before averaging.

We then compare inner recovery across the same tolerances under both reporting methods.
\par\nopagebreak[4]

\begin{SASORIECblock}{figure}
\begingroup\centering\singlespacing
\setlength{\abovecaptionskip}{4pt}\setlength{\belowcaptionskip}{4pt}
\includegraphics[width=.45\linewidth]{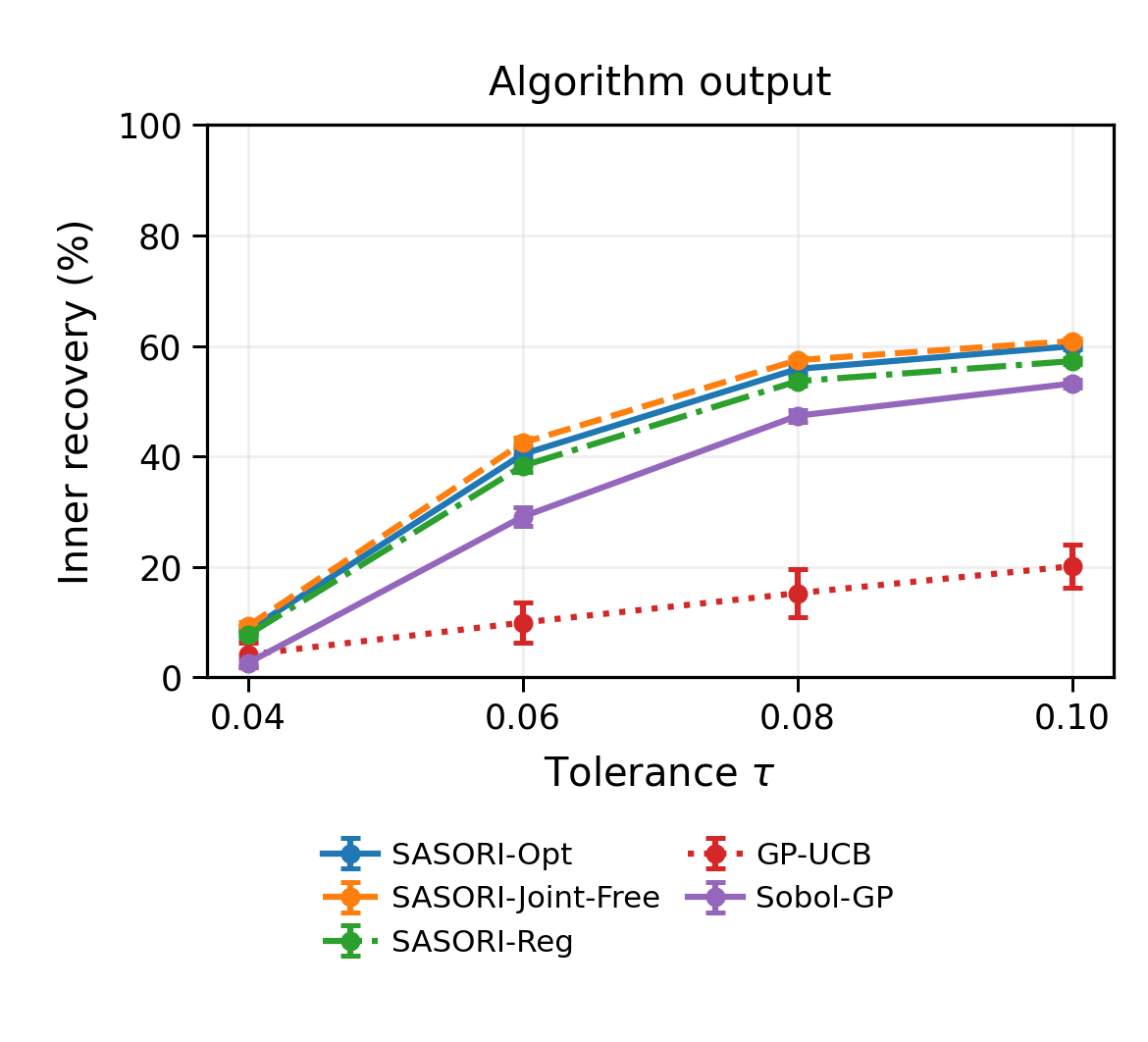}\hfill
\includegraphics[width=.45\linewidth]{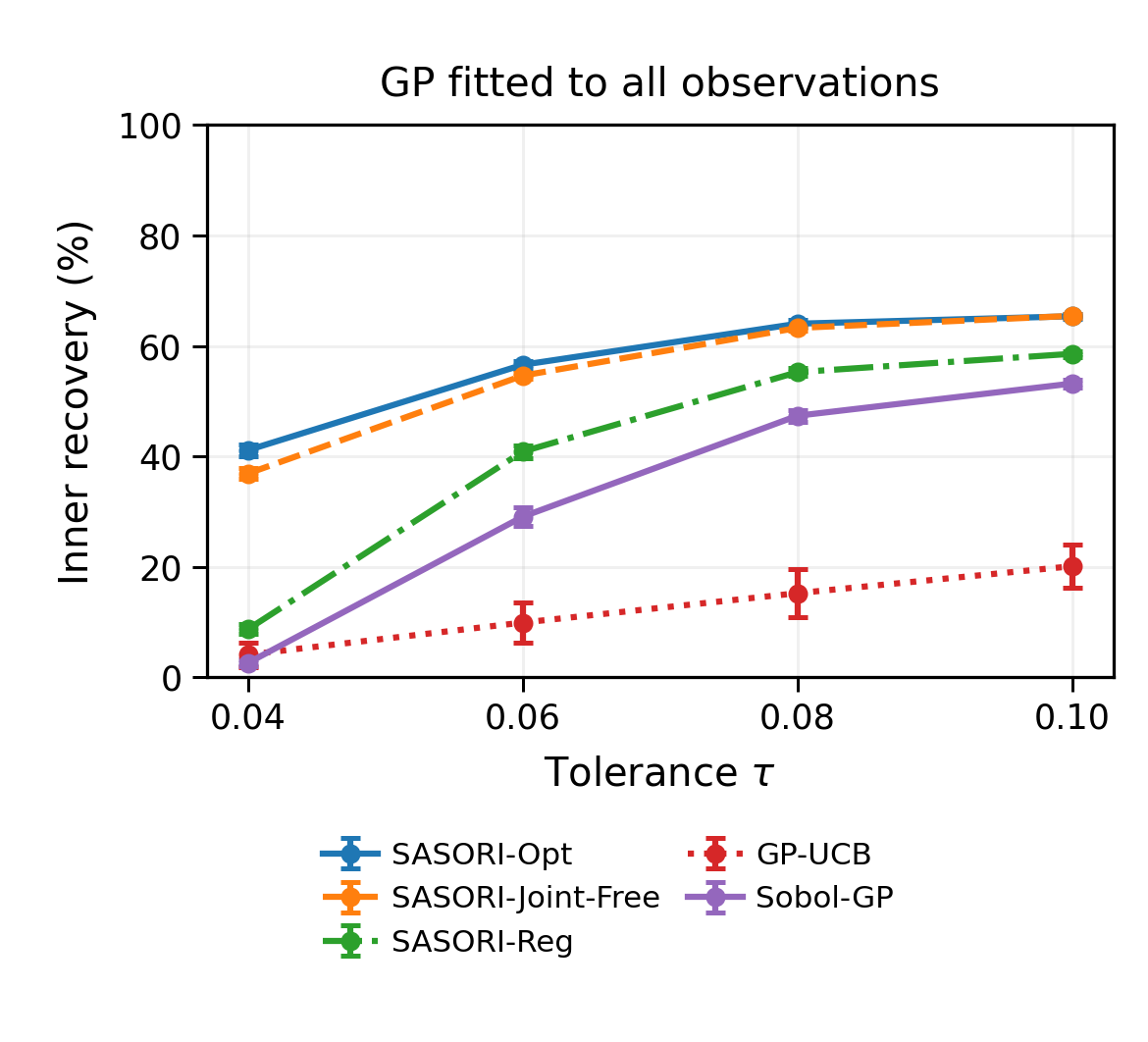}
\caption{Branin-2 inner recovery across tolerances: algorithm output (left) and GP fitted to all observations (right). Error bars show one standard error.}\label{fig:ec-num-branin-recovery}
\par\endgroup
\end{SASORIECblock}

\noindent
Figure~\ref{fig:ec-num-branin-recovery} shows that all three SASORI variants recover more acceptable designs than either baseline across the tested tolerances. Fitting one GP to all observations increases their mean recovery, with Opt and Joint-Free achieving similar recovery at the largest tolerance.

We also compare the fractions of the domain that remain unclassified.
\par\nopagebreak[4]

\begin{SASORIECblock}{figure}
\begingroup\centering\singlespacing
\setlength{\abovecaptionskip}{4pt}\setlength{\belowcaptionskip}{4pt}
\includegraphics[width=.45\linewidth]{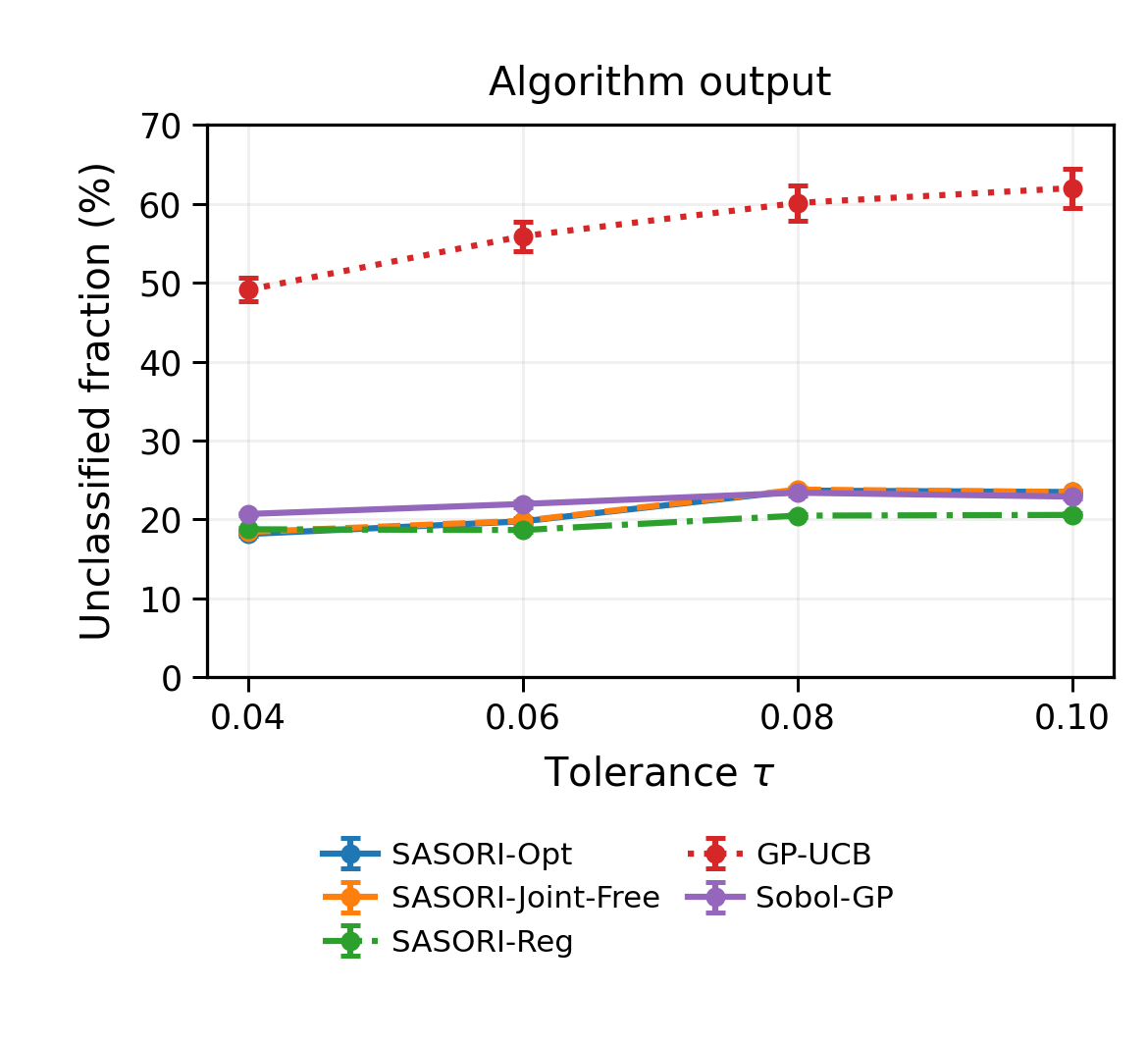}\hfill
\includegraphics[width=.45\linewidth]{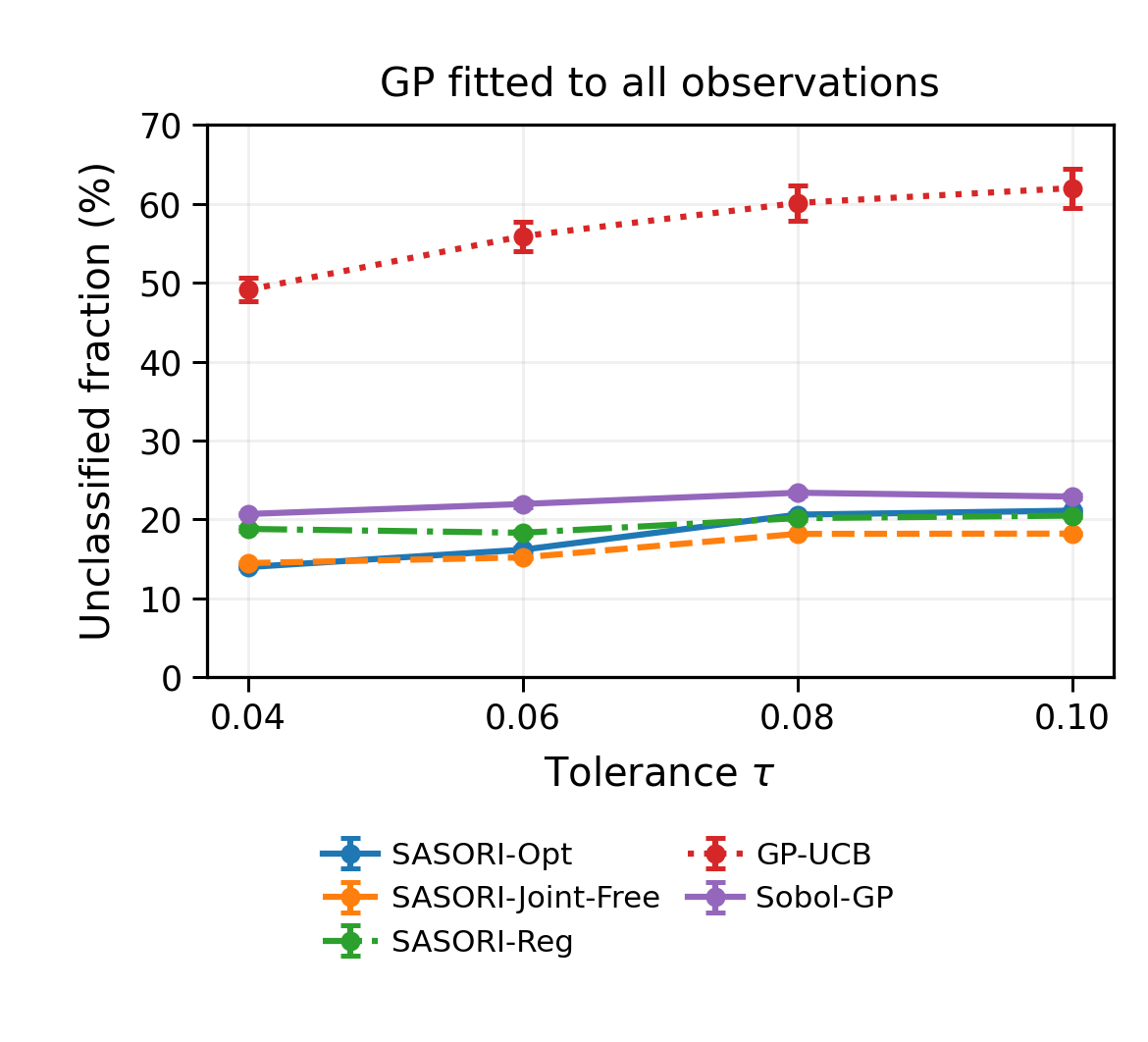}
\caption{Branin-2 unclassified fractions across tolerances: algorithm output (left) and GP fitted to all observations (right). Error bars show one standard error.}\label{fig:ec-num-branin-unclassified}
\par\endgroup
\end{SASORIECblock}

\noindent
Figure~\ref{fig:ec-num-branin-unclassified} shows that Reg has the smallest mean unclassified fraction at the three larger tolerances for the algorithm output, while Joint-Free has the smallest fraction at these tolerances when all observations are used. Resolving membership includes excluding unacceptable designs as well as enlarging the inner set, so the ranking need not match that of inner recovery.

% We summarize the three region metrics at $\tau=T=0.1$ for the GP fitted to all observations.
% \par\nopagebreak[4]

% \begin{SASORIECblock}{table}
% \begingroup\centering\small\singlespacing
% \setlength{\abovecaptionskip}{4pt}\setlength{\belowcaptionskip}{4pt}
% \setlength{\tabcolsep}{4pt}\renewcommand{\arraystretch}{1.12}
% \caption{Branin-2 regions from a GP fitted to all observations at $\tau=T=0.1$ (mean $\pm$ standard error). Lower is better for $\eta_Q$ and Unclassified; larger is better for Recovery.}\label{tab:ec-num-branin-all}
% \begin{tabular*}{\linewidth}{@{\extracolsep{\fill}}lccc@{}}
% \toprule
% Method & $\eta_Q(\tau)$ & Recovery (\%) & Unclassified (\%)\\
% \midrule
% Opt & $0.0555\pm0.0011$ & $65.38 \pm 0.42$ & $21.12 \pm 0.17$\\
% Joint-Free & $0.0418\pm0.0006$ & $65.45 \pm 0.37$ & $18.21 \pm 0.12$\\
% Reg & $0.0497\pm0.0009$ & $58.58 \pm 0.53$ & $20.49 \pm 0.19$\\
% GP-UCB & $0.7024\pm0.0716$ & $20.12 \pm 3.94$ & $62.01 \pm 2.51$\\
% Sobol-GP & $0.0661\pm0.0010$ & $53.20 \pm 0.74$ & $22.91 \pm 0.27$\\
% \bottomrule
% \end{tabular*}
% \par\endgroup
% \end{SASORIECblock}

% \noindent
% Table~\ref{tab:ec-num-branin-all} complements the algorithm output in Table~\ref{tab:num-branin-summary}. Opt and Joint-Free have similar mean inner recovery, while Joint-Free has the smallest empirical region error and unclassified fraction. This comparison uses a common construction of the bounds to assess the regional information contained in each method's observations.

To illustrate these region estimates, we show the algorithm's sets at $\tau\in\{0.04,0.06,0.08\}$ for the same macro-replication as in Figure~\ref{fig:num-branin-regions}.
\par\nopagebreak[4]

\begin{SASORIECblock}{figure}
\begingroup\centering\singlespacing
\setlength{\abovecaptionskip}{4pt}\setlength{\belowcaptionskip}{4pt}
\setlength{\tabcolsep}{2pt}
\begin{tabular}{@{}cccc@{}}
{\small $\tau=0.04$} & \raisebox{-.5\height}{\includegraphics[width=.26\linewidth]{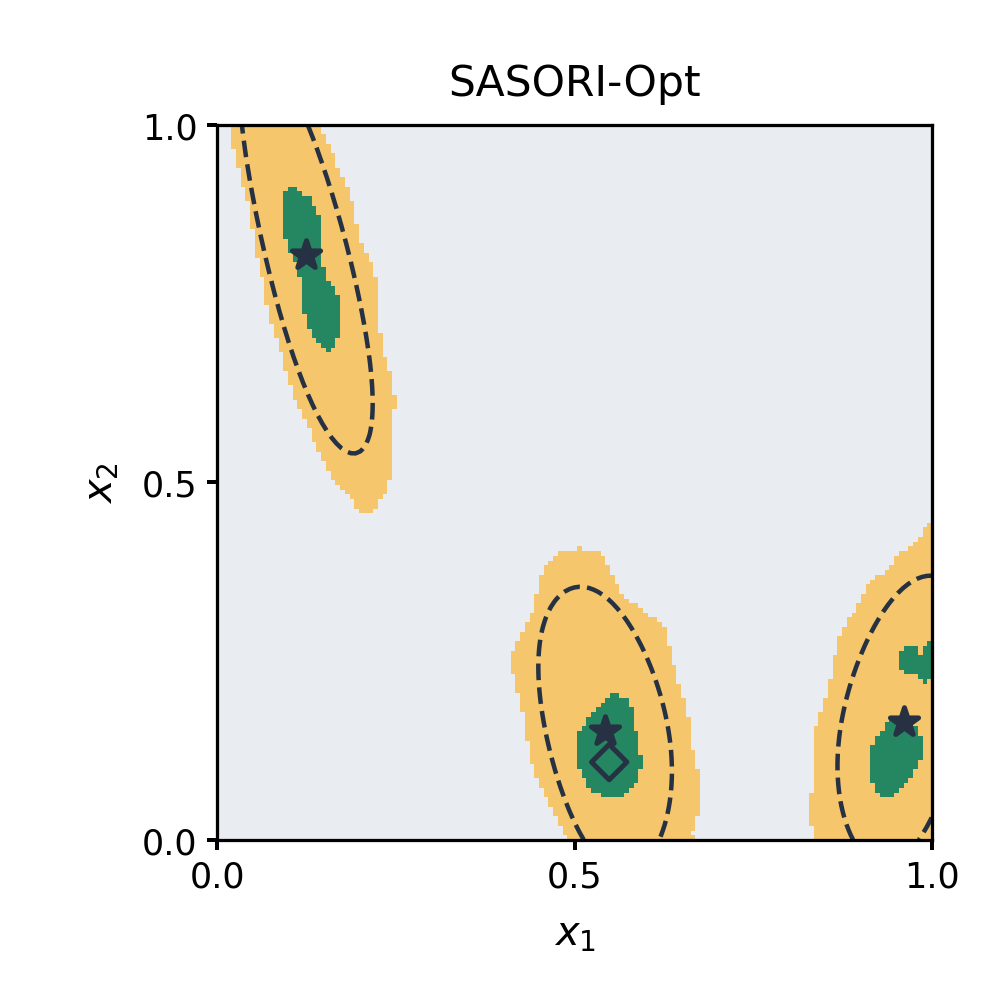}} & \raisebox{-.5\height}{\includegraphics[width=.26\linewidth]{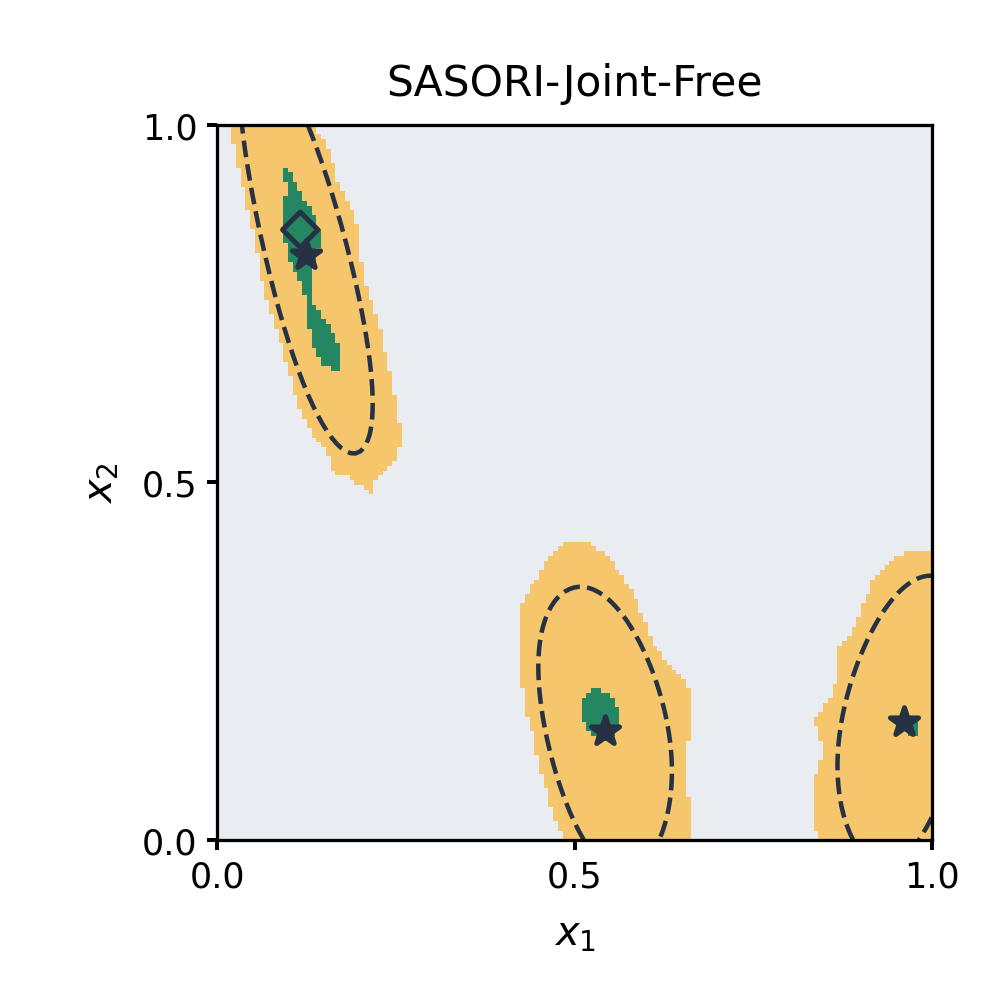}} & \raisebox{-.5\height}{\includegraphics[width=.26\linewidth]{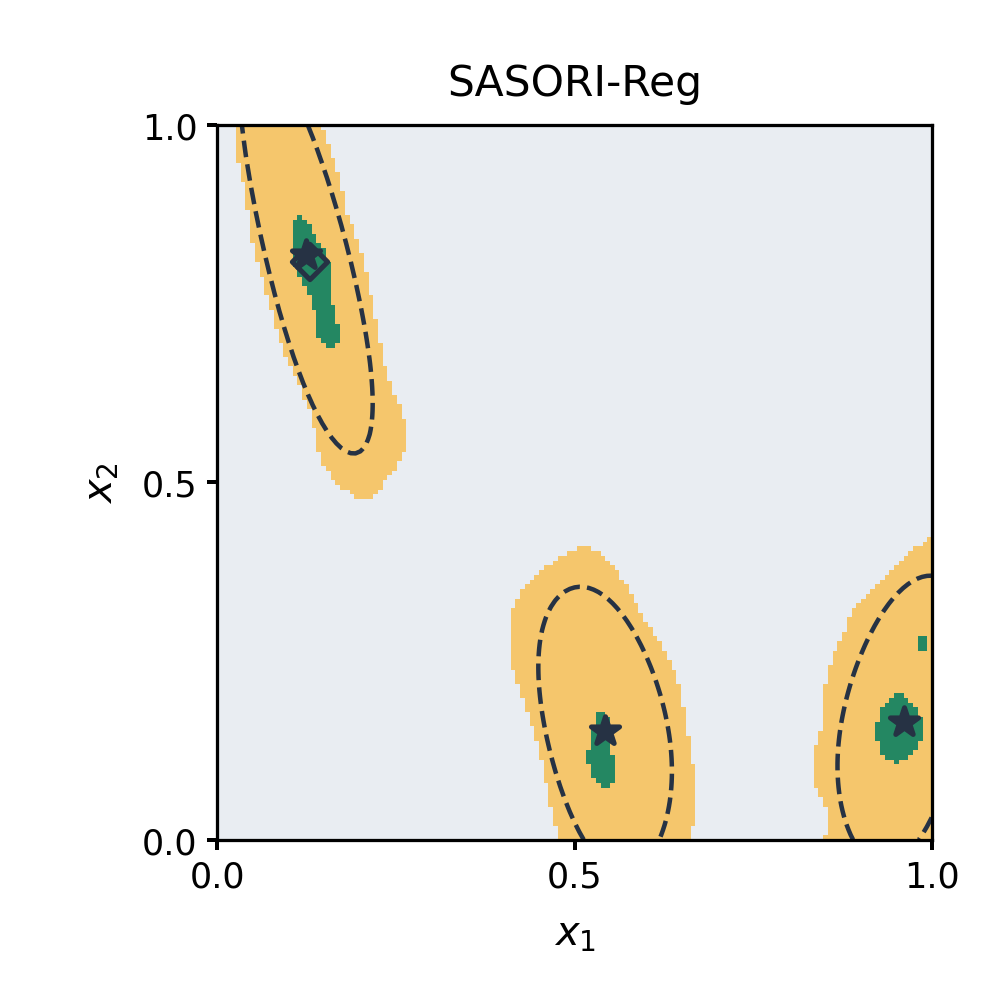}}\\[5pt]
{\small $\tau=0.06$} & \raisebox{-.5\height}{\includegraphics[width=.26\linewidth]{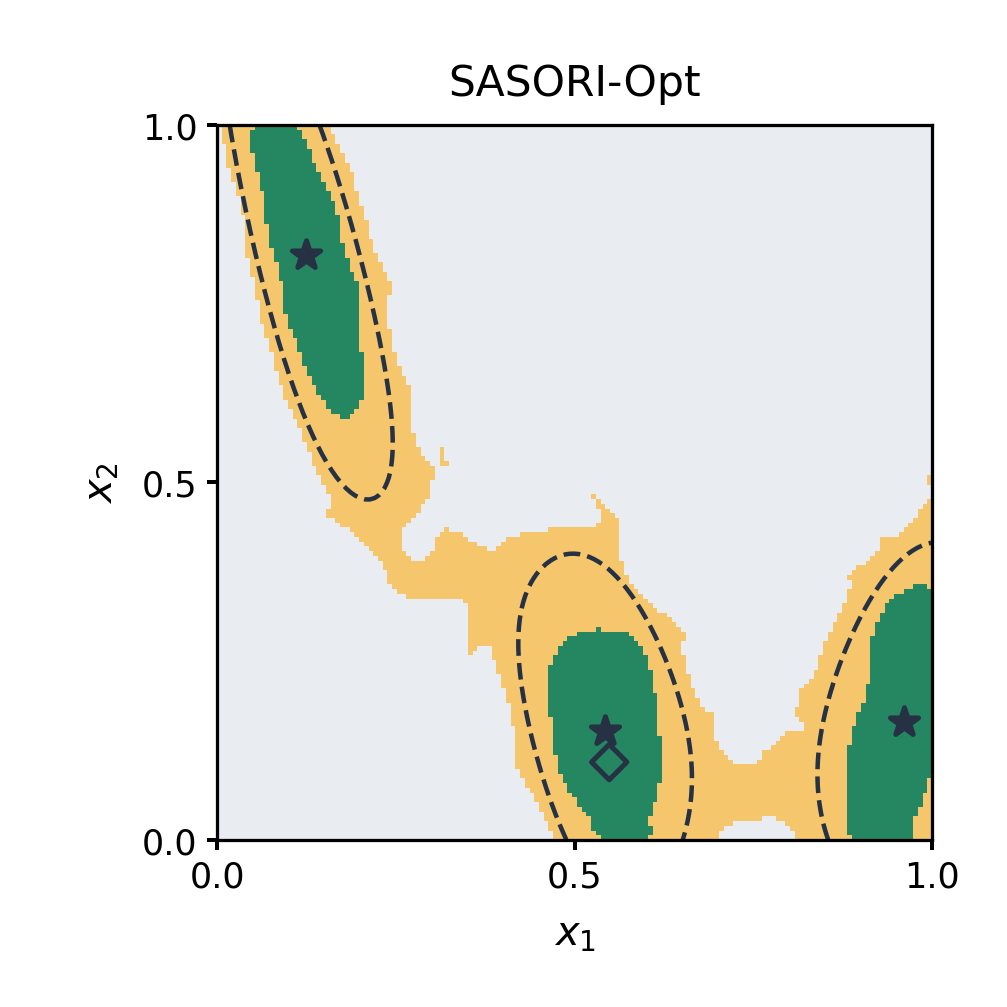}} & \raisebox{-.5\height}{\includegraphics[width=.26\linewidth]{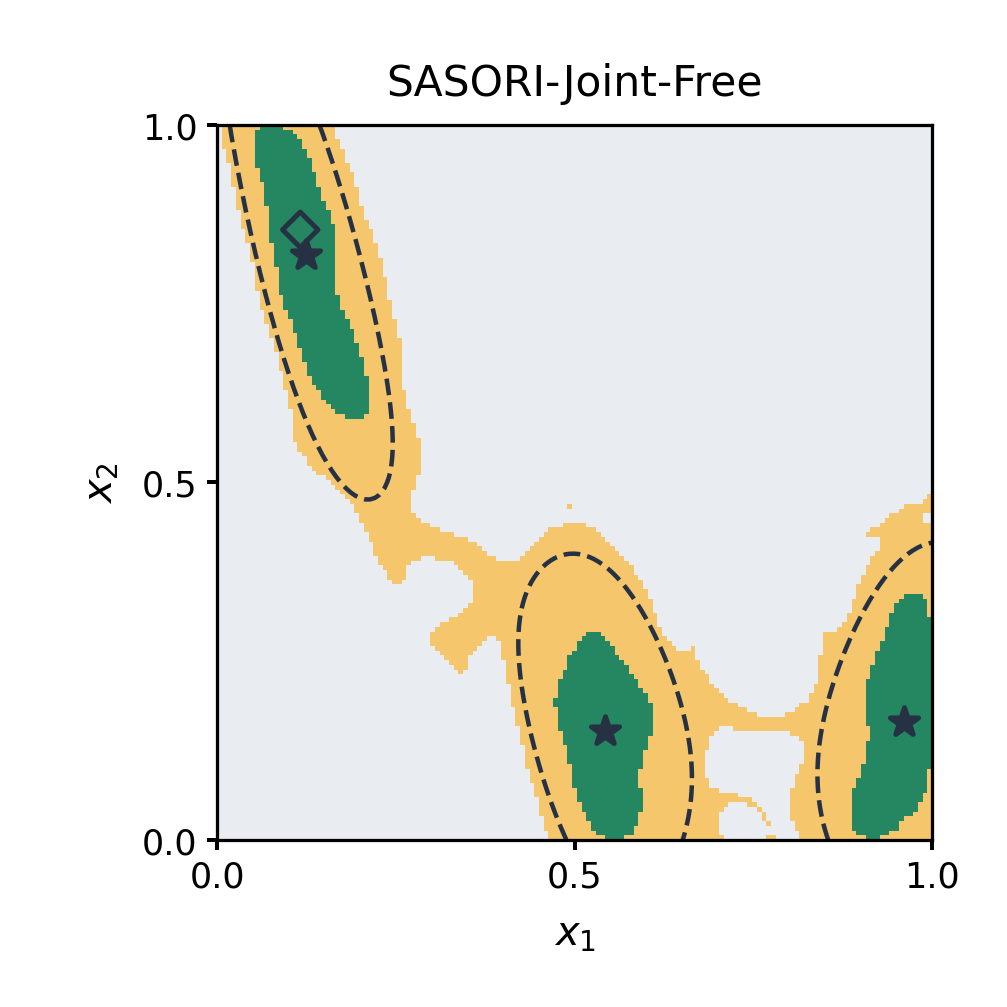}} & \raisebox{-.5\height}{\includegraphics[width=.26\linewidth]{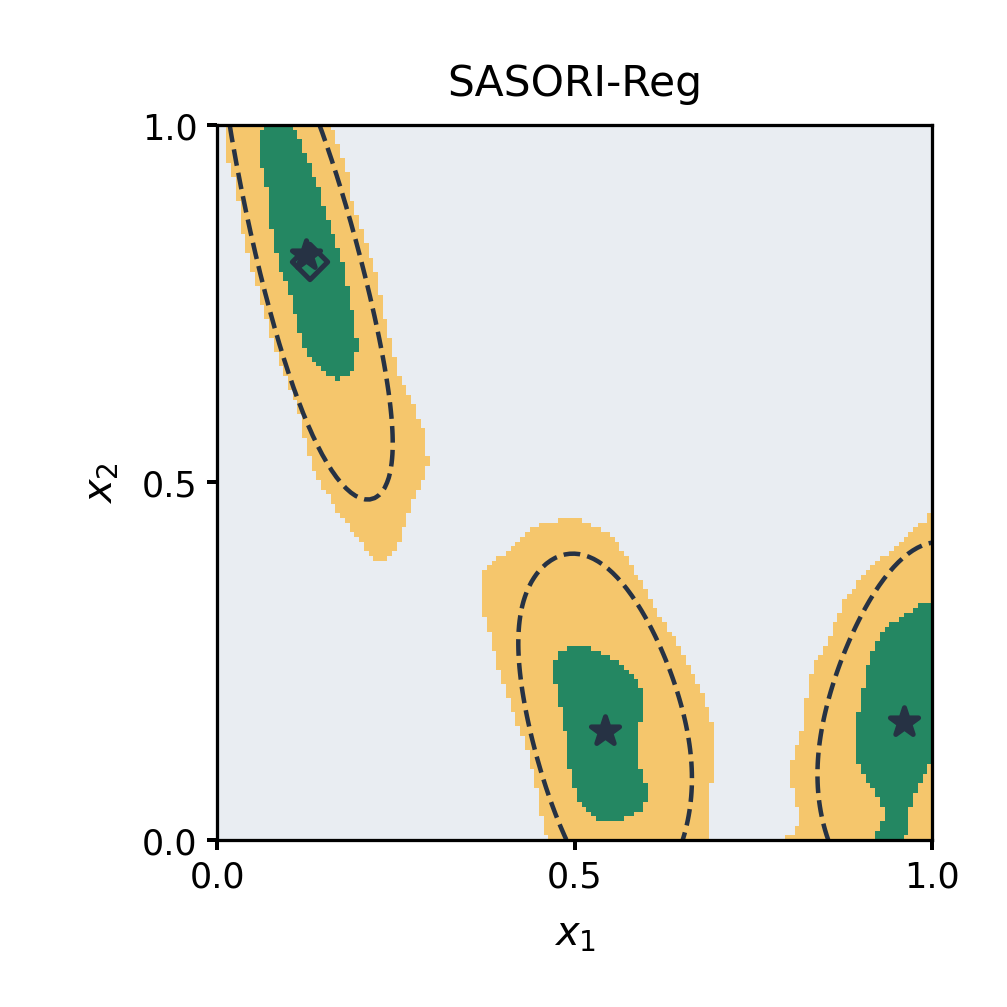}}\\[5pt]
{\small $\tau=0.08$} & \raisebox{-.5\height}{\includegraphics[width=.26\linewidth]{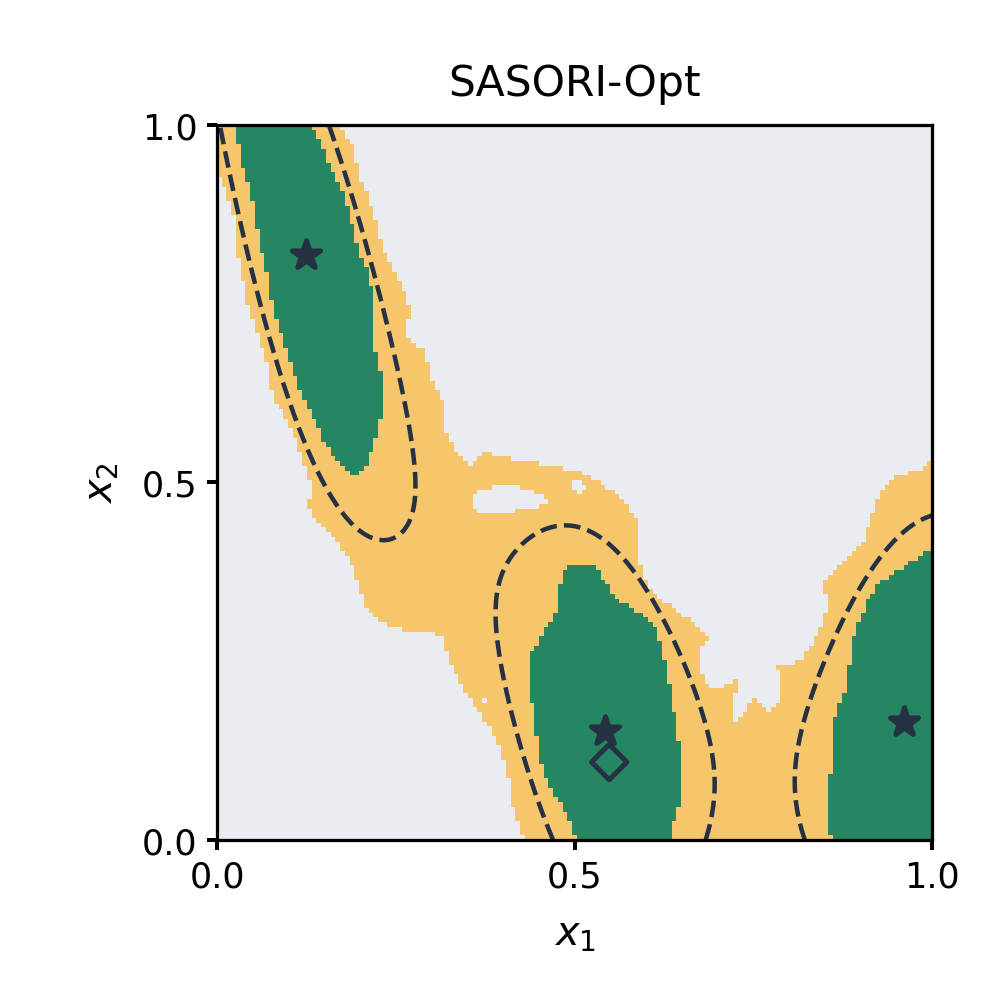}} & \raisebox{-.5\height}{\includegraphics[width=.26\linewidth]{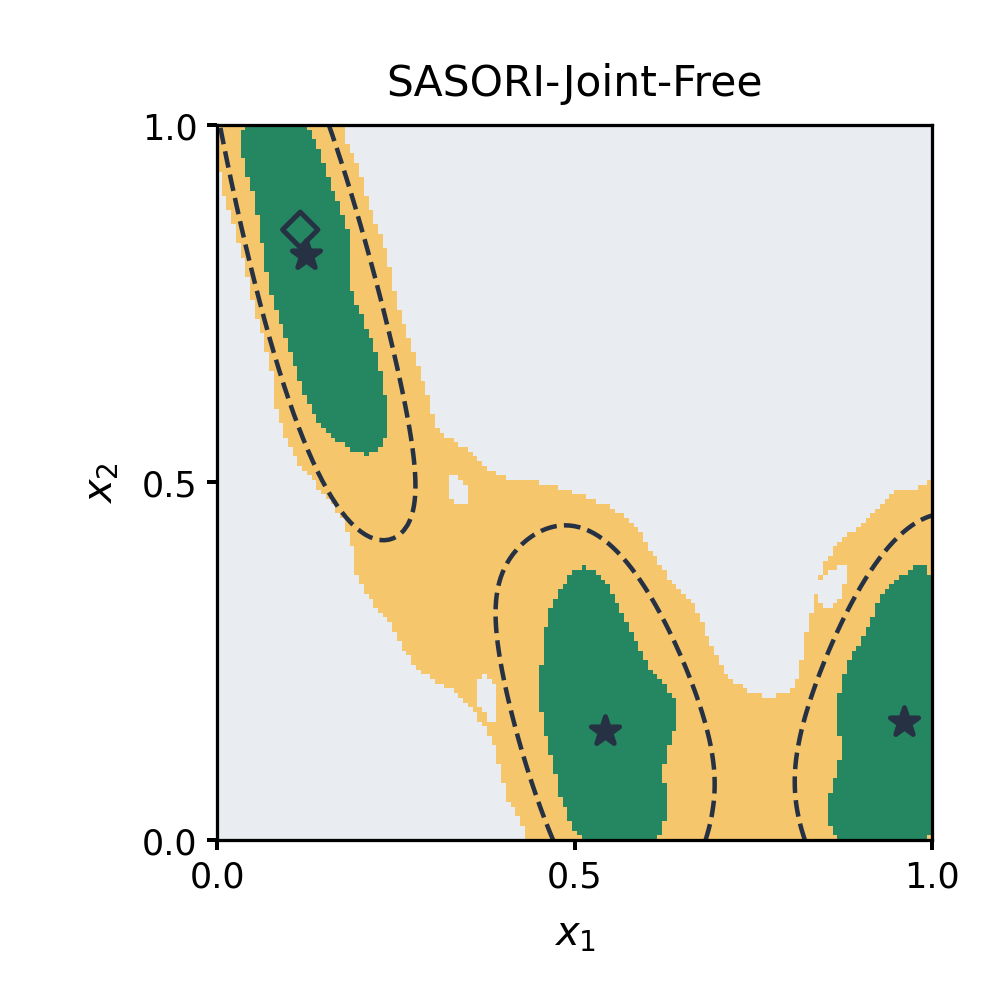}} & \raisebox{-.5\height}{\includegraphics[width=.26\linewidth]{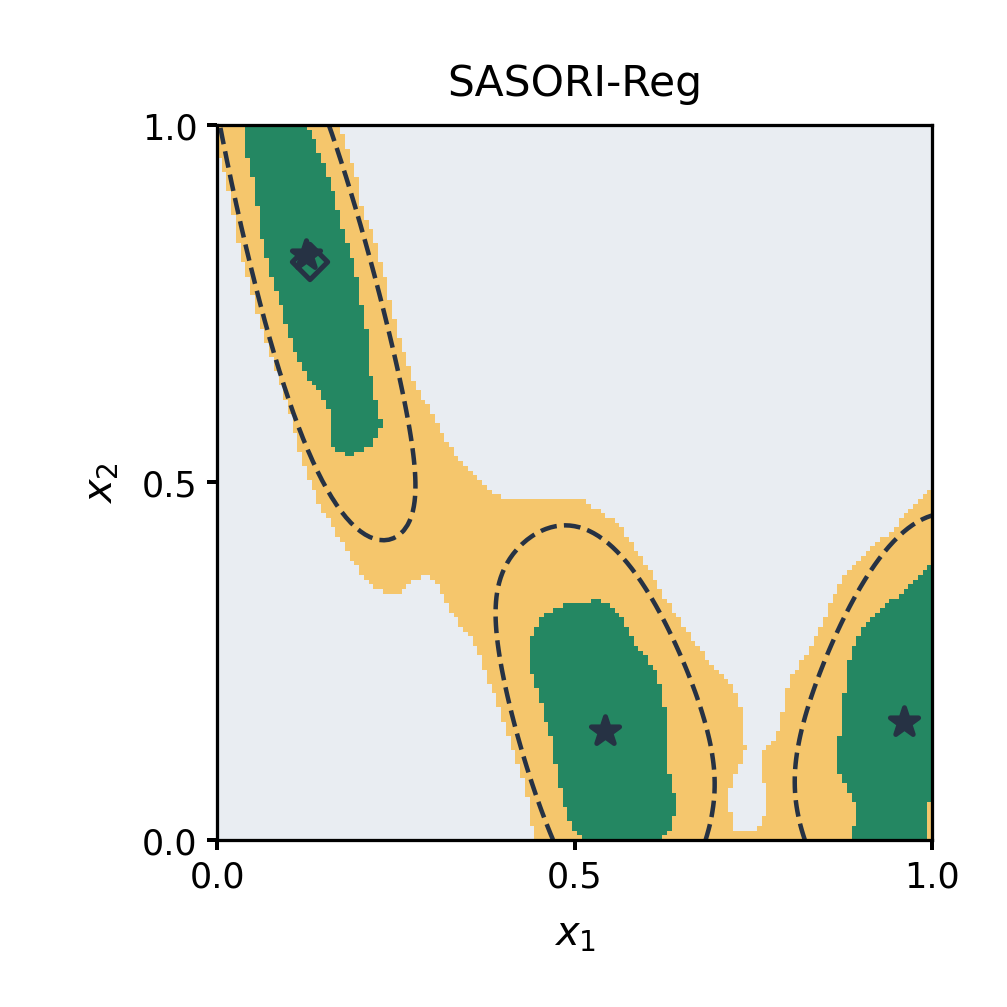}}\\[5pt]
\end{tabular}
\par\smallskip
\includegraphics[width=.97\linewidth]{figures/branin_region_legend.png}
\caption{Branin-2 algorithm regions at three additional tolerances.}\label{fig:ec-num-branin-maps}
\par\endgroup
\end{SASORIECblock}

\noindent
Figure~\ref{fig:ec-num-branin-maps} shows that the inner sets expand around the three maxima as $\tau$ increases. Opt recovers the largest portion of the acceptable region in this example, while Reg leaves less of the domain unclassified at $\tau=0.06$ and $0.08$. As in the main illustration, the green sets identify acceptable alternatives and the yellow sets show unresolved membership. 

\subsubsection{Final Regret}

We next report the final regret values corresponding to the Branin-2 curves in Figure~\ref{fig:num-branin-regret}.
\par\nopagebreak[4]

\begin{SASORIECblock}{table}
\begingroup\centering\small\singlespacing
\setlength{\abovecaptionskip}{4pt}\setlength{\belowcaptionskip}{4pt}
\setlength{\tabcolsep}{4pt}\renewcommand{\arraystretch}{1.12}
\caption{Branin-2 final regret (mean $\pm$ standard error).}\label{tab:num-branin}
\begin{tabular*}{\linewidth}{@{\extracolsep{\fill}}lccc@{}}
\toprule
Method & $10^3r_N$ & $10^5r_N^{\rm best}$ & $R_N$\\
\midrule
Opt & $1.938\pm0.356$ & $3.850\pm0.580$ & $152.77 \pm 2.18$\\
Joint-Free & $2.243\pm0.437$ & $4.854\pm0.955$ & $231.55 \pm 2.94$\\
Reg & $1.244\pm0.250$ & $4.536\pm0.964$ & $318.72 \pm 1.69$\\
GP-UCB & $0.356\pm0.081$ & $0.146\pm0.048$ & $17.37 \pm 0.46$\\
Sobol-GP & $2.124\pm0.431$ & $5.107\pm0.835$ & $943.34 \pm 0.11$\\
\bottomrule
\end{tabular*}
\par\endgroup
\end{SASORIECblock}

\noindent
Table~\ref{tab:num-branin} shows that Opt has the lowest mean best query and cumulative regret among the SASORI variants, while Reg has the lowest mean regret at the final recommendation. The latter may reflect how regional exploration also improves the confidence bounds used to select the recommended point.

\subsubsection{Running Time}\label{ec:num-phases}

We finally report the recorded sampling times for the five methods on Branin-2.
\par\nopagebreak[4]

\begin{SASORIECblock}{table}
\begingroup\centering\small\singlespacing
\setlength{\abovecaptionskip}{4pt}\setlength{\belowcaptionskip}{4pt}
\setlength{\tabcolsep}{4pt}\renewcommand{\arraystretch}{1.12}
\caption{Branin-2 recorded sampling time in minutes (mean $\pm$ standard error).}\label{tab:ec-num-time}
\begin{tabular*}{\linewidth}{
  @{\hspace{.2\linewidth}}
  >{\centering\arraybackslash}p{.3\linewidth}
  @{}
  >{\centering\arraybackslash}p{.3\linewidth}
  @{\hspace{.2\linewidth}}
}
\toprule
Method & Minutes\\
\midrule
Opt & $47.89 \pm 0.38$\\
Joint-Free & $44.75 \pm 0.40$\\
Reg & $48.42 \pm 0.40$\\
GP-UCB & $119.01 \pm 4.29$\\
Sobol-GP & $0.24 \pm 0.01$\\
\bottomrule
\end{tabular*}
\par\endgroup
\end{SASORIECblock}

\noindent
Table~\ref{tab:ec-num-time} shows that the three SASORI variants have similar recorded sampling times and take less time than GP-UCB in these runs. Sobol-GP requires the least computation time because it does not optimize an acquisition function.

\newpage
\subsection{Additional Hartmann-3 Results}\label{ec:num-hartmann}

\subsubsection{Other Tolerances}\label{ec:num-hartmann-tolerances}

We first compare Hartmann-3 empirical region error across $\tau\in\{0.12,0.18,0.24,0.30\}$ for the algorithm output and the GP fitted to all observations.
\par\nopagebreak[4]

\begin{SASORIECblock}{figure}
\begingroup\centering\singlespacing
\setlength{\abovecaptionskip}{4pt}\setlength{\belowcaptionskip}{4pt}
\includegraphics[width=.45\linewidth]{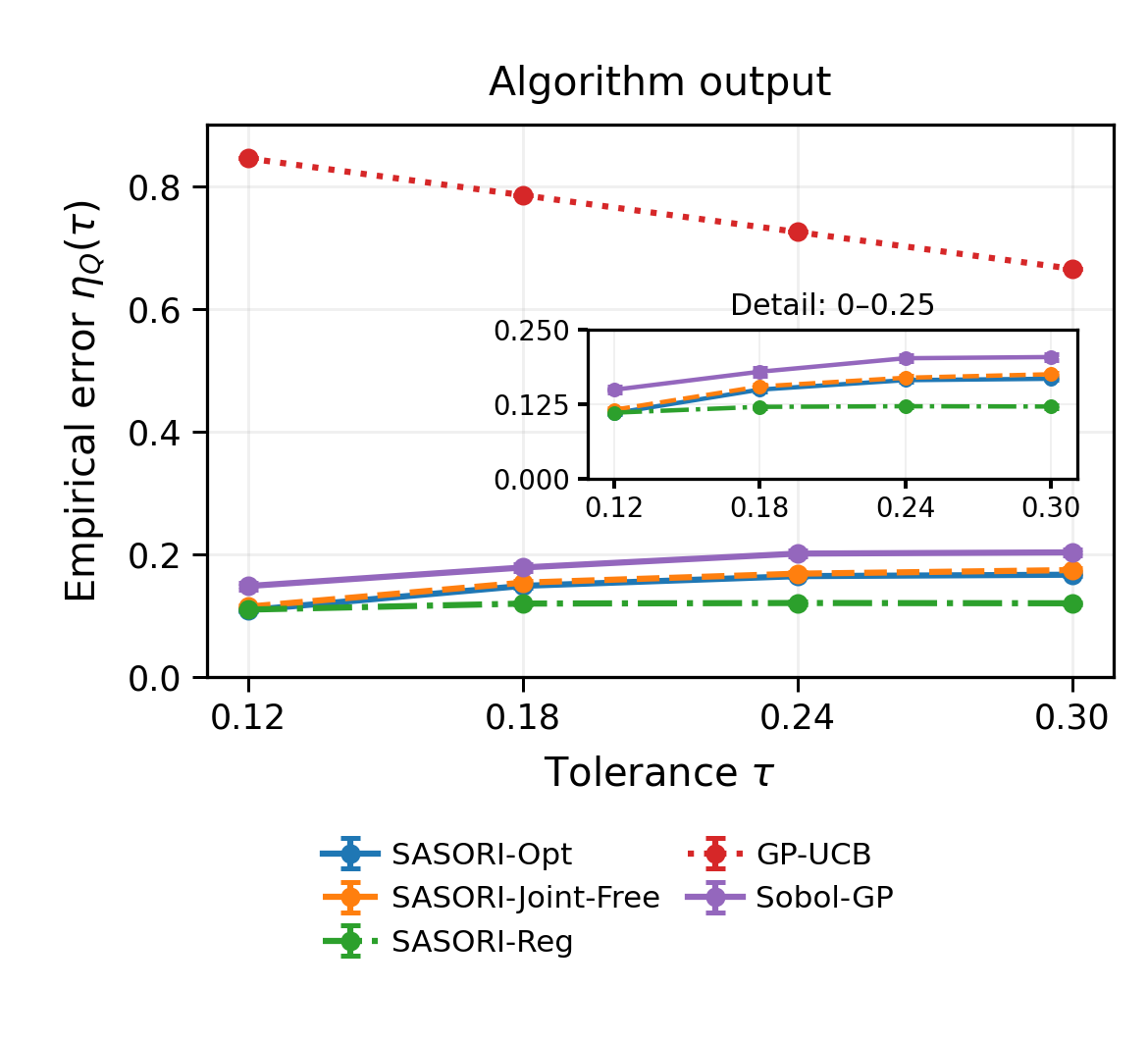}\hfill
\includegraphics[width=.45\linewidth]{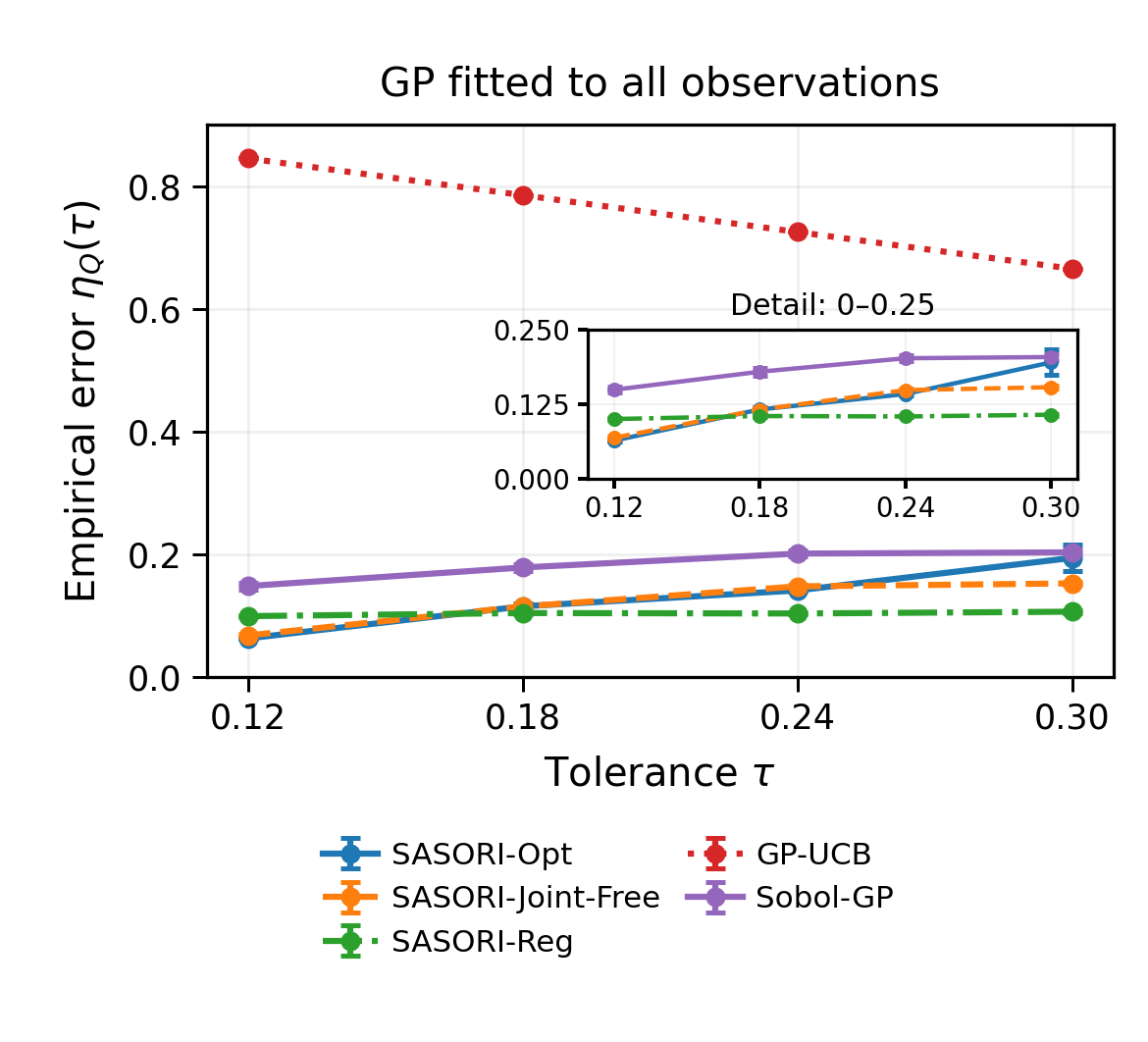}
\caption{Hartmann-3 empirical region error $\eta_Q(\tau)$: algorithm output (left) and GP fitted to all observations (right). Error bars show one standard error.}\label{fig:ec-num-hartmann-eta}
\par\endgroup
\end{SASORIECblock}

\noindent
Figure~\ref{fig:ec-num-hartmann-eta} shows that all three SASORI variants have smaller mean empirical region errors than either baseline across the tested tolerances under both reporting methods. Opt has the smallest mean error at the smallest tolerance, whereas Reg has the smallest error at the three larger tolerances. Reg's errors vary little across these larger tolerances, consistent with its emphasis on region estimation accuracy.

We next summarize the maximum error across these tolerances as $\eta_{Q,\max}/T$.
\par\nopagebreak[4]

\begin{SASORIECblock}{table}
\begingroup\centering\small\singlespacing
\setlength{\abovecaptionskip}{4pt}\setlength{\belowcaptionskip}{4pt}
\setlength{\tabcolsep}{4pt}\renewcommand{\arraystretch}{1.12}
\caption{Hartmann-3 maximum empirical region error across the four tested tolerances (mean $\pm$ standard error).}\label{tab:ec-num-hartmann-regions}
\begin{tabular*}{\linewidth}{@{\extracolsep{\fill}}lcc@{}}
\toprule
Method & Algorithm output: $\eta_{Q,\max}/T$ & All observations: $\eta_{Q,\max}/T$\\
\midrule
Opt & $0.571\pm0.006$ & $0.655\pm0.071$\\
Joint-Free & $0.593\pm0.009$ & $0.522\pm0.008$\\
Reg & $0.419\pm0.006$ & $0.366\pm0.005$\\
GP-UCB & $2.818815\pm0.000028$ & $2.818815\pm0.000028$\\
Sobol-GP & $0.733\pm0.017$ & $0.733\pm0.017$\\
\bottomrule
\end{tabular*}
\par\endgroup
\end{SASORIECblock}

\noindent
Table~\ref{tab:ec-num-hartmann-regions} shows that Reg has the smallest mean maximum error under both reporting methods. With all observations, Joint-Free also has a smaller maximum error than Opt. These values summarize the errors in Figure~\ref{fig:ec-num-hartmann-eta}, with the maximum taken within each run before averaging.

We then compare inner recovery across the same tolerances under both reporting methods.
\par\nopagebreak[4]

\begin{SASORIECblock}{figure}
\begingroup\centering\singlespacing
\setlength{\abovecaptionskip}{4pt}\setlength{\belowcaptionskip}{4pt}
\includegraphics[width=.45\linewidth]{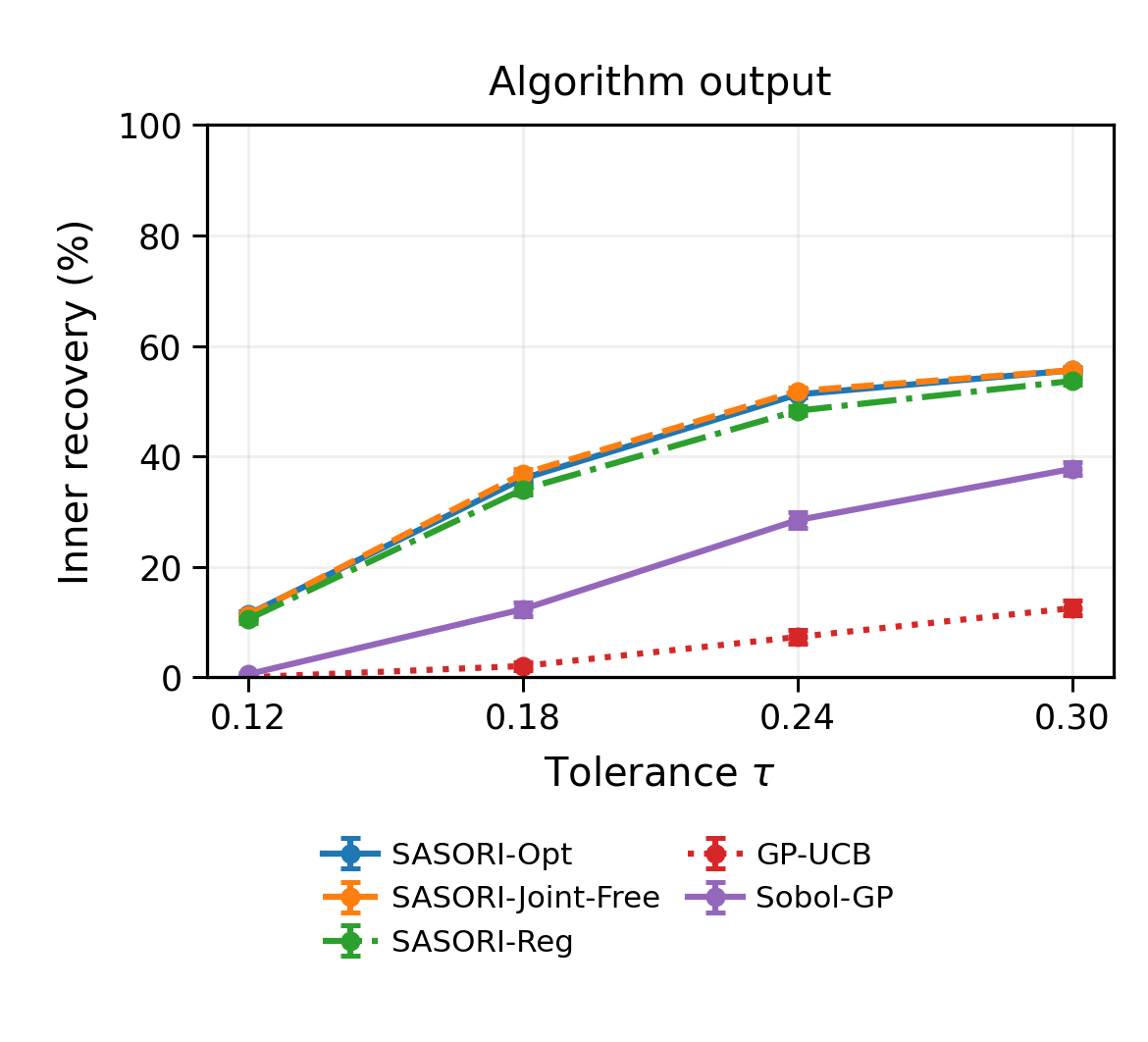}\hfill
\includegraphics[width=.45\linewidth]{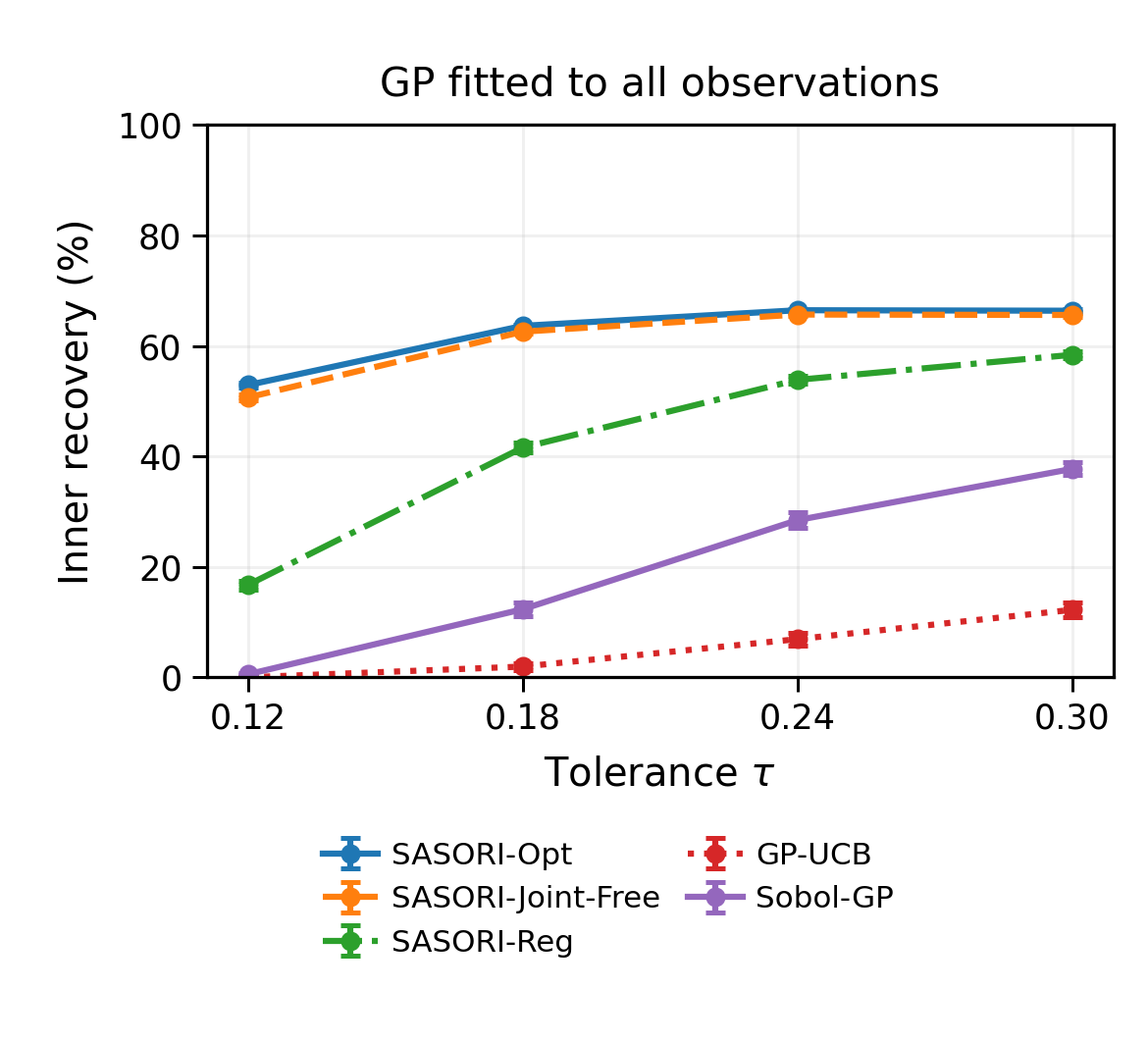}
\caption{Hartmann-3 inner recovery across tolerances: algorithm output (left) and GP fitted to all observations (right). Error bars show one standard error.}\label{fig:ec-num-hartmann-recovery}
\par\endgroup
\end{SASORIECblock}

\noindent
Figure~\ref{fig:ec-num-hartmann-recovery} shows that all three SASORI variants recover more acceptable designs than either baseline across the tested tolerances. Fitting one GP to all observations increases their mean recovery, with Opt and Joint-Free performing similarly.

We also compare the fractions of the domain that remain unclassified.
\par\nopagebreak[4]

\begin{SASORIECblock}{figure}
\begingroup\centering\singlespacing
\setlength{\abovecaptionskip}{4pt}\setlength{\belowcaptionskip}{4pt}
\includegraphics[width=.45\linewidth]{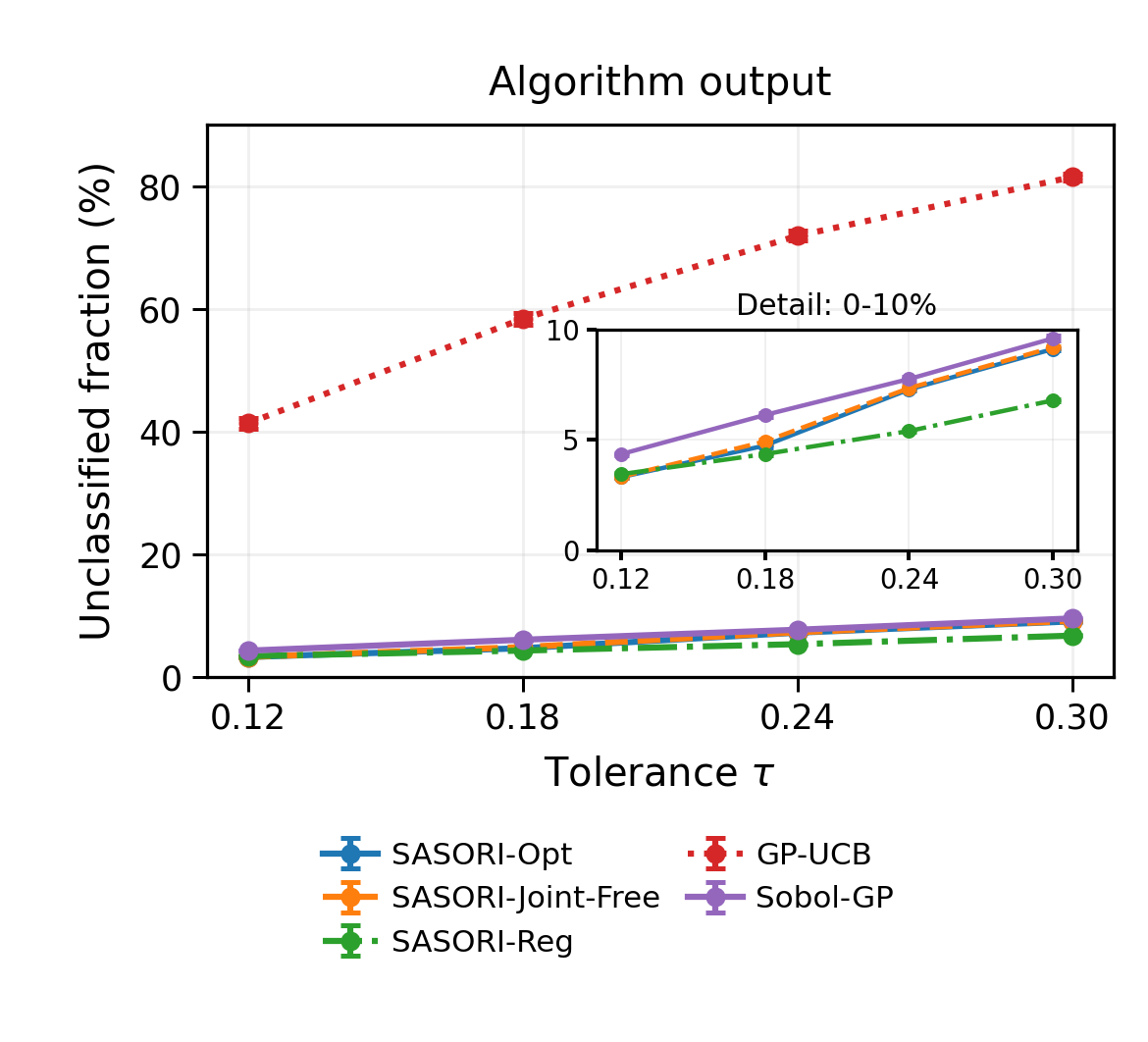}\hfill
\includegraphics[width=.45\linewidth]{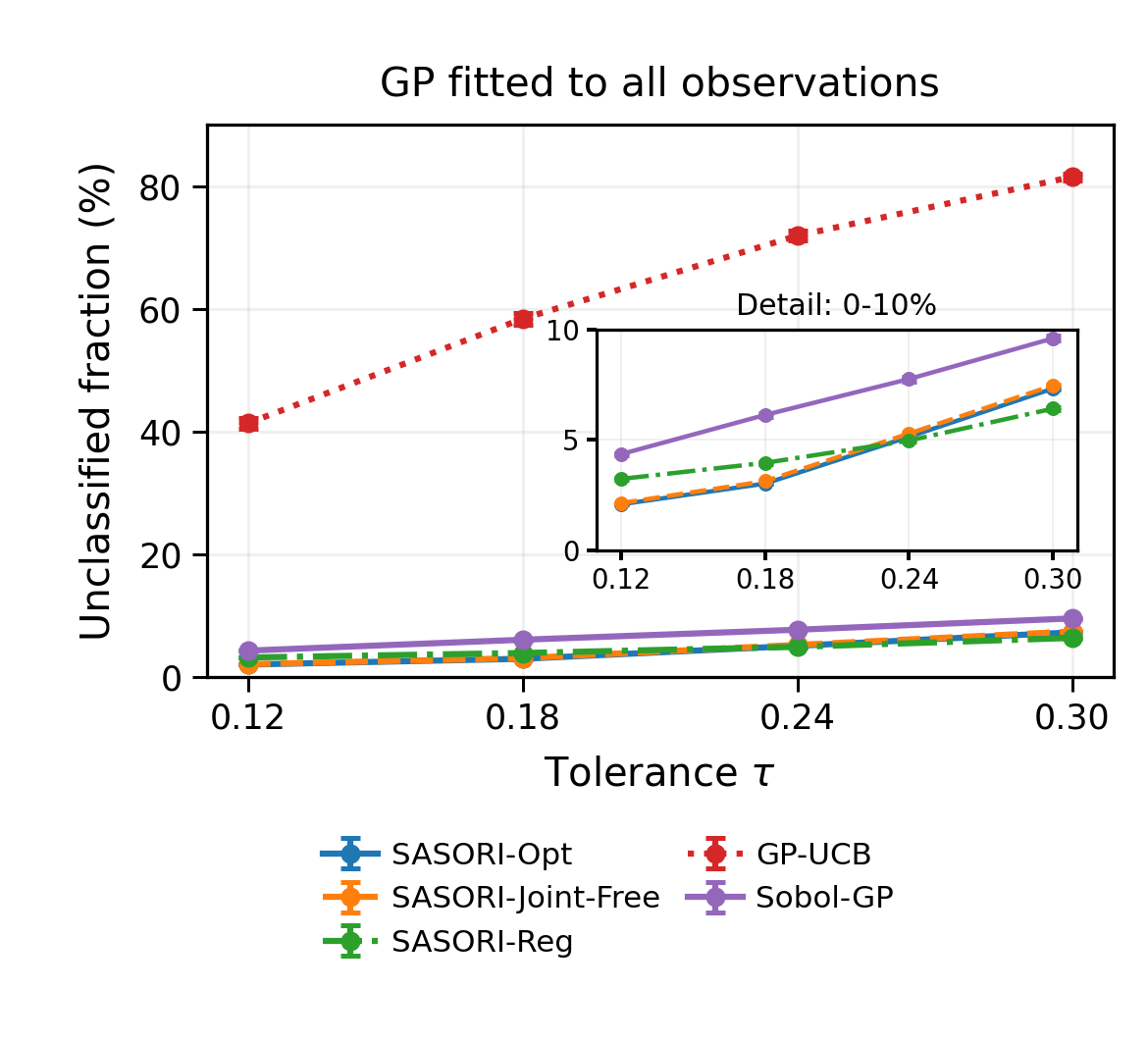}
\caption{Hartmann-3 unclassified fractions across tolerances: algorithm output (left) and GP fitted to all observations (right). Error bars show one standard error.}\label{fig:ec-num-hartmann-unclassified}
\par\endgroup
\end{SASORIECblock}

\noindent
Figure~\ref{fig:ec-num-hartmann-unclassified} shows that Reg has the smallest mean unclassified fraction at the three larger tolerances for the algorithm output and at the two largest tolerances when all observations are used. All three variants leave less of the domain unclassified than either baseline. As in Branin-2, Reg resolves membership effectively even when its inner recovery is lower than that of Opt or Joint-Free.

\subsubsection{Final Regret}

We next report the final regret values corresponding to the Hartmann-3 curves in Figure~\ref{fig:num-hartmann-regret}.
\par\nopagebreak[4]

\begin{SASORIECblock}{table}
\begingroup\centering\small\singlespacing
\setlength{\abovecaptionskip}{4pt}\setlength{\belowcaptionskip}{4pt}
\setlength{\tabcolsep}{4pt}\renewcommand{\arraystretch}{1.12}
\caption{Hartmann-3 final regret (mean $\pm$ standard error).}\label{tab:num-hartmann}
\begin{tabular*}{\linewidth}{@{\extracolsep{\fill}}lccc@{}}
\toprule
Method & $10^3r_N$ & $10^5r_N^{\rm best}$ & $R_N$\\
\midrule
Opt & $4.345\pm0.753$ & $69.990\pm7.690$ & $687.27 \pm 5.24$\\
Joint-Free & $4.591\pm0.951$ & $71.250\pm9.490$ & $824.05 \pm 5.83$\\
Reg & $5.038\pm0.916$ & $185.300\pm22.500$ & $1711.04 \pm 6.84$\\
GP-UCB & $0.500\pm0.104$ & $3.048\pm0.844$ & $90.72 \pm 0.65$\\
Sobol-GP & $2.685\pm0.477$ & $386.300\pm43.800$ & $2554.40 \pm 0.09$\\
\bottomrule
\end{tabular*}
\par\endgroup
\end{SASORIECblock}

\noindent
Table~\ref{tab:num-hartmann} shows that Opt has the lowest mean recommendation, best query, and cumulative regret among the SASORI variants. Joint-Free maintains similar recommendation quality while incurring substantially less cumulative regret than Reg.

\subsubsection{Running Time}\label{ec:num-hartmann-phases}

We finally report the recorded sampling times for the five methods on Hartmann-3.
\par\nopagebreak[4]

\begin{SASORIECblock}{table}
\begingroup\centering\small\singlespacing
\setlength{\abovecaptionskip}{4pt}\setlength{\belowcaptionskip}{4pt}
\setlength{\tabcolsep}{4pt}\renewcommand{\arraystretch}{1.12}
\caption{Hartmann-3 recorded sampling time in minutes (mean $\pm$ standard error).}\label{tab:ec-num-hartmann-time}
\begin{tabular*}{\linewidth}{
  @{\hspace{.2\linewidth}}
  >{\centering\arraybackslash}p{.3\linewidth}
  @{}
  >{\centering\arraybackslash}p{.3\linewidth}
  @{\hspace{.2\linewidth}}
}
\toprule
Method & Minutes\\
\midrule
Opt & $47.62 \pm 0.63$\\
Joint-Free & $44.33 \pm 0.60$\\
Reg & $45.82 \pm 0.57$\\
GP-UCB & $141.33 \pm 4.91$\\
Sobol-GP & $0.31 \pm 0.01$\\
\bottomrule
\end{tabular*}
\par\endgroup
\end{SASORIECblock}

\noindent
Table~\ref{tab:ec-num-hartmann-time} shows similar recorded sampling times for the three SASORI variants, all below GP-UCB in these runs. Sobol-GP is fastest because its query locations do not require acquisition optimization.

\endgroup

\end{document}